%% file: main.tex
\pdfoutput=1
\documentclass[10pt,twocolumn,letterpaper]{article}

\usepackage[pagenumbers]{wacv}              %

\input{preamble}
\usepackage{multicol}

\definecolor{wacvblue}{rgb}{0.21,0.49,0.74}
\usepackage[pagebackref,breaklinks,colorlinks,allcolors=wacvblue]{hyperref}

\def\wacvPaperID{460} %
\def\confName{WACV}
\def\confYear{2027}

\title{RefRef: A Dataset and Benchmark for Reconstructing\\ Refractive and Reflective Objects}

\author{Yue Yin$^{1}$ \hspace{3em} Enze Tao$^{1}$ \hspace{3em} Weijian Deng$^{2}$ \hspace{3em} Dylan Campbell$^{1}$\\
{\normalsize $^{1}$Australian National University \quad $^{2}$Tsinghua Shenzhen International Graduate School, Tsinghua University}\\
{\tt\small \{yue.yin1, enze.tao, dylan.campbell\}@anu.edu.au \quad dengwj16@sz.tsinghua.edu.cn}
}

\usepackage{tikz}

\newlength{\teaserdatasetgap}
\newlength{\teaseropticalgap}
\newlength{\teaserdatasetgroupwidth}
\newsavebox{\teaserimagebox}

\newlength{\teaserimagewidth}
\newlength{\teaserimageheight}
\newlength{\teaserimagehalfwidth}

\newlength{\teasercropleft}
\newlength{\teasercropbottom}
\newlength{\teasercropright}
\newlength{\teasercroptop}

\newlength{\teasercroppedwidth}
\newlength{\teasercroppedheight}
\newlength{\teaserherocroptopamount}
\newlength{\teaserherocropbottomamount}

\newcommand{\teaserherocroptopfraction}{0.03}
\newcommand{\teaserherocropbottomfraction}{0.06}

\newcommand{\teaserdatasethalfwidth}{%
    \dimexpr(\linewidth-\teaserdatasetgap)/2\relax
}

\pgfkeys{
    /teasercrop/.is family,
    /teasercrop,
    left/.initial=0,
    bottom/.initial=0,
    right/.initial=0,
    top/.initial=0,
}

\newcommand{\teaserlabelledimage}[5]{%
    \begingroup

    \sbox{\teaserimagebox}{%
        \includegraphics[width=#1]{#2}%
    }%

    \setlength{\teaserimagewidth}{\wd\teaserimagebox}%
    \setlength{\teaserimageheight}{\ht\teaserimagebox}%

    \pgfkeys{
        /teasercrop,
        left=0,
        bottom=0,
        right=0,
        top=0,
        #5
    }%

    \setlength{\teasercropleft}
        {\pgfkeysvalueof{/teasercrop/left}\teaserimagewidth}%
    \setlength{\teasercropbottom}
        {\pgfkeysvalueof{/teasercrop/bottom}\teaserimageheight}%
    \setlength{\teasercropright}
        {\pgfkeysvalueof{/teasercrop/right}\teaserimagewidth}%
    \setlength{\teasercroptop}
        {\pgfkeysvalueof{/teasercrop/top}\teaserimageheight}%

    \setlength{\teasercroppedwidth}{\teaserimagewidth}%
    \addtolength{\teasercroppedwidth}{-\teasercropleft}%
    \addtolength{\teasercroppedwidth}{-\teasercropright}%

    \setlength{\teasercroppedheight}{\teaserimageheight}%
    \addtolength{\teasercroppedheight}{-\teasercropbottom}%
    \addtolength{\teasercroppedheight}{-\teasercroptop}%

    \begin{tikzpicture}[baseline=0pt]
        \path[use as bounding box]
            (0,0) rectangle
            (\teaserimagewidth,\teaserimageheight);

        \begin{scope}
            \clip
                (0,0) rectangle
                (\teaserimagewidth,\teaserimageheight);

            \node[
                anchor=south west,
                inner sep=0pt,
                minimum width=\teaserimagewidth,
                minimum height=\teaserimageheight
            ] at (0,0) {%
                \resizebox{\teaserimagewidth}{\teaserimageheight}{%
                    \begin{tikzpicture}
                        \clip
                            (0,0) rectangle
                            (\teasercroppedwidth,\teasercroppedheight);

                        \node[
                            anchor=south west,
                            inner sep=0pt
                        ] at (-\teasercropleft,-\teasercropbottom) {%
                            \usebox{\teaserimagebox}%
                        };
                    \end{tikzpicture}%
                }%
            };
        \end{scope}

        \if\relax\detokenize{#3}\relax
        \else
            \node[
                anchor=south,
                yshift=1.5pt,
                fill=black,
                fill opacity=0.68,
                text opacity=1,
                text=white,
                rounded corners=1pt,
                inner xsep=1.8pt,
                inner ysep=0.8pt,
                font={#4}
            ] at (.5\teaserimagewidth,0) {#3};
        \fi
    \end{tikzpicture}%

    \endgroup
}

\newcommand{\teasersmalllabelfont}{%
    \fontsize{6.5}{7.0}\selectfont
}

\newcommand{\teasersmallcell}[3]{%
    \begin{minipage}[t]{\teaserdatasethalfwidth}
        \centering
        \teaserlabelledimage
            {\linewidth}
            {#1}
            {#2}
            {\teasersmalllabelfont}
            {#3}
    \end{minipage}%
}

\newcommand{\teaserheroimage}[4]{%
    \begingroup

    \sbox{\teaserimagebox}{%
        \includegraphics[width=#1]{#2}%
    }%

    \setlength{\teaserimagewidth}{\wd\teaserimagebox}%
    \setlength{\teaserimageheight}{\ht\teaserimagebox}%
    \setlength{\teaserimagehalfwidth}{.5\teaserimagewidth}%
    \setlength{\teaserherocroptopamount}
        {\teaserherocroptopfraction\teaserimageheight}%
    \setlength{\teaserherocropbottomamount}
        {\teaserherocropbottomfraction\teaserimageheight}%

    \setlength{\teasercroppedheight}{\teaserimageheight}%
    \addtolength{\teasercroppedheight}{-\teaserherocroptopamount}%
    \addtolength{\teasercroppedheight}{-\teaserherocropbottomamount}%

    \begin{tikzpicture}[baseline=0pt]
        \path[use as bounding box]
            (0,0) rectangle
            (\teaserimagewidth,\teasercroppedheight);

        \begin{scope}
            \clip
                (0,0) rectangle
                (\teaserimagewidth,\teasercroppedheight);

            \node[
                anchor=south west,
                inner sep=0pt
            ] at (0,-\teaserherocropbottomamount) {%
                \usebox{\teaserimagebox}%
            };
        \end{scope}

        \if\relax\detokenize{#3}\relax
        \else
            \node[
                anchor=south,
                yshift=1.5pt,
                fill=black,
                fill opacity=0.68,
                text opacity=1,
                text=white,
                rounded corners=1pt,
                inner xsep=1.8pt,
                inner ysep=0.8pt,
                font={#4}
            ] at (\teaserimagehalfwidth,0) {#3};
        \fi
    \end{tikzpicture}%

    \endgroup
}

\newcommand{\teaserdomainband}[9]{%
    \begin{minipage}[t]{\teaserdatasethalfwidth}
        \vspace{0pt}
        \centering

        {\footnotesize #1\par}
        \vspace{1pt}

        \teaserheroimage
            {\linewidth}
            {#2}
            {#3}
            {\scriptsize}\par

        \vspace{\teaserdatasetgap}

        \teasersmallcell{#4}{#5}{#6}%
        \hspace{\teaserdatasetgap}%
        \teasersmallcell{#7}{#8}{#9}
    \end{minipage}%
}

\newlength{\teaserlabelimagewidth}

\newcommand{\teaserimagewithlabel}[3]{%
    \setlength{\teaserlabelimagewidth}{#1}%
    \begin{tikzpicture}[baseline=0pt]
        \node[
            anchor=south west,
            inner sep=0pt
        ] at (0,0) {%
            \includegraphics[width=\teaserlabelimagewidth]{#2}%
        };

        \node[
            anchor=south,
            yshift=1.5pt,
            fill=black,
            fill opacity=0.68,
            text opacity=1,
            text=white,
            rounded corners=1pt,
            inner xsep=1.8pt,
            inner ysep=0.8pt,
            font={\scriptsize}
        ] at (.5\teaserlabelimagewidth,0) {#3\vphantom{fp}};
    \end{tikzpicture}%
}

\newcommand{\refrefteaser}{%
    \par
    \begin{minipage}{\textwidth}
        \captionsetup{type=figure}
        \centering

        \setlength{\teaserdatasetgroupwidth}{%
            \dimexpr
                (\linewidth
                -0.235\linewidth
                -\teaseropticalgap
                -\teaserdatasetgap)/2
            \relax
        }%

        \noindent\makebox[\linewidth][c]{%
            \begin{minipage}[b]{0.235\linewidth}
                \centering

                \includegraphics[
                    width=0.75\linewidth
                ]{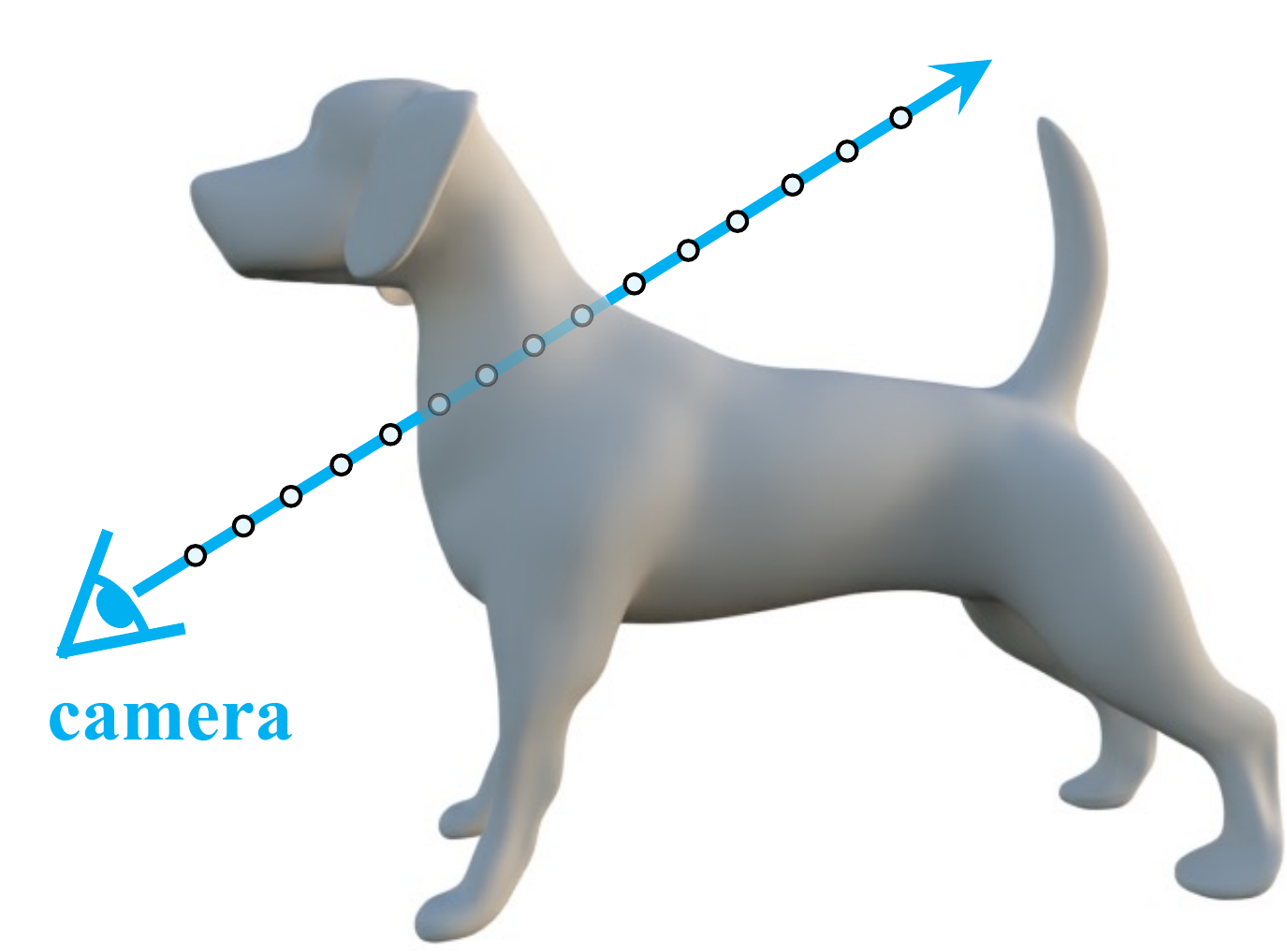}\par
                \vspace{-2pt}
                {\small (a) Opaque and Lambertian\par}

                \vspace{3pt}

                \includegraphics[
                    width=\linewidth
                ]{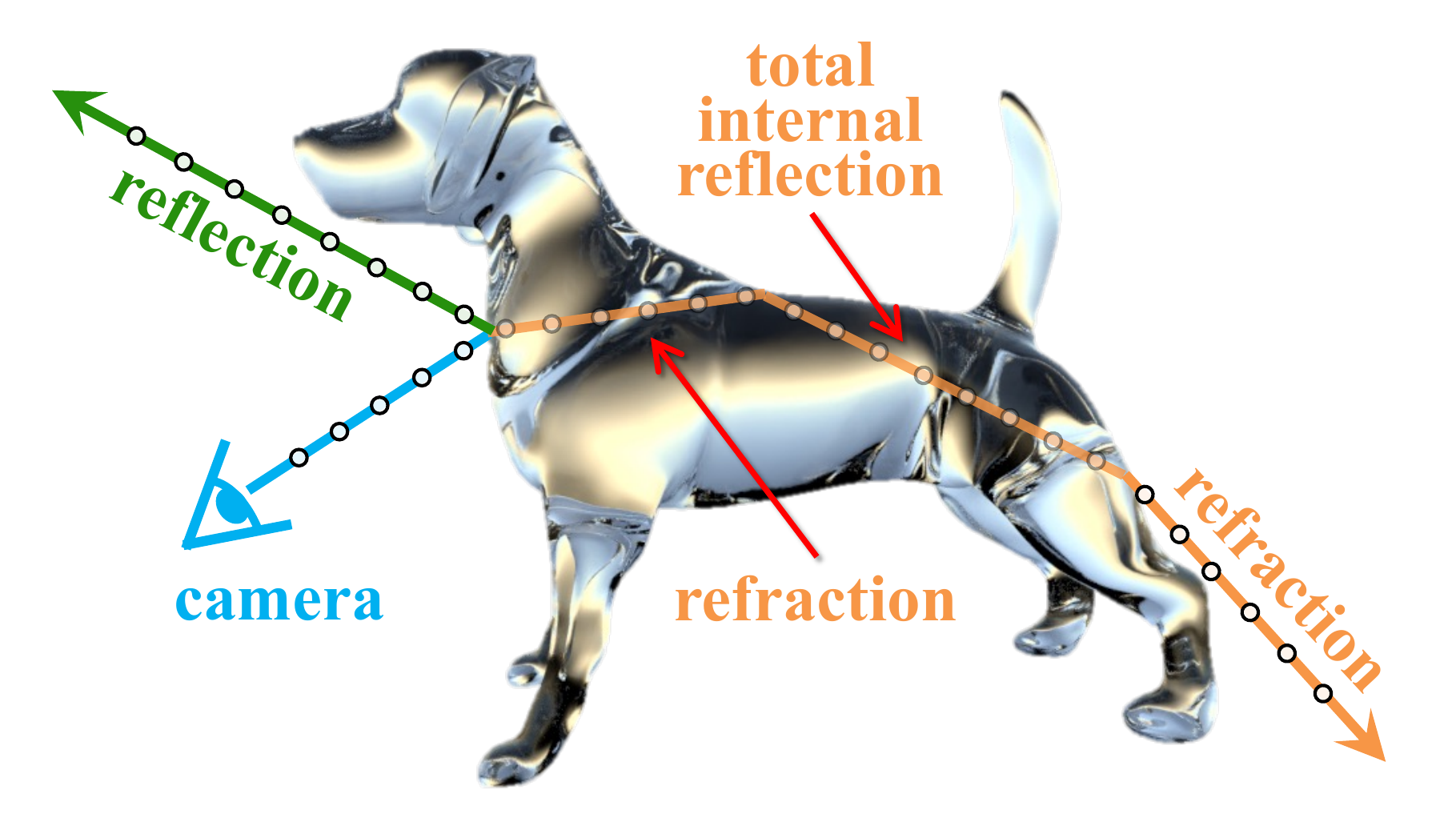}\par
                \vspace{-2pt}
                {\small (b) Refractive and Reflective\par}
            \end{minipage}%
            \hspace{\teaseropticalgap}%
            \begin{minipage}[b]{\teaserdatasetgroupwidth}
                \centering
            
                \teaserdomainband
                    {Bounded}
                    {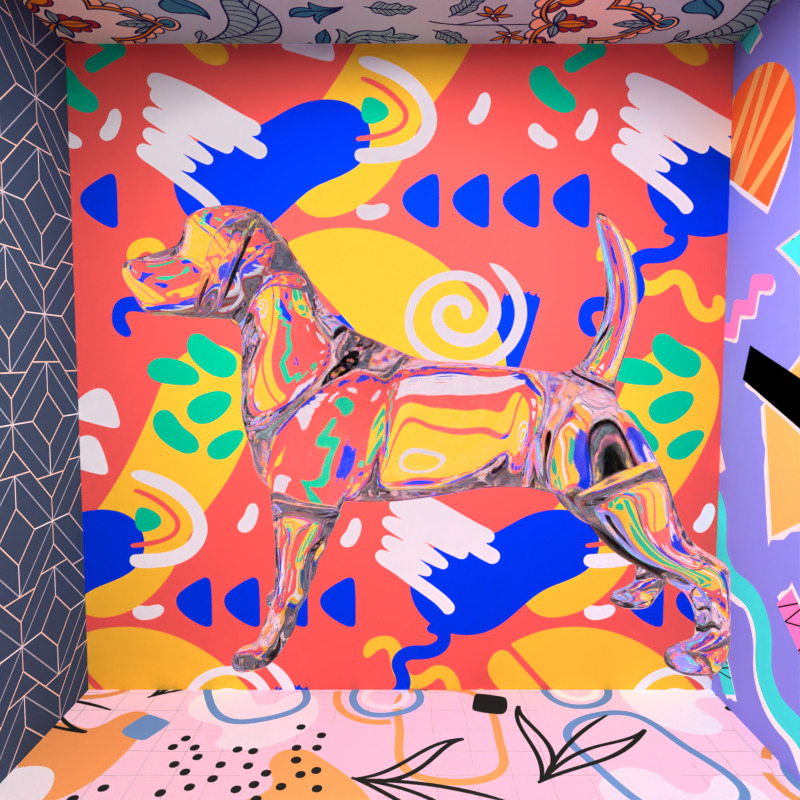}
                    {}
                    {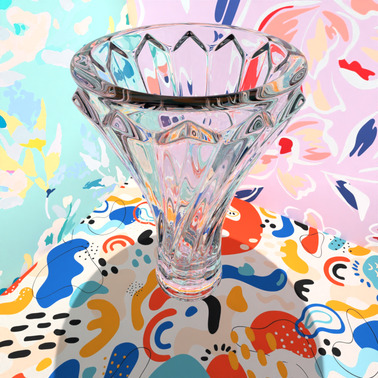}
                    {}
                    {}
                    {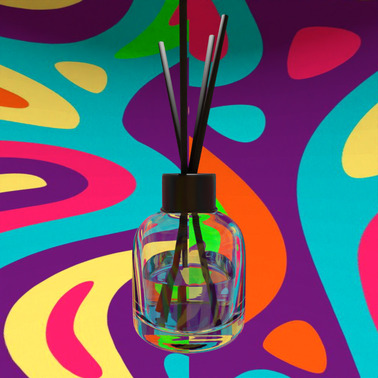}
                    {}
                    {}%
                \hspace{\teaserdatasetgap}%
                \teaserdomainband
                    {Unbounded}
                    {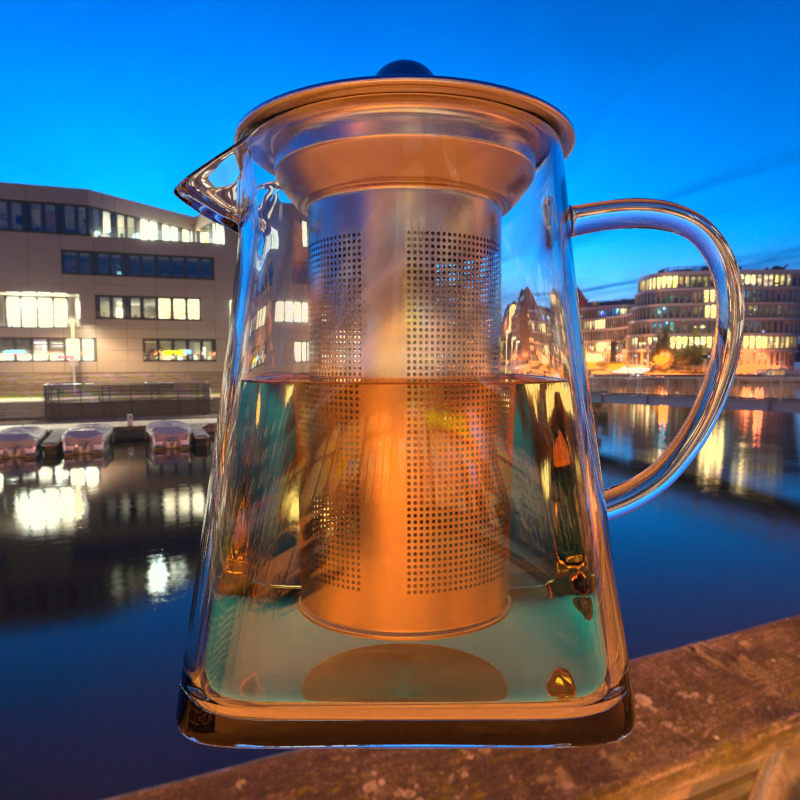}
                    {}
                    {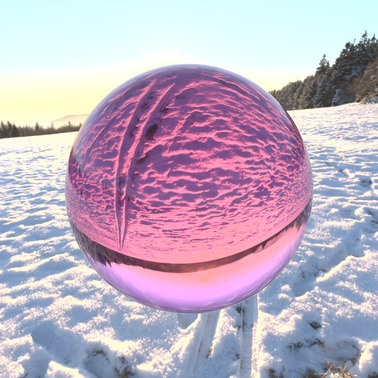}
                    {}
                    {}
                    {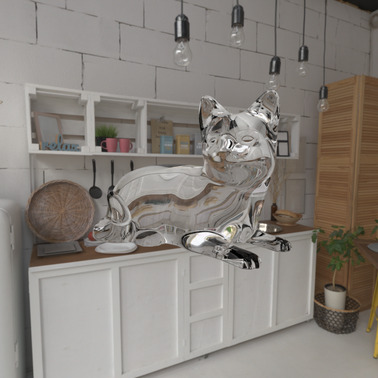}
                    {}
                    {}%
            
                \par\vspace{\teaserdatasetgap}
                {\small (c) Synthetic Scenes\par}
            \end{minipage}%
            \hspace{-1pt}
            \begin{minipage}[b]{\teaserdatasetgroupwidth}
                \centering

                \teaserdomainband
                    {Indoor}
                    {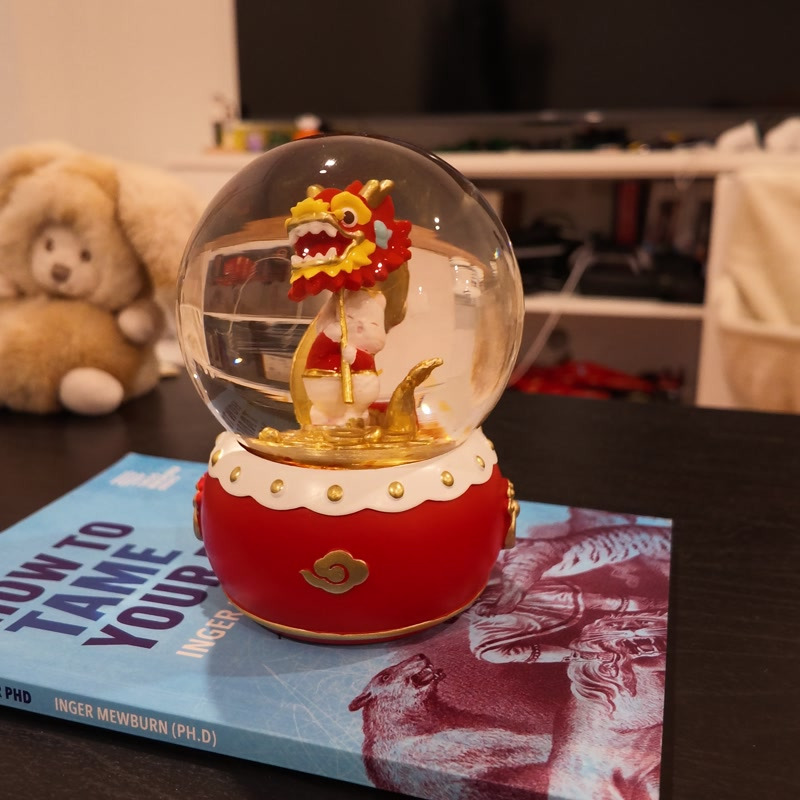} {}
                    {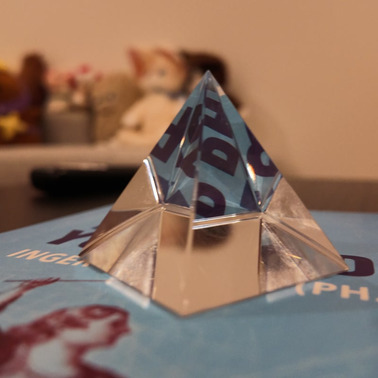} {} {}
                    {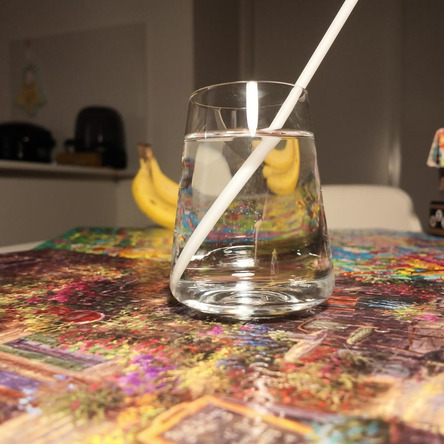} {} {left=0.15,right=0.0,bottom=0.15}%
                \hspace{\teaserdatasetgap}%
                \teaserdomainband
                    {Outdoor}
                    {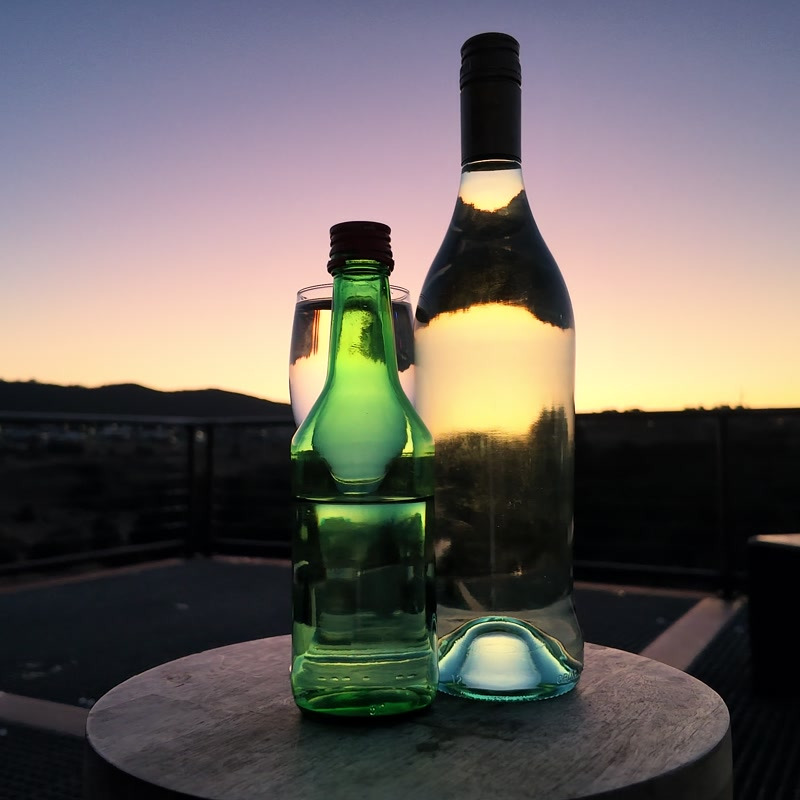} {}
                    {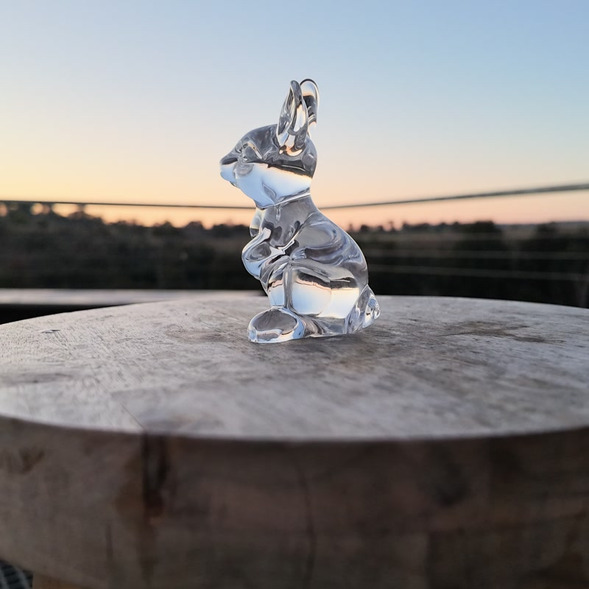} {} {left=0.2,bottom=0.29, top=0.06, right=0.15}
                    {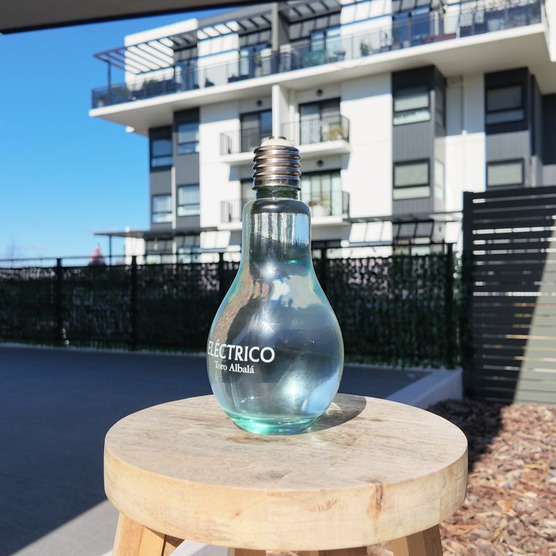} {} {bottom=0.12,left=0.18, right=0.14,top=0.2}%

                \par\vspace{\teaserdatasetgap}
                {\small (d) Real-world Scenes\par}
            \end{minipage}%
        }

        \begin{minipage}{\linewidth}
            \centering
            \teaserimagewithlabel
                {0.108\linewidth}
                {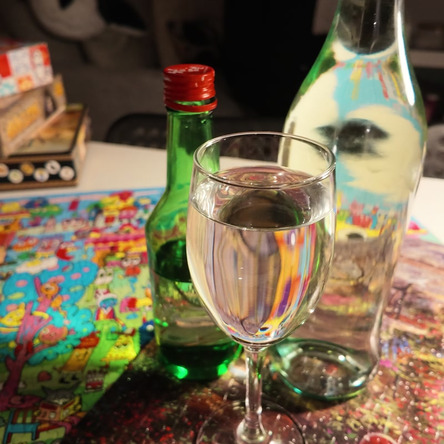}
                {GT}%
            \hspace{1pt}%
            \teaserimagewithlabel
                {0.108\linewidth}
                {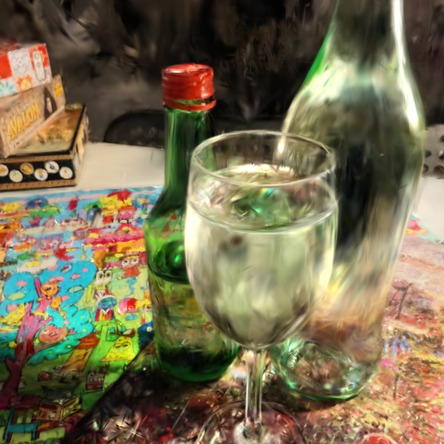}
                {Splatfacto}%
            \hspace{1pt}%
            \teaserimagewithlabel
                {0.108\linewidth}
                {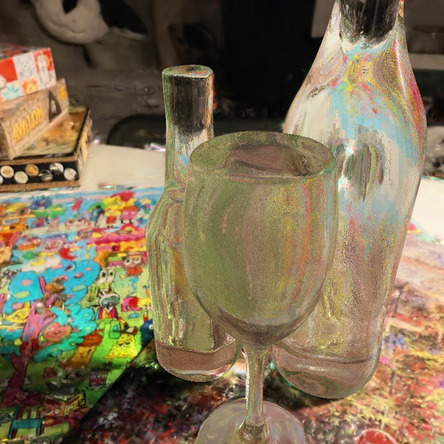}
                {TransparentGS}%
            \hspace{1pt}%
            \teaserimagewithlabel
                {0.108\linewidth}
                {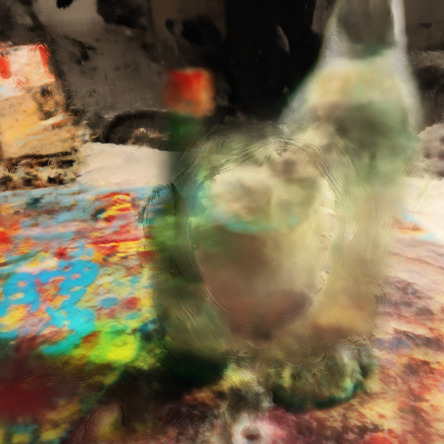}
                {NeuS}%
            \hspace{1pt}%
            \teaserimagewithlabel
                {0.108\linewidth}
                {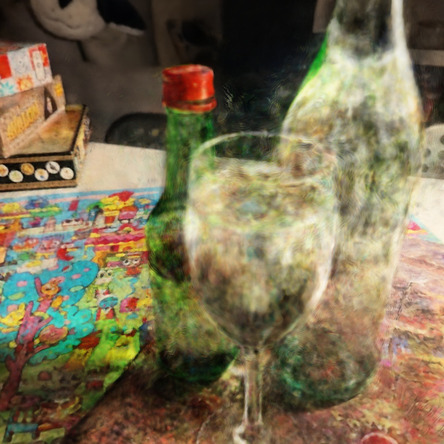}
                {Zip-NeRF}%
            \hspace{1pt}%
            \teaserimagewithlabel
                {0.108\linewidth}
                {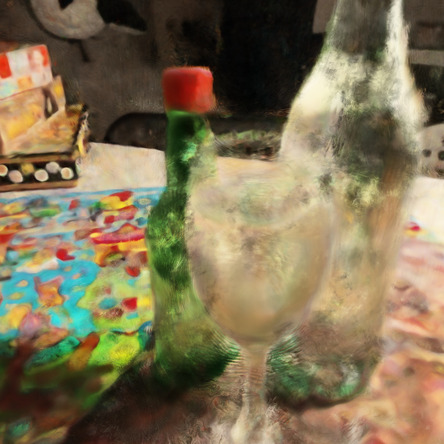}
                {MS-NeRF}%
            \hspace{1pt}%
            \teaserimagewithlabel
                {0.108\linewidth}
                {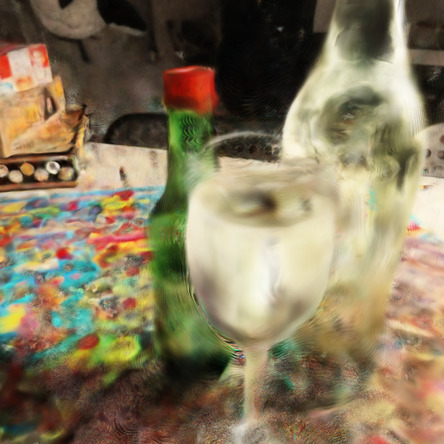}
                {RayDef}%
            \hspace{1pt}%
            \teaserimagewithlabel
                {0.108\linewidth}
                {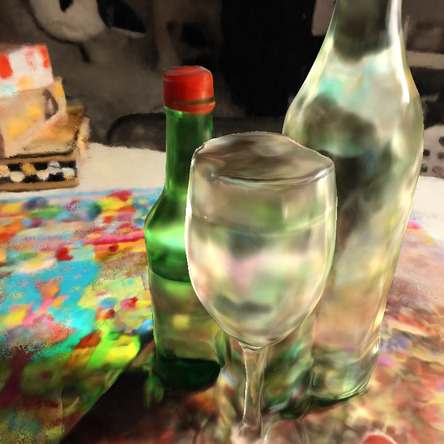}
                {NU-NeRF}%
            \hspace{1pt}%
            \teaserimagewithlabel
                {0.108\linewidth}
                {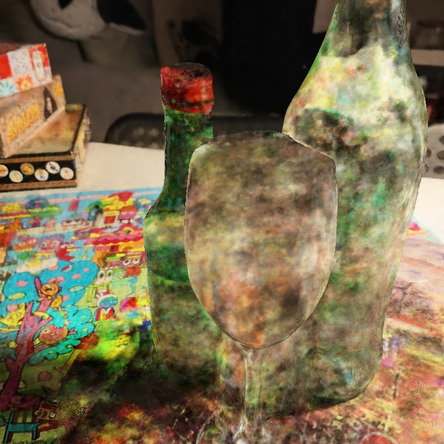}
                {R3F}%
            \par\vspace{-1pt}
            {\small (e) An Unsolved Open Problem\par}
        \end{minipage}
        \vspace{-9pt}
        \caption{
            Modeling refractive light transport for reconstruction and novel view synthesis is challenging, because while
            (a)~opaque Lambertian objects allow modeling of light paths as linear,
            (b)~refractive and reflective objects bend and branch light paths.
            While many methods address this issue in simplified settings, they are not evaluated on a comprehensive dataset that considers all of the factors that make this problem challenging.
            We present the \dataset{} dataset that contains
            (c)~150 synthetic scenes with diverse bounded and unbounded backgrounds, and
            (d)~60 real scenes from varied indoor and outdoor environments.
            Different levels of object complexity are considered, including convex, nonconvex, and multi-material objects, as well as tinted materials and multi-object configurations.
            Our benchmark reveals that the
            task is far from solved for refractive objects: not a single existing method can handle real scenes with multiple objects~(e).
        }
        \label{fig:teaser}
    \end{minipage}

    \par\vspace*{12pt}
}

\makeatletter
\g@addto@macro\@maketitle{\refrefteaser%
    \begingroup
    \setlength{\multicolsep}{0pt}%
    \renewenvironment{abstract}
        {\begin{multicols}{2}\centerline{\large\bf Abstract}\vspace{0pt}\noindent\it\ignorespaces}
        {\end{multicols}}%
    \input{abstract}    \endgroup
    \vspace*{12pt}%
\thispagestyle{empty}}
\makeatother

\begin{document}

\maketitle
\input{main_paper}
{
    \small
    \bibliographystyle{ieeenat_fullname}
    \bibliography{main}
}
\input{appendix}

\end{document}

%% file: preamble.tex
\usepackage{multirow}
\usepackage{tabularx}
\usepackage{xparse}

\usepackage{enumitem}
\usepackage{tikz}
\usepackage{adjustbox}
\usepackage{makecell}
\usepackage{pdflscape}

\newcolumntype{C}{>{\centering\arraybackslash}X}

\input{math_commands.tex}

\newcommand{\dataset}{RefRef\xspace}
\newcommand{\datasetlong}{RefRef dataset\xspace}

\newcommand{\method}{R3F\xspace}
\newcommand{\methodlong}{Refractive--Reflective Radiance Field\xspace}

\newcommand{\refl}{\text{\reflectbox{R}}}
\newcommand{\refr}{\text{\rotatebox[origin=c]{20}{R}}}

\makeatletter
\renewcommand\paragraph{%
  \@startsection{paragraph}%
                {4}%
                {\z@}%
                {-1.0ex \@plus -.2ex \@minus -0.2ex}%
                {-1.5ex \@plus -0.2ex}%
                {\normalfont\normalsize\bfseries}}%
\makeatother

%% file: math_commands.tex
\usepackage{amsmath,amsfonts,bm}

\def\1{\bm{1}}

\def\rvc{{\mathbf{c}}}
\def\rvd{{\mathbf{d}}}

\def\rvn{{\mathbf{n}}}
\def\rvo{{\mathbf{o}}}
\def\rvp{{\mathbf{p}}}

\def\rvr{{\mathbf{r}}}
\def\rvs{{\mathbf{s}}}

\def\rvw{{\mathbf{w}}}
\def\rvx{{\mathbf{x}}}

\DeclareMathAlphabet{\mathsfit}{\encodingdefault}{\sfdefault}{m}{sl}
\SetMathAlphabet{\mathsfit}{bold}{\encodingdefault}{\sfdefault}{bx}{n}

\def\gI{{\mathcal{I}}}

\def\gL{{\mathcal{L}}}

\def\gS{{\mathcal{S}}}

\newcommand{\transpose}{^\mathsf{T}}

\newcommand{\ind}[1]{\ensuremath{\makebox[.3ex][l]{[}\makebox{[}#1\makebox[.3ex][l]{]}\makebox{]}}}

\usepackage{amssymb}
\usepackage{pifont}
\newcommand{\cmark}{\ding{51}}%
\newcommand{\xmark}{\ding{55}}%

%% file: abstract.tex
\begin{abstract}
Modern 3D reconstruction and novel view synthesis approaches have demonstrated strong performance on scenes with opaque, non-refractive objects.
However, most assume straight light paths and therefore cannot properly handle refractive and reflective materials.
The lack of datasets specialized for these effects has impeded efforts to fairly and thoroughly evaluate performance and thereby make progress in this domain.
Most existing datasets focus on opaque scenes, while those targeting refractive objects are often limited to thin glass with negligible ray bending, untinted single-material objects, or backgrounds treated as infinitely distant, leaving key failure modes under strong refraction untested.
To address this gap, we introduce the \datasetlong.
It contains 150 synthetic and 60 real scenes spanning varying levels of geometric complexity and diverse background types that expose the limitations of existing methods.
For the purpose of benchmarking, we also provide an oracle method that, given the object geometry and refractive indices, calculates accurate light paths for neural rendering, and a simple two-stage baseline that relaxes these assumptions.
We evaluate these against state-of-the-art methods and show that the task is far from solved.
The dataset and code are available at \href{https://huggingface.co/datasets/yinyue27/RefRef}{\texttt{huggingface.co/datasets/yinyue27/RefRef}} and \href{https://github.com/YueYin27/refref}{\texttt{github.com/YueYin27/refref}}.
\end{abstract}

%% file: main_paper.tex
\section{Introduction}
\label{sec:intro}

Refractive and reflective objects, such as glass and water, pose significant challenges for 3D reconstruction and novel view synthesis due to the complex behavior of light as it passes through or reflects off these materials.
Accurate modeling of these optical phenomena is essential for precise 3D reconstruction and photorealistic rendering.
However, most existing neural radiance approaches~\cite{mildenhall2021nerf, barron2022mip, tensorf, mueller2022instant} and Gaussian splatting methods~\cite{kerbl3Dgaussians,2dgs,yu2024mipgs} struggle with refractive or reflective objects~\cite{raydeform}.
These methods typically assume that light travels in a straight path, which is valid for Lambertian objects, as illustrated in \cref{fig:teaser}, but fail in scenarios with complex optical behavior.

A key obstacle hindering the development and evaluation of suitable methods is the lack of datasets that expose the full difficulty of reconstructing scenes with refractive and reflective objects.
Standard 3D reconstruction datasets contain mostly opaque scenes, whereas datasets for refractive objects typically simplify many of the challenges: thin glass produces negligible ray bending; single-material objects omit interfaces between different refractive media; a limited range of object geometries prevents evaluation across diverse shapes; untinted materials omit wavelength-dependent attenuation; and infinitely distant backgrounds mean that the observed color only depends on the direction of the outgoing refracted ray, not its origin.
These simplifications make the problem considerably easier and meet the baked-in assumptions of specialized methods.
Even general-purpose methods perform well since limited view-dependent modeling is sufficient for such scenes.
Larger real-world datasets of transparent objects from robotics~\cite{chen2022clearpose,fang2022transcg,kim2024transpose} mostly contain thin-walled containers and target pose estimation, depth completion, or grasping rather than reconstruction under strong refraction.
As a result, these datasets do not reveal how the methods perform under strong refraction, complex geometry, multiple materials, tinting, or diverse backgrounds.

To address this dataset limitation, we introduce \dataset, a dataset and benchmark designed for the task of reconstructing scenes with complex refractive and reflective objects, as shown in \cref{fig:teaser}.
\dataset consists of 150 synthetic scenes with 50 objects and 60 real scenes with 15 objects.
Objects span three categories: convex single-material, nonconvex single-material, and nonconvex multi-material, with varied colors, opacities, and refractive indices.
These categories allow us to evaluate methods across diverse geometric and material configurations with different levels of complexity and varied combinations of refraction, reflection, total internal reflection, and absorption.
Each synthetic object is rendered against two bounded backgrounds and one unbounded background, while each real-world configuration is captured in two indoor and two outdoor environments with varied lighting 
conditions.
Bounded backgrounds contain finite-depth geometry near the refractive object, so the observed color depends on both the origin and direction of the outgoing refracted ray. 
Unbounded backgrounds use infinitely distant environment maps, so the color depends only on the ray direction, which simplifies the problem considerably.
Together, these scenes provide diverse optical challenges for evaluating 3D reconstruction methods.

For benchmarking, we also provide an oracle method that traces physically accurate light paths given the ground-truth object geometry and refractive indices, together with a simple two-stage baseline, \method, that relaxes these requirements by estimating object geometry and properties.
We evaluate state-of-the-art methods against these references, and show that they are largely unable to handle scenes with strongly refractive objects.
Our contributions are:
\begin{enumerate}[left=0pt, nosep]
    \item a dataset of 150 synthetic and 60 real scenes with strongly refractive and reflective objects;
    \item an oracle and a baseline that establish reference performance with and without ground-truth object geometry and refractive indices; and
    \item a benchmark evaluating state-of-the-art methods on this dataset, revealing the limits of existing approaches at handling complex optical phenomena.
\end{enumerate}

\section{Related Work}
\label{sec:relwork}

\begin{table}[!t]
  \renewcommand{\arraystretch}{0.92}
  \caption{Comparison of datasets by counts of refractive scenes $|\gS_{\mathrm{refr}}|$ and images $|\gI_{\mathrm{refr}}|$,
  and the presence of refractive scenes with
  predominantly thick or solid refractive objects (Strong),
  real data (Real),
  multi-material refractive objects (Multi),
  bounded backgrounds (Bounded),
  and tinted refractive materials (Tinted).
  $^*$Our estimates: the paper does not specify and the publicly released synthetic subset contains no strongly refractive scenes.}
  \label{tab:datasets}
  \vspace{-6pt}
  \footnotesize
  \setlength{\tabcolsep}{0.15em}{
  \begin{tabular*}{\linewidth}{@{\extracolsep{\fill}}l *{7}{c}@{}}
    \toprule
    Datasets & $|\gS_{\mathrm{refr}}|$ & $|\gI_{\mathrm{refr}}|$ & Strong & Real & Multi & Bounded & Tinted\\
    \midrule
    DTU~\cite{DTU} & 0 & 0 & \xmark & \xmark & \xmark & \xmark & \xmark \\
    T \& T~\cite{tanksNtemples} & 0 & 0 & \xmark & \xmark & \xmark & \xmark & \xmark \\
    ShapeNet~\cite{chang2015shapenet} & 0 & 0 & \xmark & \xmark & \xmark & \xmark & \xmark \\
    OmniObject3D~\cite{wu2023omniobject3d} & 0 & 0 & \xmark & \xmark & \xmark & \xmark & \xmark \\
    Objaverse-XL~\cite{objaverseXL} & 0 & 0 & \xmark & \xmark & \xmark & \xmark & \xmark \\
    Shiny~\cite{shinydataset} & 2 & 348 & \cmark & \cmark & \cmark & \cmark & \cmark \\
    OpenMaterial~\cite{dang2024openmaterial} & 143 & 12.9k & \cmark & \xmark & \xmark & \xmark & \xmark \\
    3DReflecNet~\cite{liang20263dreflecnet} & 11k$^*$ & 655k$^*$ & \cmark & \cmark & \cmark & \xmark & \xmark \\
    ClearPose~\cite{chen2022clearpose} & 51 & 354k & \xmark & \cmark & \xmark & \cmark & \xmark \\
    TransCG~\cite{fang2022transcg} & 130 & 57.7k & \xmark & \cmark & \xmark & \cmark & \xmark \\
    TRansPose~\cite{kim2024transpose} & 87 & 100k & \xmark & \cmark & \cmark & \cmark & \xmark \\
    Dex-NeRF~\cite{ichnowski2021dex} & 16 & 1.1k & \xmark & \cmark & \xmark & \cmark & \xmark \\
    \dataset (ours) & 210 & 57k & \cmark & \cmark & \cmark & \cmark & \cmark \\[-0.5ex]
    \bottomrule
  \end{tabular*}}
  \vspace{-6pt}
\end{table}

\paragraph{General 3D reconstruction datasets.}
Widely adopted 3D reconstruction datasets (\cref{tab:datasets}) include DTU~\cite{DTU}, which provides multiview images and ground-truth geometry for object reconstruction, and Tanks and Temples~\cite{tanksNtemples}, which provides complex real-world scenes captured with high-precision scanners.
Large-scale datasets such as Objaverse-XL~\cite{objaverseXL}, ShapeNet~\cite{chang2015shapenet}, and OmniObject3D~\cite{wu2023omniobject3d} provide extensive collections of 3D objects, but none of them contains strongly refractive objects.

\paragraph{Refractive and reflective object datasets.}
Several datasets (\cref{tab:datasets}) address the unique challenges posed by refractive and reflective objects.
The Shiny dataset~\cite{shinydataset} introduces complex view-dependent effects, such as rainbow reflections and refractions through glassware, to evaluate view synthesis under challenging conditions, but only includes two scenes with refractive objects.
In contrast, OpenMaterial~\cite{dang2024openmaterial} contains 143 synthetic scenes with strong refraction,
but only includes single-material, untinted objects, and infinitely distant backgrounds that greatly simplify the light transport estimation problem.
More recently, 3DReflecNet~\cite{liang20263dreflecnet} introduced a large-scale dataset with 2 transparent material types.
However, its released synthetic subset lacks strongly refractive scenes, and its real refractive objects are filmed in front of plain backdrops that are masked out before evaluation, leaving few refracted background patterns.
Moreover, robotics datasets of transparent objects~\cite{chen2022clearpose,fang2022transcg,kim2024transpose,ichnowski2021dex} offer posed multi-view captures for pose estimation, depth completion, and grasping, but mostly feature thin-walled containers with little ray bending.
Compared with these datasets, ours provides more challenging and diverse scene settings for benchmarking 3D reconstruction of refractive and reflective objects.
It contains 150 synthetic and 60 real scenes, featuring strongly refractive objects, some with multiple materials or tinting, and bounded backgrounds.
These properties appear in different combinations across scenes, supporting evaluation under varying levels of geometric, material, and optical complexity.

\paragraph{Neural 3D reconstruction.}
A neural radiance field (NeRF)~\cite{mildenhall2021nerf} optimizes the parameters of a coordinate network to map from spatial position and viewpoint to color and density.
To improve efficiency and accuracy, Zip-NeRF~\cite{barron2023zipnerf} combines a feature grid with anti-aliasing.
Other approaches fit implicit geometry representations through volume rendering, including UNISURF~\cite{unisurf}, VolSDF~\cite{volsdf}, and NeuS~\cite{wang2021neus}, which model 3D surfaces using occupancy or signed distance fields.
Another line of approaches rasterize 3D Gaussians~\cite{kerbl3Dgaussians,szymanowicz2024flash3d}, allowing for real-time rendering with high visual quality.
However, while some methods have explored single reflections, including NeRF-based approaches~\cite{verbin2022refnerf,verbin2024nerf,MaAT00SZR24,TiwaryDBKVR23,QiuJZYC023,Han0FSSOMJ24}, as well as recent Gaussian splatting techniques~\cite{jiang2024gaussianshader, yang2024spec, ye20243d, yao2024reflective}, none of the works in this section explicitly handle the curved light paths necessary for modeling refraction.

\paragraph{Transparent object modeling.}
Modeling refractive and reflective materials is challenging due to the complexity of the light paths~\cite{ZhangDCHW00024, LiYC20, msnerf, raydeform, wang2023nemto, GaoZWCZ23, QiuJZYC023, huang2025transparentgs}.
Prior work has explored this challenge across a range of settings, including dynamic water surfaces~\cite{XiongH21} and opaque objects observed through a single refractive medium~\cite{TongMMNL23, ZhanNNZ23, SunWYG24, CassidyMQLD20}.
For transparent object reconstruction, existing methods can be categorized by how they model or constrain light transport.
Controlled approaches exploit known illumination or structured patterns to recover ray--position correspondences~\cite{LiYC20, LyuWLC020, WuZQG018, LiLWCW0X23}, improving reconstruction accuracy but limiting applicability to general scenes.
Among methods without such controlled setups, several implicitly model curved light paths.
Eikonal Fields~\cite{bemana2022eikonal} learns a spatially varying refractive-index field and traces curved light paths.
MS-NeRF~\cite{msnerf} simplifies this by introducing multi-space feature fields to jointly model multiple subspaces.
NEMTO~\cite{wang2023nemto} predicts the exiting light direction directly, treating internal refraction and reflection as a black box, but assumes an infinitely distant background.
Ray Deformation Networks~\cite{raydeform} instead learns a deformation field that predicts light bending, but struggles with the high-frequency detail of complex shapes.
Another set of work explicitly traces reflected and refracted paths through estimated geometry.
Visual-hull-based reconstruction~\cite{yoon2024neural} and NeRRF~\cite{nerrf} infer geometry from silhouettes, but suffer from voxelization artifacts or simplified light transport.
In contrast, TNSR~\cite{Deng:tnsr} refines the surface geometry by backpropagating through the shape-dependent ray geometry;
a similar strategy is taken by Gao et al.~\cite{GaoZWCZ23}.
Methods for objects with internal structure further extend this formulation to multiple layers.
Opti-NeuS~\cite{yang2026opti} models a transparent shell surrounding an opaque interior, while NU-NeRF~\cite{SunWYG24} reconstructs nested transparent objects progressively, causing errors in the outer surface to propagate inward.
Alternative representations have also been explored.
3D Gaussian Ray Tracing~\cite{loccoz20243dgrt} traces volumetric Gaussian particles, whereas 3DGUT~\cite{wu20253dgut} retains splatting efficiency while supporting secondary rays for reflection and refraction.
TransparentGS~\cite{huang2025transparentgs} proposes Gaussian primitives with transparency attributes, but assumes only two refractions and no internal structure.
GLINT~\cite{na2026glint} separates interface, reflection, and transmission components, but assumes optically thin surfaces with negligible refractive bending.
Overall, most methods assume simplified settings and struggle with complex light interactions such as total internal reflection and multiple successive refractions, which frequently occur in even simple transparent objects.

\section{The Refractive--Reflective Object Dataset}
\label{sec:dataset}

In this section, we introduce \dataset, a dataset for reconstructing scenes with refractive and reflective objects from posed images, as shown in \cref{fig:teaser}.
It contains 150 synthetic scenes rendered against three background types and 60 real scenes captured in diverse indoor and outdoor environments under different lighting conditions.
As summarized in \cref{tab:datasets}, most existing 3D reconstruction datasets focus on opaque objects.
Those that target transparent objects either contain predominantly thin glass surfaces with negligible refraction or consider volumetric objects under simplified conditions, such as single-material configurations, low-transmittance materials, or distant backgrounds.
In contrast, \dataset includes objects with substantial refractive volumes, highly transparent and tinted materials, both single-material and multi-material configurations, and backgrounds ranging from nearby geometry to distant environments.
These properties appear in different combinations across scenes, introducing varying levels of optical complexity, such as significant light bending, multiple refractions, total internal reflections, and light absorption.

\begin{figure*}[!t]
    \centering
    \includegraphics[width=\linewidth]{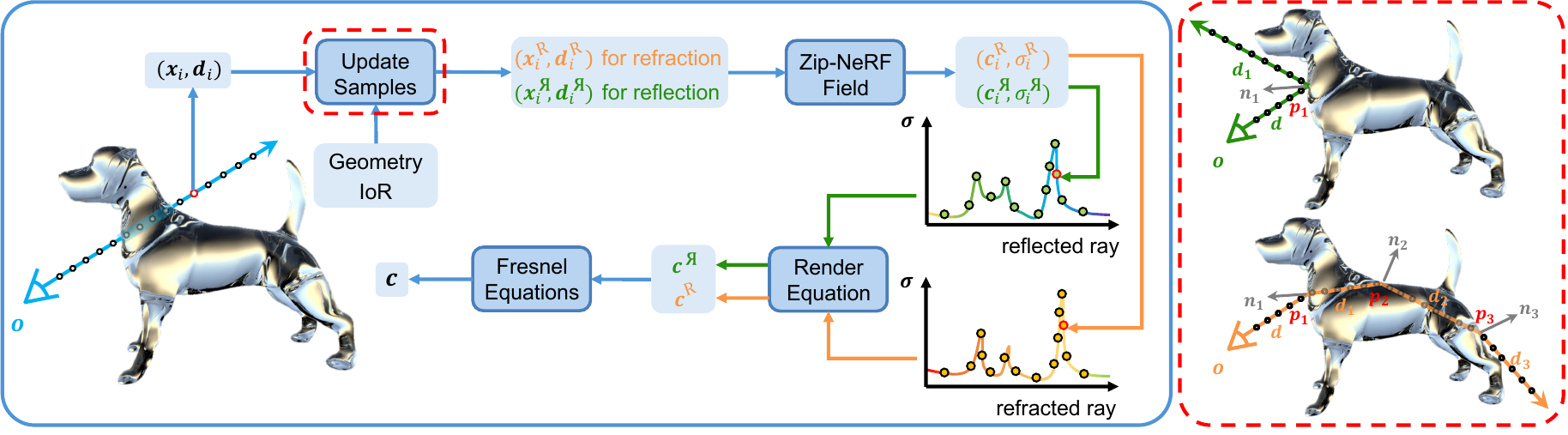}%
    \vspace{-6pt}
    \caption{
    Overview of the benchmark's oracle method.
    Starting with a straight ray (blue arrow), sample points are generated along its path.
    The scene's geometry and refractive index update the ray trajectory, as shown in the red dashed box.
    Refracted and reflected rays are processed separately, producing updated sample positions and directions.
    The Zip-NeRF field then predicts color $\rvc_i$ and density $\sigma_i$ for each sample.
    Using \cref{eq:rendering}, the final color along each ray is rendered and combined via \cref{eq:fresnel}.
    }
    \label{fig:oracle}
    \vspace{-4pt}
\end{figure*}

\subsection{Dataset Overview and Categories}

\paragraph{Geometry and material complexity.}
\dataset is organized into three categories according to geometric complexity and the number of refractive materials:
\begin{enumerate}[left=0pt, nosep]
\item \textbf{Convex single-material (27 synthetic, 8 real).}
This category includes objects with convex geometries, each composed of a single refractive material, such as a glass sphere and a glass pyramid.
These relatively simple geometries provide a controlled setting for evaluating refraction and reflection through a single medium.

\item \textbf{Nonconvex single-material (60 synthetic, 4 real).}
This category includes objects with nonconvex geometries, each composed of a single refractive material, such as glass sculptures, glass jars, and magnifiers.
Their nonconvex geometries introduce additional challenges, including self-occlusion, multiple surface intersections, and total internal reflection.

\item \textbf{Nonconvex multi-material (63 synthetic, 48 real).}
This category includes objects with nonconvex geometries and multiple refractive materials, such as reed diffusers, wine bottles, and snow globes.
Differences in refractive indices and colors create nested interfaces, multiple refractions, and light absorption.
\end{enumerate}

\paragraph{Background diversity.}
The synthetic subset includes 100 bounded scenes with textured cube or sphere backgrounds and 50 unbounded scenes represented by different HDR environment maps.
The bounded backgrounds contain finite-distance geometry and high-frequency textures, producing parallax and spatially varying refracted patterns.
In contrast, the unbounded backgrounds evaluate methods under the assumption that the background of the scene is infinitely distant from the transparent object.
The real-world subset comprises 60 scenes captured across two indoor and two outdoor environments under different lighting conditions.
The indoor environments include one illuminated by ceiling lights and another captured in a darker room with illumination from a lamp.
The outdoor environments include one captured under direct sunlight with static shadows and another captured after sunset.
Together, these environments provide diverse backgrounds and lighting conditions for evaluating reconstruction methods in both controlled and real-world settings.

\paragraph{Material tinting.}
To capture variations in material appearance, \dataset includes both colorless and tinted transparent objects, with 40\% of scenes containing tinted materials.
Tinting occurs when the material absorbs part of the light passing through it, making the observed color dependent on both the material properties and the distance traveled within the object.
This creates a challenge for existing methods, many of which do not account for internal absorption.

\subsection{Data Collection}

\paragraph{The synthetic scenes}
are rendered in Blender using the Cycles path tracer~\cite{blender}, which models refraction, reflection, total internal reflection, and light absorption.
Each object is placed in one of three background environments:
a bounded cube textured with six high-frequency images randomly selected from a collection of 24;
a bounded (finite-radius) sphere textured with one high-frequency image selected from a collection of 13;
or an unbounded (infinite-radius) spherical environment map randomly selected from a collection of 5, which provides both the background appearance and illumination.
Refractive materials are assigned appropriate transparency properties and refractive indices from a set of six, with tint added where applicable.
Each scene contains 300 posed images at a resolution of $800 \times 800$ pixels, resulting in 45,000 synthetic images across 150 scenes.
The images are divided into training (used for per-scene fitting), validation, and test sets, each containing 100 views.
Training and validation views are sampled around the object at a fixed radius, while test views follow a separate trajectory with a different radius, preventing evaluation from relying on interpolation between nearby training views.
For each scene, we provide RGB images, camera parameters, object masks, depth maps, object geometry, and the refractive indices used during rendering.
Further preprocessing details, rendering configurations, and object licensing information are in the appendix (\cref{sec:rendering_details}).

\paragraph{The real scenes}
subset contains 60 scenes produced by capturing 15 object configurations across four environments, with each configuration consisting of either a single object or multiple objects, as detailed in the appendix (\cref{sec:real_capture_details}).
For each scene, we record two RGB videos using a DJI OSMO Pocket 4 at a resolution of $4096 \times 2160$.
The camera moves around the object while the object and surrounding scene remain stationary.
The first video provides training and validation views, while the second follows a trajectory with a different radius and provides held-out test views.
We evenly sample 100 frames from each trajectory and jointly estimate camera intrinsics and poses for all 200 frames using Agisoft Metashape~\cite{agisoftmetashape}, with foreground refractive objects masked out using SAM~3~\cite{sam3segmentconcepts}.
Across all scenes, the estimated poses achieve a subpixel median epipolar error of 0.36 pixels on a set of 3.23M SIFT~\cite{lowe2004distinctive} feature matches from 25,934 image pairs, indicating high pose accuracy.
The images are downsampled and cropped to $800 \times 800$ pixels to match the resolution of the synthetic scenes, with camera intrinsics adjusted accordingly.
Of the 100 frames from the first trajectory, 90\% are used for training and 10\% for validation.

Outdoor scenes are captured under cloudless skies and windless conditions to ensure the scene is near-static throughout acquisition.
For scenes captured in direct sunlight, we ensure that shadows from the camera and operator do not enter the scene. 
For liquid-filled objects, we remove visible bubbles and begin recording only after the liquid has completely settled.
Despite these controls, real-world captures still contain surface imperfections, camera reflections, sensor noise, and inaccuracies in the estimated camera poses.
All details are in the appendix (\cref{sec:real_capture_details}).

\subsection{Supported Tasks and Applications}

\dataset primarily supports 3D reconstruction and novel view synthesis of scenes containing refractive and reflective objects.
Our benchmark evaluates novel view synthesis and depth estimation on test views from a held-out trajectory.
Additional annotations also support geometry evaluation, surface normal estimation, transparent object segmentation and detection, material estimation, and light transport modeling.
The synthetic and real scenes further enable research on physically plausible image and video generation, material editing, synthetic-to-real transfer, and robustness across capture conditions, with applications in robotics, augmented reality, and 3D content creation.

\section{A Volume Rendering Oracle Method}
\label{sec:method}

To evaluate how well reconstruction methods perform on \dataset, we introduce an oracle method as a reference point for the benchmark.
Given posed RGB images, NeRF-based methods learn a volumetric representation of scene geometry and appearance that can render novel views.
However, most methods struggle with scenes involving refractive and reflective objects due to their assumption of straight light paths~\cite{raydeform}.
Our oracle method instead uses the ground-truth geometry and refractive indices to trace refracted and reflected light paths, as shown in \cref{fig:oracle}.
It assumes piecewise linear light paths (\ie, piecewise constant refractive indices), a single explicit reflection occurring at the first surface intersection, and a maximum of 10 refractions or total internal reflections.
Additional reflections may be modeled implicitly through NeRF's view-dependent color.

\paragraph{NeRF preliminaries.}
The Neural Radiance Field (NeRF) method~\cite{mildenhall2021nerf} 
represents a scene with a coordinate network $\phi_\theta$ that maps each 3D sample position $\rvx_i$ and view direction $\rvd$ to density $\sigma_i$ and view-dependent color $\rvc_i$.
For a camera ray $\rvr(t) = \rvo + t\rvd$, the pixel color is approximated by integrating samples between the near and far planes,
\begin{equation}
\label{eq:rendering}
\rvc(\rvr) = \sum_{i=1}^{N} \underbrace{T_i (1 - \exp(-\sigma_i \Delta t_i))}_{w_i} \rvc_i,
\end{equation}
where $T_i=\exp \left( -\sum_{j=1}^{i-1} \sigma_j \Delta t_j \right)$ denotes the accumulated transmittance up to sample $i$, and $\Delta t_i$ denotes the distance between consecutive samples.

\subsection{Modeling Refractions and Reflections}
\label{sec:modeling_refr_refl}
The oracle method extends Zip-NeRF~\cite{barron2023zipnerf} to handle refraction and explicit reflection, as shown in \cref{fig:oracle}.
It represents light paths as piecewise linear functions parametrized by $K+1$ points $\{\rvp_i\}_{i=0}^K$ and unit direction vectors $\{\rvd_i\}_{i=0}^K$,
\begin{equation}
\label{eq:piecewise_linear}
\rvr(t) = \sum_{i = 0}^K \ind{t \in [\tau_i, \tau_{i+1})} \left( \rvp_i + (t - \tau_i) \rvd_i \right),
\end{equation}
where $\ind{\cdot}$ is an Iverson bracket, $\rvp_0 = \rvo$, $\rvd_0 = \rvd$ and the cumulative distance is given by
\begin{equation}
\tau_i =
\begin{cases} 
0 & \text{for } i=0 \\
\infty & \text{for } i=K+1 \\
\sum_{j=1}^{i} \|\rvp_j - \rvp_{j-1}\| & \text{otherwise.}
\end{cases}
\end{equation}

The method considers two paths: a (multiple) refraction path $\rvr^\refr$ and a (single) reflection path $\rvr^\refl$.
We next show how to compute the parameters for each.

\paragraph{Refraction and reflection parameters.}
For the refraction ray $\rvr^\refr$, the next intersection $\rvp_{i+1}$ from position $\rvp_i$ along direction $\rvd_i$ is computed using the ground-truth geometry.
Let
$\alpha_i = \nu_i / \nu_{i+1}$,
$\beta_i = -\rvd_{i}\transpose\rvn(\rvp_{i+1})$,
$\gamma_i^2 = 1 - \alpha_i^2 (1 - \beta_i^2 )$,
$\nu_i$ be the refractive index of the $i$\textsuperscript{th} medium, and
$\rvn(\rvx)$ be the ground-truth unit normal at location $\rvx$.
If $\gamma_i^2 \geqslant 0$, the next refracted direction $\rvd_{i+1}$ is computed using Snell's Law~\cite{born2013principles},
\begin{equation}
\label{eq:snell}
    \rvd_{i+1} = \alpha_i\rvd_{i} + \left(\alpha_i \beta_i - \gamma_i \right)\rvn(\rvp_{i+1}).
\end{equation}
Otherwise, total internal reflection will occur and the direction is given by the Law of Reflection,
\begin{equation}
    \label{eq:reflection}
    \rvd_{i+1} = \rvd_{i} - 2(\rvd_{i}\transpose \rvn(\rvp_{i+1})) \rvn(\rvp_{i+1}). 
\end{equation}
For the reflection ray $\rvr^\refl$, we model only the first surface reflection, with $\rvd_1^\refl$ given by \cref{eq:reflection}.

\paragraph{Sampling and rendering.}
For neural rendering, we sample points along the refraction and reflection paths using the Mip-NeRF 360 proposal sampler~\cite{barron2022mip}, which concentrates samples in high-density regions.
At each sample $\rvx_i$, the neural field is queried with the corresponding local direction $\rvd_i$, rather than a constant camera direction:
$(\rvc_i^\refr, \sigma_i^\refr) = \phi(\rvx_i^\refr, \rvd_i^\refr)$
and
$(\rvc_i^\refl, \sigma_i^\refl) = \phi(\rvx_i^\refl, \rvd_i^\refl)$.
We then apply \cref{eq:rendering} to obtain the colors $\rvc^\refr$ and $\rvc^\refl$,
which are combined using the Fresnel equations~\cite{hecht2012optics},
\begin{subequations}
    \label{eq:fresnel}
    \begin{align}
        \rvc' &= R (\rvc^\refl - \rvc^\refr) + \rvc^\refr, \quad \text{where} \quad R = \frac{1}{2}(R_p + R_s), \\
        R_p &= \left( \frac{\nu_1 \beta_0 - \nu_0 \gamma_0}{\nu_1 \beta_0 + \nu_0 \gamma_0} \right)^2, \quad R_s = \left( \frac{\nu_0 \beta_0 - \nu_1 \gamma_0}{\nu_0 \beta_0 + \nu_1 \gamma_0} \right)^2,
    \end{align}
\end{subequations}
where $R_p$ and $R_s$ are the reflection coefficients for parallel and perpendicular polarized light; $\rvc'$ is converted to a non-linear sRGB space to obtain the predicted color $\hat{\rvc}$~\cite{verbin2022refnerf}.

\subsection{Optimization}
\label{sec:loss}
The parameters $\theta$ of the coordinate network $\phi_\theta$ are optimized with respect to a photometric loss $\gL_\text{rgb}$, an anti-aliased interlevel loss $\gL_\text{int}$, and a modified distortion loss $\gL_\text{dist}$.
The per-pixel photometric loss is given by
\begin{equation}
\label{eq:rgb_loss}
    \gL_\text{rgb}(\hat{\rvc}, \rvc) = \frac{1}{3} \|\hat{\rvc} - \rvc\|^2_2,
\end{equation}
where $\rvc$ is the ground-truth pixel color.
The interlevel loss $\gL_{\text{int}}$~\cite{barron2023zipnerf} encourages consistency between the proposal network
and the main network.
The original Mip-NeRF 360 distortion loss~\cite{barron2022mip} encourages a single compact weight peak, reducing floaters for opaque surfaces.
However, for transparent objects, it suppresses object density and proposal samples within the object.
We therefore exclude samples inside the refractive object for this loss,
\begin{equation}
\label{eq:distortion_loss}
\begin{aligned}
    \gL_{\text{dist}}({\rvs,\rvw}) = &\sum_{i,j \in \gI_\text{out}} w_i w_j \left| \frac{s_i + s_{i+1}}{2} - \frac{s_j + s_{j+1}}{2} \right| \\
    &+ \frac{1}{3} \sum_{i \in \gI_\text{out}} w_i^2 (s_{i+1} - s_i),
\end{aligned}
\end{equation}
where $s_i$ is the normalized ray distance~\cite{barron2022mip} to sample point $i$, and $\gI_\text{out}$ contains indices outside the refractive object.
This modification retains more high-weight samples within the object, improving reconstruction.
The overall loss is then %
\begin{equation}
\label{eq:loss}
    \gL=\gL_\text{rgb} + \lambda_1 \gL_\text{int} + \lambda_2 \gL_\text{dist},
\end{equation}
where $\lambda_1$ and $\lambda_2$ weight the loss contributions, each averaged across pixels in the dataset.

\begin{table*}[!t]
    \centering
    \renewcommand{\arraystretch}{0.95}
    \caption{
    Quantitative results on the \dataset test set for novel view synthesis (PSNR, masked PSNR, SSIM, and LPIPS) and geometry estimation (DMAE).
    A dash denotes a result that cannot be reported due to method or ground-truth limitations.
    On the synthetic data, there is a large performance gap between the Oracle reference point and the next best approach (our simple baseline R3F), while all methods perform poorly on the real scenes with very blurry results from even the top-performing method Zip-NeRF.
    This demonstrates that refractive object reconstruction is very far from being solved.
    $^\dagger$Has access to ground-truth foreground geometry and material properties.
    }
    \setlength{\tabcolsep}{1pt}
    \label{tab:result_quantitative}

    \resizebox{\linewidth}{!}{%
        \begin{tabular}{@{}llccccc@{\hspace{0.5em}}ccccc@{\hspace{0.5em}}ccccc@{\hspace{0.5em}}ccccc@{}}
        \toprule
         & Method
         & \multicolumn{5}{c}{Convex single-material}
         & \multicolumn{5}{c}{Nonconvex single-material}
         & \multicolumn{5}{c}{Nonconvex multi-material}
         & \multicolumn{5}{c}{Entire dataset} \\
         \cmidrule(lr){3-7}
         \cmidrule(lr){8-12}
         \cmidrule(lr){13-17}
         \cmidrule(lr){18-22}

         & &
         \scriptsize{PSNR}\newline$\uparrow$ &
         \scriptsize{PSNR{\raisebox{-0.6ex}{\fontsize{3.5pt}{4pt}\selectfont \hspace{-0.5pt}M}}}\newline$\uparrow$ &
         \scriptsize{SSIM}\newline$\uparrow$ &
         \scriptsize{LPIPS}\newline$\downarrow$ &
         \scriptsize{DMAE}\newline$\downarrow$ &

         \scriptsize{PSNR}\newline$\uparrow$ &
         \scriptsize{PSNR{\raisebox{-0.6ex}{\fontsize{3.5pt}{4pt}\selectfont \hspace{-0.5pt}M}}}\newline$\uparrow$ &
         \scriptsize{SSIM}\newline$\uparrow$ &
         \scriptsize{LPIPS}\newline$\downarrow$ &
         \scriptsize{DMAE}\newline$\downarrow$ &

         \scriptsize{PSNR}\newline$\uparrow$ &
         \scriptsize{PSNR{\raisebox{-0.6ex}{\fontsize{3.5pt}{4pt}\selectfont \hspace{-0.5pt}M}}}\newline$\uparrow$ &
         \scriptsize{SSIM}\newline$\uparrow$ &
         \scriptsize{LPIPS}\newline$\downarrow$ &
         \scriptsize{DMAE}\newline$\downarrow$ &

         \scriptsize{PSNR}\newline$\uparrow$ &
         \scriptsize{PSNR{\raisebox{-0.6ex}{\fontsize{3.5pt}{4pt}\selectfont \hspace{-0.5pt}M}}}\newline$\uparrow$ &
         \scriptsize{SSIM}\newline$\uparrow$ &
         \scriptsize{LPIPS}\newline$\downarrow$ &
         \scriptsize{DMAE}\newline$\downarrow$ \\

        \midrule
        \multirow{12}{*}{\rotatebox{90}{\makecell{Synthetic Scenes}}}
         & NeuS
         & 19.62 & 14.91 & 0.63 & \underline{0.17} & 1.50
         & 19.91 & 14.04 & 0.65 & 0.23 & 1.57
         & 19.34 & 15.07 & 0.62 & 0.20 & 1.65
         & 19.65 & 14.65 & 0.63 & 0.21 & 1.58 \\

         & Splatfacto
         & 18.63 & 14.32 & 0.79 & 0.33 & 16.38
         & 19.06 & 14.32 & 0.74 & 0.33 & 12.23
         & 19.95 & 15.41 & 0.77 & 0.33 & 10.70
         & 19.39 & 14.81 & 0.76 & 0.33 & 12.31 \\

         & Zip-NeRF
         & 19.84 & 16.42 & 0.76 & 0.35 & 1.35
         & 21.08 & 16.03 & 0.78 & 0.24 & 1.05
         & \underline{22.24} & 16.99 & \underline{0.82} & 0.21 & 1.72
         & 21.50 & 16.68 & 0.80 & 0.24 & 1.52 \\

         & TNSR
         & 18.96 & 11.79 & 0.78 & 0.22 & 1.26
         & 19.15 & 11.55 & \underline{0.83} & 0.24 & 1.47
         & 17.68 & 10.34 & 0.76 & \underline{0.18} & 1.70
         & 18.65 & 11.15 & 0.78 & 0.19 & 1.49 \\

         & MS-NeRF
         & 18.86 & 15.48 & 0.73 & 0.35 & 2.18
         & 20.06 & 15.38 & 0.74 & 0.29 & 2.26
         & 21.41 & 16.41 & 0.78 & 0.26 & 1.86
         & 20.34 & 15.78 & 0.76 & 0.30 & 2.08 \\

         & RayDef
         & 20.42 & 16.59 & 0.73 & 0.40 & 0.80
         & 20.71 & 16.06 & 0.70 & 0.36 & 0.88
         & 21.92 & \underline{17.07} & 0.74 & 0.32 & 0.94
         & 21.16 & 16.58 & 0.73 & 0.34 & 0.88 \\

         & RoseNeRF
         & 18.24 & 11.63 & 0.67 & 0.43 & --
         & 18.89 & 11.80 & 0.65 & 0.42 & --
         & 18.48 & 14.34 & 0.65 & 0.43 & --
         & 18.54 & 12.59 & 0.66 & 0.43 & -- \\

         & NU-NeRF
         & 20.30 & \underline{21.28} & 0.73 & 0.32 & 0.73
         & 20.46 & \textbf{20.89} & 0.71 & 0.28 & 0.51
         & 19.96 & 17.00 & 0.72 & 0.33 & 0.75
         & 20.24 & \underline{19.72} & 0.72 & 0.31 & 0.66 \\

         & NeRRF
         & 13.49 & 12.96 & 0.45 & 0.74 & 1.73
         & 14.28 & 13.34 & 0.52 & 0.65 & 1.94
         & 13.10 & 12.06 & 0.53 & 0.74 & 2.01
         & 13.65 & 12.74 & 0.51 & 0.70 & 1.93 \\

         & TransparentGS & 14.82 & 11.58 & 0.69 & 0.36 & 0.21 & 16.44 & 12.25 & 0.68 & 0.31 & 0.20 & 15.69 & 10.62 & 0.72 & 0.31 & 0.24 & 15.83 & 11.44 & 0.70 & 0.32 & 0.22 \\

         & \method (ours)
         & \underline{23.36} & 19.53 & \underline{0.83} & 0.18 & \underline{0.09}
         & \underline{21.76} & 16.32 & 0.79 & \underline{0.20} & \underline{0.18}
         & 20.90 & 16.88 & 0.81 & \underline{0.18} & \underline{0.12}
         & \underline{22.01} & 17.57 & \underline{0.82} & \underline{0.18} & \underline{0.13} \\

         & Oracle$^\dagger$
         & \textbf{32.66} & \textbf{26.12} & \textbf{0.96} & \textbf{0.05} & \textbf{0.08}
         & \textbf{28.65} & \underline{20.49} & \textbf{0.93} & \textbf{0.08} & \textbf{0.07}
         & \textbf{28.68} & \textbf{21.23} & \textbf{0.92} & \textbf{0.10} & \textbf{0.06}
         & \textbf{29.34} & \textbf{21.76} & \textbf{0.93} & \textbf{0.08} & \textbf{0.07} \\

        \midrule
        \multirow{9}{*}{\rotatebox{90}{\makecell{Real Scenes}}}
         & NeuS
         & 19.16 & 15.90 & 0.62 & \underline{0.34} & --
         & 19.75 & 14.79 & 0.62 & \underline{0.32} & --
         & 18.73 & 15.87 & 0.61 & 0.36 & --
         & 18.86 & 15.80 & 0.61 & 0.35 & -- \\

         & Splatfacto
         & 16.23 & 15.45 & 0.68 & 0.42 & --
         & 16.30 & 14.39 & 0.67 & 0.40 & --
         & 16.19 & 15.32 & 0.67 & 0.43 & --
         & 16.20 & 15.28 & 0.67 & 0.43 & -- \\

         & Zip-NeRF
         & \textbf{20.57} & \textbf{16.94} & \textbf{0.77} & \textbf{0.31} & --
         & \textbf{21.32} & \textbf{16.26} & \textbf{0.76} & \textbf{0.27} & --
         & \textbf{20.30} & \textbf{17.09} & \textbf{0.75} & \textbf{0.32} & --
         & \textbf{20.40} & \textbf{17.01} & \textbf{0.75} & \textbf{0.32} & -- \\

         & TNSR & 17.08 & 13.33 & 0.63 & 0.42 & -- & 17.23 & 13.06 & 0.61 & 0.41 & -- & 15.87 & 13.16 & 0.58 & 0.48 & -- & 16.12 & 13.18 & 0.59 & 0.47 & -- \\

         & MS-NeRF
         & 19.41 & 16.13 & \underline{0.72} & 0.41 & --
         & \underline{20.10} & 15.20 & \underline{0.71} & 0.38 & --
         & 19.28 & 15.97 & \underline{0.71} & 0.40 & --
         & 19.35 & 15.94 & \underline{0.71} & 0.40 & -- \\

         & RayDef
         & 19.24 & 16.34 & 0.71 & 0.44 & --
         & 19.83 & \underline{15.56} & 0.69 & 0.42 & --
         & 18.80 & \underline{16.27} & 0.69 & 0.46 & --
         & 18.93 & 16.23 & 0.69 & 0.45 & -- \\

         & NU-NeRF
         & 18.96 & \underline{16.53} & 0.65 & 0.38 & -- & 19.35 & 15.48 & 0.64 & 0.39 & -- & 18.45 & \underline{16.27} & 0.65 & 0.38 & -- & 18.59 & \underline{16.25} & 0.65 & 0.38 & -- \\

         & TransparentGS & 17.00 & 13.40 & 0.61 & 0.36 & -- & 16.88 & 12.67 & 0.58 & 0.33 & -- & 15.62 & 12.15 & 0.57 & 0.40 & -- & 15.89 & 12.36 & 0.58 & 0.39 & -- \\

         & \method (ours)
         & \underline{19.72} & 16.45 & \underline{0.72} & \underline{0.34} & --
         & 18.75 & 14.12 & 0.65 & 0.38 & --
         & \underline{19.39} & 16.24 & \underline{0.71} & \underline{0.34} & --
         & \underline{19.39} & 16.13 & \underline{0.71} & \underline{0.34} & -- \\

        \bottomrule
        \end{tabular}%
    }
\end{table*}

\subsection{\method: A Simple Two-Stage Baseline}
\label{sec:method_relaxation}

Alongside the oracle, we introduce a simple two-stage baseline that relaxes the requirements for ground-truth object geometry and refractive index, providing a second point of reference for the benchmark.
\method (\methodlong) relaxes the oracle's ground-truth requirements with a two-stage reconstruction pipeline.
First, a background radiance field is optimized using Zip-NeRF~\cite{barron2023zipnerf} with foreground rays masked out using object masks predicted by SAM~3~\cite{sam3segmentconcepts}.
Then, the geometry and refractive index of the object are estimated using NU-NeRF~\cite{SunWYG24}, and a separate in-object field is optimized using the oracle's ray tracing, with the background field frozen, preventing artifacts like floaters outside the object, which are otherwise common when modeling refractive and reflective objects~\cite{barron2023zipnerf,Deng:tnsr,raydeform,msnerf}.
While effective,
\method's performance entirely depends on the quality of the geometry prediction.

\section{Benchmark and Experiments}
\label{sec:experiments}

\begin{figure*}[!t]
\centering

\begin{subfigure}{\textwidth}
\centering

\makebox[0.030\linewidth]{}
\begin{minipage}[c]{0.119\linewidth}
    \centering
    \fontsize{8pt}{9.6pt}\selectfont
    Ground~Truth
\end{minipage}\hfill
\begin{minipage}[c]{0.119\linewidth}
    \centering
    \fontsize{8pt}{9.6pt}\selectfont
    Zip-NeRF
\end{minipage}\hfill
\begin{minipage}[c]{0.119\linewidth}
    \centering
    \fontsize{8pt}{9.6pt}\selectfont
    MS-NeRF
\end{minipage}\hfill
\begin{minipage}[c]{0.119\linewidth}
    \centering
    \fontsize{8pt}{9.6pt}\selectfont
    NU-NeRF
\end{minipage}\hfill
\begin{minipage}[c]{0.119\linewidth}
    \centering
    \fontsize{8pt}{9.6pt}\selectfont
    NeRRF
\end{minipage}\hfill
\begin{minipage}[c]{0.119\linewidth}
    \centering
    \fontsize{8pt}{9.6pt}\selectfont
    TransparentGS
\end{minipage}\hfill
\begin{minipage}[c]{0.119\linewidth}
    \centering
    \fontsize{8pt}{9.6pt}\selectfont
    \method~(ours)
\end{minipage}\hfill
\begin{minipage}[c]{0.119\linewidth}
    \centering
    \fontsize{8pt}{9.6pt}\selectfont
    Oracle
\end{minipage}\vfill

\begin{minipage}[c]{0.030\linewidth}
    \centering
    \rotatebox{90}{%
        \fontsize{8pt}{9.6pt}\selectfont Cube BG%
    }
\end{minipage}%
\begin{subfigure}[c]{0.119\linewidth}
    \centering
    \clipbox{0pt \height/30 0pt \height/30}{\includegraphics[width=\linewidth]{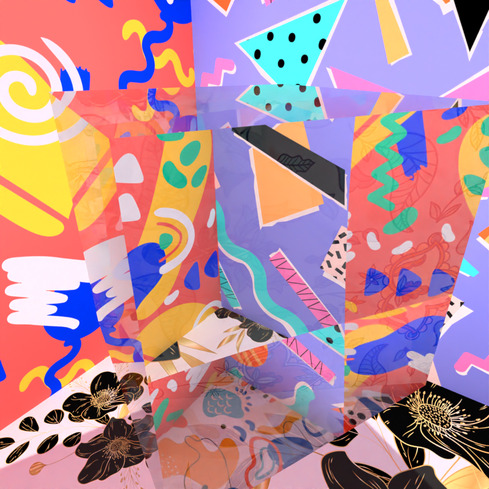}}
\end{subfigure}\hfill
\begin{subfigure}[c]{0.119\linewidth}
    \centering
    \clipbox{0pt \height/30 0pt \height/30}{\includegraphics[width=\linewidth]{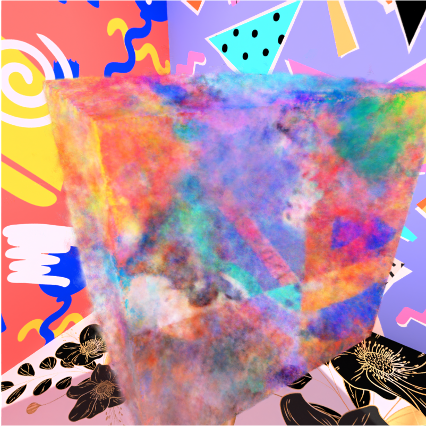}}
\end{subfigure}\hfill
\begin{subfigure}[c]{0.119\linewidth}
    \centering
    \clipbox{0pt \height/30 0pt \height/30}{\includegraphics[width=\linewidth]{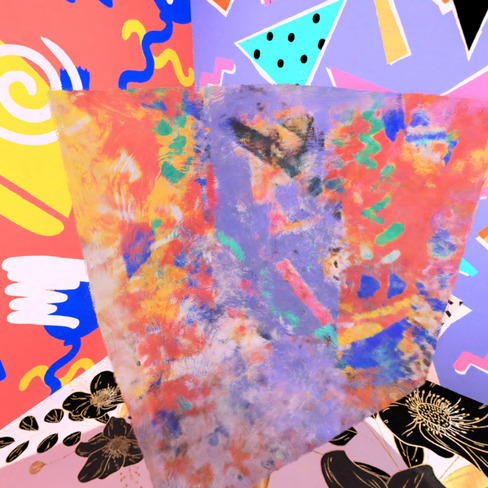}}
\end{subfigure}\hfill
\begin{subfigure}[c]{0.119\linewidth}
    \centering
    \clipbox{0pt \height/30 0pt \height/30}{\includegraphics[width=\linewidth]{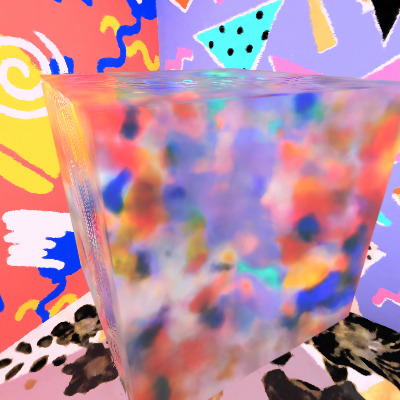}}
\end{subfigure}\hfill
\begin{subfigure}[c]{0.119\linewidth}
    \centering
    \clipbox{0pt \height/30 0pt \height/30}{\includegraphics[width=\linewidth]{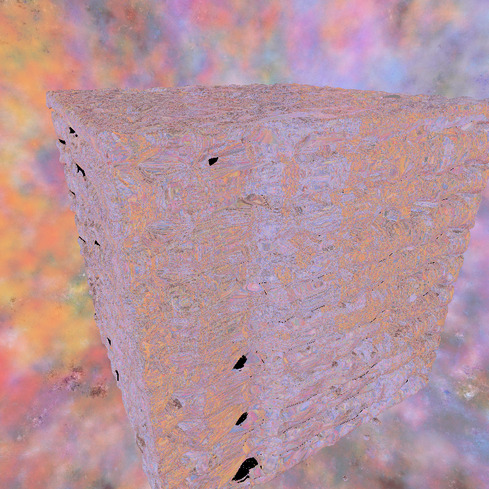}}
\end{subfigure}\hfill
\begin{subfigure}[c]{0.119\linewidth}
    \centering
    \clipbox{0pt \height/30 0pt \height/30}{\includegraphics[width=\linewidth]{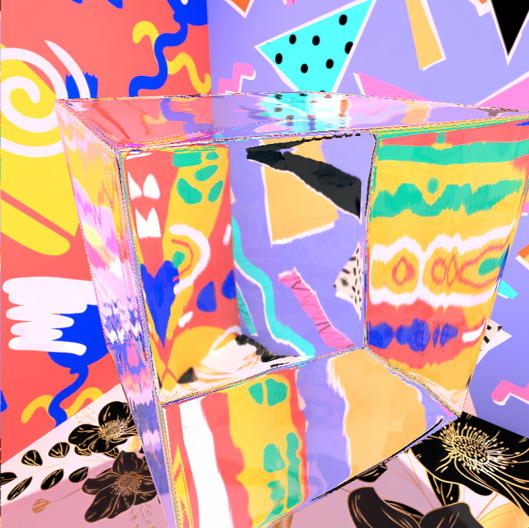}}
\end{subfigure}\hfill
\begin{subfigure}[c]{0.119\linewidth}
    \centering
    \clipbox{0pt \height/30 0pt \height/30}{\includegraphics[width=\linewidth]{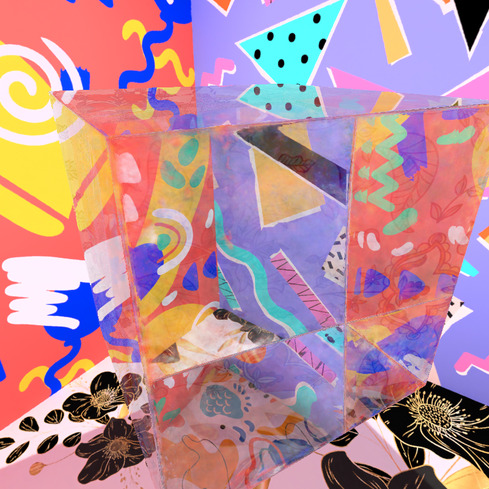}}
\end{subfigure}\hfill
\begin{subfigure}[c]{0.119\linewidth}
    \centering
    \clipbox{0pt \height/30 0pt \height/30}{\includegraphics[width=\linewidth]{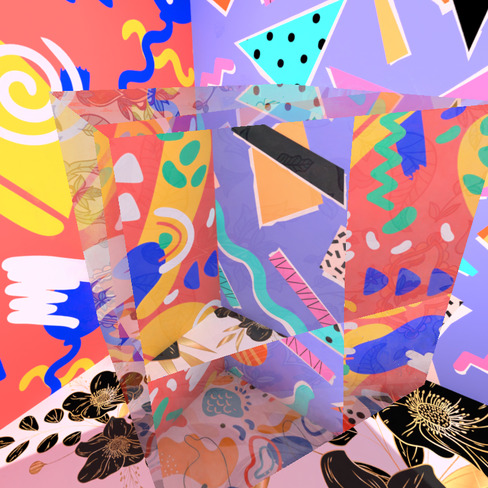}}
\end{subfigure}\vfill
\begin{minipage}[c]{0.030\linewidth}
    \centering
    \rotatebox{90}{%
        \fontsize{8pt}{9.6pt}\selectfont Sphere BG%
    }
\end{minipage}%
\begin{subfigure}[c]{0.119\linewidth}
    \centering
    \clipbox{0pt \height/30 0pt \height/30}{\includegraphics[width=\linewidth]{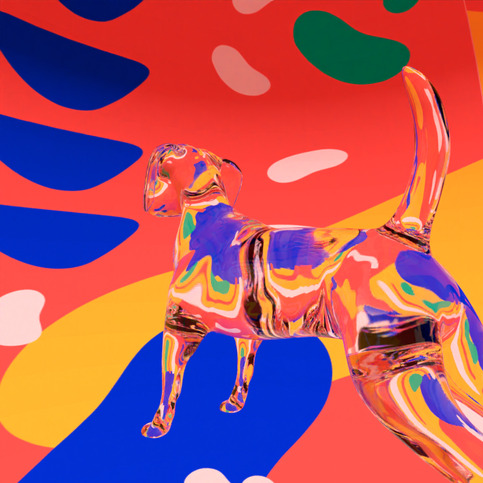}}
\end{subfigure}\hfill
\begin{subfigure}[c]{0.119\linewidth}
    \centering
    \clipbox{0pt \height/30 0pt \height/30}{\includegraphics[width=\linewidth]{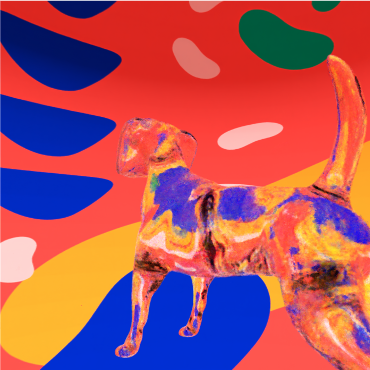}}
\end{subfigure}\hfill
\begin{subfigure}[c]{0.119\linewidth}
    \centering
    \clipbox{0pt \height/30 0pt \height/30}{\includegraphics[width=\linewidth]{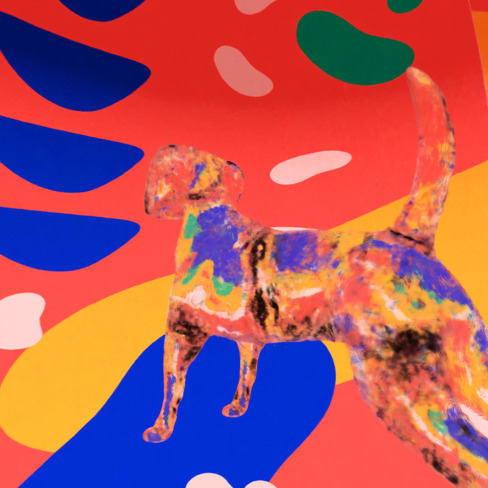}}
\end{subfigure}\hfill
\begin{subfigure}[c]{0.119\linewidth}
    \centering
    \clipbox{0pt \height/30 0pt \height/30}{\includegraphics[width=\linewidth]{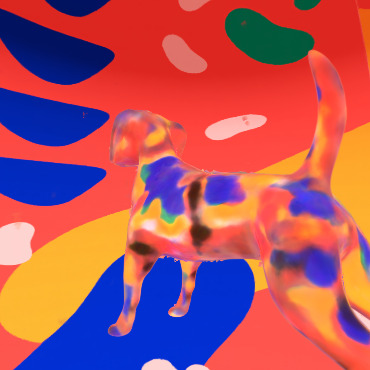}}
\end{subfigure}\hfill
\begin{subfigure}[c]{0.119\linewidth}
    \centering
    \clipbox{0pt \height/30 0pt \height/30}{\includegraphics[width=\linewidth]{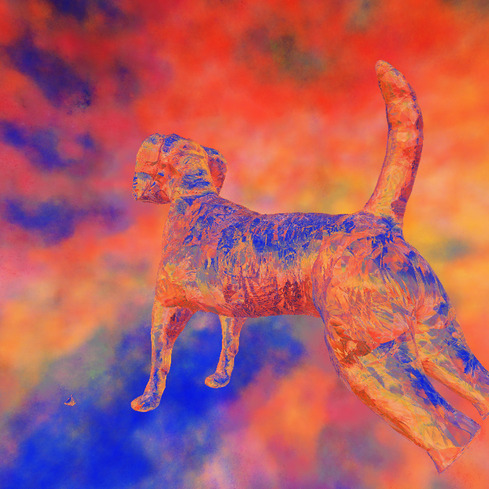}}
\end{subfigure}\hfill
\begin{subfigure}[c]{0.119\linewidth}
    \centering
    \clipbox{0pt \height/30 0pt \height/30}{\includegraphics[width=\linewidth]{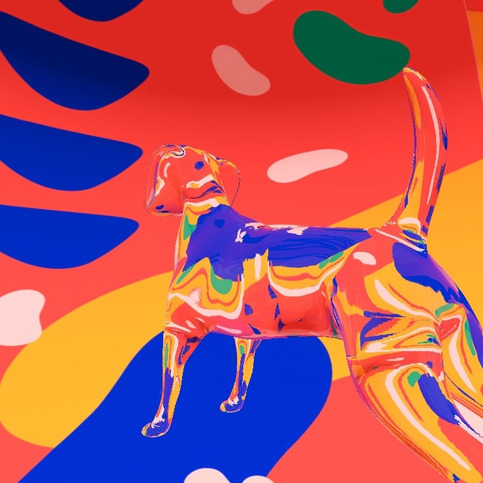}}
\end{subfigure}\hfill
\begin{subfigure}[c]{0.119\linewidth}
    \centering
    \clipbox{0pt \height/30 0pt \height/30}{\includegraphics[width=\linewidth]{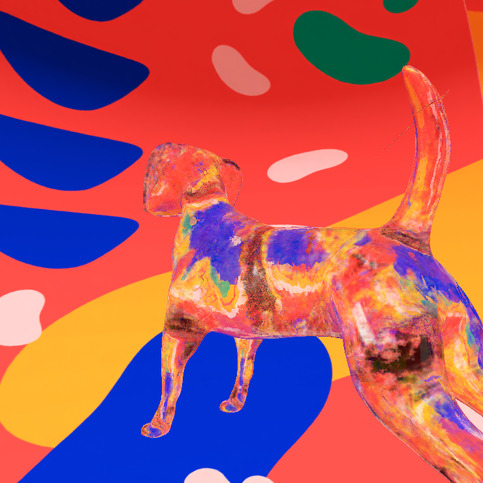}}
\end{subfigure}\hfill
\begin{subfigure}[c]{0.119\linewidth}
    \centering
    \clipbox{0pt \height/30 0pt \height/30}{\includegraphics[width=\linewidth]{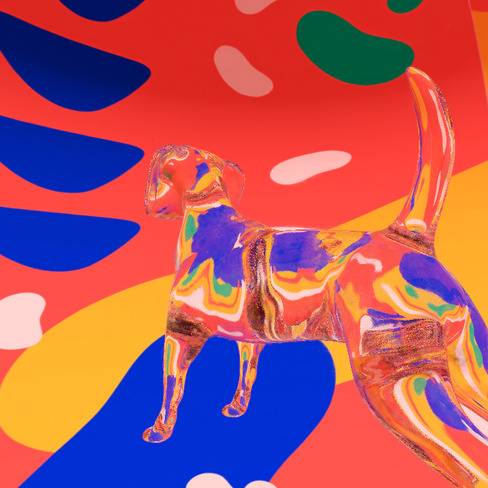}}
\end{subfigure}%

\begin{minipage}[c]{0.030\linewidth}
    \centering
    \rotatebox{90}{%
        \fontsize{8pt}{9.6pt}\selectfont Env Map BG%
    }
\end{minipage}%
\begin{subfigure}[c]{0.119\linewidth}
    \centering
    \clipbox{0pt \height/30 0pt \height/30}{\includegraphics[width=\linewidth]{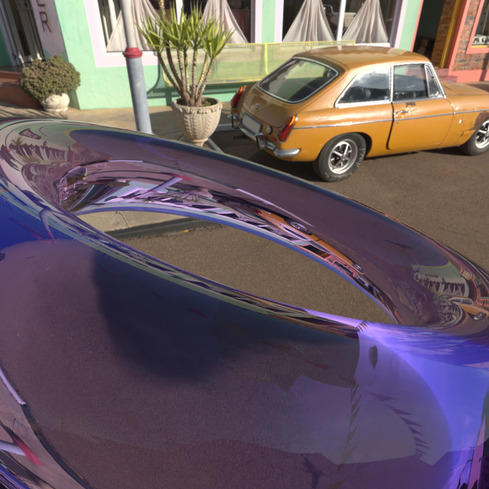}}
\end{subfigure}\hfill
\begin{subfigure}[c]{0.119\linewidth}
    \centering
    \clipbox{0pt \height/30 0pt \height/30}{\includegraphics[width=\linewidth]{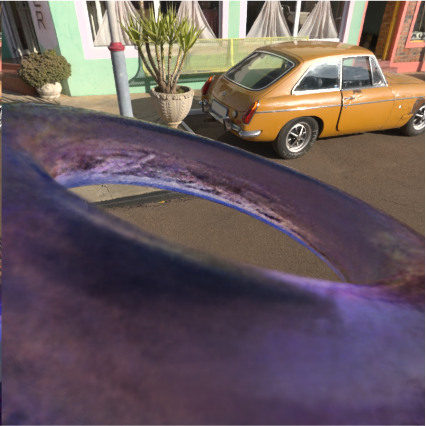}}
\end{subfigure}\hfill
\begin{subfigure}[c]{0.119\linewidth}
    \centering
    \clipbox{0pt \height/30 0pt \height/30}{\includegraphics[width=\linewidth]{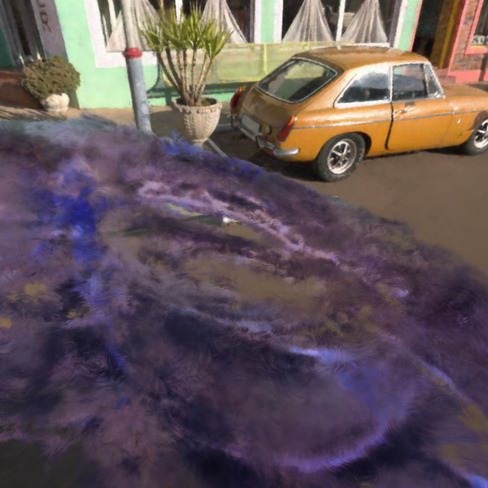}}
\end{subfigure}\hfill
\begin{subfigure}[c]{0.119\linewidth}
    \centering
    \clipbox{0pt \height/30 0pt \height/30}{\includegraphics[width=\linewidth]{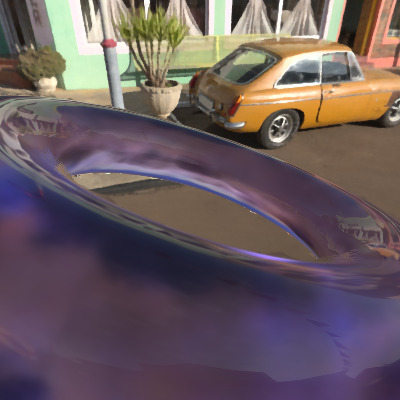}}
\end{subfigure}\hfill
\begin{subfigure}[c]{0.119\linewidth}
    \centering
    \clipbox{0pt \height/30 0pt \height/30}{\includegraphics[width=\linewidth]{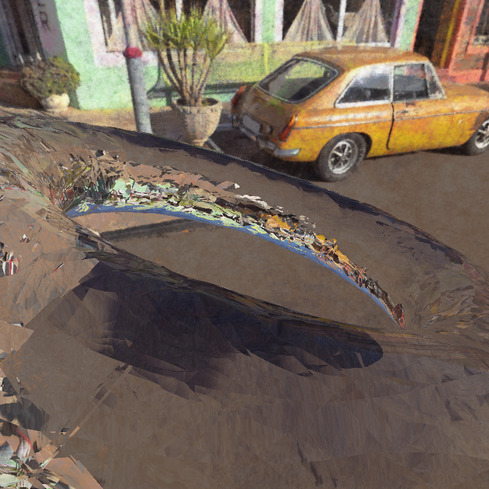}}
\end{subfigure}\hfill
\begin{subfigure}[c]{0.119\linewidth}
    \centering
    \clipbox{0pt \height/30 0pt \height/30}{\includegraphics[width=\linewidth]{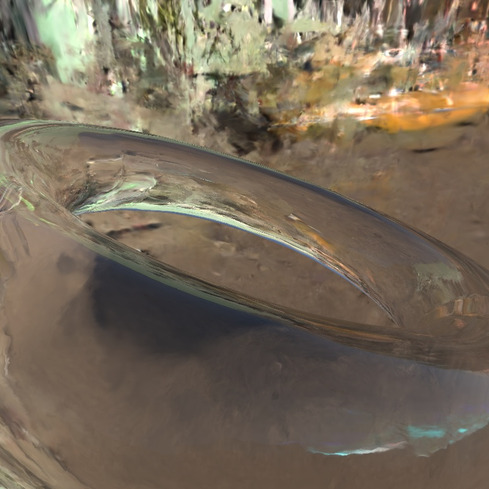}}
\end{subfigure}\hfill
\begin{subfigure}[c]{0.119\linewidth}
    \centering
    \clipbox{0pt \height/30 0pt \height/30}{\includegraphics[width=\linewidth]{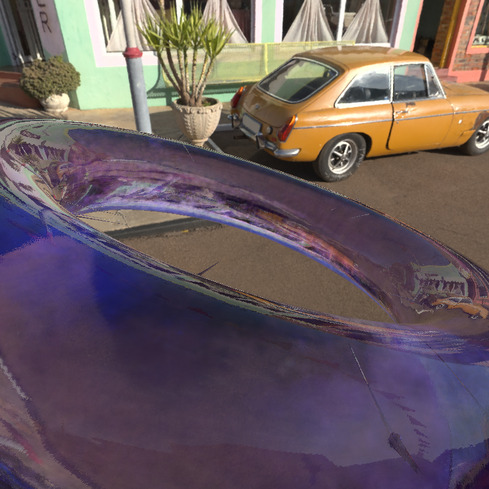}}
\end{subfigure}\hfill
\begin{subfigure}[c]{0.119\linewidth}
    \centering
    \clipbox{0pt \height/30 0pt \height/30}{\includegraphics[width=\linewidth]{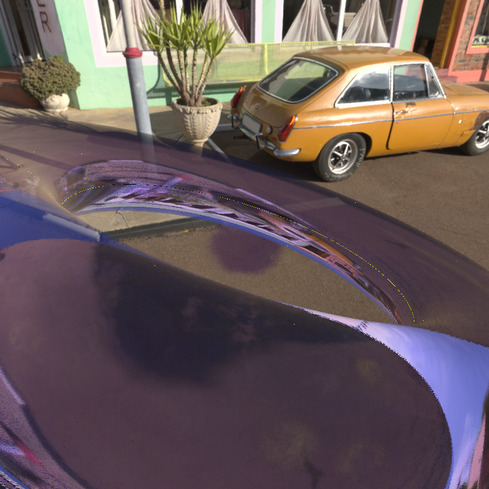}}
\end{subfigure}%
\caption{Synthetic Scenes}
\label{fig:results_synthetic}
\end{subfigure}\par\vspace{2pt}
\begin{subfigure}{\textwidth}
\centering
\makebox[0.030\linewidth]{}
\begin{minipage}[c]{0.119\linewidth}
    \centering
    \fontsize{8pt}{9.6pt}\selectfont
    Ground~Truth
\end{minipage}\hfill
\begin{minipage}[c]{0.119\linewidth}
    \centering
    \fontsize{8pt}{9.6pt}\selectfont
    Splatfacto
\end{minipage}\hfill
\begin{minipage}[c]{0.119\linewidth}
    \centering
    \fontsize{8pt}{9.6pt}\selectfont
    TransparentGS
\end{minipage}\hfill
\begin{minipage}[c]{0.119\linewidth}
    \centering
    \fontsize{8pt}{9.6pt}\selectfont
    Zip-NeRF
\end{minipage}\hfill
\begin{minipage}[c]{0.119\linewidth}
    \centering
    \fontsize{8pt}{9.6pt}\selectfont
    MS-NeRF
\end{minipage}\hfill
\begin{minipage}[c]{0.119\linewidth}
    \centering
    \fontsize{8pt}{9.6pt}\selectfont
    Ray Deformation
\end{minipage}\hfill
\begin{minipage}[c]{0.119\linewidth}
    \centering
    \fontsize{8pt}{9.6pt}\selectfont
    NU-NeRF
\end{minipage}\hfill
\begin{minipage}[c]{0.119\linewidth}
    \centering
    \fontsize{8pt}{9.6pt}\selectfont
    \method~(ours)
\end{minipage}\vfill

\begin{minipage}[c]{0.030\linewidth}
    \centering
    \rotatebox{90}{%
        \fontsize{8pt}{9.6pt}\selectfont Courtyard%
    }
\end{minipage}%
\begin{subfigure}[c]{0.119\linewidth}
    \centering
    \includegraphics[width=\linewidth,trim={0 20bp 0 20bp},clip]{
        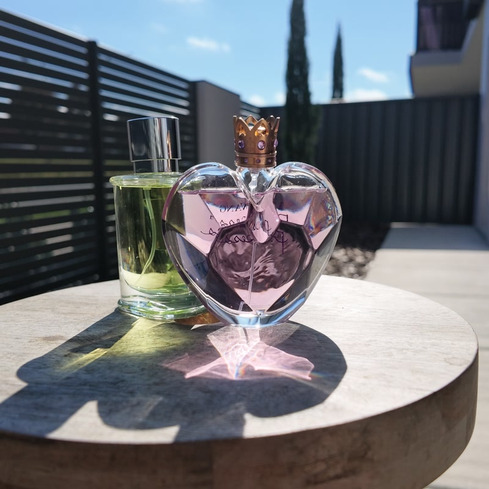
    }
\end{subfigure}\hfill
\begin{subfigure}[c]{0.119\linewidth}
    \centering
    \includegraphics[width=\linewidth,trim={0 20bp 0 20bp},clip]{
        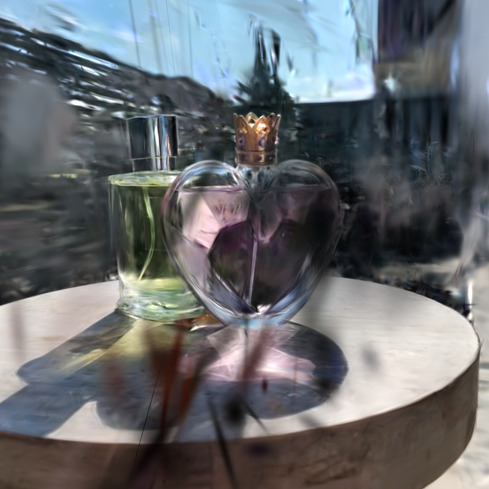
    }
\end{subfigure}\hfill
\begin{subfigure}[c]{0.119\linewidth}
    \centering
    \includegraphics[width=\linewidth,trim={0 20bp 0 20bp},clip]{
        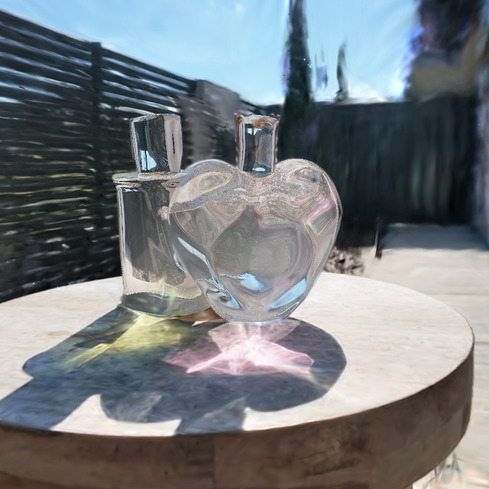
    }
\end{subfigure}\hfill
\begin{subfigure}[c]{0.119\linewidth}
    \centering
    \includegraphics[width=\linewidth,trim={0 20bp 0 20bp},clip]{
        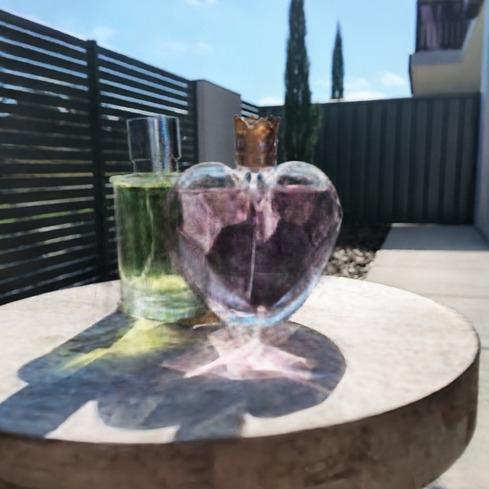
    }
\end{subfigure}\hfill
\begin{subfigure}[c]{0.119\linewidth}
    \centering
    \includegraphics[width=\linewidth,trim={0 20bp 0 20bp},clip]{
        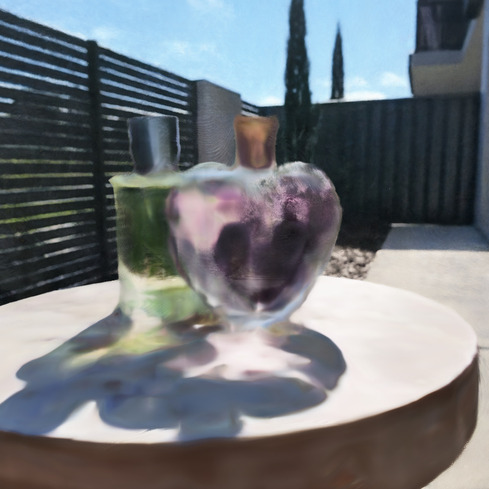
    }
\end{subfigure}\hfill
\begin{subfigure}[c]{0.119\linewidth}
    \centering
    \includegraphics[width=\linewidth,trim={0 20bp 0 20bp},clip]{
        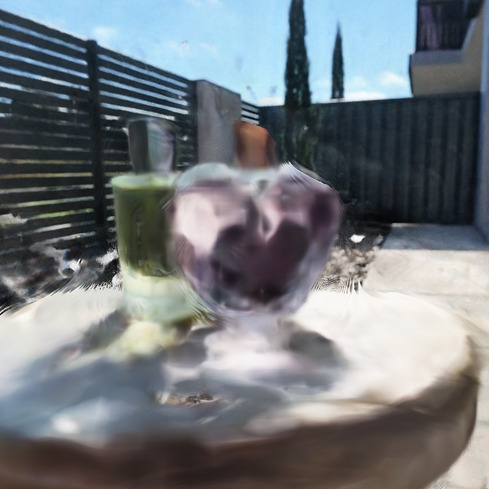
    }
\end{subfigure}\hfill
\begin{subfigure}[c]{0.119\linewidth}
    \centering
    \includegraphics[width=\linewidth,trim={0 20bp 0 20bp},clip]{
        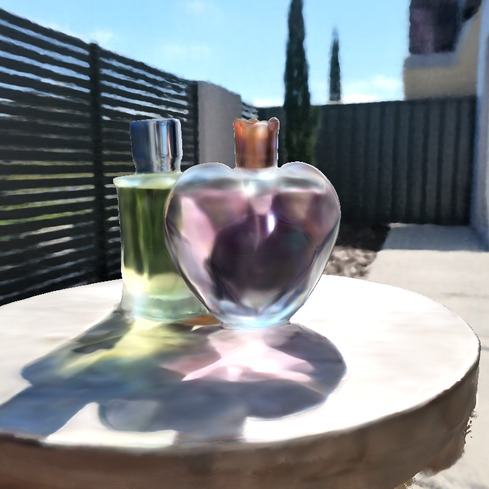
    }
\end{subfigure}\hfill
\begin{subfigure}[c]{0.119\linewidth}
    \centering
    \includegraphics[width=\linewidth,trim={0 20bp 0 20bp},clip]{
        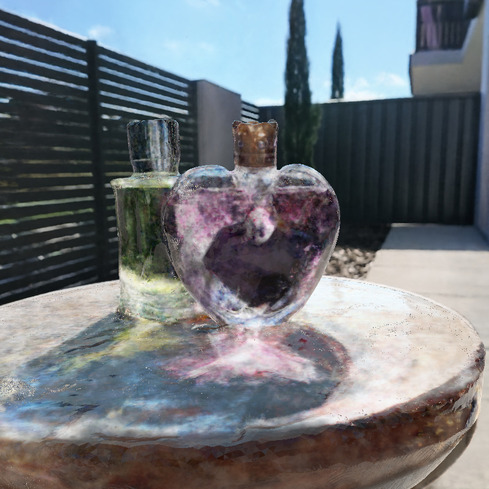
    }
\end{subfigure}\vfill

\begin{minipage}[c]{0.030\linewidth}
    \centering
    \rotatebox{90}{%
        \fontsize{8pt}{9.6pt}\selectfont Lookout Deck%
    }
\end{minipage}%
\begin{subfigure}[c]{0.119\linewidth}
    \centering
    \includegraphics[width=\linewidth,trim={0 20bp 0 20bp},clip]{
        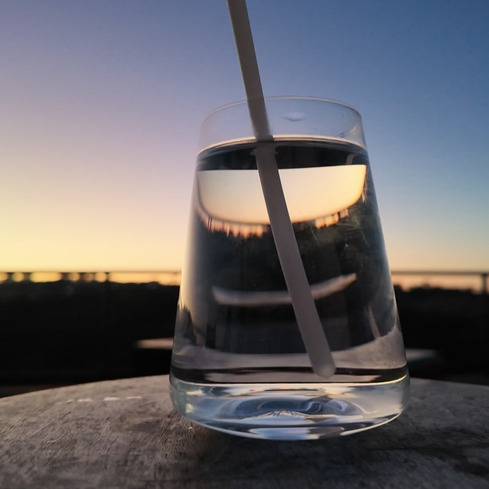
    }
\end{subfigure}\hfill
\begin{subfigure}[c]{0.119\linewidth}
    \centering
    \includegraphics[width=\linewidth,trim={0 20bp 0 20bp},clip]{
        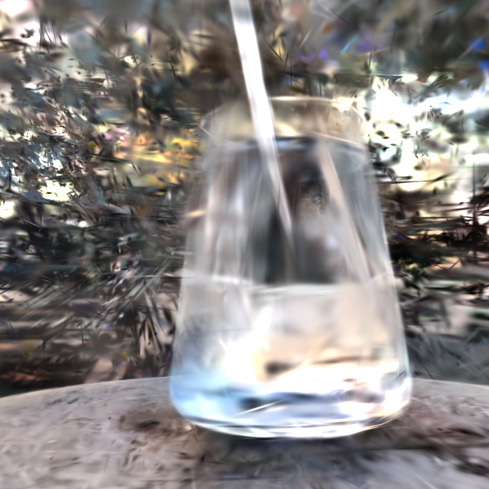
    }
\end{subfigure}\hfill
\begin{subfigure}[c]{0.119\linewidth}
    \centering
    \includegraphics[width=\linewidth,trim={0 20bp 0 20bp},clip]{
        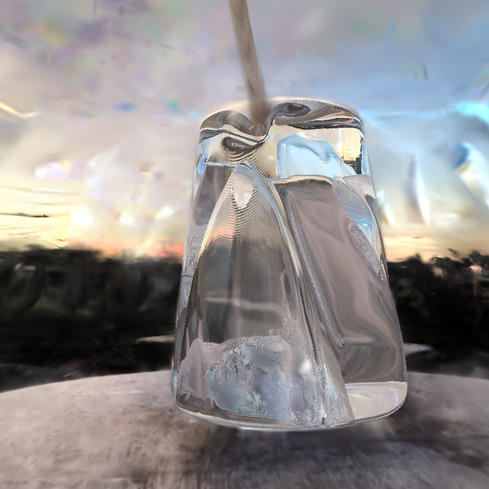
    }
\end{subfigure}\hfill
\begin{subfigure}[c]{0.119\linewidth}
    \centering
    \includegraphics[width=\linewidth,trim={0 20bp 0 20bp},clip]{
        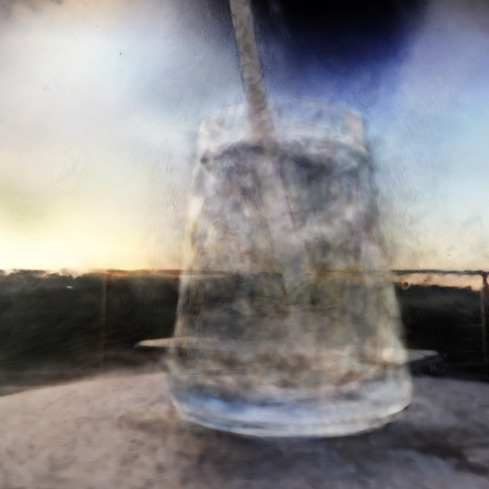
    }
\end{subfigure}\hfill
\begin{subfigure}[c]{0.119\linewidth}
    \centering
    \includegraphics[width=\linewidth,trim={0 20bp 0 20bp},clip]{
        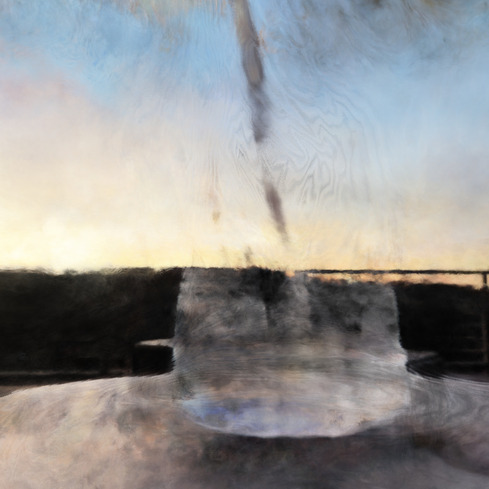
    }
\end{subfigure}\hfill
\begin{subfigure}[c]{0.119\linewidth}
    \centering
    \includegraphics[width=\linewidth,trim={0 20bp 0 20bp},clip]{
        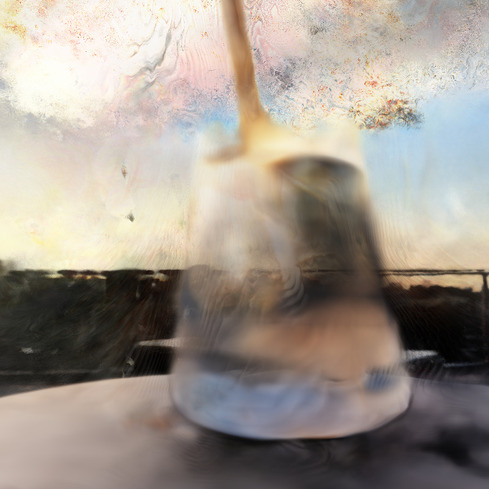
    }
\end{subfigure}\hfill
\begin{subfigure}[c]{0.119\linewidth}
    \centering
    \includegraphics[width=\linewidth,trim={0 20bp 0 20bp},clip]{
        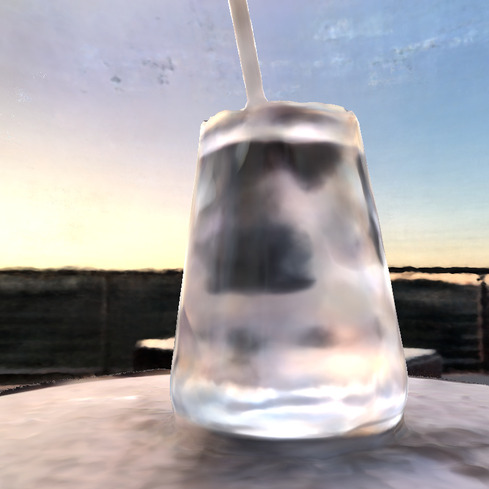
    }
\end{subfigure}\hfill
\begin{subfigure}[c]{0.119\linewidth}
    \centering
    \includegraphics[width=\linewidth,trim={0 20bp 0 20bp},clip]{
        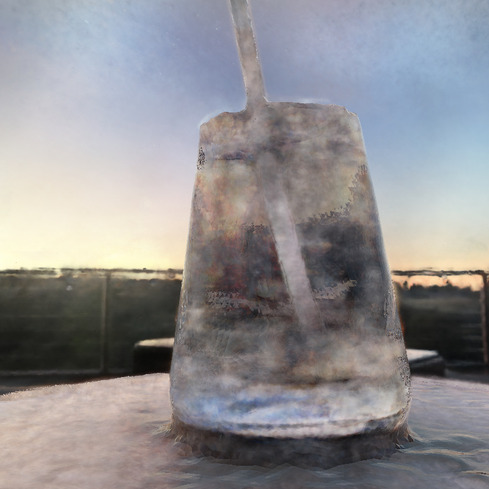
    }
\end{subfigure}\vfill

\begin{minipage}[c]{0.030\linewidth}
    \centering
    \rotatebox{90}{%
        \fontsize{8pt}{9.6pt}\selectfont Living Room%
    }
\end{minipage}%
\begin{subfigure}[c]{0.119\linewidth}
    \centering
    \includegraphics[width=\linewidth,trim={0 20bp 0 20bp},clip]{
        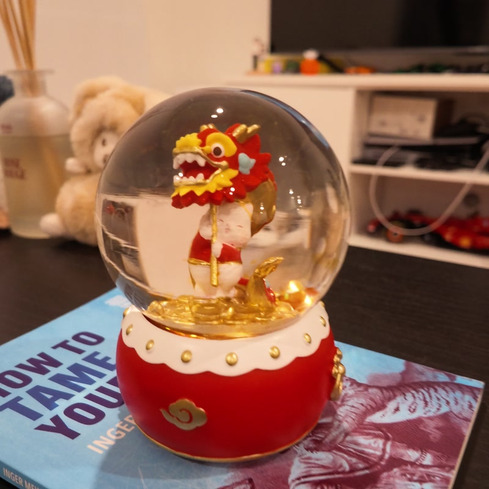
    }
\end{subfigure}\hfill
\begin{subfigure}[c]{0.119\linewidth}
    \centering
    \includegraphics[width=\linewidth,trim={0 20bp 0 20bp},clip]{
        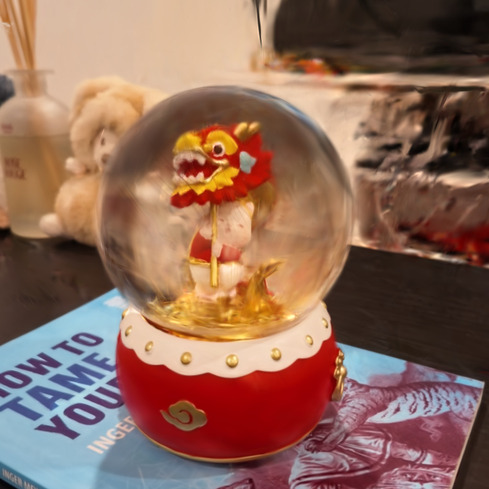
    }
\end{subfigure}\hfill
\begin{subfigure}[c]{0.119\linewidth}
    \centering
    \includegraphics[width=\linewidth,trim={0 20bp 0 20bp},clip]{
        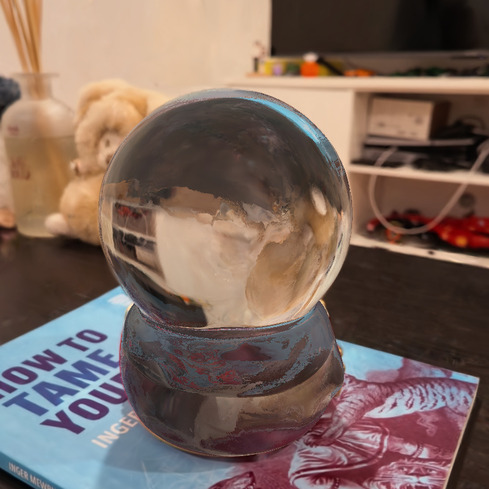
    }
\end{subfigure}\hfill
\begin{subfigure}[c]{0.119\linewidth}
    \centering
    \includegraphics[width=\linewidth,trim={0 20bp 0 20bp},clip]{
        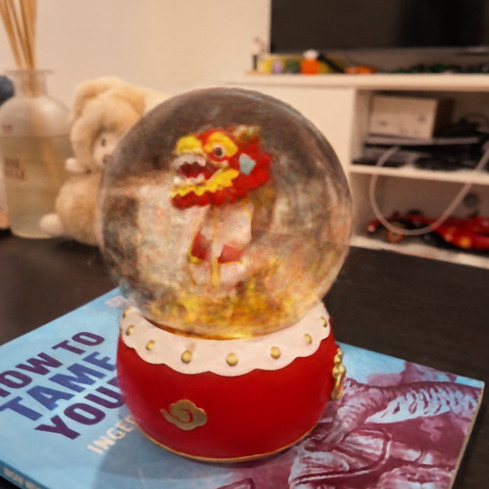
    }
\end{subfigure}\hfill
\begin{subfigure}[c]{0.119\linewidth}
    \centering
    \includegraphics[width=\linewidth,trim={0 20bp 0 20bp},clip]{
        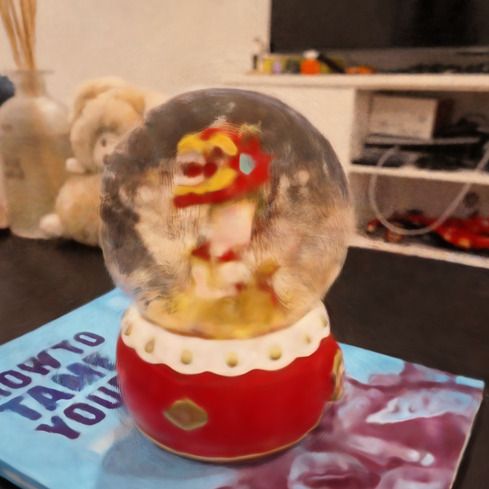
    }
\end{subfigure}\hfill
\begin{subfigure}[c]{0.119\linewidth}
    \centering
    \includegraphics[width=\linewidth,trim={0 20bp 0 20bp},clip]{
        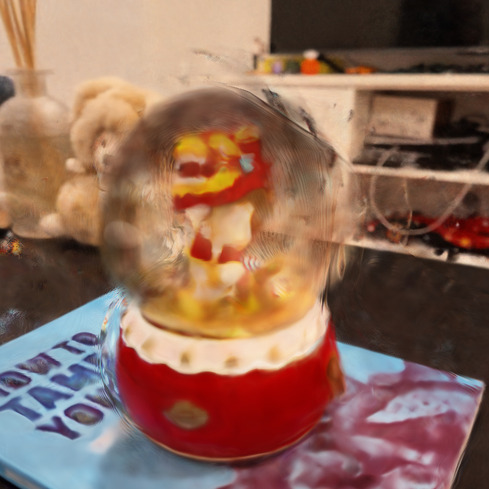
    }
\end{subfigure}\hfill
\begin{subfigure}[c]{0.119\linewidth}
    \centering
    \includegraphics[width=\linewidth,trim={0 20bp 0 20bp},clip]{
        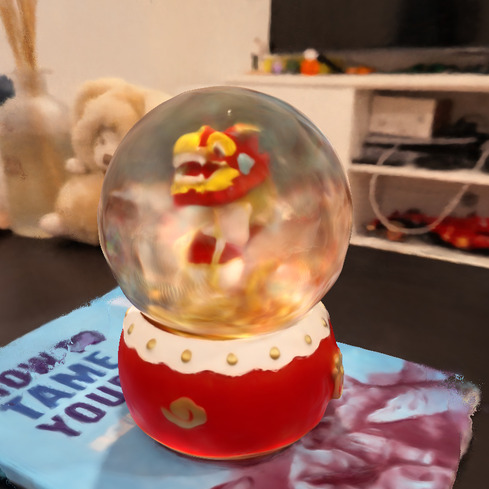
    }
\end{subfigure}\hfill
\begin{subfigure}[c]{0.119\linewidth}
    \centering
    \includegraphics[width=\linewidth,trim={0 20bp 0 20bp},clip]{
        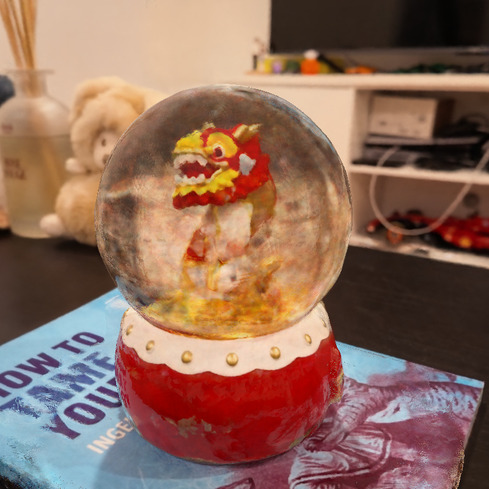
    }
\end{subfigure}%
\caption{Real Scenes}
\label{fig:results_realworld}
\end{subfigure}%
\vspace{-6pt}%
\caption{
    Qualitative results on (a) synthetic 
    and (b) real scenes.
    On the synthetic scenes, the refracted patterns are best recovered by oracle and \method,
    while no method successfully recovers them on the real scenes.
    Reconstruction results are in the appendix (\cref{sec:experiment_supp}).
}
\label{fig:results}
\end{figure*}

\paragraph{Evaluated tasks and compared methods.}
We benchmark novel view synthesis and depth estimation on the synthetic \dataset scenes, and novel view synthesis only on the real scenes, since accurate geometry is not available for this data.
We evaluate general-purpose novel view synthesis methods, including NeuS~\cite{wang2021neus}, Splatfacto~\cite{kerbl3Dgaussians,nerfstudio}, and Zip-NeRF~\cite{barron2023zipnerf}, as well as methods designed for refractive or reflective scenes, including TNSR~\cite{Deng:tnsr}, MS-NeRF~\cite{msnerf}, Ray Deformation Networks~\cite{raydeform}, RoseNeRF~\cite{rosenerf}, NU-NeRF~\cite{SunWYG24}, NeRRF~\cite{nerrf}, and TransparentGS~\cite{huang2025transparentgs}.
We do not evaluate GLINT~\cite{na2026glint}, as it is designed for optically thin glass and approximates the transmission direction by the viewing direction, an assumption that does not account for strong refraction.
Opti-NeuS~\cite{yang2026opti} is not evaluated because its implementation is not publicly available.

\paragraph{Metrics.}
We evaluate novel view synthesis using the
peak signal-to-noise ratio (PSNR), which measures the overall pixel-level fidelity,
masked PSNR (PSNR\textsubscript{M}), which restricts evaluation to the refractive object so that the background does not dominate,
structural similarity index measure (SSIM), which measures the structural consistency of refracted patterns,
and learned perceptual image patch similarity (LPIPS), which is more sensitive to blur than PSNR.
To assess geometric fidelity, we report the distance mean absolute error (DMAE) between the estimated and ground-truth distance maps, which reveals whether accurate appearance is accompanied by accurate geometry, since some methods may reproduce refracted appearances by deforming the geometry or introducing floaters.
Importantly, standard weighted-sum distance rendering techniques~\cite{mildenhall2021nerf} are not reliable for refractive objects, since weights may be distributed across multiple refractive surfaces and the opaque background.
Therefore, we take the median distance of the weight samples along each ray, as a robust estimate of the distance to the nearest (potentially translucent) surface, avoiding an average over disconnected surfaces.

\paragraph{Implementation details.}
The oracle method and R3F are based on Zip-NeRF~\cite{barron2023zipnerf} with added support for piecewise linear light paths and explicit reflection.
For the oracle,
the distortion loss weight is set to $0.01$,
the bounded scene near and far planes are set to $t_n{=}0.05$ and $t_f{=}15$,
the unbounded scene far plane is $t_f{=}1000$ with a contraction warp function,
and the radiance fields are optimized for $25$k iterations on a single A6000 GPU with a batch size of~$4096$.
R3F follows a two-stage reconstruction process:
the background stage uses the settings above for $10$k iterations; and
the foreground stage optimizes a second in-object neural field for $10$k iterations.
For the latter, the rays are traced through the refractive object to the background, capturing all light interactions, but only the in-object field is unfrozen.
Full implementation details are in the appendix (\cref{sec:app_implementation,sec:hyperparameters}).

\paragraph{Synthetic scene results.}
Aggregate results are reported in \cref{tab:result_quantitative}, with breakdowns by background type provided in \cref{tab:result_quantitative_breakdown}.
Among non-oracle methods, \method achieves the highest PSNR and SSIM and the lowest DMAE, while NU-NeRF~\cite{SunWYG24} achieves the highest masked PSNR.
However, the oracle is substantially better overall.
The qualitative examples in \cref{fig:results} show scenes with relatively simple geometry and material composition.
The cube exhibits extensive total internal reflection, whereas the dog's nonconvex geometry causes repeated surface intersections and refractions, producing highly nonlinear light transport that is sensitive to small geometric errors.
The torus further introduces tinting effects.
Zip-NeRF~\cite{barron2023zipnerf} and NU-NeRF blur the refracted patterns.
NeRRF~\cite{nerrf} assumes an infinitely distant background and cannot reconstruct nearby geometry,
while TransparentGS~\cite{huang2025transparentgs} produces sharp-looking but incorrect structures; neither models tinting.
In contrast, \method and the oracle recover refracted patterns more accurately.
This shows that accurate physics-based reconstruction requires precise geometry and optical properties.
Without them, even simple refractive objects remain challenging.

\paragraph{Real scene results.}
The failures become more severe on real-world captures.
The examples in \cref{fig:results} include tinting, multiple interfaces, fine internal structures, and opaque content viewed through refractive surfaces.
Although Zip-NeRF leads all reported novel view synthesis metrics, its rendering results remain heavily blurred, and all methods perform poorly.
The glass-and-straw scene is particularly challenging, with most methods producing floaters around the object.
These consistent failures demonstrate that real-world refractive reconstruction remains far from solved.

\section{Conclusion}
\label{sec:conclusion}

In this work, we present \dataset{}, a dataset and benchmark comprising 150 synthetic and 60 real scenes for 3D reconstruction and novel view synthesis of refractive and reflective objects.
It combines diverse geometries, material compositions, and tinting with bounded and unbounded synthetic backgrounds and indoor and outdoor real-world environments, allowing us to reveal the failure modes of existing methods across diverse challenges.
As reference points for the benchmark, we introduce an oracle method using ground-truth geometry and refractive indices and a simple two-stage baseline using estimated geometry. %
Benchmarking state-of-the-art methods, including those designed for nonlinear light paths, revealed significant performance gaps.
This shows that the task remains far from solved, highlighting the value of \dataset{} as a challenging benchmark against which to develop more reliable methods.

\section*{Acknowledgments}
Dr Campbell is the recipient of an Australian Research Council Discovery Early Career Award (project number DE250100542) funded by the Australian Government.

%% file: appendix.tex
\clearpage

\twocolumn[
\begin{center}
    {\Large\bfseries Supplementary Material}
\end{center}
\vspace{1em}
]

\setcounter{section}{0}
\renewcommand{\thesection}{\Alph{section}}
\renewcommand{\theHsection}{app.\Alph{section}}

\section{Synthetic Scene Rendering Details}
\label{sec:rendering_details}
We render all synthetic scenes using the Cycles path tracer in Blender \cite{blender}, which employs Monte Carlo integration to solve the rendering equation using BSDFs (Bidirectional Scattering Distribution Functions) \cite{kajiya1986rendering}. Light transport is approximated by stochastically tracing paths of light rays and recursively evaluating their interactions with surfaces and transmissive materials, as follows:
\begin{equation}
\begin{aligned}
L_o(x, \omega_o)
&=
L_e(x, \omega_o)
\\
&
+ \int_{\Omega} f_s(x, \omega_i, \omega_o)\, L_i(x, \omega_i)\, |\omega_i \cdot n|\, d\omega_i,
\end{aligned}
\end{equation}
where $L_o(x, \omega_o)$ is the radiance leaving point $x$ in direction $\omega_o$, $L_e$ is the emitted radiance, $f_s$ is the BSDF, $L_i$ is the incoming radiance from direction $\omega_i$, $n$ is the surface normal at $x$, and $\Omega$ is the unit sphere of incoming directions. The absolute cosine term $|\omega_i \cdot n|$ accounts for the cosine falloff of incoming light. Unlike the BRDF, the BSDF $f_s$ accounts for both surface reflection and transmission, which is necessary for modeling materials such as glass and transparent plastics used in our scenes.

Refractive materials use Blender’s Glass BSDF with zero roughness and a material-specific refractive index, and tinted materials additionally contain a Volume Absorption shader, so that the observed color depends on the distance traveled through the material. Opaque components, such as metal and plastic parts, use the Principled BSDF. All objects are smooth-shaded Catmull--Clark subdivision surfaces, and the released meshes are these subdivided surfaces. Object placements, lighting, and camera positions are randomized to encourage variation across samples. To ensure rendering quality, we configure Cycles as follows:
\begin{itemize}[left=0pt, noitemsep]
    \item Adaptive sampling \cite{bucher1988adaptive} is enabled to improve rendering efficiency by dynamically adjusting the number of samples per pixel based on estimated noise. Pixels with low variance receive fewer samples, while high-frequency regions are sampled more densely. We set the noise threshold to 0.05, which defines the per-pixel error tolerance for terminating sampling. The minimum number of samples is set to 0, allowing Cycles to automatically determine this value from the threshold. The maximum number of samples per pixel is capped at 600.
    \item The maximum number of light bounces is set to 12 to support complex global illumination. This includes up to 4 diffuse bounces for soft indirect light, 4 glossy bounces for specular reflections, 12 transmission bounces to handle multiple refractive interfaces, and 8 transparent bounces for rays passing through alpha-masked geometry. Volume scattering is disabled (volume bounces = 0), since the only volumes in our scenes are purely absorbing and do not scatter light.
    \item To reduce bright noise artifacts (``fireflies''), we clamp both direct and indirect light contributions to 1.0. We also enable reflective and refractive caustics to preserve high-frequency light transport effects such as focused reflections or refractions through transparent objects.
    \item The Glossy Filter is set to 0.0 to retain sharp caustic and highlight details without blurring.
    \item Images are denoised with OpenImageDenoise using albedo and normal guide passes. We use a Gaussian pixel filter with a width of 1.5 pixels and the Standard view transform for color management.
\end{itemize}

\paragraph{Re-rendering the scenes.}
Methods that re-render the released meshes with their own physically based renderer should match the Gaussian pixel filter and, to a lesser extent, the denoiser, since both noticeably affect the appearance inside refractive objects.

\begin{table*}[!t]
    \centering
    \footnotesize
    \caption{
        Source and license information for objects used in our dataset.
        Objects are obtained from BlenderKit, Blender, CGTrader, Free3D, Keenan's 3D Model Repository, or created by the authors.
        Website-sourced objects were further modified (\eg, their materials and refractive indices) to better suit our dataset requirements.
    }
    \label{tab:object_licenses}
    \vspace{-6pt}
    \setlength{\tabcolsep}{5pt}
    \renewcommand{\arraystretch}{1.0}
    \begin{tabularx}{\textwidth}{@{}l l r >{\raggedright\arraybackslash}X@{}}
        \toprule
        Source & License & \# & Objects \\
        \midrule
        BlenderKit & Royalty Free & 31 & ampoule, beaker, candle holder, cat, cola bottle, conical flask, crystal vase, demijohn vase, diamond, flower vase, fox, household item set, Korken jar, lab equipment set, light bulb, magnifier, man sculpture, perfume red, perfume yellow, plastic bottle, reed diffuser, skull bottle, sleeping dragon, star-shaped bottle, teacup, teapot, Vardagen jar, vial, water pitcher, wine glass, wisolt kettle \\
        \addlinespace[1pt]
         & CC0 & 2 & generic sculpture, woman sculpture \\
        \addlinespace[3pt]
        Blender & GPL & 1 & monkey \\
        \addlinespace[3pt]
        CGTrader & Royalty Free & 3 & graduated cylinder, round bottom flask, test tube \\
        \addlinespace[3pt]
        Free3D & Royalty Free & 1 & dog \\
        \addlinespace[3pt]
        Keenan's & CC0 & 1 & cow \\
        \addlinespace[3pt]
        Authors & CC0 & 11 & ball, cube, cylinder, pyramid, and torus (one clear and one tinted version of each), syringe \\
        \midrule
        Total & & 50 & \\
        \bottomrule
    \end{tabularx}
\end{table*}

\paragraph{Licenses.}
\label{sec:licenses}
Detailed information on the sources and licensing of objects in our dataset is provided in \cref{tab:object_licenses}, covering both objects created by our team and those sourced from online repositories with specific licensing terms. For sourced objects, we made material modifications where necessary to better align with the dataset's requirements.

\section{Real Scene Capture Details}
\label{sec:real_capture_details}

\paragraph{Object configurations and environments.}
The real-world subset contains 60 scenes, obtained by capturing 15 object configurations in four environments.
The configurations comprise a glass sphere, pyramid, rabbit, bulb-shaped bottle, dishwashing-liquid bottle, two different snow globes, cocktail shaker, measuring jug, perfume bottle, perfume set, glass of water, glass of water with a straw, wine bottle, and wine-bottle set.
The perfume set and wine-bottle set contain multiple objects, while the remaining configurations contain a single object.
The four environments include two outdoor and two indoor settings.
The outdoor settings are a courtyard under direct sunlight and a lookout deck captured after sunset, while the indoor settings are a bright living room illuminated by ceiling lights and a darker room containing a puzzle table illuminated by a lamp.
Each object configuration is captured once in every environment, yielding $15 \times 4 = 60$ scenes.

\paragraph{Image acquisition.}
For each scene, we record two RGB videos using a DJI OSMO Pocket 4 at a resolution of $4096 \times 2160$.
The camera moves around the object while the object and surrounding environment remain stationary.
The first video follows the training trajectory, while the second follows a trajectory with a different radius and is used for testing.
This separation reduces evaluation based on interpolation between nearby training views.
We uniformly sample 100 frames from each video.
Of the 100 frames from the first trajectory, 90 are used for training and 10 for validation, while all 100 frames from the second trajectory are used for testing.
The images are downsampled and center-cropped to $800 \times 800$ pixels to match the synthetic scenes, and the camera intrinsics are adjusted accordingly.

\paragraph{Camera calibration and pose estimation.}
We jointly estimate the camera parameters of all 200 frames using Agisoft Metashape~\cite{agisoftmetashape}.
Joint estimation places the training and test cameras in a common coordinate system.
During image alignment, the foreground refractive objects are excluded using masks generated by SAM~3~\cite{sam3segmentconcepts}.
This prevents view-dependent refractions and reflections from being treated as stationary features during feature matching.

\paragraph{Capture controls.}
Outdoor scenes are captured under cloudless skies and windless conditions to minimize changes in illumination and background motion.
For captures under direct sunlight, we ensure that shadows cast by the camera and operator do not enter the scene.
For liquid-filled objects, visible bubbles are removed and recording begins only after the liquid has completely settled.
Despite these controls, the real scenes retain realistic imperfections, including surface defects, reflections of the camera, sensor noise, and residual pose-estimation errors.

\paragraph{Pose-consistency evaluation.}
We assess pose consistency using SIFT~\cite{lowe2004distinctive}
correspondences from the static background.
Image pairs are selected using viewing-angle and camera-baseline filters, including a $60^\circ$ viewing-angle criterion.
We exclude features within dilated foreground-object masks, match SIFT descriptors using bidirectional brute-force $L_2$ search, and retain matches that pass Lowe's ratio test and a mutual consistency check.
The matched points are undistorted using the estimated camera distortion parameters.
We then derive the fundamental matrix directly from the estimated intrinsics and poses~\cite{hartley2003multiple}, after converting from the OpenGL to OpenCV camera convention, and compute the symmetric point-to-epipolar-line distance for every match.
We estimate fundamental matrices using USAC-MAGSAC~\cite{raguram2012usac,barath2020magsac++}.
Across all scenes, we obtain 3.23 million valid correspondences from 25,934 image pairs, with a median epipolar error of 0.36 pixels.

\begin{figure*}[!t]
    \centering
    \includegraphics[width=\linewidth]{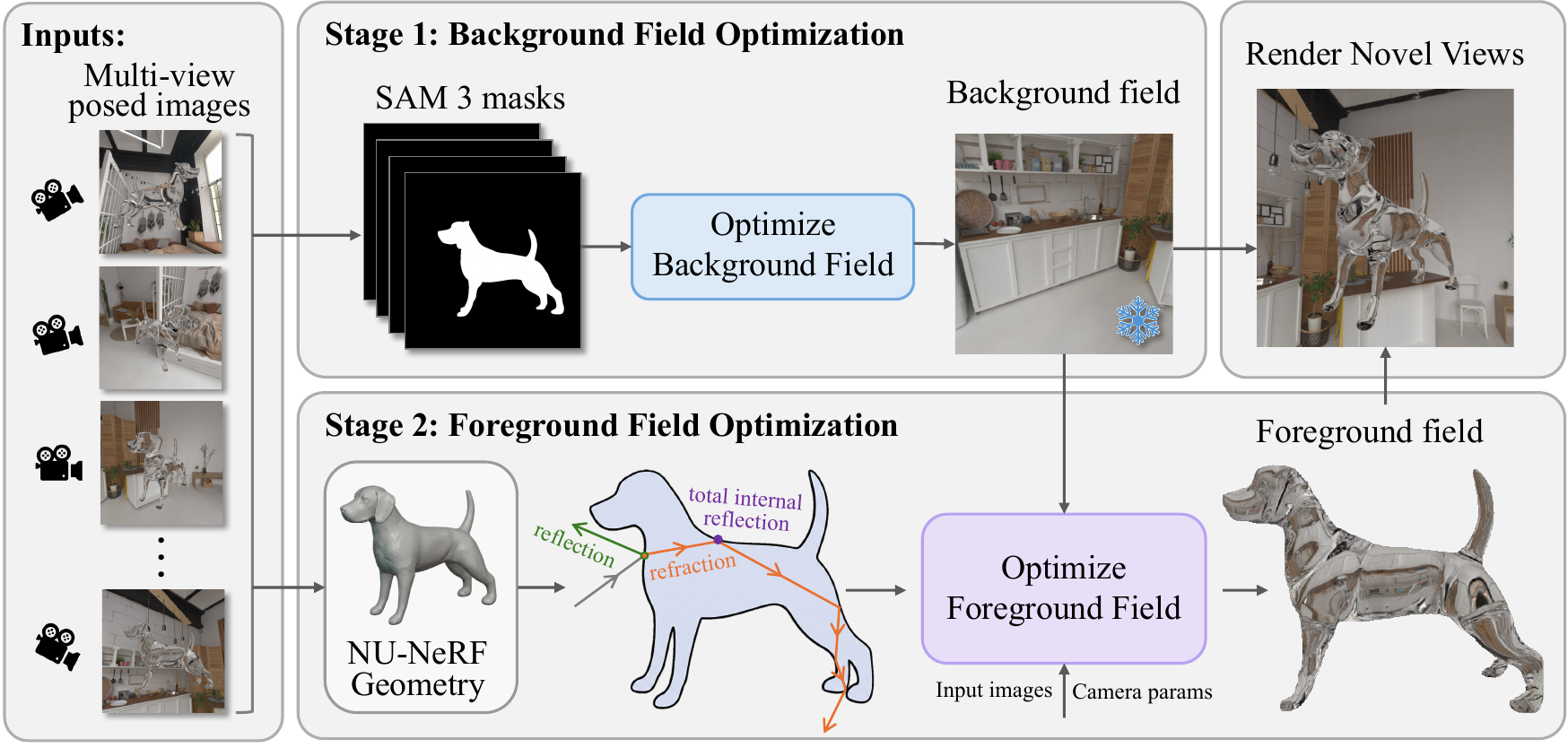}
    \caption{
        \method pipeline.
        Given multi-view posed images, \method performs two-stage reconstruction.
        In Stage 1, object masks predicted by SAM~3~\cite{sam3segmentconcepts} are used to mask foreground rays while optimizing a background radiance field with Zip-NeRF~\cite{barron2023zipnerf}.
        In Stage 2, the object geometry and refractive index are estimated using NU-NeRF~\cite{SunWYG24}.
        These estimates are then used to optimize a separate in-object foreground field via full ray tracing that models complex light interactions such as reflection, multi-bounce refraction, and total internal reflection, while keeping the background field frozen.
        The learned foreground field captures absorption and tinting effects within the refractive object and is combined with the background field to render novel views.
    }
    \label{fig:r3f_pipeline}
    \vspace{-9pt}
\end{figure*}

\section{Additional Methodological Details}

In this section, we provide additional details on our sampling strategy and how updated sample points are processed for density and color prediction in our oracle method.
These extensions complement the core concepts outlined in the main paper and focus on implementation-specific details.

\paragraph{Sampling strategy.}
We begin by generating an initial straight ray $\rvr(t) = \rvo + t\rvd$, which is then split into two paths: a refraction path $\rvr^\refr$ and a reflection path $\rvr^\refl$.
To sample along these paths, we adopt the proposal sampling strategy introduced in Zip-NeRF \cite{barron2023zipnerf}, which builds upon the hierarchical sampling approach of Mip-NeRF 360 \cite{barron2022mip}. Zip-NeRF samples inside a cone following a spiral path, we update the center line of the cone while preserving the sampling pattern. It starts with a uniform sampler that generates $N$ sample points $\rvx_i$ along the initial straight ray.
These points are then updated, resulting in new positions $\rvx_i^\refr$ and $\rvx_i^\refl$ with corresponding direction vectors $\rvd_i^\refr$ and $\rvd_i^\refl$, which follow curved paths for refractions and reflections, respectively.
Next, the density $\sigma_i$ and weight $w_i$ of each sample point along these curved paths are computed.
These weights, which represent the contribution of each sample, are further passed through a probability density function sampler to concentrate samples in higher density regions to enhance the accuracy of rendering.

\paragraph{Processing updated sample points.}
After each sampling stage, the updated sample point positions $\rvx_i^\refr$ and $\rvx_i^\refl$ are processed through a spatial encoding function $\xi$ for efficient representation.
These encoded spatial points are then fed into a multi-layer perceptron (MLP) $f_\theta$, which predicts the scene density $\sigma_i$.
To obtain the predicted color at each 3D position, another MLP $g_\theta$ processes the refined viewing directions $\rvd_i^\refr$ and $\rvd_i^\refl$, encoded using spherical harmonics $\psi$, alongside the computed density $\sigma_i$.
This yields the emitted color $\rvc_i = (R, G, B)$ for each sample point.

\begin{figure*}[!t]
    \centering
    \begin{subfigure}[]{0.25\linewidth}\centering
        \includegraphics[width=0.98\linewidth,height=0.65\linewidth]{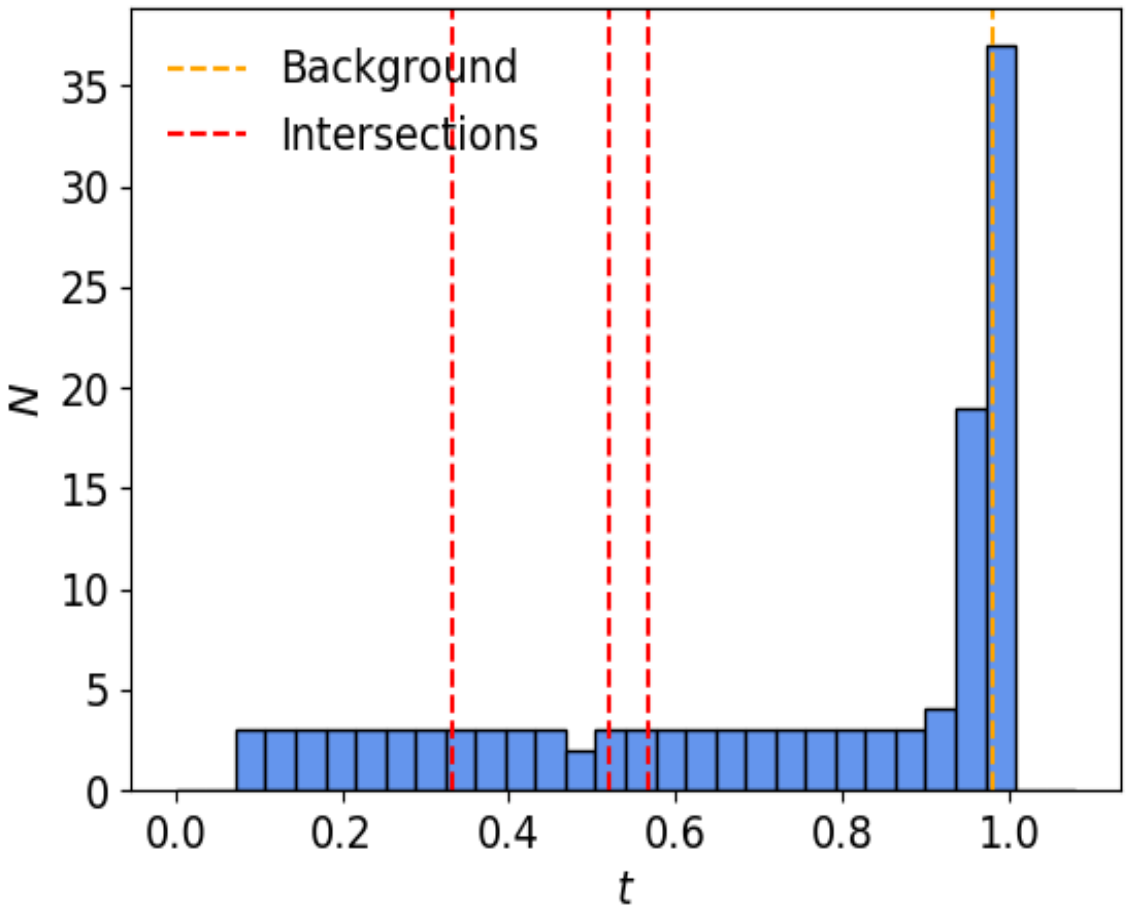}
        \caption{$\!\gL_{\text{dist}}^\text{orig}\!$ sample distr.}
        \label{fig:distortion_a}
    \end{subfigure}\hfill
    \begin{subfigure}[]{0.25\linewidth}\centering
        \includegraphics[width=0.98\linewidth,height=0.65\linewidth]{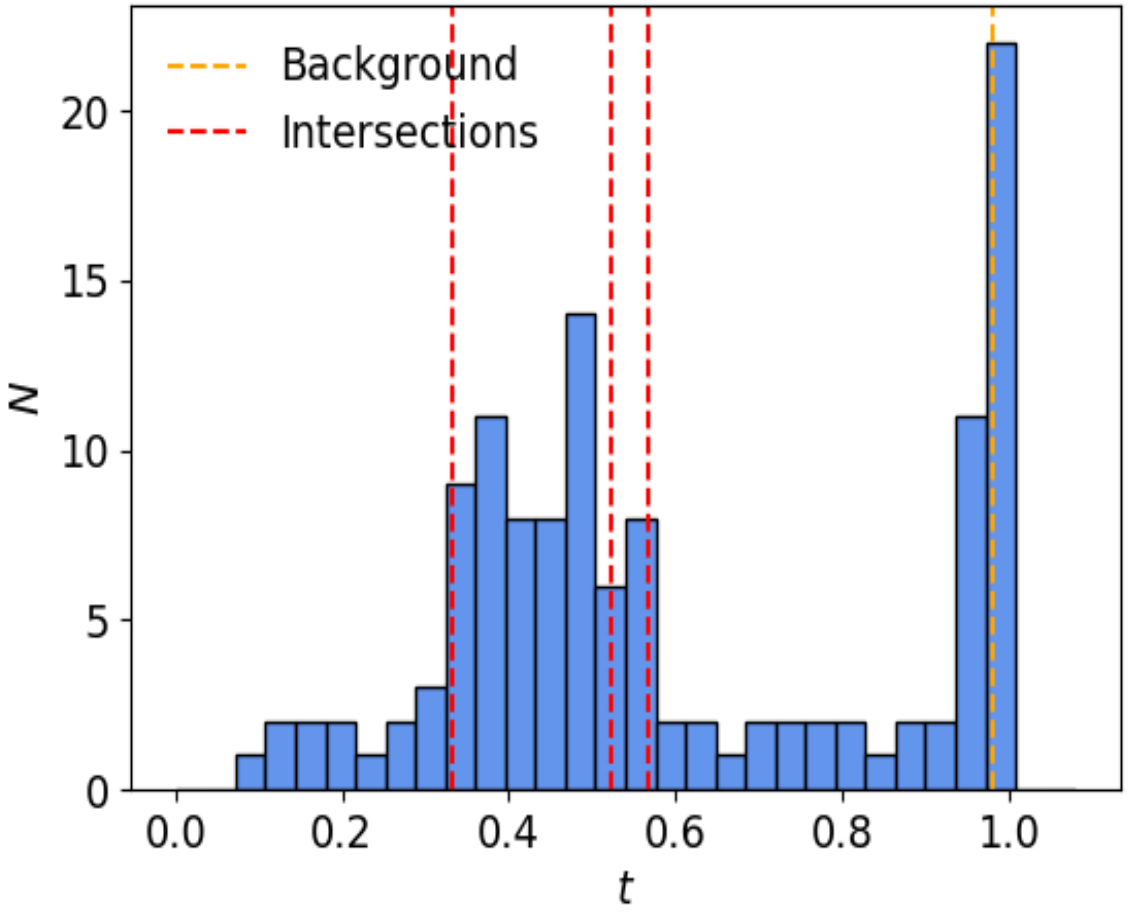}
        \caption{$\!\gL_\text{dist}^{\vphantom{\text{orig}}}\!$ sample distr.}
        \label{fig:distortion_b}
    \end{subfigure}\hfill
    \begin{subfigure}[]{0.25\linewidth}\centering
        \includegraphics[width=\linewidth,height=0.65\linewidth]{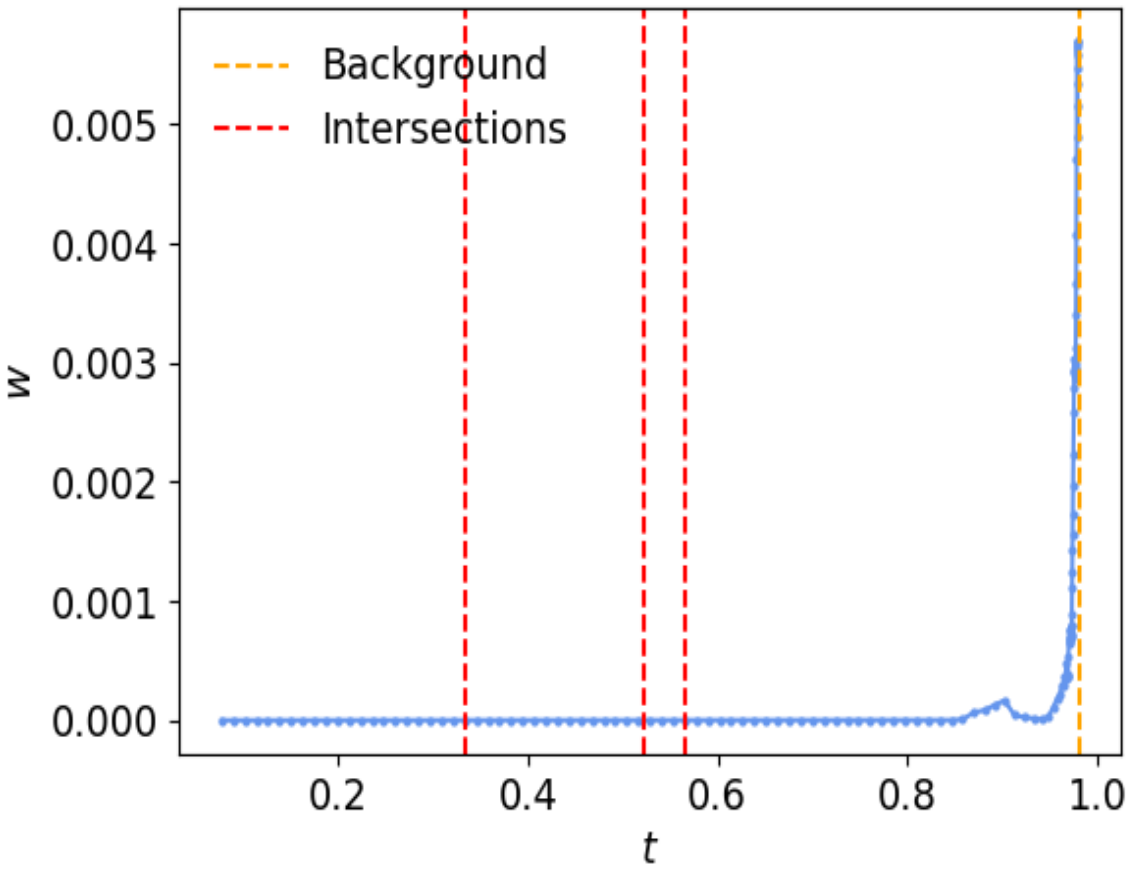}
        \caption{$\!\gL_{\text{dist}}^\text{orig}\!$ weight distr.}
        \label{fig:distortion_c}
    \end{subfigure}\hfill
    \begin{subfigure}[]{0.25\linewidth}\centering
        \includegraphics[width=\linewidth,height=0.65\linewidth]{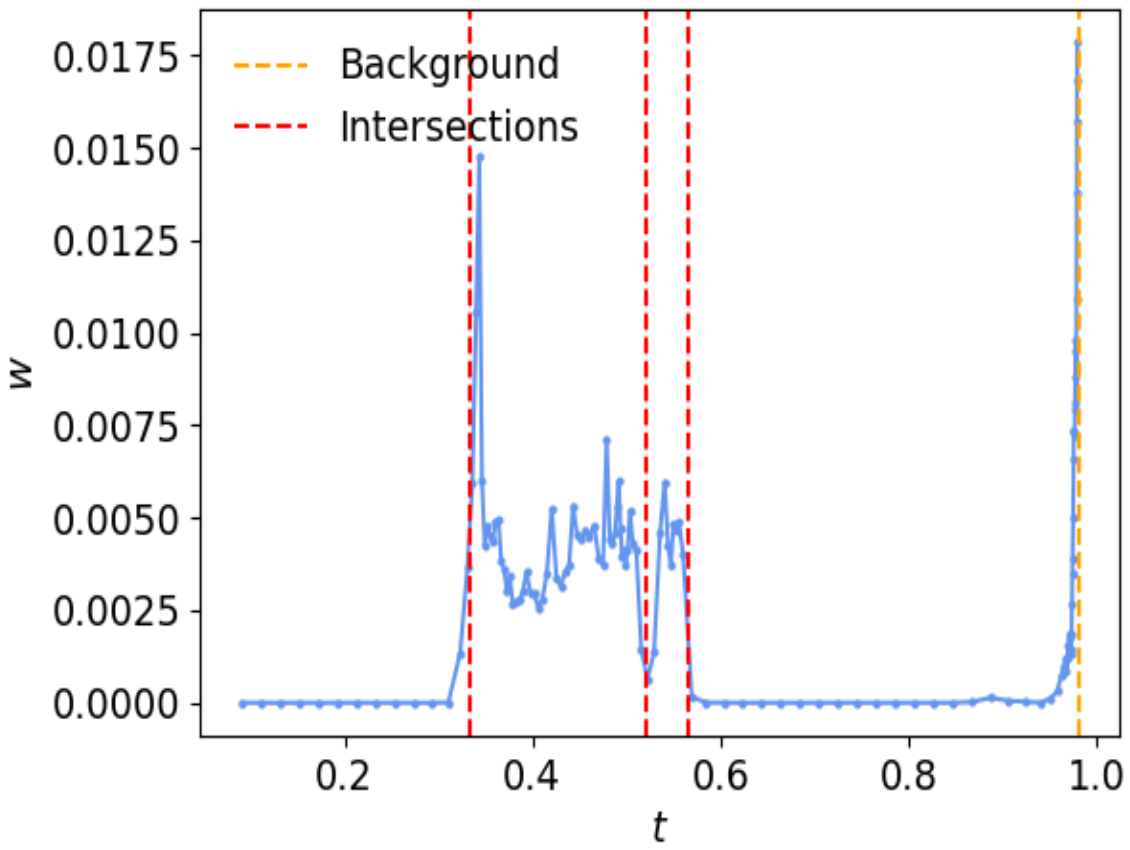}
        \caption{$\!\gL_{\text{dist}}^{\vphantom{\text{orig}}}\!$ weight distr.}
        \label{fig:distortion_d}
    \end{subfigure}%
    \vspace{-3pt}%
    \caption{
        Comparison of the original~\cite{barron2022mip} distortion loss $\gL_\text{dist}^\text{orig}$ and the translucency-corrected distortion loss $\gL_{\text{dist}}$.
        (\subref{fig:distortion_a}--\subref{fig:distortion_b})~Sample point and
        (\subref{fig:distortion_c}--\subref{fig:distortion_d})~weight distributions.
    }
    \label{fig:distortion}
\end{figure*}

\paragraph{Corrected distortion loss.}
The distortion loss $\gL_\text{dist}^\text{orig}$ introduced in Mip-NeRF 360~\cite{barron2022mip} encourages the weight distribution along a ray to coalesce and otherwise tend to zero.
This leads to a preference for a single high-weight peak, that is, a single opaque surface.
This is desirable for standard NeRF settings, since it reduces floaters and background collapse.
However, it is not applicable for translucent objects, where color contributions along a ray arise from both the translucent and opaque media.
As shown in \cref{fig:distortion_a,fig:distortion_c}, the unmodified loss tends to reduce the density of translucent objects to zero, which also reduces samples within the object due to proposal sampling.
After applying the corrected distortion loss in \cref{eq:distortion_loss}, which excludes samples inside the refractive object, the model allocates more samples with higher weights within the refractive object (\cref{fig:distortion_b,fig:distortion_d}), leading to more accurate reconstruction.

\section{Further Implementation Details}
\label{sec:app_implementation}

For optimizing the oracle method, we use an Adam optimizer with an initial learning rate of $8 \times 10^{-3}$, $\epsilon = 1 \times 10^{-15}$, and an exponential decay to $1 \times 10^{-3}$ over $2.5 \times 10^4$ steps, with a 1000-step warm-up. Scene contraction is applied for HDR environment map backgrounds.  

The \method pipeline is illustrated in \cref{fig:r3f_pipeline}. The initial geometry and refractive index of the object are estimated using NU-NeRF~\cite{SunWYG24}.
To smooth and simplify the geometry for convex shapes, we perform a convex hull check by computing the chamfer distance between the estimated mesh and its convex hull.
If the chamfer distance falls below a predefined threshold of 0.01, the estimated geometry is replaced by its convex hull, as the object is likely to be close to convex.
This particularly simplifies the mesh for piecewise-planar shapes, increasing the efficiency of the ray tracing.

\section{Hyperparameters}
\label{sec:hyperparameters}
In this section, we detail the hyperparameter settings used for training the models evaluated in our experiments.
The configurations, including learning rates, optimizers, and scheduler settings, were carefully chosen to ensure stable convergence and performance across different methods. 
Specific hyperparameter choices for the evaluated approaches are described below.

\begin{table*}[!t]
    \centering
    \renewcommand{\arraystretch}{0.9}
    \caption{
    Detailed quantitative breakdown of the \dataset test set by background type and object geometry.
    The synthetic scenes comprise three background types: two categories of bounded backgrounds, including high-frequency cube and sphere patterns, and one category of unbounded backgrounds consisting of multiple environment maps.
    The indoor real scenes consist of the living room and puzzle table captures, while the outdoor real scenes consist of the courtyard and lookout-deck captures.
    For environment-map scenes, DMAE is evaluated only within the foreground object mask, as the background has no finite depth. Real-scene DMAE is unavailable.
    The best and second-best results within each background/geometry block are shown in bold and underlined, respectively.
    A dash denotes an unavailable result.
    $^\dagger$ Has access to ground-truth foreground geometry and material properties.
    }
    \setlength{\tabcolsep}{1pt}
    \label{tab:result_quantitative_breakdown}

    \resizebox{\linewidth}{!}{%
        \begin{tabular}{@{}llccccc@{\hspace{0.5em}}ccccc@{\hspace{0.5em}}ccccc@{\hspace{0.5em}}ccccc@{}}
        \toprule
         & Method
         & \multicolumn{5}{c}{Convex single-material}
         & \multicolumn{5}{c}{Nonconvex single-material}
         & \multicolumn{5}{c}{Nonconvex multi-material}
         & \multicolumn{5}{c}{Entire dataset} \\
         \cmidrule(lr){3-7}
         \cmidrule(lr){8-12}
         \cmidrule(lr){13-17}
         \cmidrule(lr){18-22}

         & &
         \scriptsize{PSNR}\newline$\uparrow$ &
         \scriptsize{PSNR{\raisebox{-0.6ex}{\fontsize{3.5pt}{4pt}\selectfont \hspace{-0.5pt}M}}}\newline$\uparrow$ &
         \scriptsize{SSIM}\newline$\uparrow$ &
         \scriptsize{LPIPS}\newline$\downarrow$ &
         \scriptsize{DMAE}\newline$\downarrow$ &
         \scriptsize{PSNR}\newline$\uparrow$ &
         \scriptsize{PSNR{\raisebox{-0.6ex}{\fontsize{3.5pt}{4pt}\selectfont \hspace{-0.5pt}M}}}\newline$\uparrow$ &
         \scriptsize{SSIM}\newline$\uparrow$ &
         \scriptsize{LPIPS}\newline$\downarrow$ &
         \scriptsize{DMAE}\newline$\downarrow$ &
         \scriptsize{PSNR}\newline$\uparrow$ &
         \scriptsize{PSNR{\raisebox{-0.6ex}{\fontsize{3.5pt}{4pt}\selectfont \hspace{-0.5pt}M}}}\newline$\uparrow$ &
         \scriptsize{SSIM}\newline$\uparrow$ &
         \scriptsize{LPIPS}\newline$\downarrow$ &
         \scriptsize{DMAE}\newline$\downarrow$ &
         \scriptsize{PSNR}\newline$\uparrow$ &
         \scriptsize{PSNR{\raisebox{-0.6ex}{\fontsize{3.5pt}{4pt}\selectfont \hspace{-0.5pt}M}}}\newline$\uparrow$ &
         \scriptsize{SSIM}\newline$\uparrow$ &
         \scriptsize{LPIPS}\newline$\downarrow$ &
         \scriptsize{DMAE}\newline$\downarrow$ \\

        \midrule
        \multirow{12}{*}{\rotatebox{90}{\makecell{Cube BG}}}
         & NeuS
         & 18.48 & 14.94 & 0.66 & 0.19 & 1.20
         & 19.11 & 14.86 & 0.67 & 0.33 & 1.17
         & 19.19 & 15.97 & 0.63 & 0.20 & 1.31
         & 19.15 & 15.33 & 0.65 & 0.25 & 1.23 \\

         & Splatfacto
         & \underline{22.32} & 14.90 & \underline{0.87} & \underline{0.11} & 14.56
         & \underline{24.40} & 16.02 & \underline{0.88} & \underline{0.07} & 9.42
         & \underline{24.70} & \underline{17.72} & \underline{0.86} & \underline{0.10} & 11.03
         & \underline{24.26} & 16.69 & \underline{0.87} & \underline{0.09} & 10.96 \\

         & Zip-NeRF
         & 17.98 & 14.52 & 0.69 & 0.37 & 1.31
         & 20.28 & 15.20 & 0.75 & 0.23 & 1.57
         & 21.43 & 16.02 & 0.79 & 0.19 & 1.88
         & 20.40 & 15.44 & 0.76 & 0.24 & 1.66 \\

         & TNSR
         & 19.30 & 11.92 & 0.71 & 0.26 & 0.90
         & 18.92 & 11.25 & 0.83 & 0.22 & 1.31
         & 17.66 & 10.06 & 0.71 & 0.16 & 1.19
         & 18.78 & 11.02 & 0.75 & 0.20 & 1.10 \\

         & MS-NeRF
         & 17.77 & 14.26 & 0.67 & 0.34 & 0.40
         & 19.76 & 14.86 & 0.73 & 0.24 & 0.37
         & 21.48 & 16.61 & 0.78 & 0.22 & 0.45
         & 20.07 & 15.44 & 0.74 & 0.25 & 0.41 \\

         & RayDef
         & 18.17 & 14.83 & 0.69 & 0.38 & 0.25
         & 19.56 & 14.96 & 0.70 & 0.28 & 0.57
         & 21.60 & 16.53 & 0.76 & 0.23 & 0.47
         & 20.18 & 15.61 & 0.73 & 0.27 & 0.43 \\

         & RoseNeRF
         & 16.17 & 10.11 & 0.67 & 0.42 & --
         & 18.61 & 11.01 & 0.65 & 0.31 & --
         & 19.34 & 12.12 & 0.66 & 0.27 & --
         & 18.04 & 11.08 & 0.66 & 0.33 & -- \\

         & NU-NeRF
         & 17.61 & 17.39 & 0.61 & 0.40 & 0.14
         & 18.53 & \underline{17.72} & 0.63 & 0.30 & 0.22
         & 17.38 & 16.55 & 0.62 & 0.40 & 0.61
         & 17.84 & \underline{17.22} & 0.62 & 0.37 & 0.32 \\

         & NeRRF
         & 12.15 & 12.78 & 0.37 & 0.85 & 3.03
         & 12.37 & 13.19 & 0.42 & 0.79 & 3.40
         & 11.72 & 12.13 & 0.43 & 0.86 & 3.51
         & 12.06 & 12.67 & 0.42 & 0.83 & 3.39 \\

         & TransparentGS
         & 14.78 & 11.16 & 0.68 & 0.32 & 0.28
         & 16.19 & 11.19 & 0.70 & 0.25 & 0.18
         & 15.72 & 9.91 & 0.73 & 0.24 & 0.20
         & 15.74 & 10.65 & 0.71 & 0.26 & 0.21 \\

         & \method (ours)
         & 21.27 & \underline{17.83} & 0.78 & 0.21 & \underline{0.11}
         & 19.69 & 14.62 & 0.73 & 0.24 & \underline{0.13}
         & 17.40 & 16.11 & 0.75 & 0.13 & \underline{0.06}
         & 19.45 & 16.19 & 0.76 & 0.19 & \underline{0.10} \\

         & Oracle$^\dagger$
         & \textbf{31.64} & \textbf{25.37} & \textbf{0.96} & \textbf{0.03} & \textbf{0.04}
         & \textbf{27.81} & \textbf{19.86} & \textbf{0.92} & \textbf{0.06} & \textbf{0.04}
         & \textbf{27.45} & \textbf{20.02} & \textbf{0.91} & \textbf{0.08} & \textbf{0.04}
         & \textbf{28.28} & \textbf{20.82} & \textbf{0.92} & \textbf{0.06} & \textbf{0.04} \\

        \midrule
        \multirow{12}{*}{\rotatebox{90}{\makecell{Sphere BG}}}
         & NeuS
         & 20.77 & 14.88 & 0.60 & 0.16 & 1.81
         & 20.72 & 13.21 & 0.62 & 0.13 & 1.97
         & 19.49 & 14.16 & 0.61 & 0.20 & 1.99
         & 20.15 & 13.97 & 0.61 & 0.16 & 1.93 \\

         & Splatfacto
         & 21.67 & 15.00 & \underline{0.88} & 0.16 & 33.41
         & 22.07 & 15.30 & 0.85 & 0.17 & 26.30
         & \underline{24.53} & \underline{17.95} & \underline{0.89} & \underline{0.11} & 19.93
         & 23.03 & 16.36 & \underline{0.87} & \underline{0.14} & 24.90 \\

         & Zip-NeRF
         & 18.12 & 14.85 & 0.76 & 0.39 & 2.20
         & 21.16 & 15.57 & 0.80 & 0.21 & 0.29
         & 22.32 & 16.28 & 0.85 & 0.17 & 2.11
         & 21.38 & 15.96 & 0.82 & 0.21 & 1.90 \\

         & TNSR
         & 18.62 & 11.66 & 0.85 & 0.17 & 1.62
         & 19.38 & 11.85 & 0.83 & 0.26 & 1.62
         & 17.69 & 10.61 & 0.80 & 0.21 & 2.20
         & 18.51 & 11.28 & 0.82 & 0.18 & 1.87 \\

         & MS-NeRF
         & 19.37 & 15.03 & 0.78 & 0.29 & 0.87
         & 21.27 & 15.73 & 0.80 & 0.20 & 0.93
         & 23.45 & 16.79 & 0.86 & 0.14 & 1.07
         & 21.71 & 15.99 & 0.82 & 0.20 & 0.93 \\

         & RayDef
         & 19.90 & 15.87 & 0.79 & 0.33 & 1.98
         & 20.81 & 15.77 & 0.77 & 0.26 & 1.78
         & 22.44 & 16.70 & 0.82 & 0.21 & 1.92
         & 21.29 & 16.17 & 0.79 & 0.25 & 1.88 \\

         & RoseNeRF
         & 18.33 & 12.32 & 0.73 & 0.25 & --
         & 18.52 & 11.32 & 0.70 & 0.26 & --
         & 17.10 & 13.61 & 0.65 & 0.38 & --
         & 17.98 & 12.42 & 0.69 & 0.30 & -- \\

         & NU-NeRF
         & 19.60 & 18.78 & 0.77 & 0.28 & 1.99
         & 20.83 & \underline{19.48} & 0.77 & 0.21 & 0.32
         & 21.73 & 16.09 & 0.83 & 0.19 & 0.35
         & 20.72 & \underline{18.12} & 0.79 & 0.23 & 0.89 \\

         & NeRRF
         & 12.08 & 12.73 & 0.46 & 0.78 & 1.91
         & 12.82 & 13.37 & 0.53 & 0.71 & 2.16
         & 12.44 & 12.30 & 0.55 & 0.79 & 2.16
         & 12.54 & 12.81 & 0.53 & 0.76 & 2.12 \\

         & TransparentGS
         & 15.03 & 10.69 & 0.78 & 0.28 & \underline{0.14}
         & 16.90 & 11.55 & 0.76 & 0.23 & \underline{0.22}
         & 16.75 & 10.10 & 0.81 & 0.19 & 0.24
         & 16.50 & 10.79 & 0.79 & 0.22 & 0.21 \\

         & \method (ours)
         & \underline{24.47} & \underline{20.26} & \underline{0.88} & \underline{0.15} & \textbf{0.12}
         & \underline{23.27} & 15.03 & \underline{0.88} & \underline{0.11} & 0.25
         & 21.61 & 16.14 & 0.83 & 0.20 & \underline{0.19}
         & \underline{23.12} & 17.14 & 0.86 & 0.15 & \underline{0.18} \\

         & Oracle$^\dagger$
         & \textbf{32.87} & \textbf{26.14} & \textbf{0.96} & \textbf{0.03} & 0.19
         & \textbf{28.69} & \textbf{20.39} & \textbf{0.93} & \textbf{0.05} & \textbf{0.15}
         & \textbf{27.76} & \textbf{20.05} & \textbf{0.92} & \textbf{0.08} & \textbf{0.14}
         & \textbf{29.05} & \textbf{21.28} & \textbf{0.93} & \textbf{0.06} & \textbf{0.15} \\

        \midrule
        \multirow{12}{*}{\rotatebox{90}{\makecell{Env-map BG}}}
         & NeuS
         & -- & -- & -- & -- & --
         & -- & -- & -- & -- & --
         & -- & -- & -- & -- & --
         & -- & -- & -- & -- & -- \\

         & Splatfacto
         & 11.88 & 13.05 & 0.62 & 0.71 & 1.17
         & 10.71 & 11.64 & 0.51 & 0.76 & 0.97
         & 10.60 & 10.55 & 0.54 & 0.76 & 1.13
         & 10.86 & 11.39 & 0.54 & 0.75 & 1.08 \\

         & Zip-NeRF
         & 23.43 & 19.89 & \underline{0.84} & 0.28 & 0.54
         & 21.81 & 17.33 & \underline{0.78} & 0.28 & 1.30
         & 22.96 & \underline{18.66} & 0.83 & 0.26 & 1.18
         & 22.73 & 18.63 & 0.82 & 0.27 & 1.01 \\

         & TNSR
         & -- & -- & -- & -- & --
         & -- & -- & -- & -- & --
         & -- & -- & -- & -- & --
         & -- & -- & -- & -- & -- \\

         & MS-NeRF
         & 19.44 & 17.16 & 0.74 & 0.40 & 5.25
         & 19.16 & 15.54 & 0.68 & 0.44 & 5.48
         & 19.32 & 15.84 & 0.71 & 0.43 & 4.06
         & 19.24 & 15.92 & 0.71 & 0.43 & 4.89 \\

         & RayDef
         & 23.19 & 19.07 & 0.72 & 0.49 & 0.19
         & 21.77 & 17.43 & 0.65 & 0.53 & 0.28
         & 21.73 & 17.96 & 0.65 & 0.51 & 0.43
         & 22.03 & 17.97 & 0.66 & 0.51 & 0.32 \\

         & RoseNeRF
         & 20.23 & 12.46 & 0.62 & 0.64 & --
         & 19.56 & 13.07 & 0.59 & 0.69 & --
         & 18.99 & 17.31 & 0.62 & 0.63 & --
         & 19.59 & 14.28 & 0.61 & 0.65 & -- \\

         & NU-NeRF
         & 23.68 & \textbf{27.65} & 0.80 & 0.28 & 0.07
         & 22.02 & \textbf{25.48} & 0.73 & 0.32 & 0.99
         & 20.77 & 18.35 & 0.73 & 0.38 & 1.28
         & 22.15 & \textbf{23.83} & 0.75 & 0.33 & 0.78 \\

         & NeRRF
         & 16.22 & 13.38 & 0.50 & 0.61 & 0.26
         & 17.67 & 13.46 & 0.61 & 0.45 & 0.25
         & 15.14 & 11.75 & 0.60 & 0.57 & 0.36
         & 16.35 & 12.73 & 0.59 & 0.53 & 0.30 \\

         & TransparentGS
         & 14.65 & 12.88 & 0.61 & 0.48 & 0.20
         & 16.23 & 14.01 & 0.59 & 0.45 & 0.21
         & 14.60 & 11.84 & 0.61 & 0.49 & 0.27
         & 15.26 & 12.89 & 0.60 & 0.47 & 0.23 \\

         & \method (ours)
         & \underline{24.34} & 20.49 & \underline{0.84} & \underline{0.18} & \underline{0.05}
         & \underline{22.32} & 19.32 & 0.77 & \underline{0.27} & \underline{0.16}
         & \underline{23.70} & 18.39 & \underline{0.85} & \underline{0.20} & \underline{0.12}
         & \underline{23.45} & 19.40 & \underline{0.83} & \underline{0.20} & \underline{0.11} \\

         & Oracle$^\dagger$
         & \textbf{33.48} & \underline{26.85} & \textbf{0.96} & \textbf{0.08} & \textbf{0.00}
         & \textbf{29.46} & \underline{21.24} & \textbf{0.92} & \textbf{0.11} & \textbf{0.00}
         & \textbf{30.82} & \textbf{23.61} & \textbf{0.92} & \textbf{0.13} & \textbf{0.00}
         & \textbf{30.69} & \underline{23.17} & \textbf{0.93} & \textbf{0.11} & \textbf{0.00} \\

        \midrule
        \multirow{9}{*}{\rotatebox{90}{\makecell{Real indoor}}}
         & NeuS
         & 19.08 & 16.24 & 0.65 & \underline{0.35} & --
         & 19.85 & 15.26 & 0.65 & 0.32 & --
         & 19.08 & 16.41 & 0.64 & 0.36 & --
         & 19.13 & 16.31 & 0.64 & 0.36 & -- \\

         & Splatfacto
         & 18.69 & 16.42 & \underline{0.75} & \textbf{0.32} & --
         & 19.87 & \underline{15.95} & \underline{0.76} & \textbf{0.25} & --
         & 18.79 & 16.51 & \underline{0.74} & \underline{0.34} & --
         & 18.85 & 16.46 & \underline{0.74} & \underline{0.33} & -- \\

         & Zip-NeRF
         & \textbf{20.14} & \textbf{17.03} & \textbf{0.77} & \textbf{0.32} & --
         & \textbf{21.47} & \textbf{16.49} & \textbf{0.77} & \textbf{0.25} & --
         & \textbf{20.21} & \textbf{17.31} & \textbf{0.75} & \textbf{0.33} & --
         & \textbf{20.28} & \textbf{17.22} & \textbf{0.75} & \textbf{0.32} & -- \\

         & TNSR
         & 15.88 & 13.04 & 0.56 & 0.50 & --
         & 17.25 & 13.47 & 0.58 & 0.43 & --
         & 15.37 & 13.39 & 0.54 & 0.53 & --
         & 15.57 & 13.35 & 0.54 & 0.52 & -- \\

         & MS-NeRF
         & 18.67 & 15.79 & 0.70 & 0.43 & --
         & \underline{20.08} & 15.76 & 0.69 & 0.37 & --
         & 19.09 & 16.30 & 0.69 & 0.42 & --
         & 19.10 & 16.20 & 0.69 & 0.42 & -- \\

         & RayDef
         & 18.77 & 16.54 & 0.69 & 0.46 & --
         & 19.65 & 15.70 & 0.65 & 0.44 & --
         & 18.86 & \underline{16.67} & 0.68 & 0.48 & --
         & 18.91 & \underline{16.58} & 0.68 & 0.47 & -- \\

         & NU-NeRF
         & 18.74 & 16.69 & 0.63 & 0.38 & --
         & 19.20 & 15.75 & 0.63 & 0.38 & --
         & 18.47 & 16.49 & 0.64 & 0.39 & --
         & 18.57 & 16.46 & 0.64 & 0.39 & -- \\

         & TransparentGS
         & 16.93 & 13.71 & 0.60 & 0.36 & --
         & 17.13 & 13.39 & 0.56 & \underline{0.31} & --
         & 15.57 & 12.50 & 0.55 & 0.43 & --
         & 15.86 & 12.72 & 0.56 & 0.41 & -- \\

         & \method (ours)
         & \underline{19.61} & \underline{16.83} & 0.72 & \underline{0.35} & --
         & 18.82 & 14.45 & 0.67 & 0.37 & --
         & \underline{19.19} & 16.25 & 0.70 & 0.36 & --
         & \underline{19.22} & 16.21 & 0.70 & 0.36 & -- \\

        \midrule
        \multirow{9}{*}{\rotatebox{90}{\makecell{Real outdoor}}}
         & NeuS
         & 19.25 & 15.55 & 0.59 & \underline{0.33} & --
         & 19.65 & 14.31 & 0.59 & \underline{0.32} & --
         & 18.39 & 15.34 & 0.58 & 0.35 & --
         & 18.59 & 15.30 & 0.58 & 0.35 & -- \\

         & Splatfacto
         & 13.77 & 14.48 & 0.61 & 0.51 & --
         & 12.73 & 12.84 & 0.58 & 0.55 & --
         & 13.58 & 14.13 & 0.59 & 0.51 & --
         & 13.55 & 14.09 & 0.59 & 0.52 & -- \\

         & Zip-NeRF
         & \textbf{21.00} & \textbf{16.85} & \textbf{0.78} & \textbf{0.31} & --
         & \textbf{21.18} & \textbf{16.03} & \textbf{0.76} & \textbf{0.29} & --
         & \textbf{20.38} & \textbf{16.87} & \textbf{0.76} & \textbf{0.32} & --
         & \textbf{20.52} & \textbf{16.81} & \textbf{0.76} & \textbf{0.31} & -- \\

         & TNSR
         & 18.28 & 13.62 & 0.69 & 0.34 & --
         & 17.22 & 12.65 & 0.65 & 0.38 & --
         & 16.36 & 12.94 & 0.62 & 0.43 & --
         & 16.67 & 13.01 & 0.63 & 0.42 & -- \\

         & MS-NeRF
         & \underline{20.15} & \underline{16.46} & \underline{0.75} & 0.39 & --
         & \underline{20.12} & 14.65 & \underline{0.72} & 0.39 & --
         & 19.46 & 15.64 & \underline{0.73} & 0.38 & --
         & \underline{19.60} & 15.68 & \underline{0.73} & 0.38 & -- \\

         & RayDef
         & 19.70 & 16.15 & 0.73 & 0.42 & --
         & 20.02 & \underline{15.41} & \underline{0.72} & 0.40 & --
         & 18.76 & 15.92 & 0.71 & 0.44 & --
         & 18.97 & 15.91 & 0.71 & 0.44 & -- \\

         & NU-NeRF
         & 19.18 & 16.37 & 0.67 & 0.38 & --
         & 19.50 & 15.21 & 0.65 & 0.41 & --
         & 18.28 & 15.97 & 0.65 & 0.37 & --
         & 18.48 & 15.97 & 0.65 & 0.37 & -- \\

         & TransparentGS
         & 17.08 & 13.09 & 0.62 & 0.36 & --
         & 16.62 & 11.94 & 0.60 & 0.36 & --
         & 15.67 & 11.81 & 0.60 & 0.38 & --
         & 15.92 & 11.99 & 0.60 & 0.38 & -- \\

         & \method (ours)
         & 19.84 & 16.07 & 0.72 & \underline{0.33} & --
         & 18.61 & 13.45 & 0.61 & 0.39 & --
         & \underline{19.58} & \underline{16.22} & 0.72 & \underline{0.33} & --
         & 19.58 & \underline{16.10} & 0.72 & \underline{0.33} & -- \\

        \bottomrule
        \end{tabular}%
    }
\end{table*}

\begin{table*}[!t]
    \begin{minipage}[t]{0.485\textwidth}
        \centering
        \caption{
        Comparison of untinted and tinted single-material objects.
        Results are averaged over five object pairs (ball, cube, cylinder, pyramid, and torus) that differ only in tint, rendered with the cube and sphere backgrounds.
        $\Delta$ denotes the change from untinted to tinted objects of the same shape.
        }
        \label{tab:tinted}
        \vspace{-6pt}
        \footnotesize
        \setlength{\tabcolsep}{2pt}
        \begin{tabular*}{\linewidth}{@{\extracolsep{\fill}}l cc cc cc@{}}
            \toprule
            & \multicolumn{2}{c}{Untinted} & \multicolumn{2}{c}{Tinted} & \multicolumn{2}{c}{$\Delta$} \\
            \cmidrule(lr){2-3}\cmidrule(lr){4-5}\cmidrule(lr){6-7}
            Method & \shortstack{\scriptsize PSNR\vphantom{$_{\mathrm{M}}$}\\[-1pt]$\uparrow$} & \shortstack{\scriptsize PSNR$_{\mathrm{M}}$\\[-1pt]$\uparrow$} & \shortstack{\scriptsize PSNR\vphantom{$_{\mathrm{M}}$}\\[-1pt]$\uparrow$} & \shortstack{\scriptsize PSNR$_{\mathrm{M}}$\\[-1pt]$\uparrow$} & \shortstack{\scriptsize PSNR\vphantom{$_{\mathrm{M}}$}\\[-1pt]\phantom{$\uparrow$}} & \shortstack{\scriptsize PSNR$_{\mathrm{M}}$\\[-1pt]\phantom{$\uparrow$}} \\
            \midrule
            Zip-NeRF & 17.51 & 13.83 & 20.33 & 15.78 & +2.82 & +1.95 \\
            MS-NeRF & 18.53 & 14.06 & 19.93 & 15.43 & +1.40 & +1.37 \\
            RayDef & 18.78 & 14.86 & 20.09 & 15.83 & +1.31 & +0.97 \\
            TransparentGS & 17.29 & 12.83 & 14.94 & 10.31 & $-$2.35 & $-$2.52 \\
            \method (ours) & 24.69 & 20.40 & 22.77 & 18.31 & $-$1.92 & $-$2.09 \\
            Oracle$^\dagger$ & 33.58 & 27.34 & 31.26 & 24.57 & $-$2.32 & $-$2.77 \\
            \bottomrule
        \end{tabular*}
    \end{minipage}\hfill
    \begin{minipage}[t]{0.485\textwidth}
        \centering
        \caption{
        Comparison of unbounded and bounded backgrounds.
        Bounded results are averaged over the cube and sphere backgrounds, and unbounded results are obtained with the environment-map backgrounds, for the same 50 objects.
        $\Delta$ denotes the change from unbounded to bounded backgrounds.
        }
        \label{tab:bounded}
        \vspace{-6pt}
        \footnotesize
        \setlength{\tabcolsep}{2pt}
        \begin{tabular*}{\linewidth}{@{\extracolsep{\fill}}l cc cc cc@{}}
            \toprule
            & \multicolumn{2}{c}{Unbounded} & \multicolumn{2}{c}{Bounded} & \multicolumn{2}{c}{$\Delta$} \\
            \cmidrule(lr){2-3}\cmidrule(lr){4-5}\cmidrule(lr){6-7}
            Method & \shortstack{\scriptsize PSNR\vphantom{$_{\mathrm{M}}$}\\[-1pt]$\uparrow$} & \shortstack{\scriptsize PSNR$_{\mathrm{M}}$\\[-1pt]$\uparrow$} & \shortstack{\scriptsize PSNR\vphantom{$_{\mathrm{M}}$}\\[-1pt]$\uparrow$} & \shortstack{\scriptsize PSNR$_{\mathrm{M}}$\\[-1pt]$\uparrow$} & \shortstack{\scriptsize PSNR\vphantom{$_{\mathrm{M}}$}\\[-1pt]\phantom{$\uparrow$}} & \shortstack{\scriptsize PSNR$_{\mathrm{M}}$\\[-1pt]\phantom{$\uparrow$}} \\
            \midrule
            Zip-NeRF & 22.73 & 18.63 & 20.89 & 15.70 & $-$1.84 & $-$2.93 \\
            MS-NeRF & 19.24 & 15.92 & 20.89 & 15.72 & +1.65 & $-$0.20 \\
            RayDef & 22.03 & 17.97 & 20.74 & 15.89 & $-$1.29 & $-$2.08 \\
            TransparentGS & 15.26 & 12.89 & 16.12 & 10.72 & +0.86 & $-$2.17 \\
            \method (ours) & 23.45 & 19.40 & 21.29 & 16.67 & $-$2.16 & $-$2.73 \\
            Oracle$^\dagger$ & 30.69 & 23.17 & 28.67 & 21.05 & $-$2.02 & $-$2.12 \\
            \bottomrule
        \end{tabular*}
    \end{minipage}
\end{table*}

For MS-NeRF \cite{msnerf} and RoseNeRF \cite{rosenerf}, both the proposal networks and the field optimizer utilize an Adam optimizer with an initial learning rate of $4 \times 10^{-3}$, $\epsilon = 1 \times 10^{-15}$, and an exponential decay scheduler that reduces the learning rate to $1 \times 10^{-4}$ over $2 \times 10^5$ steps.
For Zip-NeRF \cite{barron2023zipnerf}, the model optimizer utilizes a default configuration with an Adam optimizer set to an initial learning rate of $8 \times 10^{-3}$, $\epsilon = 1 \times 10^{-15}$, and an exponential decay scheduler reducing the learning rate to $1 \times 10^{-3}$ over $2.5 \times 10^4$ steps, with 1000 warm-up steps.
For Ray Deformation Network \cite{raydeform}, both the proposal networks and the field optimizer utilize an Adam optimizer with an initial learning rate of $2 \times 10^{-3}$, $\epsilon = 1 \times 10^{-15}$, and an exponential decay scheduler reducing the learning rate to $1 \times 10^{-4}$ over $1 \times 10^5$ steps.
For NeuS \cite{wang2021neus} and TNSR \cite{Deng:tnsr}, the proposal networks utilize a default configuration with an Adam optimizer set to an initial learning rate of $1 \times 10^{-2}$, $\epsilon = 1 \times 10^{-15}$, and a multi-step decay scheduler reducing the learning rate by a factor $\gamma = 0.33$ every milestone over $2 \times 10^4$ steps. The milestones are at 10000, 15000, and 18000 steps, respectively.
For the field optimizer, we use an initial learning rate of $5 \times 10^{-4}$, and a cosine decay scheduler with 500 warm-up steps and learning rate peak value set to $5 \times 10^{-2}$.
For NeRRF~\cite{nerrf}, the optimization is conducted in two stages. 
In the first stage, the model is optimized for 30 epochs with each epoch repeated 10 times. 
In the second stage, the model is optimized for 8 epochs with each epoch repeated 100 times. 
The gradient accumulation step is set to 1 in both stages.
For NU-NeRF~\cite{SunWYG24}, we adopt the default configuration.
The optimization runs for $2 \times 10^{5}$ iterations in stage 1 and $1.5 \times 10^{5}$ iterations in stage 2.
We also include near and far planes implementation, which are set to 0.1 and 7.5, respectively, and apply a scene scale factor of 0.5.
For TransparentGS~\cite{huang2025transparentgs}, the environment is first reconstructed with Splatfacto on background pixels for $2 \times 10^{4}$ iterations on synthetic scenes and $3 \times 10^{4}$ iterations on real scenes, with a larger initialization volume for sphere and environment-map backgrounds.
The transparent object is then reconstructed using the default pipeline with 64 light probes, $2 \times 10^{4}$ SDF iterations, and $7 \times 10^{3}$ mesh iterations.
We have also included a 3D Gaussian splatting approach, Splatfacto \cite{kerbl3Dgaussians}, which is implemented in Nerfstudio~\cite{nerfstudio}. For Splatfacto, we used the default Nerfstudio configurations.

\section{Extended Experimental Results}
\label{sec:experiment_supp}

In this section, we present additional comparative results on both synthetic and real scenes from our \dataset test set, along with an ablation study of the oracle method.

\paragraph{Additional results.} We present comparative qualitative results on synthetic scenes in~\cref{fig:synthetic_bounded_results} and~\cref{fig:synthetic_unbounded_results}, and show the results on real scenes in~\cref{fig:real_qualitative_results}.
For synthetic scenes, the patterned backgrounds are more challenging due to the presence of nearby complex textures, where most methods struggle, especially in handling multiple refractions and total internal reflection.
These methods are unable to reconstruct the highly detailed patterns, either blurry or incorrect.
In contrast, the oracle and R3F methods are better able to capture these complex light interactions.
The HDR environment map background is relatively easier, and most methods perform well on simple geometries.
However, their performance fluctuates significantly on more complex shapes.
For real scenes, it is obvious to see that all methods struggle a lot compared to the synthetic scenes, which reveals that this task is far from solved.
Moreover, we include the full table that compares several state-of-the-art methods in~\cref{tab:result_quantitative_breakdown}, and geometry visualization in~\cref{fig:mesh}.

\paragraph{Effect of tinting.}
\cref{tab:tinted} compares five pairs of objects that differ only in tint, rendered with the same cube and sphere backgrounds, so that geometry and background are controlled.
Tinting has opposite effects on the two groups of methods.
Methods that explicitly trace refracted rays through the object geometry, namely the oracle, \method, and TransparentGS, perform worse on tinted objects, with PSNR drops of 1.9--2.4\,dB, as they must additionally account for distance-dependent absorption along the refracted paths.
In contrast, Zip-NeRF, which does not model refraction, and MS-NeRF and Ray Deformation Networks, which model curved light paths only implicitly, improve by 1.3--2.8\,dB, likely because absorption attenuates the refracted background, leaving less high-frequency refracted detail to be reproduced without explicit ray tracing.
Despite the drop, the oracle and \method remain the best-performing methods on both untinted and tinted objects.

\begin{table}[!t]
    \centering
    \caption{
    Ablation study of oracle on convex single-material
    objects with cube backgrounds.
    Ablated components include distortion-loss correction,
    first surface reflection, total internal reflection, and
    Zip-NeRF \cite{barron2023zipnerf} backbone replaced by
    Nerfacto \cite{nerfstudio}.
    }
    \label{tab:ablation}
    \vspace{-6pt}
    \footnotesize
    \setlength{\tabcolsep}{0.6pt}

    \begin{tabularx}{\linewidth}{
        @{}
        >{\raggedright\arraybackslash}p{0.29\linewidth}
        *{5}{>{\centering\arraybackslash}X}
        @{}
    }
        \toprule
        Method
        &
        \shortstack{PSNR\vphantom{PSNR$_{\mathrm{M}}$$\uparrow$}\\[-1pt]$\uparrow$}
        &
        \shortstack{PSNR$_{\mathrm{M}}$\vphantom{PSNR$_{\mathrm{M}}$$\uparrow$}\\[-1pt]$\uparrow$}
        &
        \shortstack{SSIM$\uparrow$\vphantom{PSNR$_{\mathrm{M}}$$\uparrow$}\\[-1pt]\scriptsize$\times 10^{2}$}
        &
        \shortstack{LPIPS$\downarrow$\vphantom{PSNR$_{\mathrm{M}}$$\uparrow$}\\[-1pt]\scriptsize$\times 10^{2}$}
        &
        \shortstack{DMAE$\downarrow$\vphantom{PSNR$_{\mathrm{M}}$$\uparrow$}\\[-1pt]\scriptsize$\times 10^{2}$}
        \\
        \midrule

        Oracle
        & \textbf{31.64}
        & \textbf{25.37}
        & \textbf{96.13}
        & \textbf{2.76}
        & 4.38
        \\

        w/o corrected $\gL_{\mathrm{dist}}$
        & 31.59
        & 25.31
        & 96.12
        & 2.78
        & 4.07
        \\

        w/o \refl
        & 29.65
        & 22.92
        & 94.69
        & 3.67
        & \textbf{2.97}
        \\

        w/o TIR
        & 25.85
        & 19.27
        & 90.84
        & 8.38
        & 15.86
        \\

        w/o Zip-NeRF
        & 24.68
        & 21.45
        & 87.79
        & 11.63
        & 27.42
        \\

        \bottomrule
    \end{tabularx}
\end{table}

\paragraph{Effect of bounded backgrounds.}
\cref{tab:bounded} compares the same 50 objects rendered with bounded (cube and sphere) and unbounded (environment-map) backgrounds.
Bounded backgrounds reduce the masked PSNR of all methods by 0.2--2.9\,dB, since the refracted appearance then depends on both the origin and the direction of each outgoing ray rather than on its direction alone.
The full-image PSNR also drops by 1.3--2.2\,dB for Zip-NeRF, Ray Deformation Networks, \method, and the oracle.
MS-NeRF and TransparentGS gain slightly in full-image PSNR, but their masked PSNR still decreases, indicating that these gains come from the background rather than from the refractive object.
Even the oracle, which has access to the ground-truth geometry, loses about 2\,dB, confirming that bounded backgrounds are substantially more challenging than infinitely distant ones.

\paragraph{Failure case analysis.}
On synthetic scenes, most methods struggle when light undergoes multiple refractions, total internal reflection, or absorption in tinted materials, which break the assumptions of most methods (\cref{fig:synthetic_bounded_results,fig:synthetic_unbounded_results}).
Even simple shapes are not reconstructed accurately.
The photometric loss is ambiguous, since distorted geometry, floaters, or view-dependent color can equally explain the training views, and it is highly non-convex for methods that trace refracted rays, as small normal changes redirect rays to far-away background locations, so the optimization easily converges to local optima.
The reconstructed geometry in \cref{fig:mesh} shows that most methods severely deform the object shape to compensate for appearance errors and largely fail to recover internal structures.

Real captures are less controlled, with surface imperfections, camera reflections, sensor noise, and camera pose errors, making the task even more challenging: all methods reach only 13.6--20.5\,dB PSNR (\cref{tab:result_quantitative_breakdown}), and almost none recovers the refracted patterns correctly as shown in \cref{fig:real_qualitative_results}.
Even methods that explicitly model refraction fail to reconstruct the refracted patterns, as they rely heavily on the estimated object shape, and small errors in geometry, masks, or poses cause large deviations in the traced light paths.
This is why Zip-NeRF achieves the highest scores on real scenes: it models refraction only through view-dependent appearance, producing blurry colors that resemble the ground truth on average but, on close inspection, are inaccurate and contain floaters around the refractive object.

\paragraph{Ablation study.}
An ablation study is presented in \cref{tab:ablation}, comparing the full oracle method with four ablated versions: without the corrected distortion loss, first surface reflection, total internal reflection (TIR), and replacing the Zip-NeRF \cite{barron2023zipnerf} backbone with Nerfacto \cite{nerfstudio}.
The full oracle method accurately models both reflections and TIR, producing highly detailed renderings and the best image quality across all metrics.
Removing the corrected distortion loss causes only a marginal drop in image quality.
Omitting the first surface reflection leads to the loss of subtle reflective details and lower image quality, although it yields a lower DMAE, while disabling TIR results in missing critical light interactions within refractive objects, and replacing the backbone causes the largest performance drop.
These results highlight the contribution of each component to rendering quality, with reflection, TIR, and the Zip-NeRF backbone being the most important.

\begin{figure*}[!t]
\centering

\setlength{\tabcolsep}{0.35pt}
\renewcommand{\arraystretch}{0}

\newcommand{\comparisonimage}[1]{%
    \includegraphics[width=0.098\linewidth]{#1}%
}

\newcommand{\rowlabel}[1]{%
    \rotatebox{90}{\makebox[\dimexpr0.098\linewidth\relax][c]{%
        \fontsize{5.5pt}{6pt}\selectfont #1%
    }}%
}

\newcommand{\rgbrow}[9]{%
    \rowlabel{RGB} &
    \comparisonimage{#1} &
    \comparisonimage{#2} &
    \comparisonimage{#3} &
    \comparisonimage{#4} &
    \comparisonimage{#5} &
    \comparisonimage{#6} &
    \comparisonimage{#7} &
    \comparisonimage{#8} &
    \comparisonimage{#9}
    \\[0.3pt]
}

\newcommand{\distrow}[9]{%
    \rowlabel{Distance Map} &
    \comparisonimage{#1} &
    \comparisonimage{#2} &
    \comparisonimage{#3} &
    \comparisonimage{#4} &
    \comparisonimage{#5} &
    \comparisonimage{#6} &
    \comparisonimage{#7} &
    \comparisonimage{#8} &
    \comparisonimage{#9}
    \\[0.3pt]
}

\begin{tabular}{@{}c@{\hspace{1.2pt}}*{9}{c}@{}}

&
{\fontsize{6.5pt}{7.5pt}\selectfont Ground Truth} &
{\fontsize{6.5pt}{7.5pt}\selectfont Splatfacto} &
{\fontsize{6.5pt}{7.5pt}\selectfont TransparentGS} &
{\fontsize{6.5pt}{7.5pt}\selectfont Zip-NeRF} &
{\fontsize{6.5pt}{7.5pt}\selectfont MS-NeRF} &
{\fontsize{6.5pt}{7.5pt}\selectfont Ray Deformation} &
{\fontsize{6.5pt}{7.5pt}\selectfont NU-NeRF} &
{\fontsize{6.5pt}{7.5pt}\selectfont \method (ours)} &
{\fontsize{6.5pt}{7.5pt}\selectfont Oracle (ours)}
\\[2pt]

\rgbrow
    {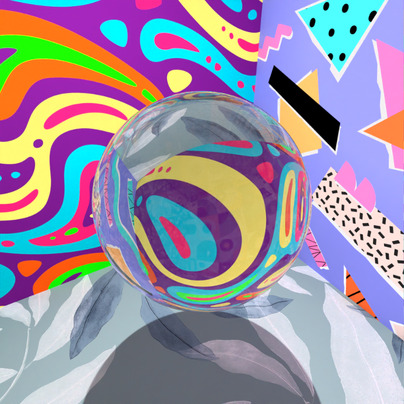}
    {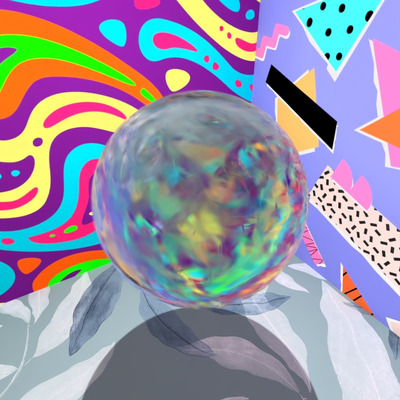}
    {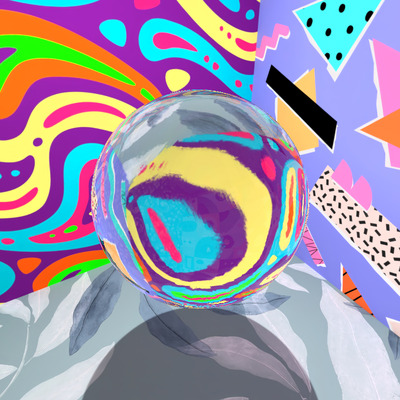}
    {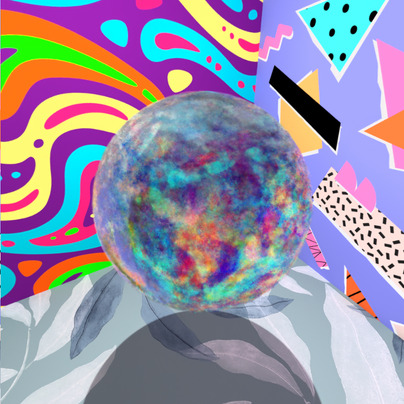}
    {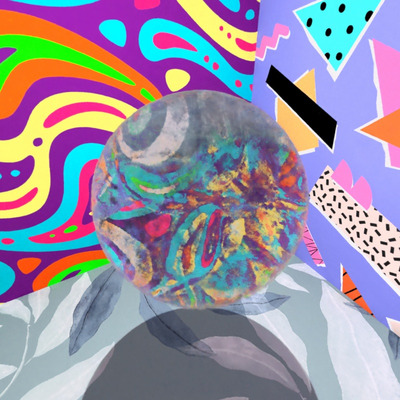}
    {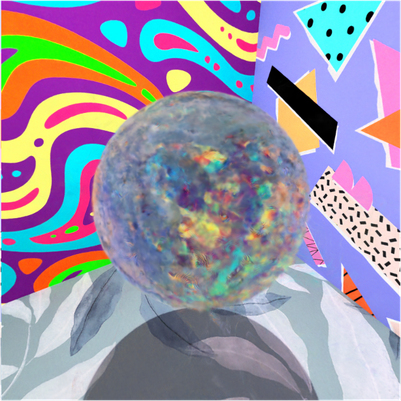}
    {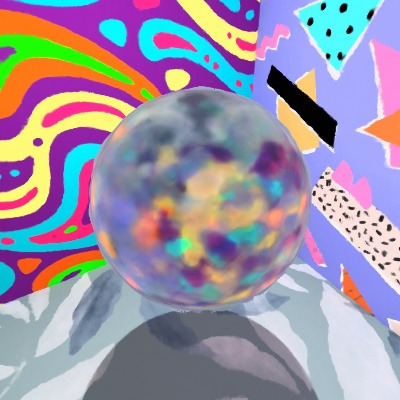}
    {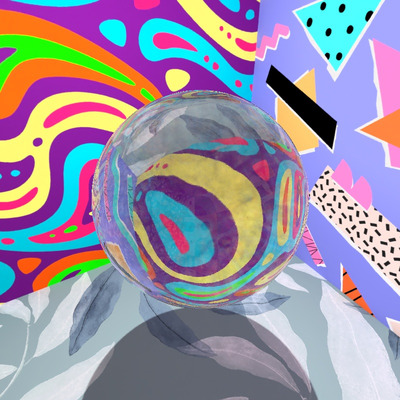}
    {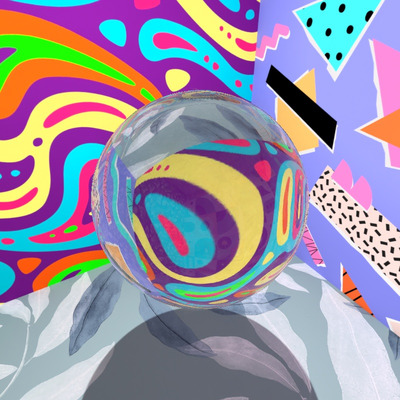}
\distrow
    {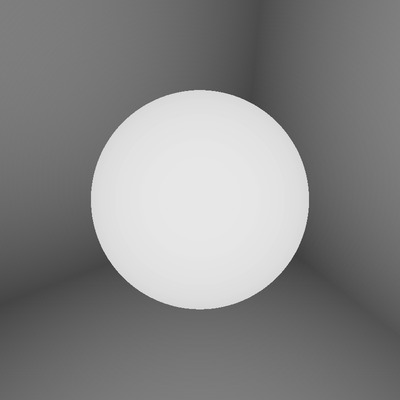}
    {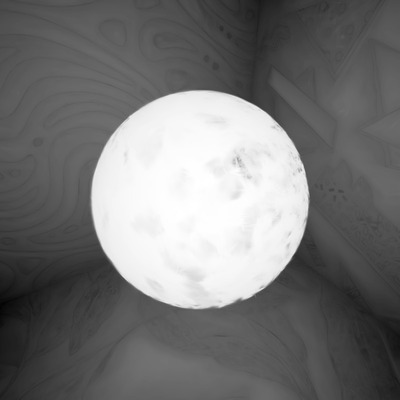}
    {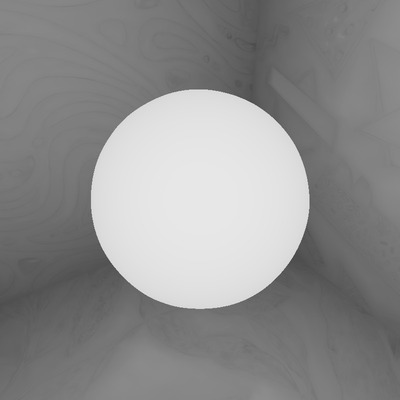}
    {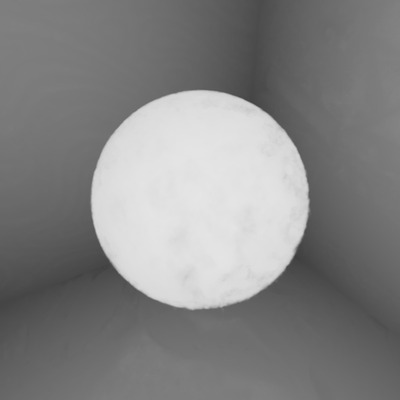}
    {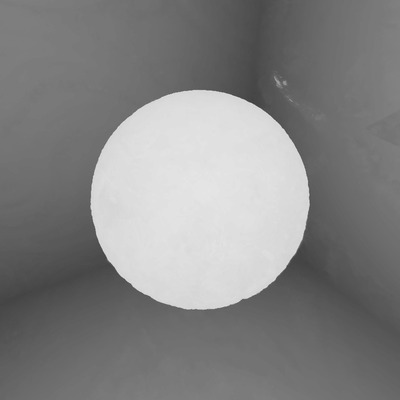}
    {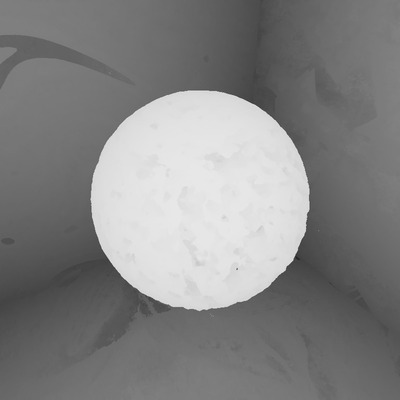}
    {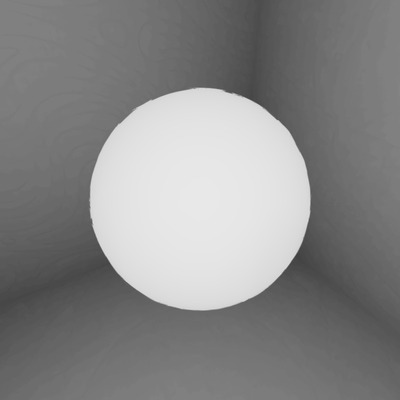}
    {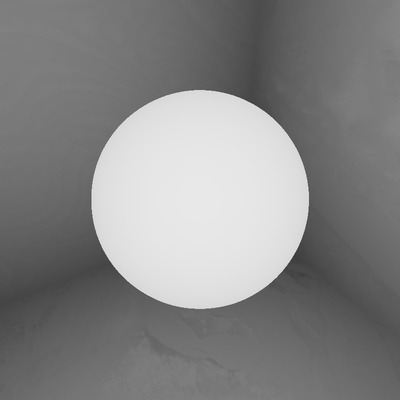}
    {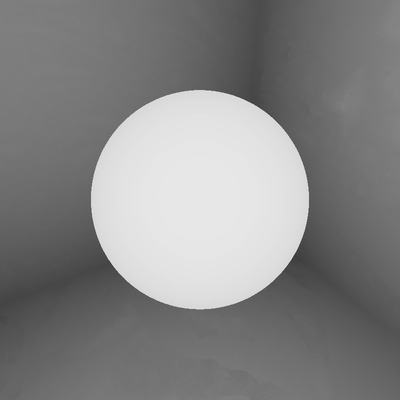}

\rgbrow
    {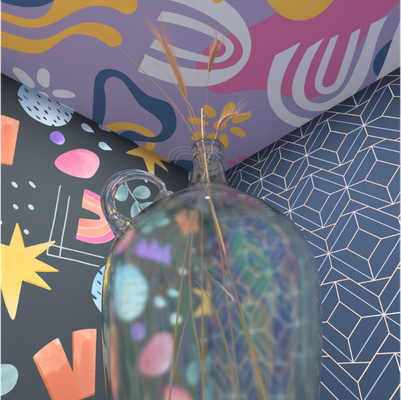}
    {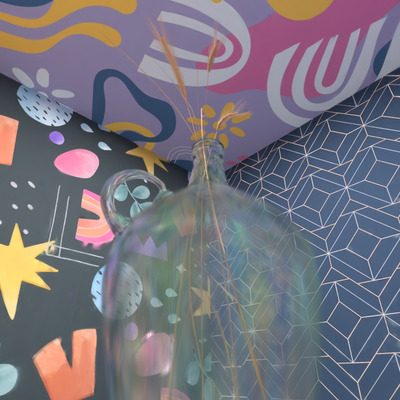}
    {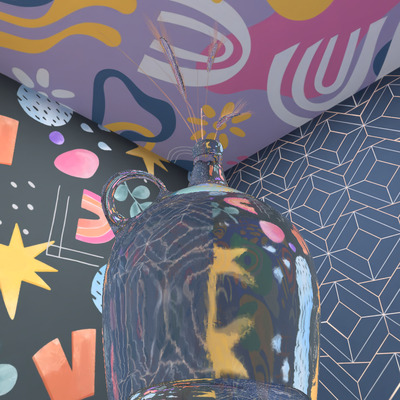}
    {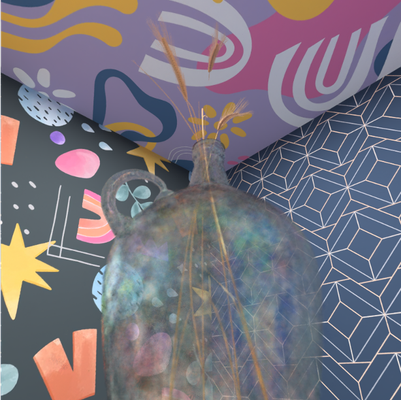}
    {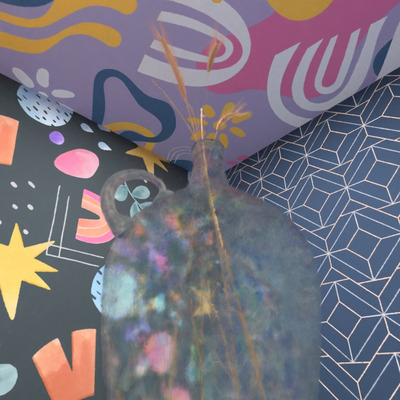}
    {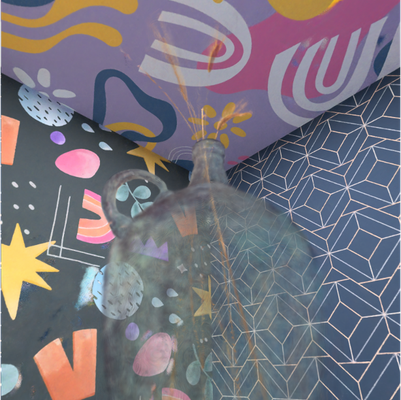}
    {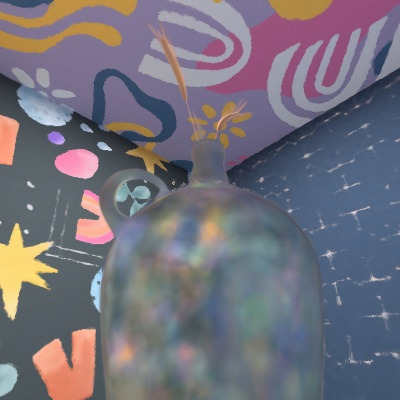}
    {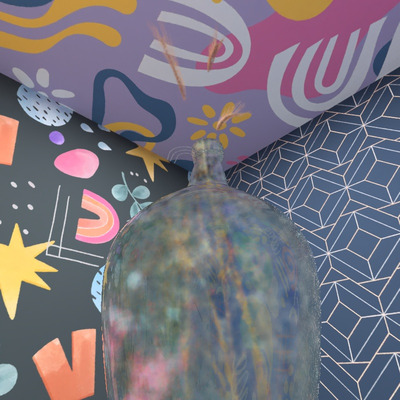}
    {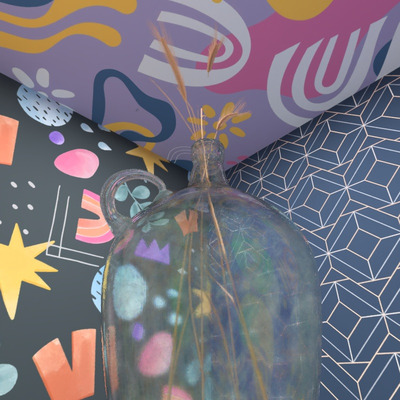}
\distrow
    {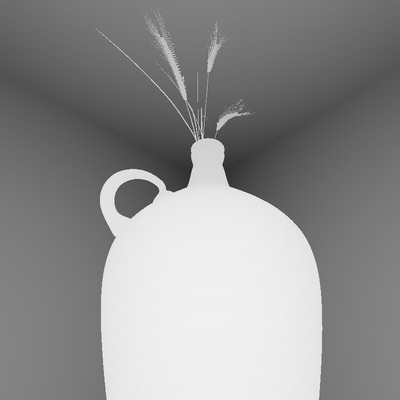}
    {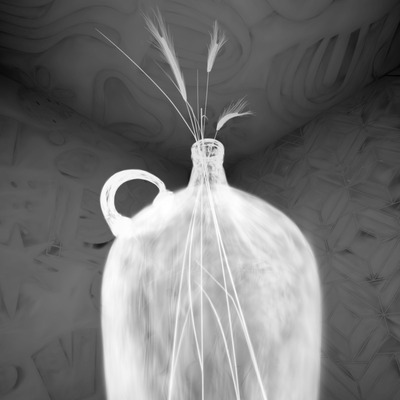}
    {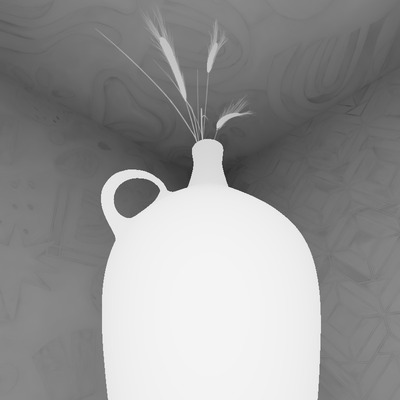}
    {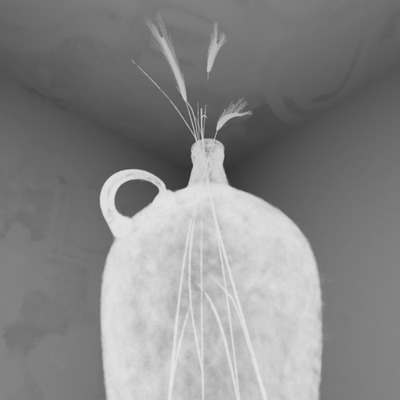}
    {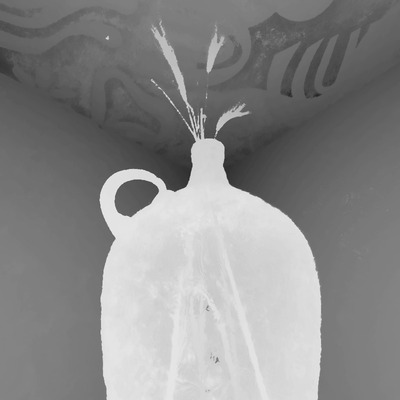}
    {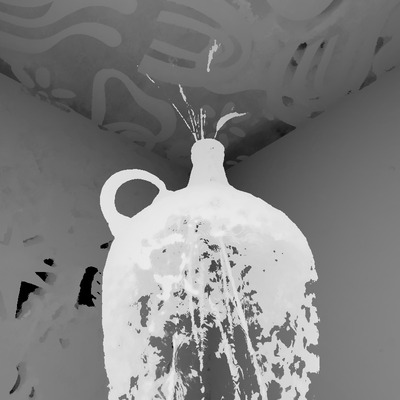}
    {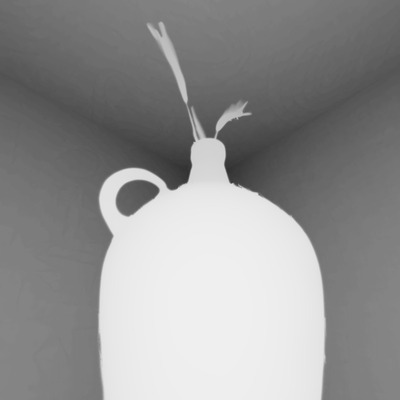}
    {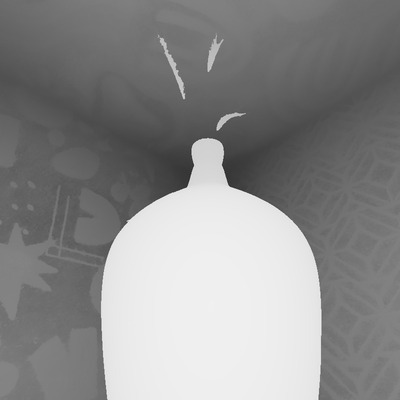}
    {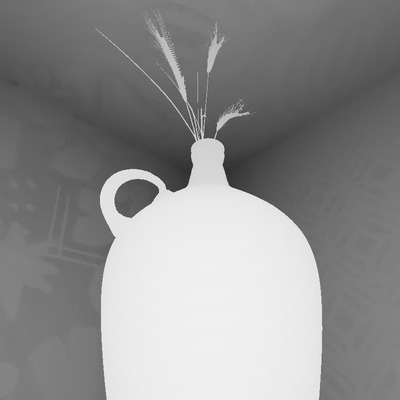}

\rgbrow
    {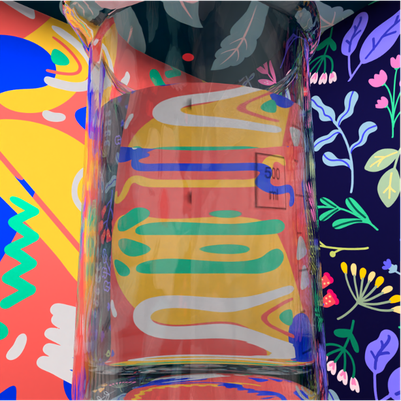}
    {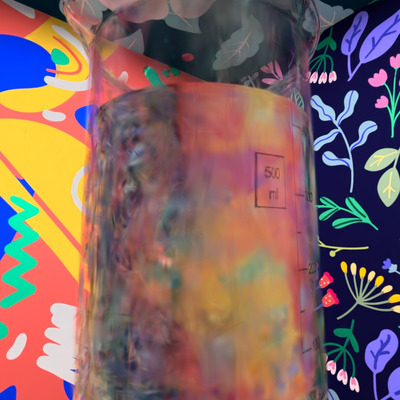}
    {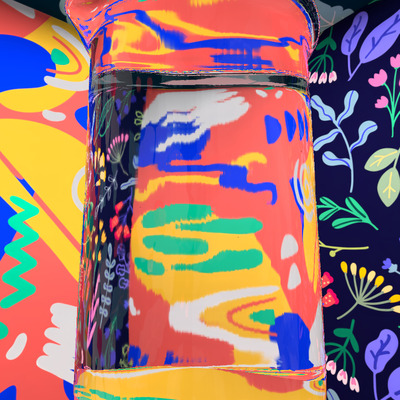}
    {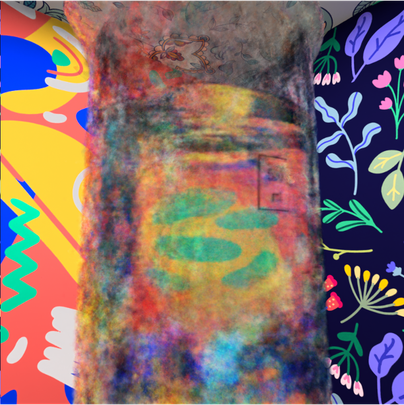}
    {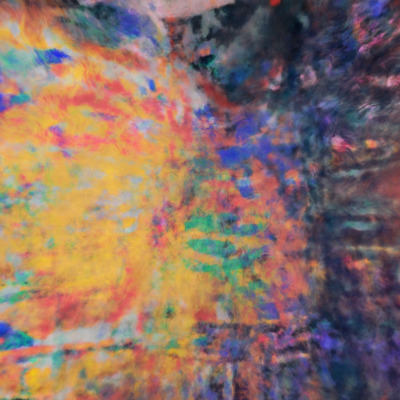}
    {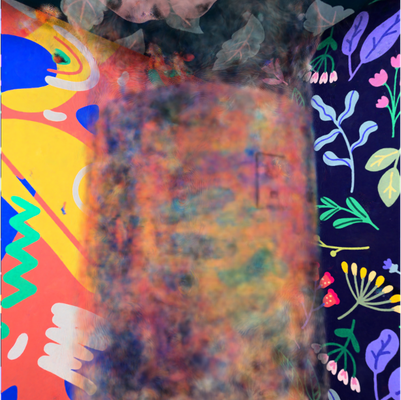}
    {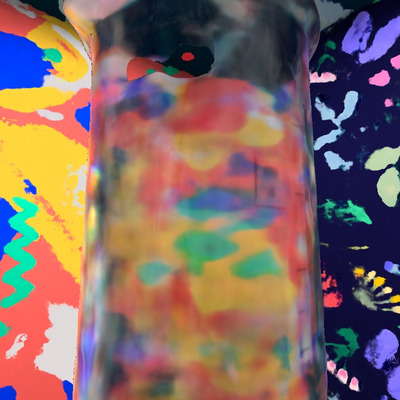}
    {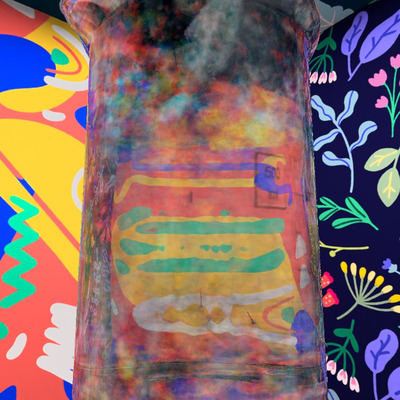}
    {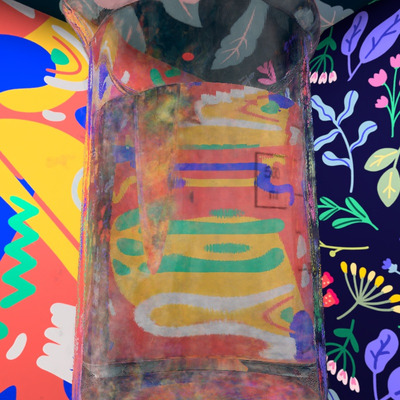}
\distrow
    {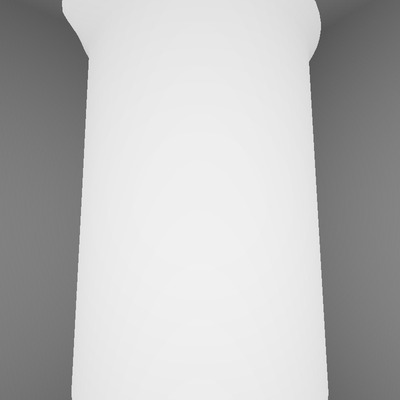}
    {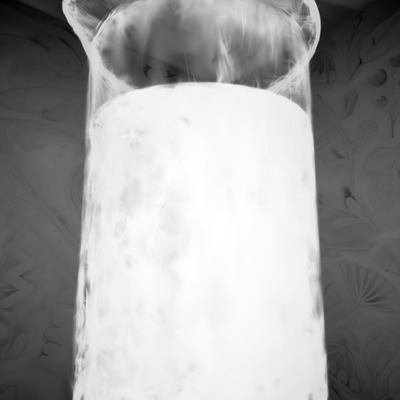}
    {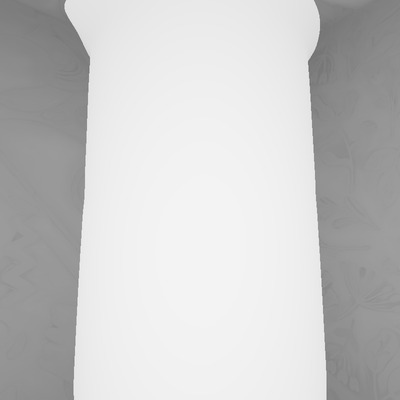}
    {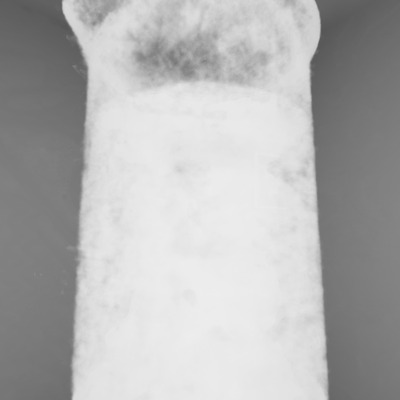}
    {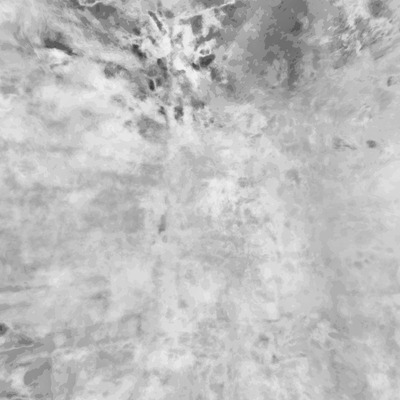}
    {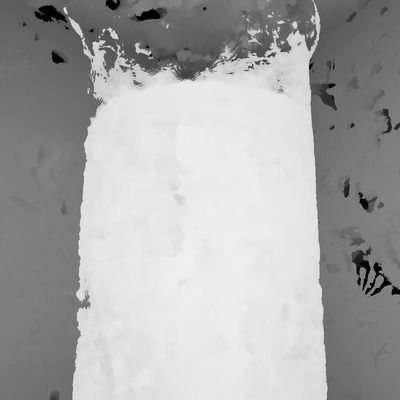}
    {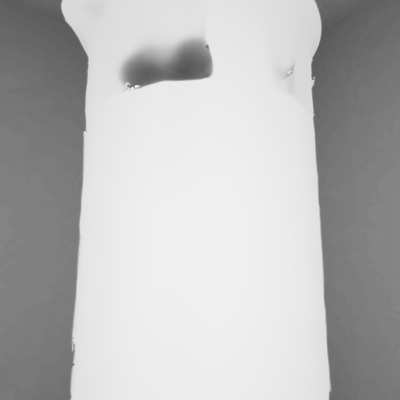}
    {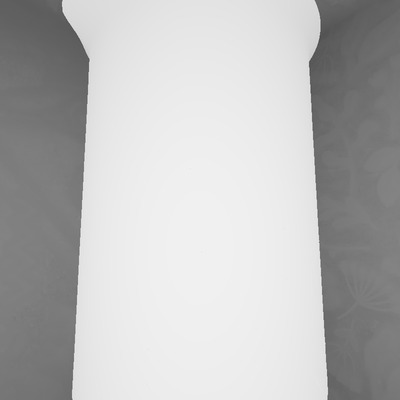}
    {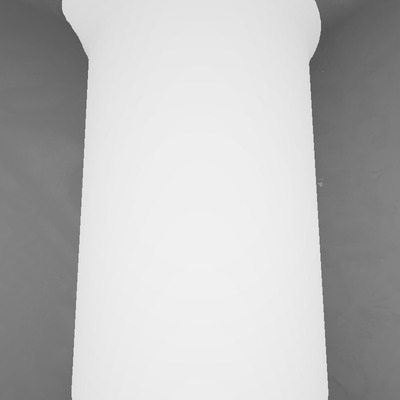}

\rgbrow
    {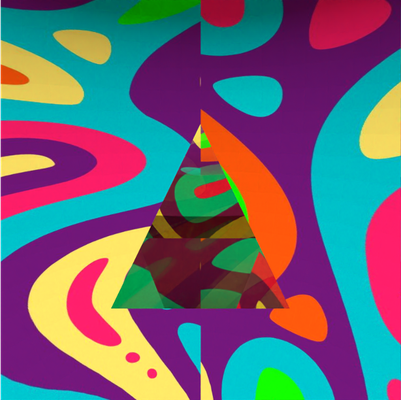}
    {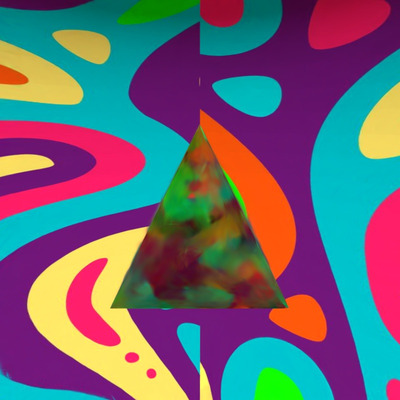}
    {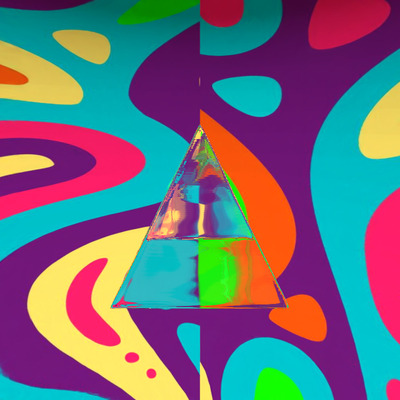}
    {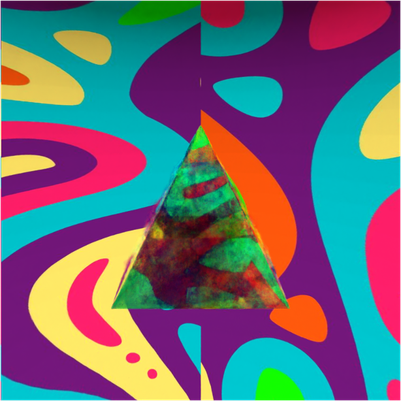}
    {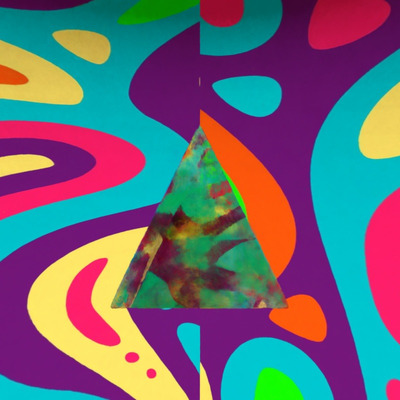}
    {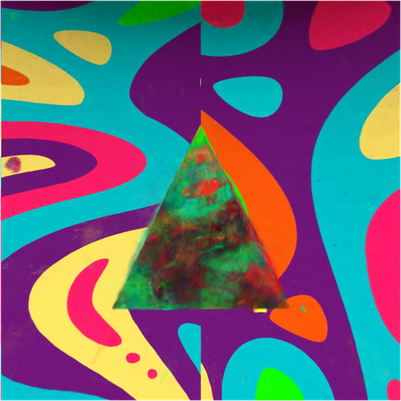}
    {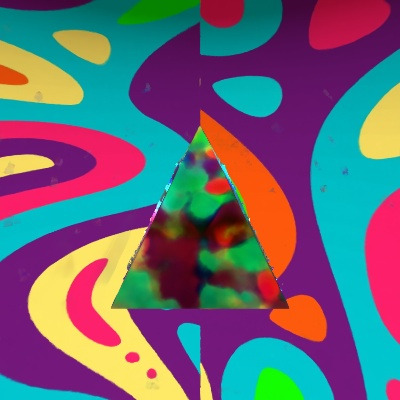}
    {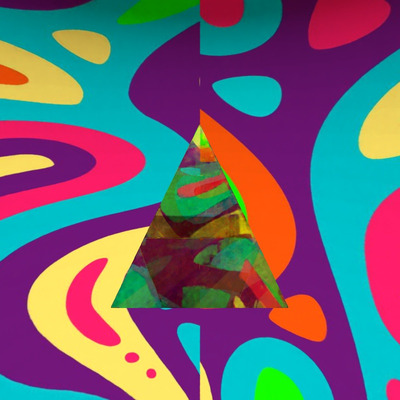}
    {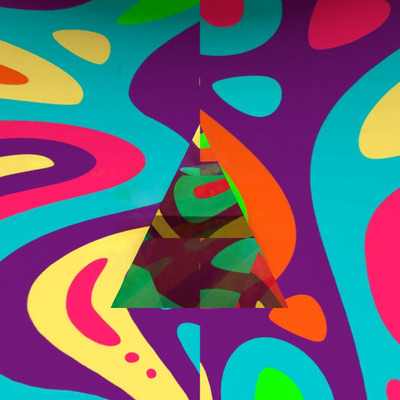}
\distrow
    {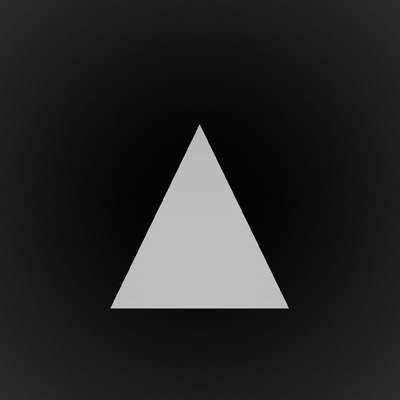}
    {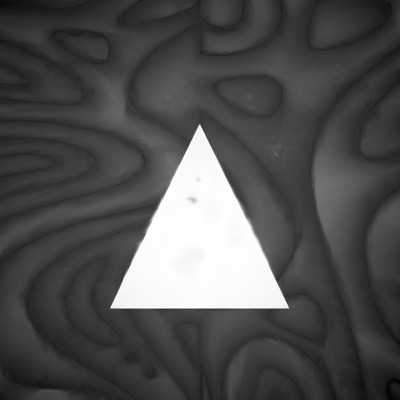}
    {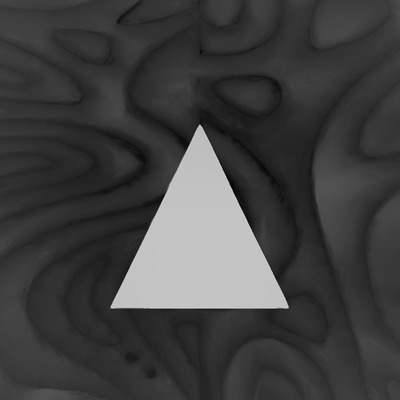}
    {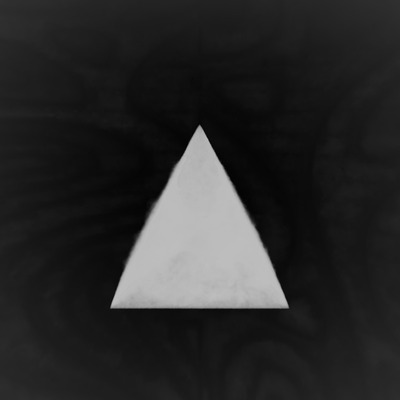}
    {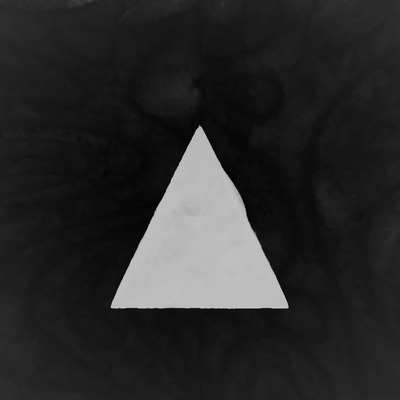}
    {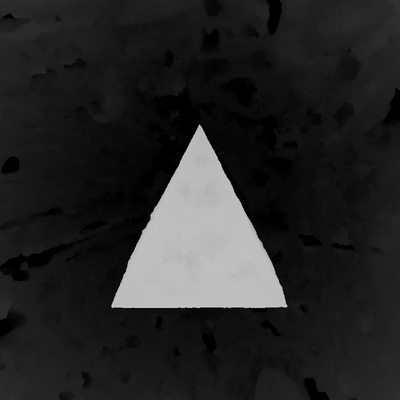}
    {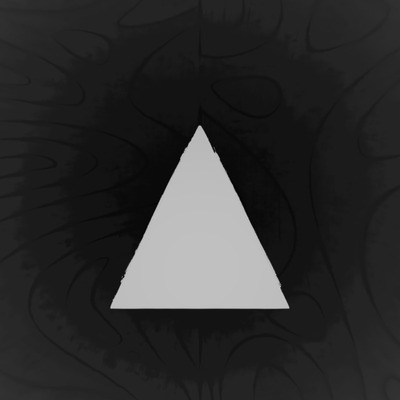}
    {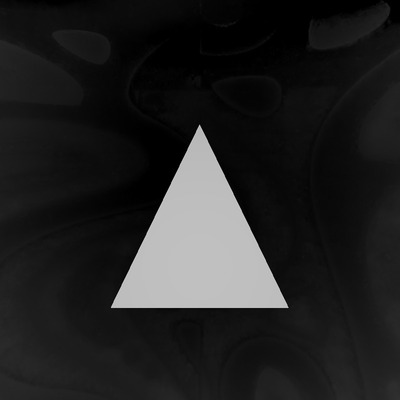}
    {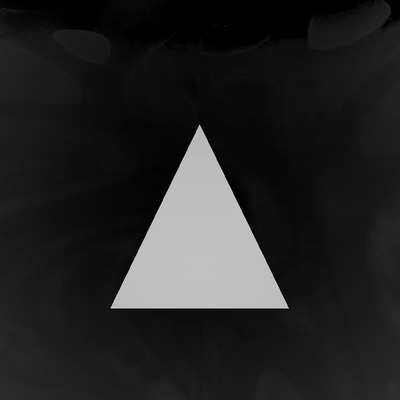}

\rgbrow
    {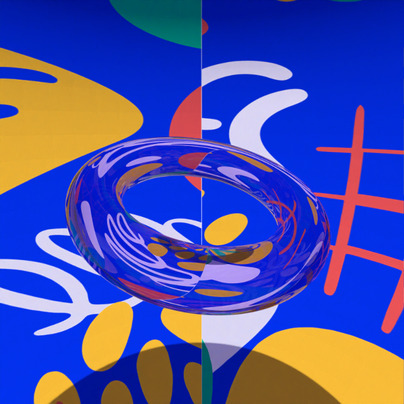}
    {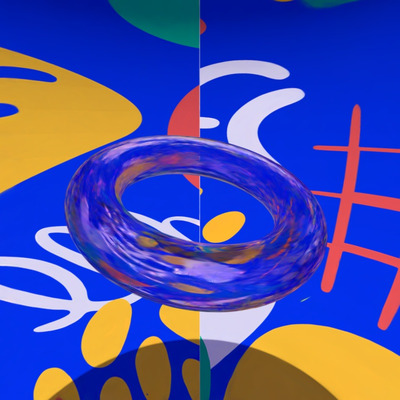}
    {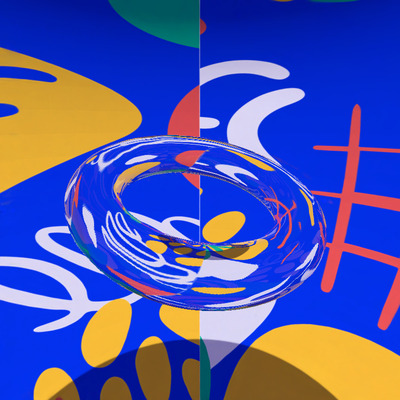}
    {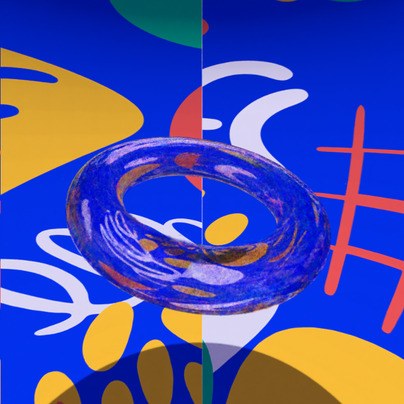}
    {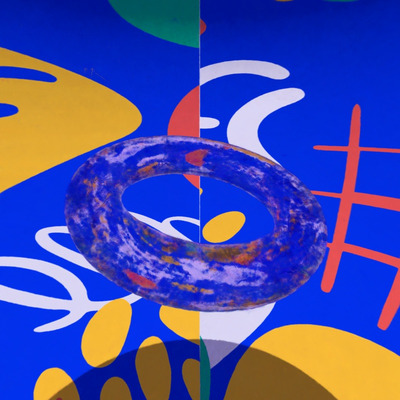}
    {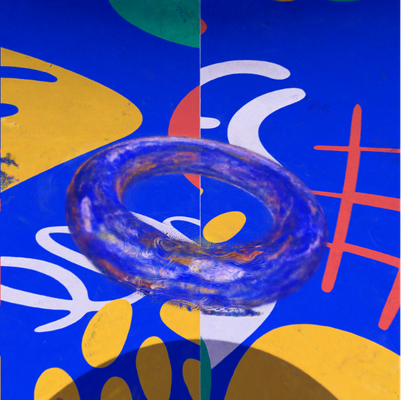}
    {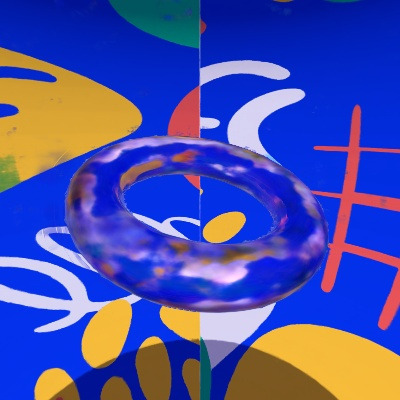}
    {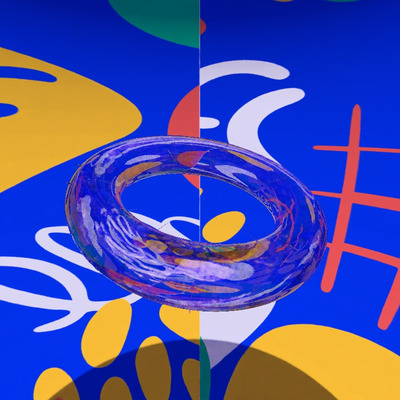}
    {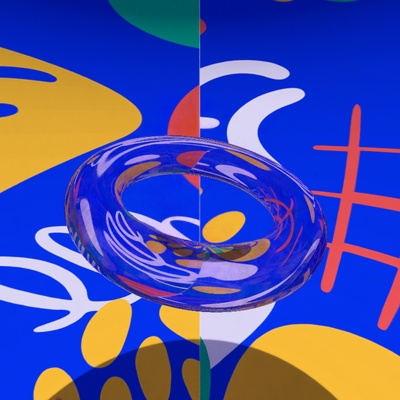}
\distrow
    {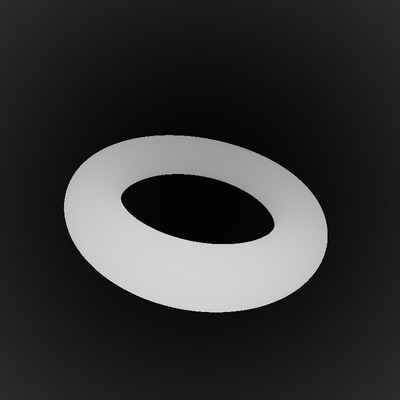}
    {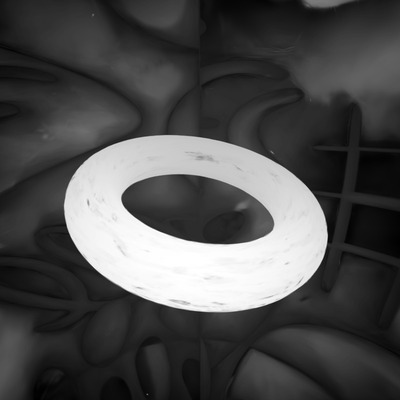}
    {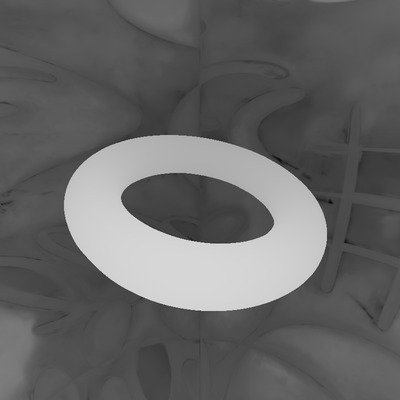}
    {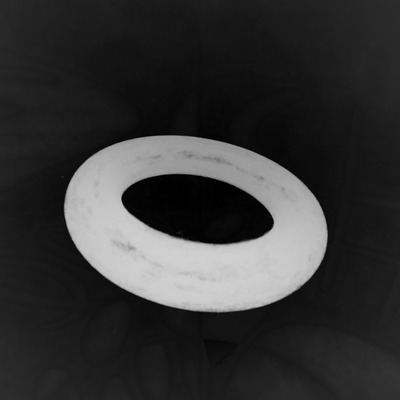}
    {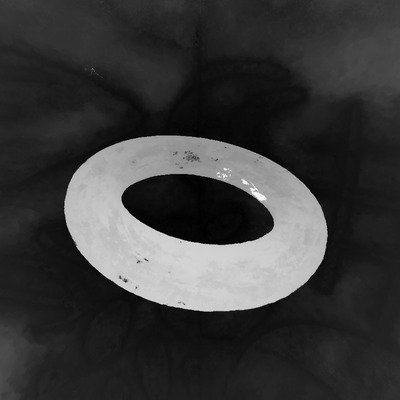}
    {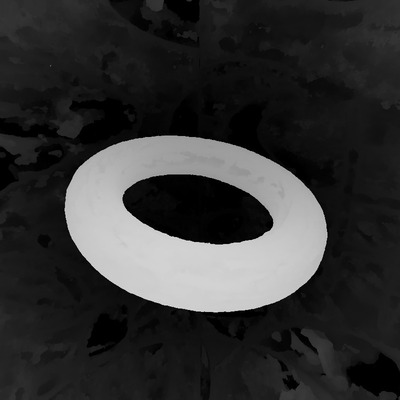}
    {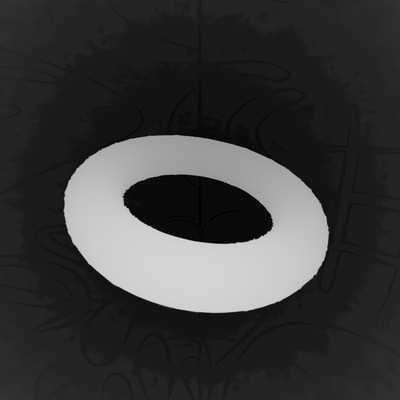}
    {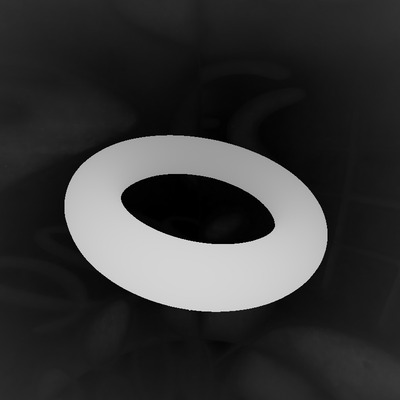}
    {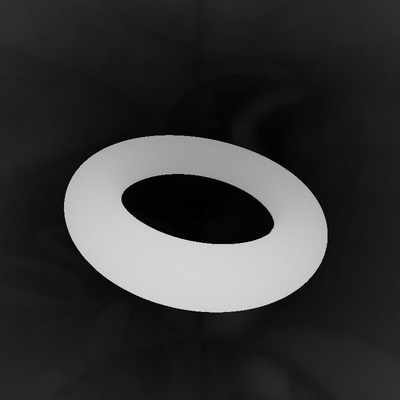}

\rgbrow
    {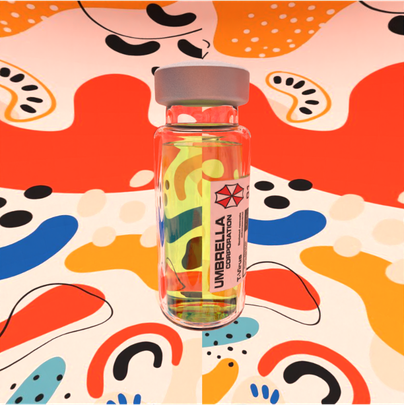}
    {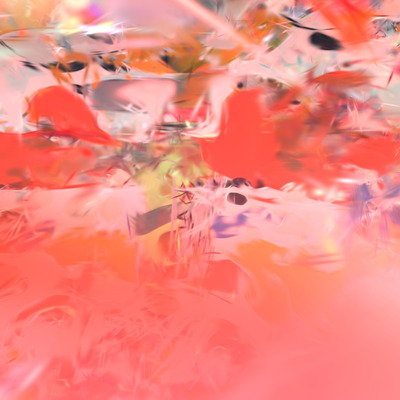}
    {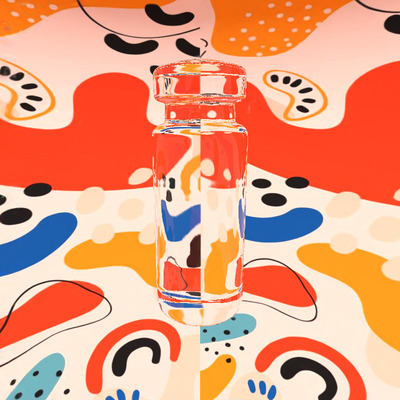}
    {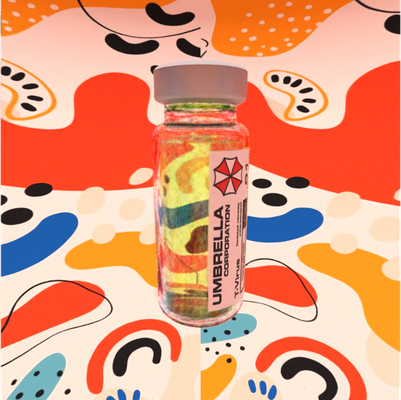}
    {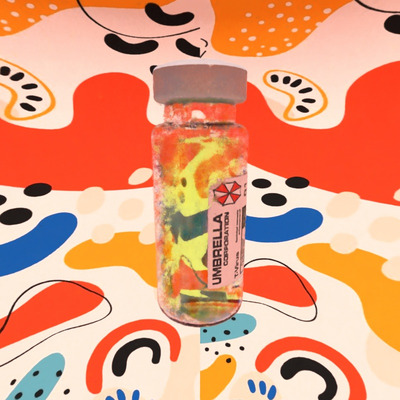}
    {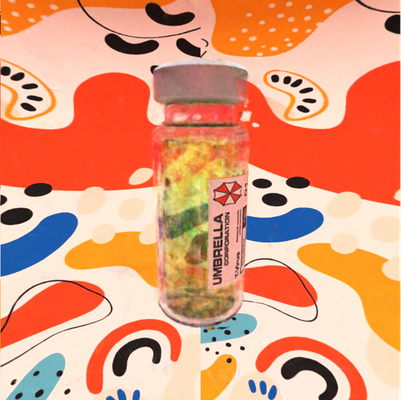}
    {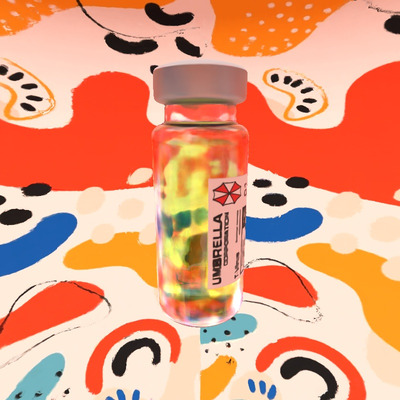}
    {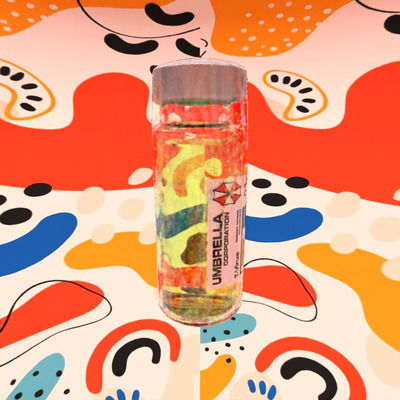}
    {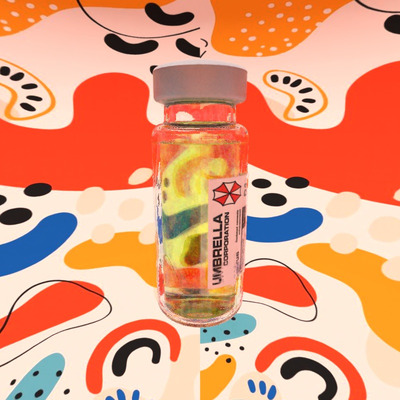}
\distrow
    {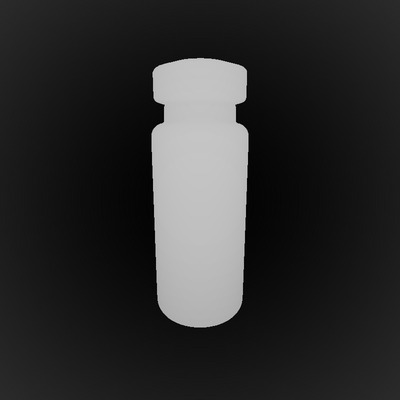}
    {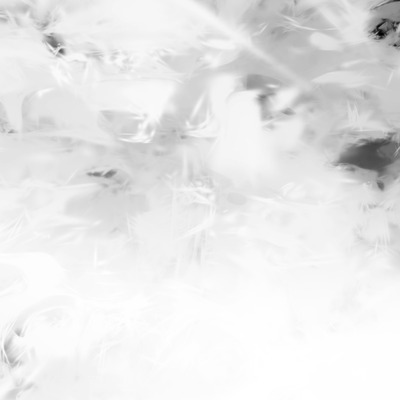}
    {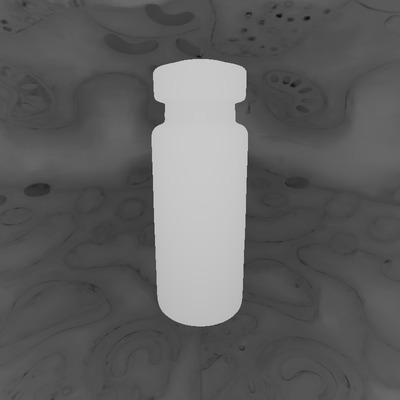}
    {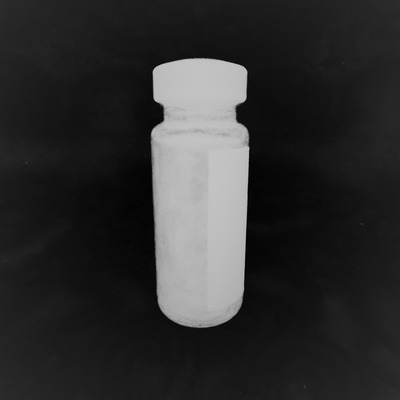}
    {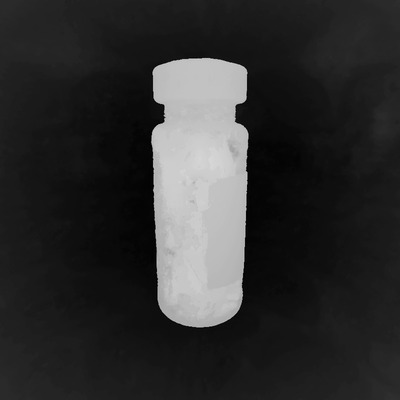}
    {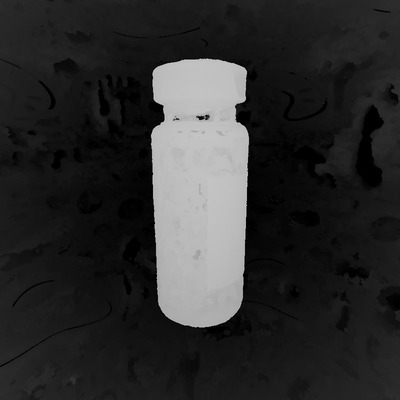}
    {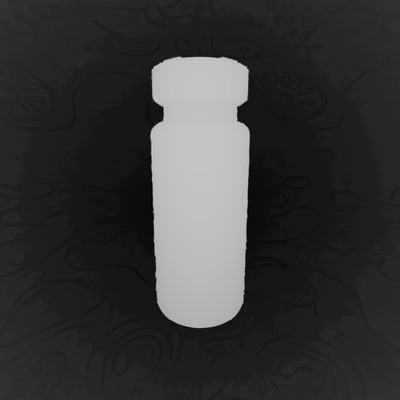}
    {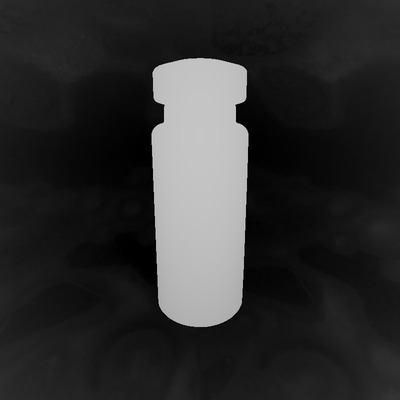}
    {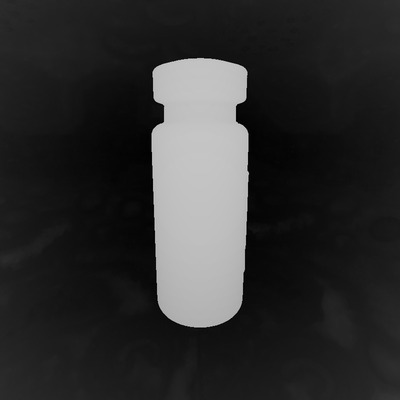}

\end{tabular}

\caption{
Additional qualitative comparison of novel view synthesis results on bounded
synthetic scenes.
The object is enclosed by a cube or a sphere textured with a
high-frequency pattern image. Each block shows one scene: the top row gives the
rendered RGB images and the bottom row the corresponding distance maps, which
expose geometric errors not apparent from RGB alone.
}
\label{fig:synthetic_bounded_results}
\end{figure*}

\begin{figure*}[!t]
\centering

\setlength{\tabcolsep}{0.35pt}
\renewcommand{\arraystretch}{0}

\newcommand{\comparisonimage}[1]{%
    \includegraphics[width=0.098\linewidth]{#1}%
}

\newcommand{\rowlabel}[1]{%
    \rotatebox{90}{\makebox[\dimexpr0.098\linewidth\relax][c]{%
        \fontsize{5.5pt}{6pt}\selectfont #1%
    }}%
}

\newcommand{\missingcell}{%
    \makebox[0.098\linewidth][c]{%
        \textcolor{black!35}{%
            \rotatebox{45}{\rule{0.1386\linewidth}{0.4pt}}%
        }%
    }%
}

\newcommand{\rgbrow}[9]{%
    \rowlabel{RGB} &
    \comparisonimage{#1} &
    \comparisonimage{#2} &
    \comparisonimage{#3} &
    \comparisonimage{#4} &
    \comparisonimage{#5} &
    \comparisonimage{#6} &
    \comparisonimage{#7} &
    \comparisonimage{#8} &
    \comparisonimage{#9}
    \\[0.3pt]
}

\newcommand{\distrow}[9]{%
    \rowlabel{Distance Map} &
    \comparisonimage{#1} &
    \comparisonimage{#2} &
    \comparisonimage{#3} &
    \comparisonimage{#4} &
    \comparisonimage{#5} &
    \comparisonimage{#6} &
    \comparisonimage{#7} &
    \comparisonimage{#8} &
    \comparisonimage{#9}
    \\[0.3pt]
}

\begin{tabular}{@{}c@{\hspace{1.2pt}}*{9}{c}@{}}

&
{\fontsize{6.5pt}{7.5pt}\selectfont Ground Truth} &
{\fontsize{6.5pt}{7.5pt}\selectfont Splatfacto} &
{\fontsize{6.5pt}{7.5pt}\selectfont TransparentGS} &
{\fontsize{6.5pt}{7.5pt}\selectfont Zip-NeRF} &
{\fontsize{6.5pt}{7.5pt}\selectfont MS-NeRF} &
{\fontsize{6.5pt}{7.5pt}\selectfont NeRRF} &
{\fontsize{6.5pt}{7.5pt}\selectfont NU-NeRF} &
{\fontsize{6.5pt}{7.5pt}\selectfont \method (ours)} &
{\fontsize{6.5pt}{7.5pt}\selectfont Oracle (ours)}
\\[2pt]

\rgbrow
    {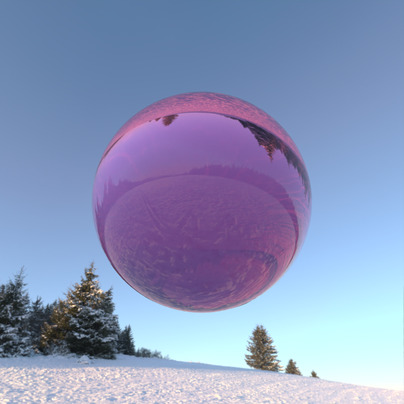}
    {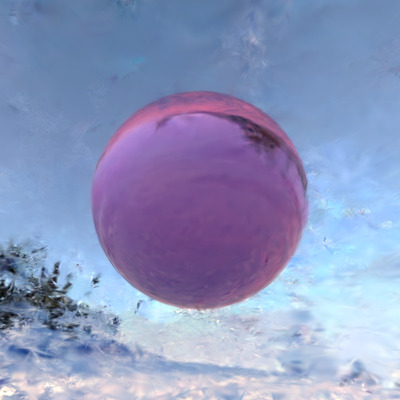}
    {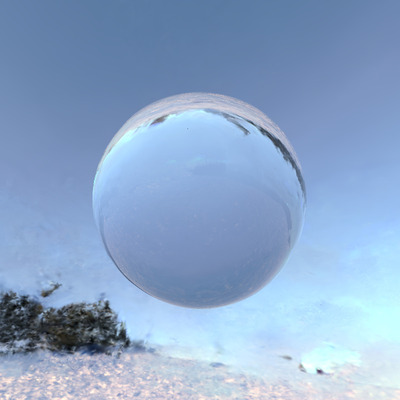}
    {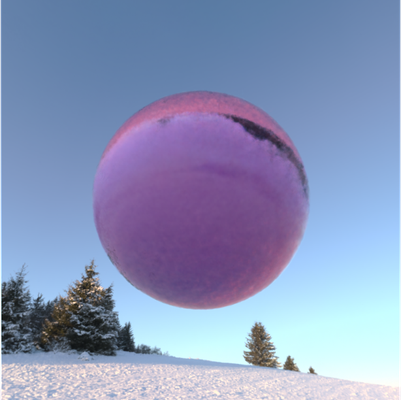}
    {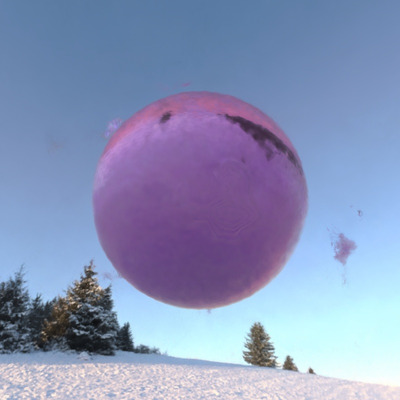}
    {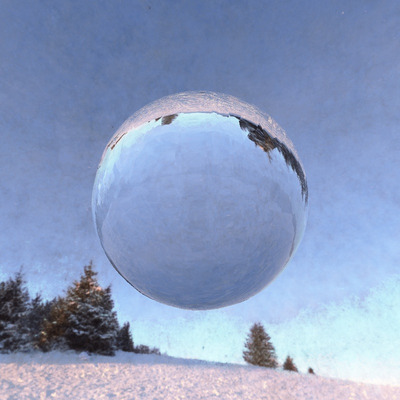}
    {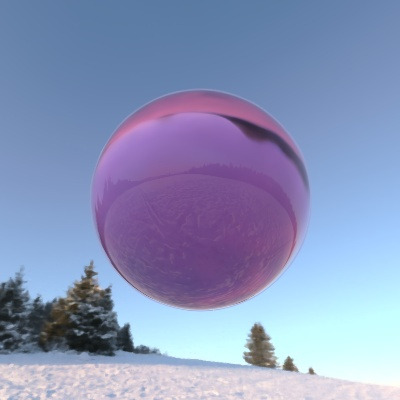}
    {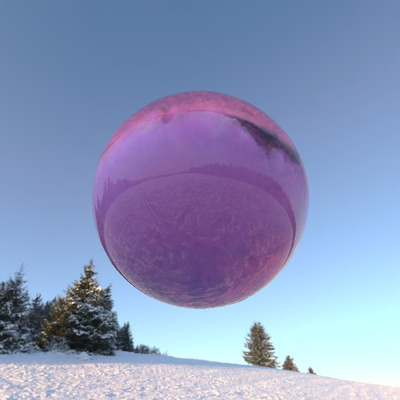}
    {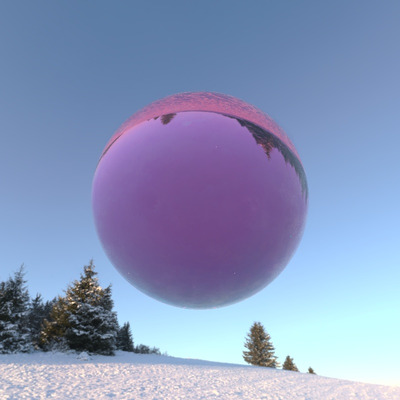}
\distrow
    {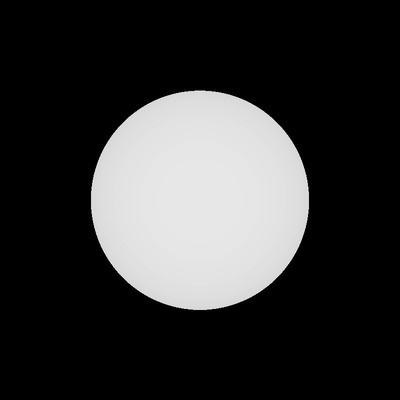}
    {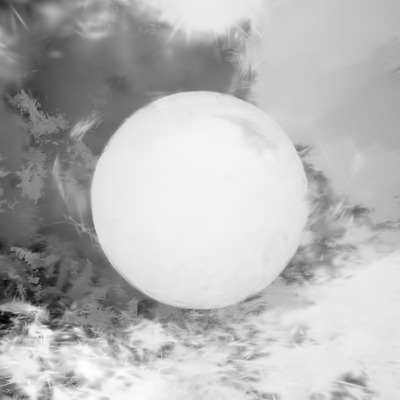}
    {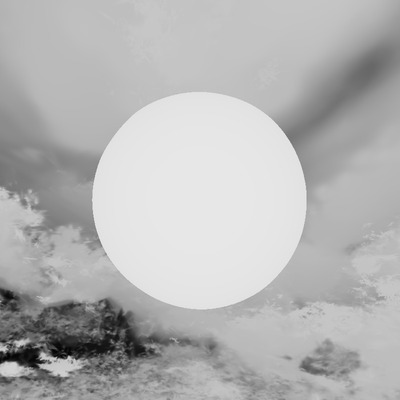}
    {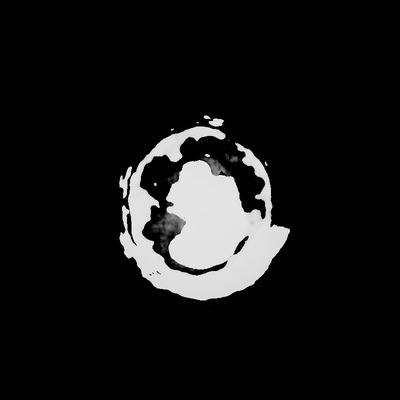}
    {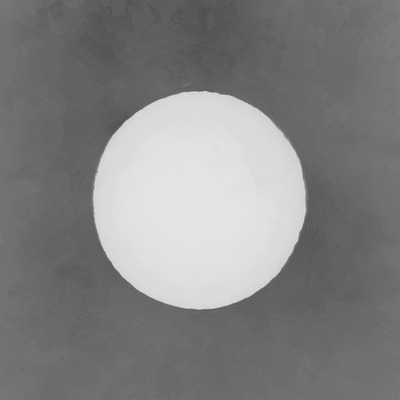}
    {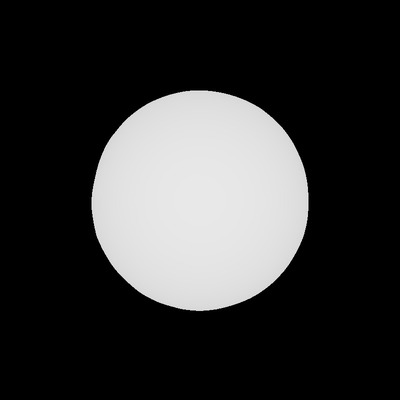}
    {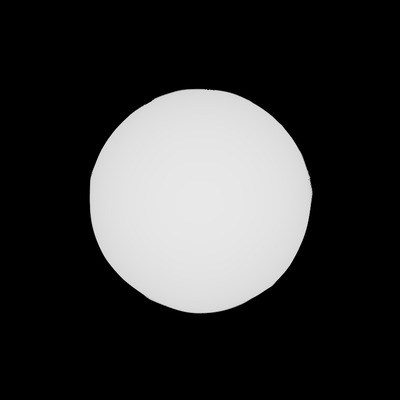}
    {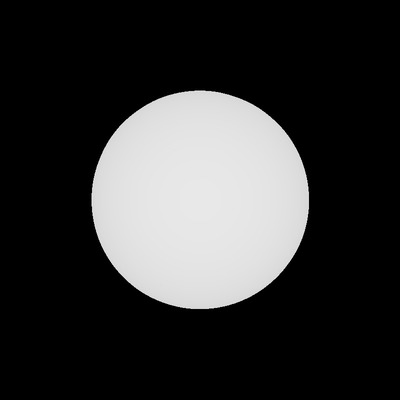}
    {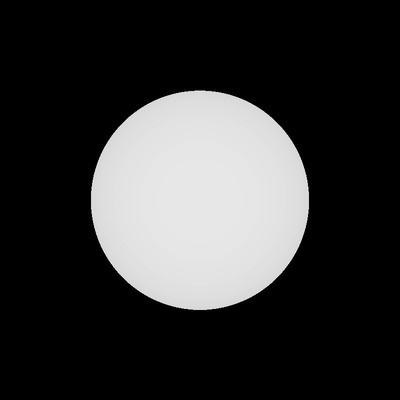}

\rgbrow
    {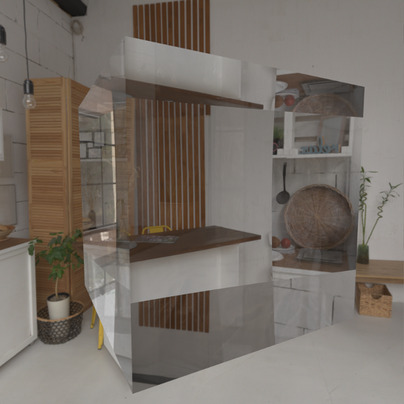}
    {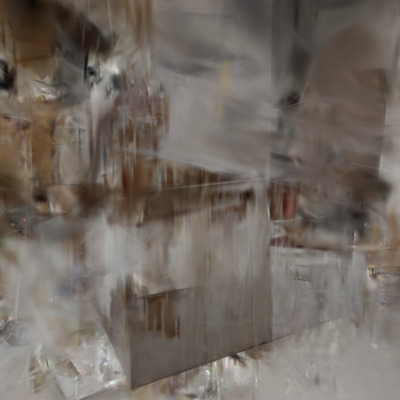}
    {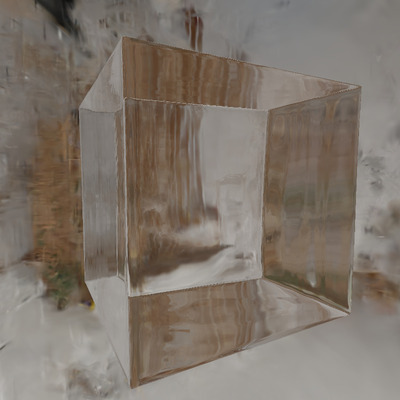}
    {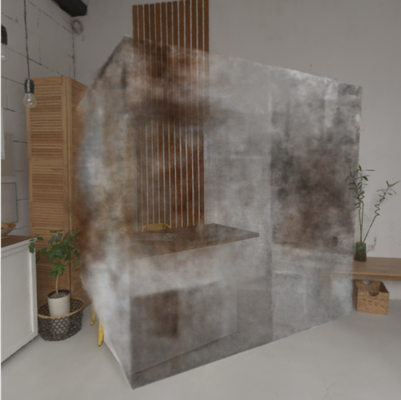}
    {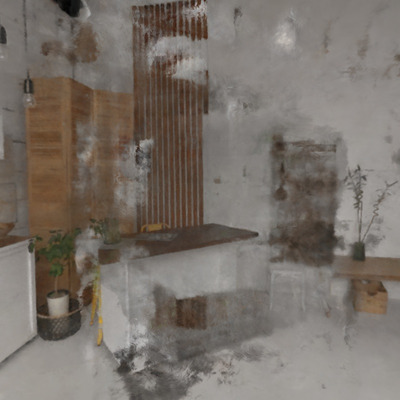}
    {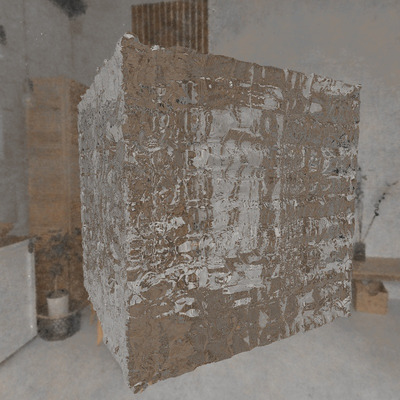}
    {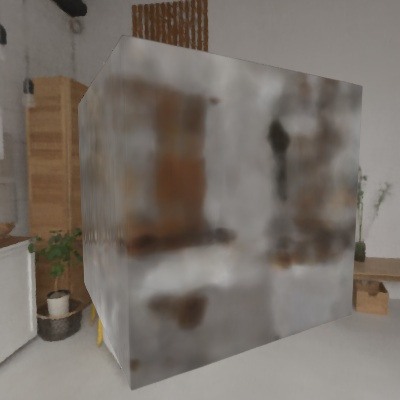}
    {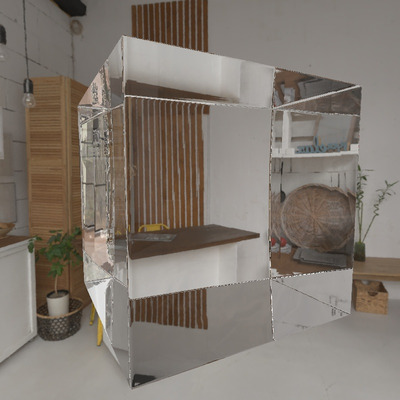}
    {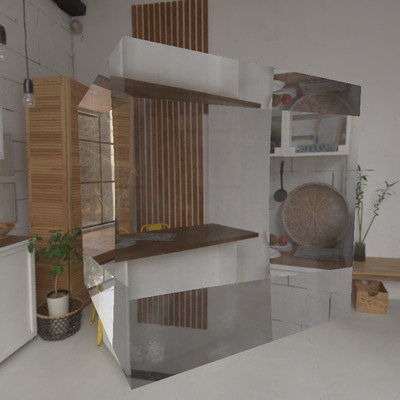}
\distrow
    {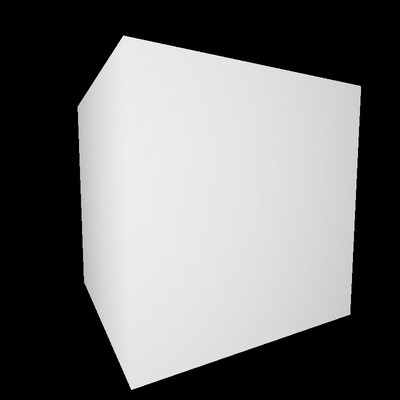}
    {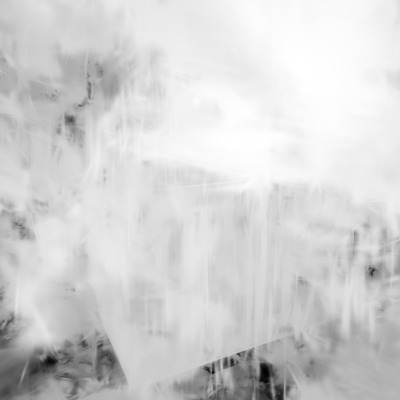}
    {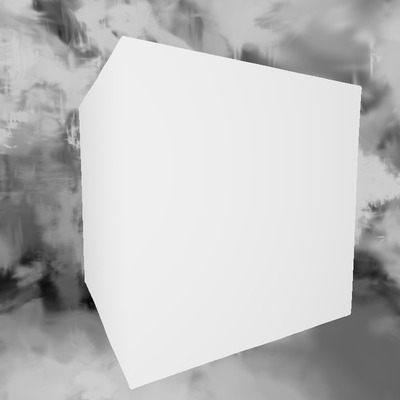}
    {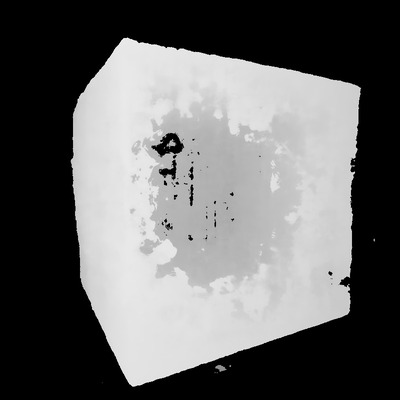}
    {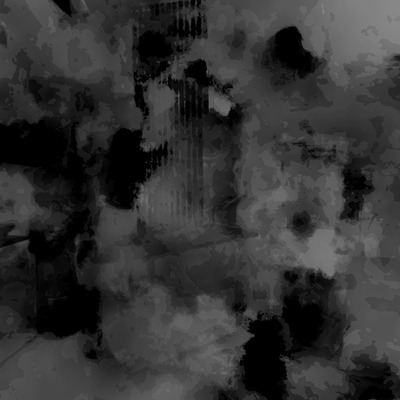}
    {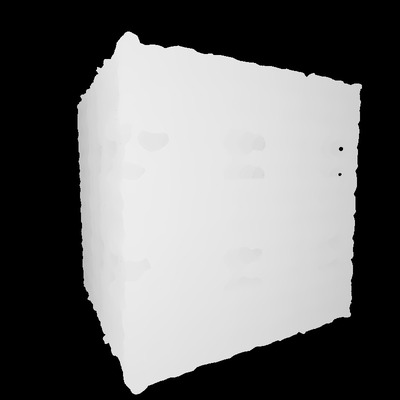}
    {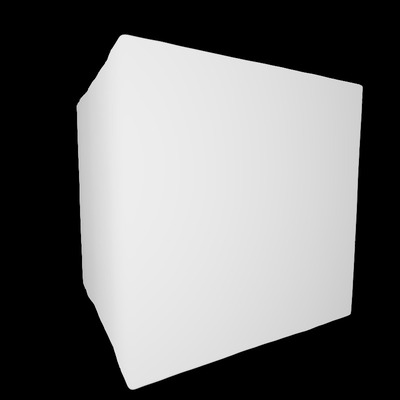}
    {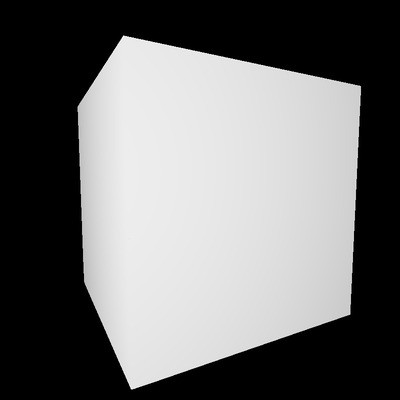}
    {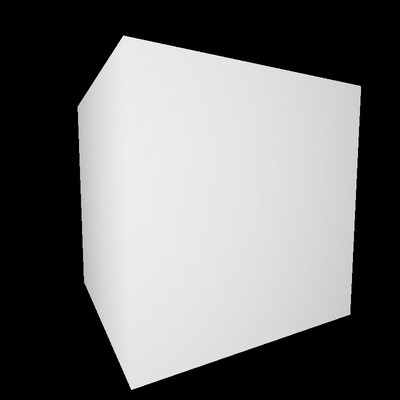}

\rgbrow
    {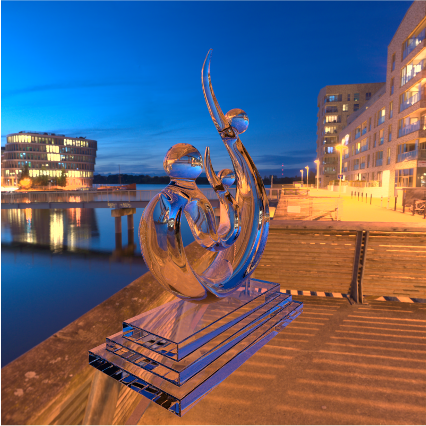}
    {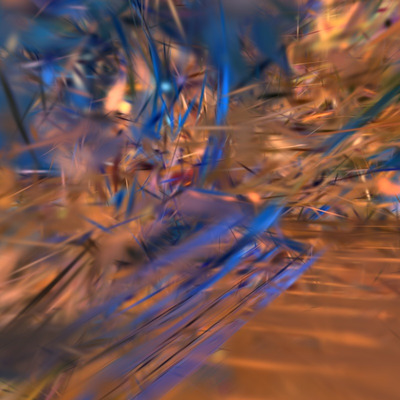}
    {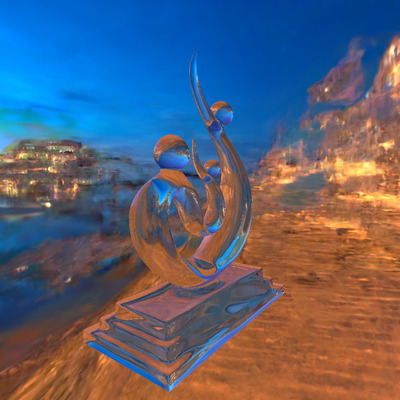}
    {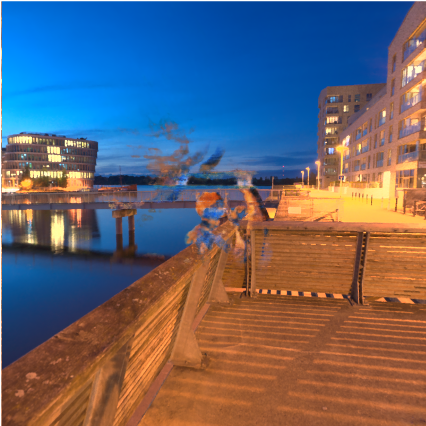}
    {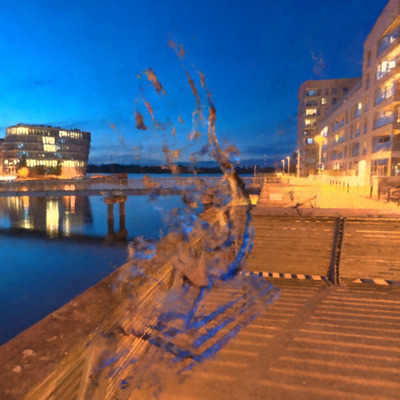}
    {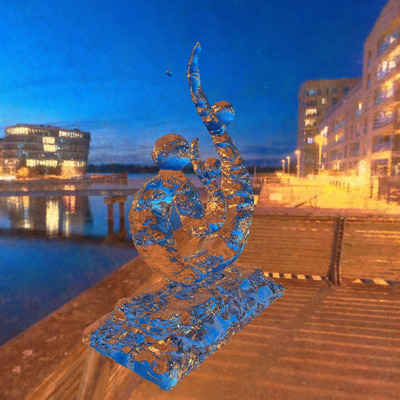}
    {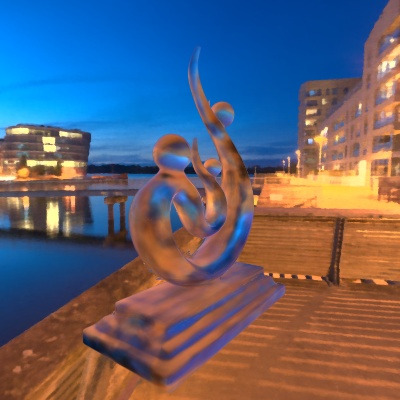}
    {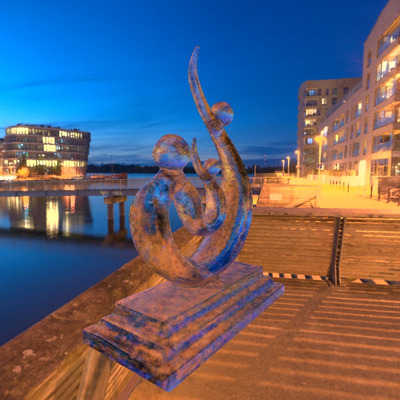}
    {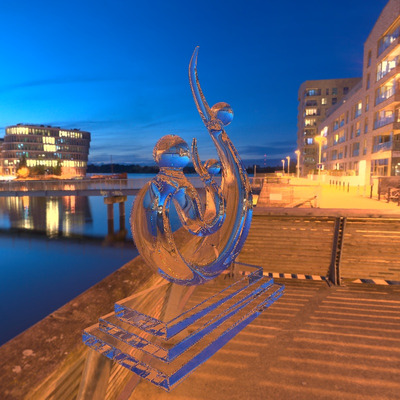}
\rowlabel{Distance Map} &
    \comparisonimage{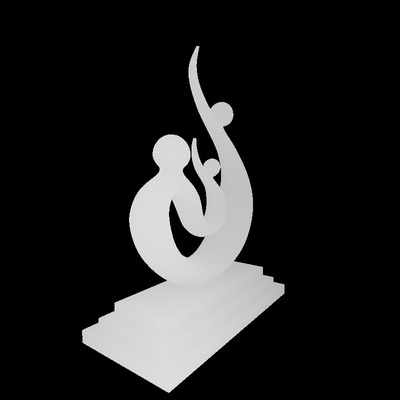} &
    \comparisonimage{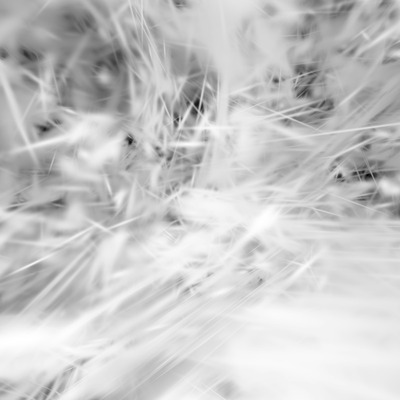} &
    \comparisonimage{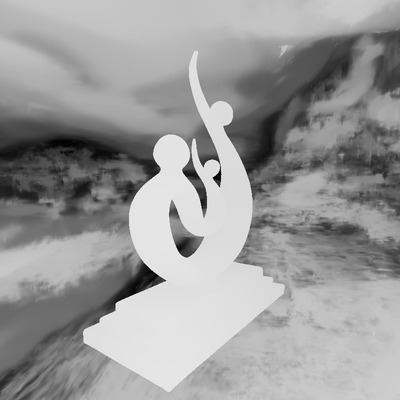} &
    \comparisonimage{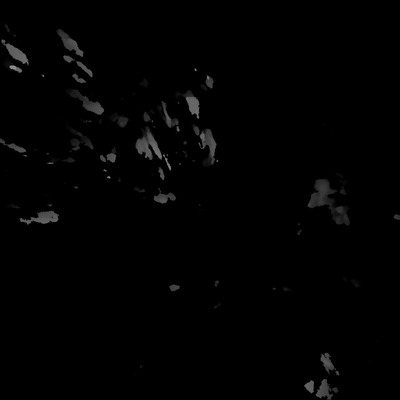} &
    \comparisonimage{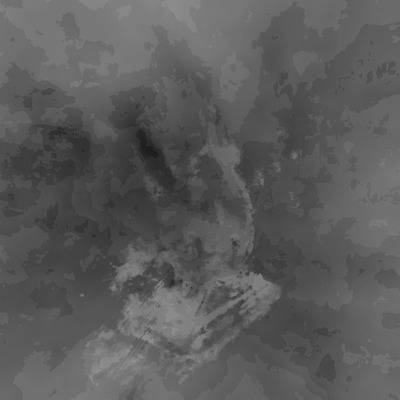} &
    \comparisonimage{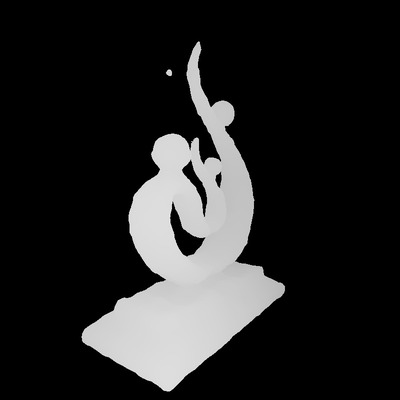} &
    \comparisonimage{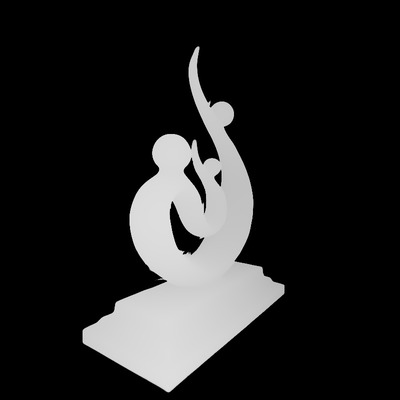} &
    \comparisonimage{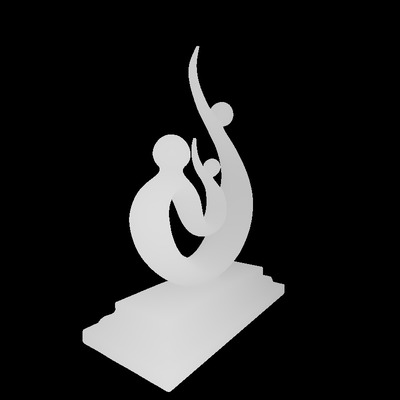} &
    \comparisonimage{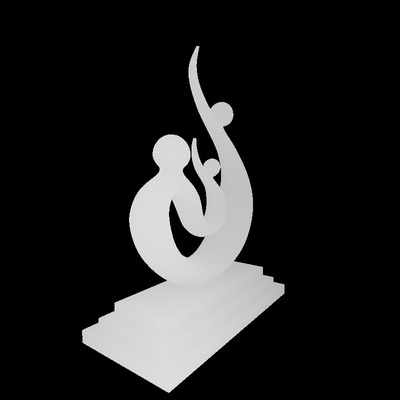}
    \\[0.3pt]

\rgbrow
    {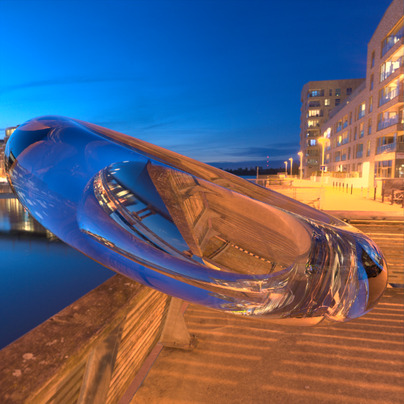}
    {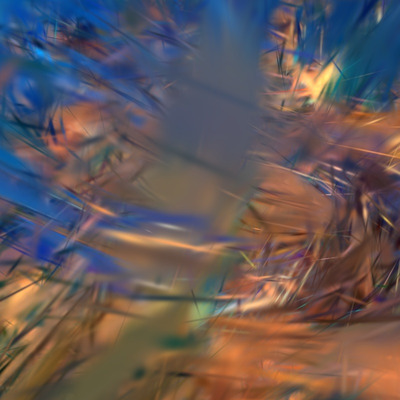}
    {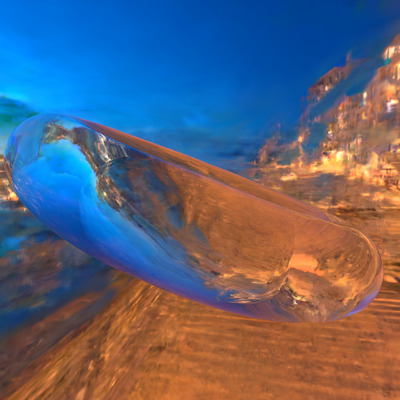}
    {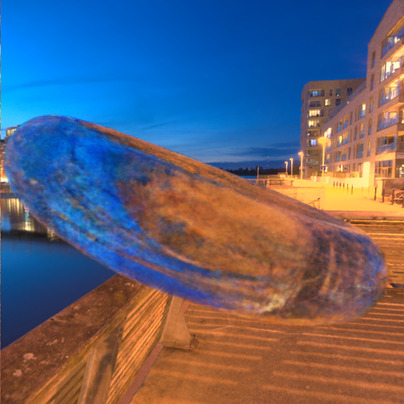}
    {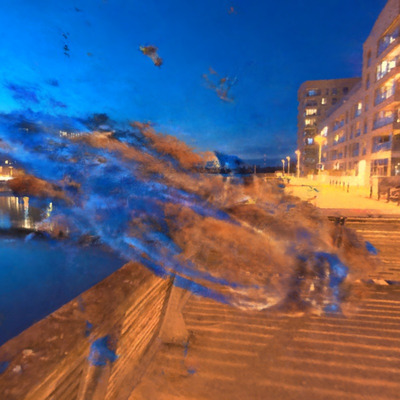}
    {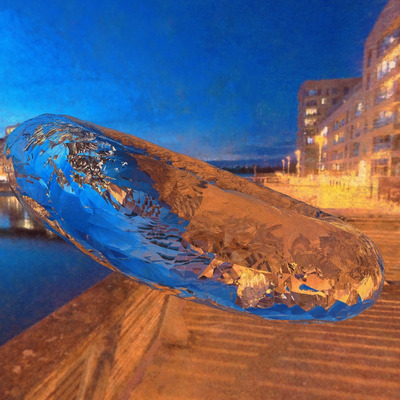}
    {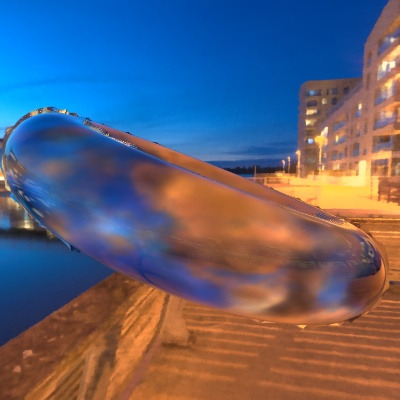}
    {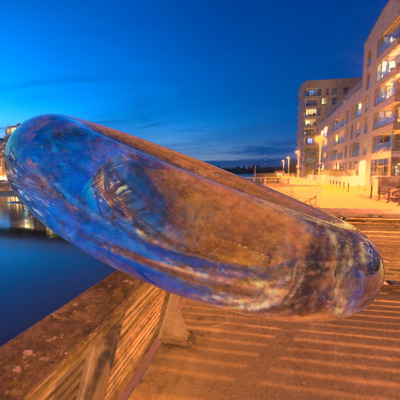}
    {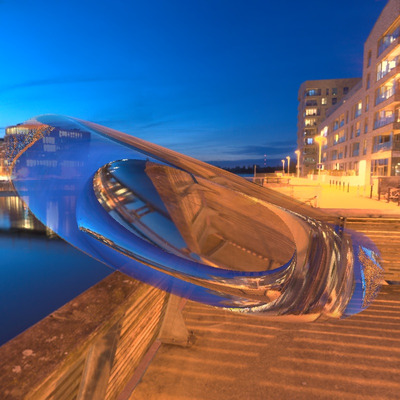}
\distrow
    {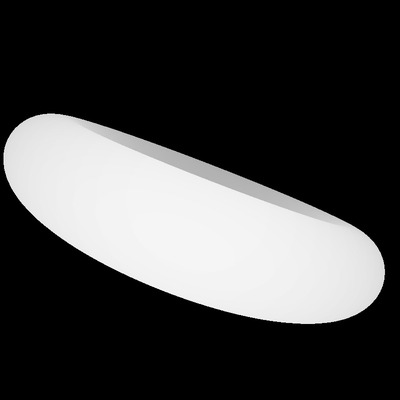}
    {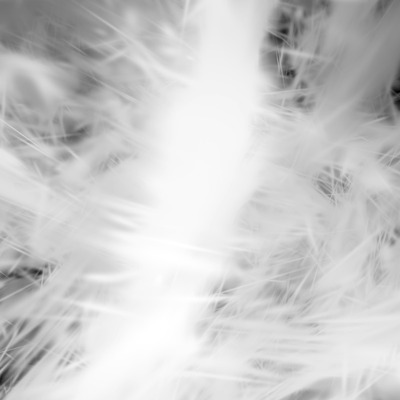}
    {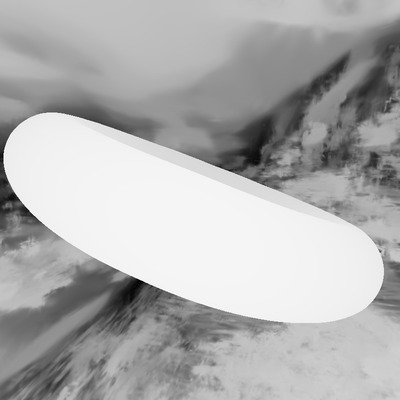}
    {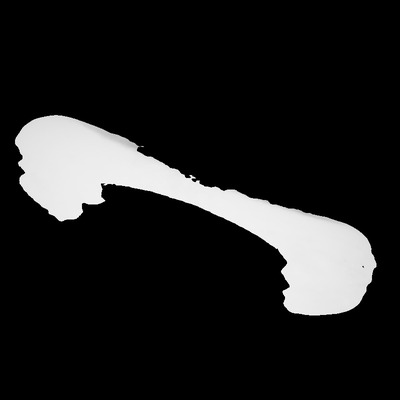}
    {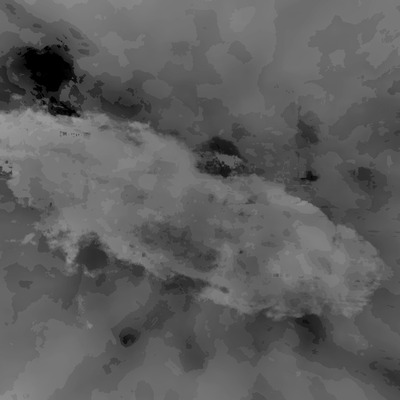}
    {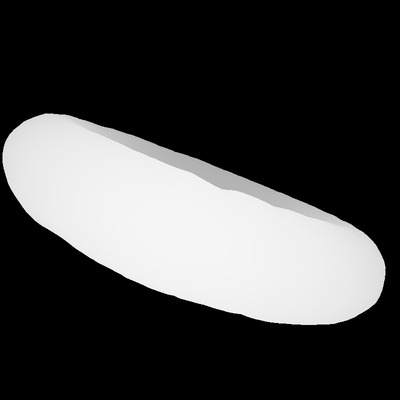}
    {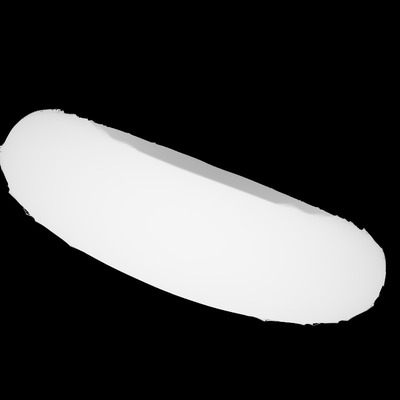}
    {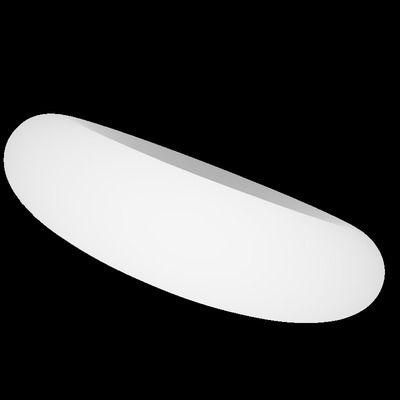}
    {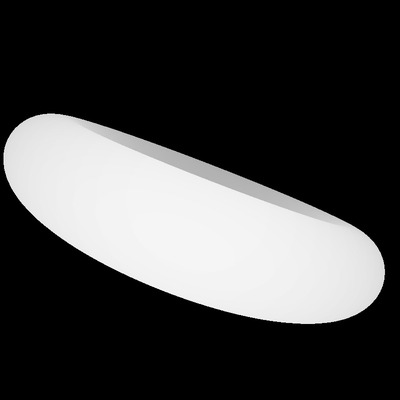}

\rgbrow
    {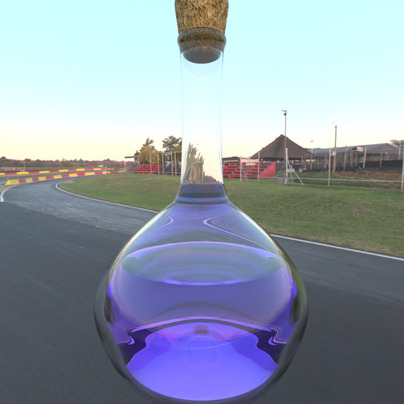}
    {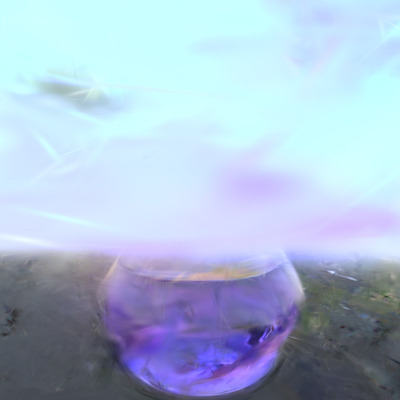}
    {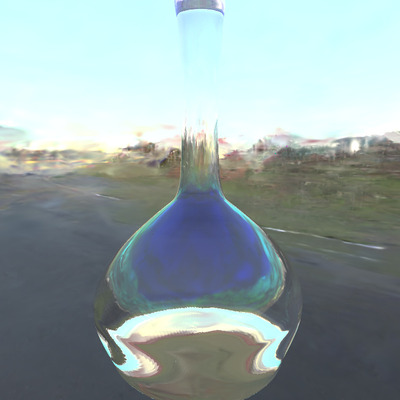}
    {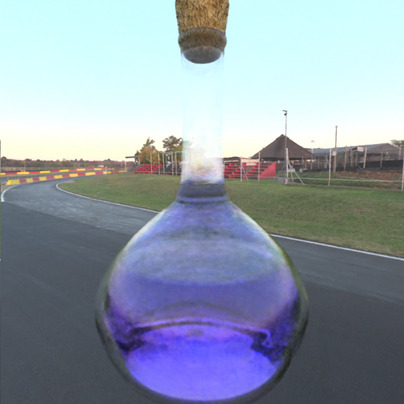}
    {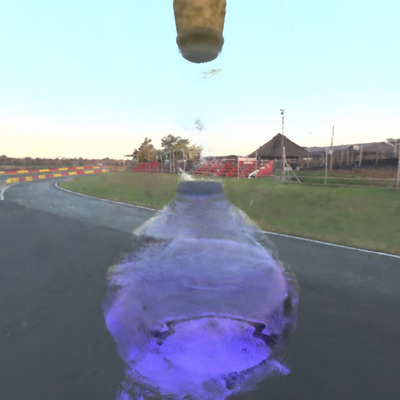}
    {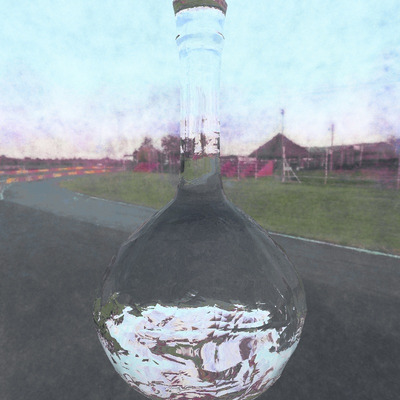}
    {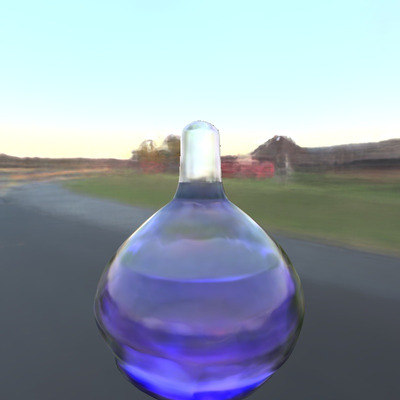}
    {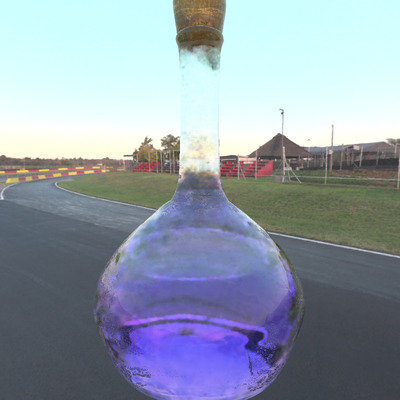}
    {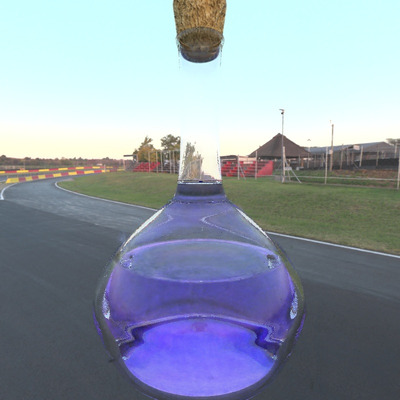}
\distrow
    {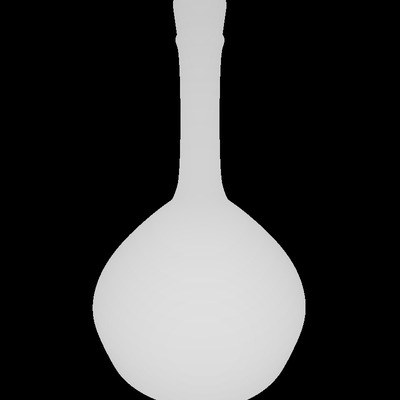}
    {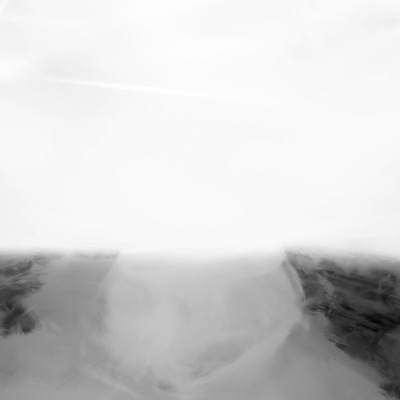}
    {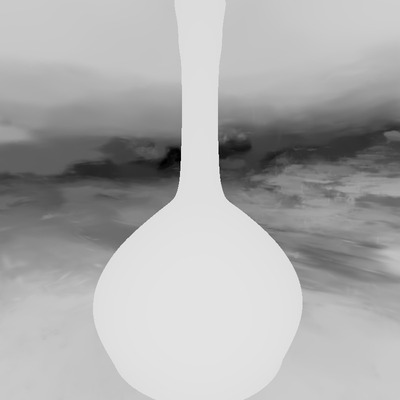}
    {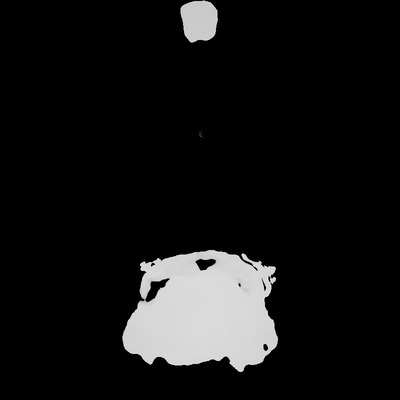}
    {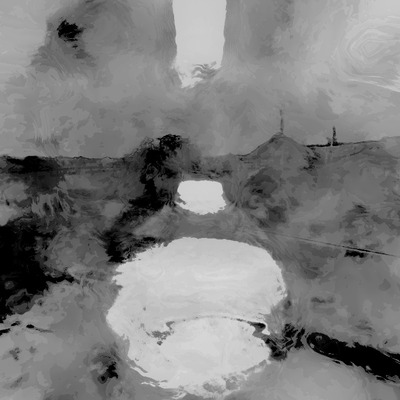}
    {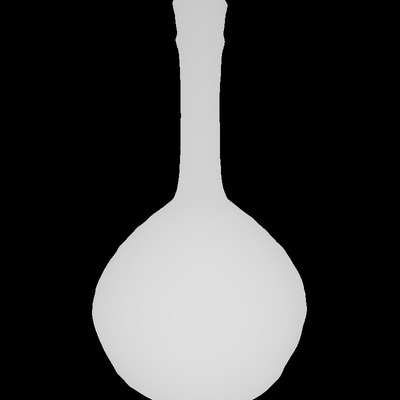}
    {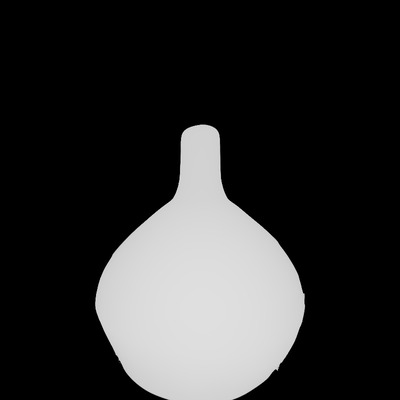}
    {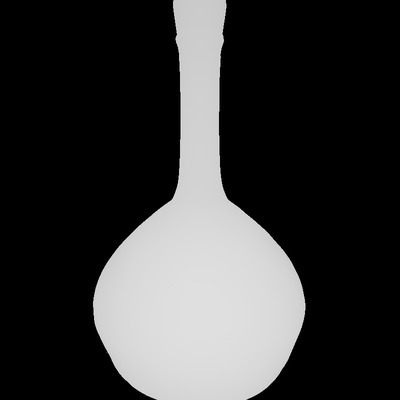}
    {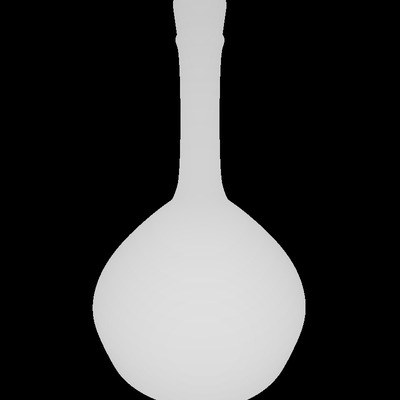}

\rgbrow
    {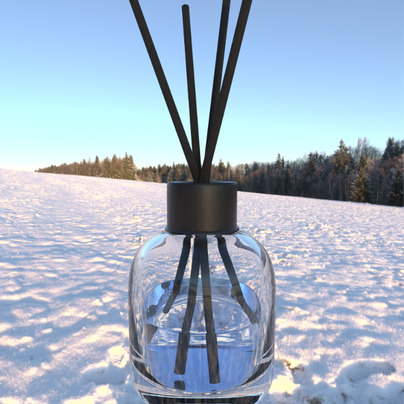}
    {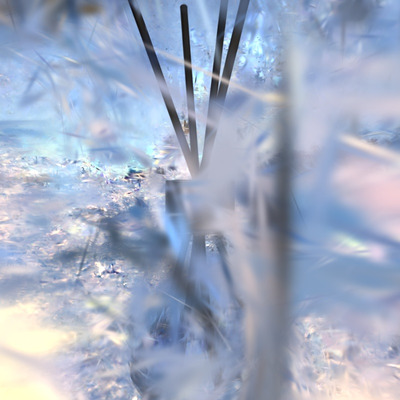}
    {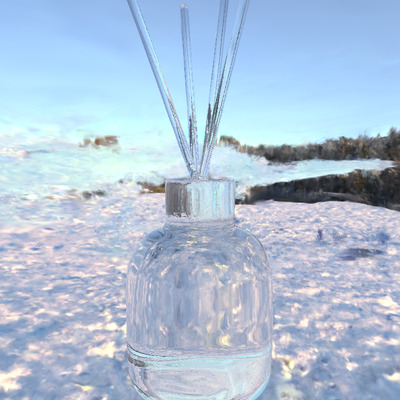}
    {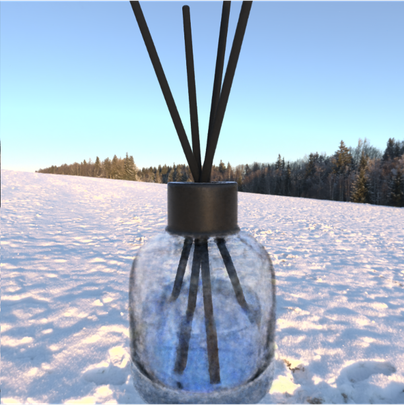}
    {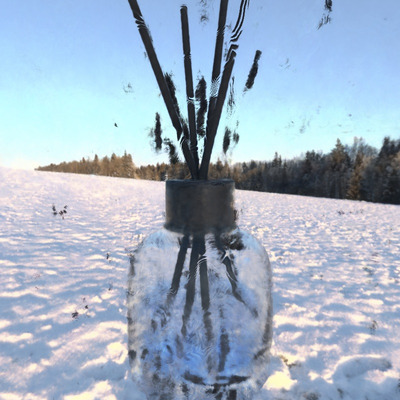}
    {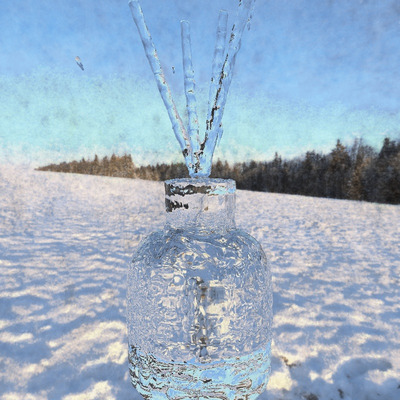}
    {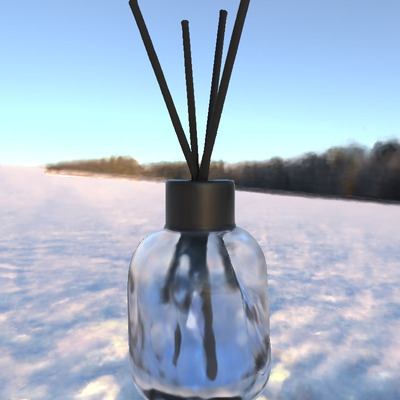}
    {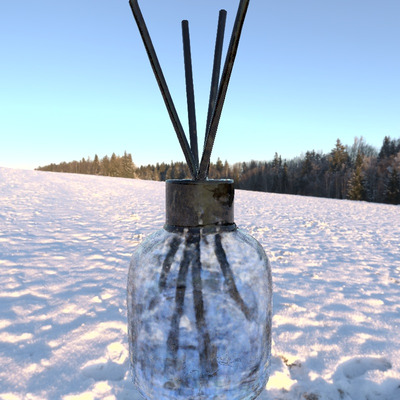}
    {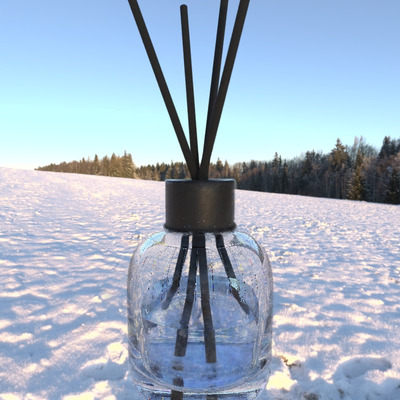}
\distrow
    {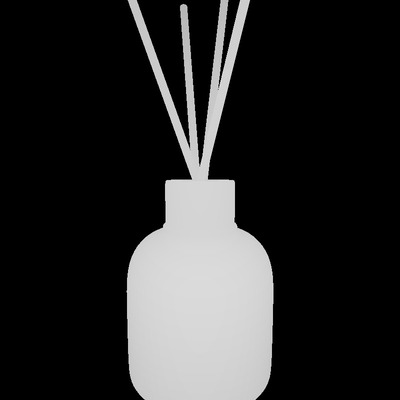}
    {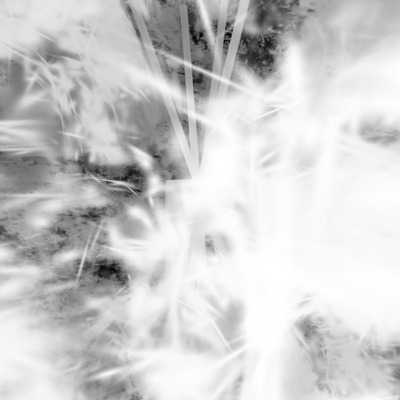}
    {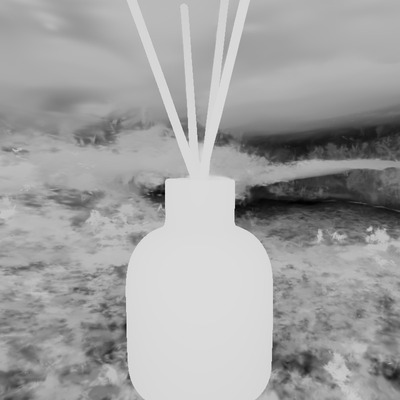}
    {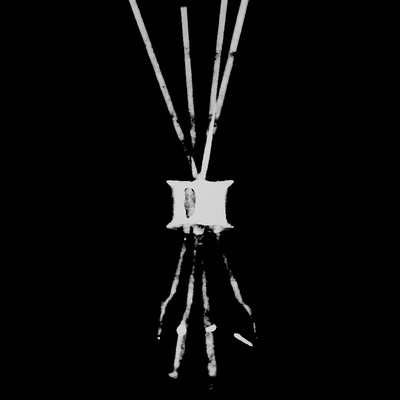}
    {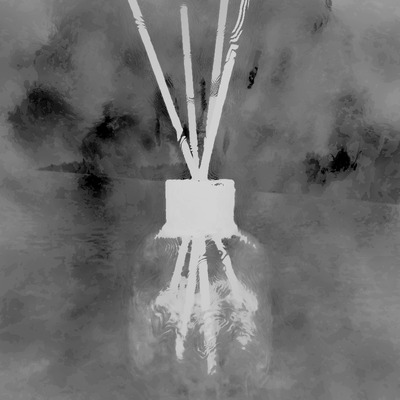}
    {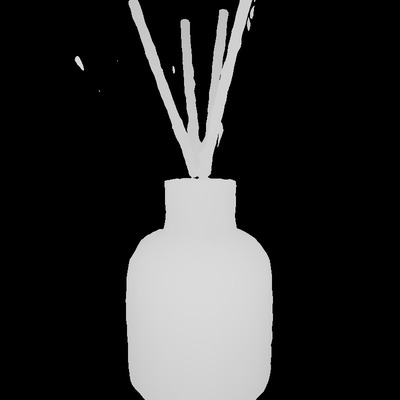}
    {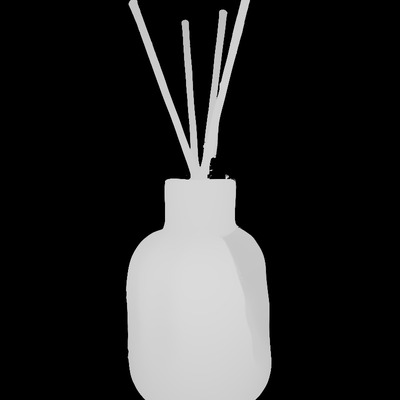}
    {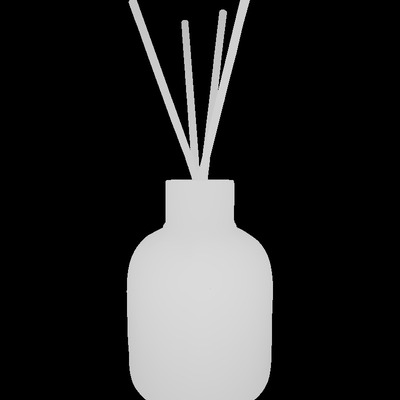}
    {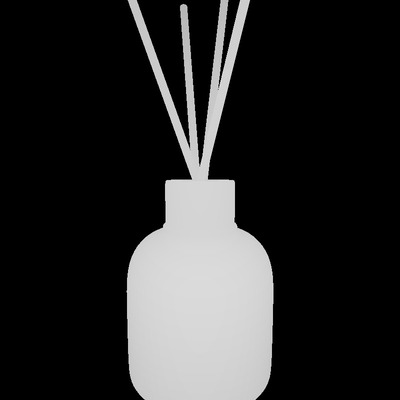}

\end{tabular}

\caption{
Additional qualitative comparison of novel view synthesis results on unbounded
synthetic scenes.
The background is an infinitely distant high-dynamic-range
environment map rather than enclosing geometry.
Each block shows one scene: the
top row gives the rendered RGB images and the bottom row the corresponding
distance maps, which expose geometric errors not apparent from RGB alone.
}
\label{fig:synthetic_unbounded_results}
\end{figure*}

\begin{figure*}[!t]
\centering

\setlength{\tabcolsep}{0.35pt}
\renewcommand{\arraystretch}{0}

\newcommand{\comparisonimage}[1]{%
    \includegraphics[width=0.1\linewidth]{#1}%
}

\newcommand{\patternrow}[9]{%
    \comparisonimage{#1} &
    \comparisonimage{#2} &
    \comparisonimage{#3} &
    \comparisonimage{#4} &
    \comparisonimage{#5} &
    \comparisonimage{#6} &
    \comparisonimage{#7} &
    \comparisonimage{#8} &
    \comparisonimage{#9} 
    \\[0.3pt]
}

\newcommand{\hdrrow}[9]{%
    \comparisonimage{#1} &
    \comparisonimage{#2} &
    \comparisonimage{#3} &
    \comparisonimage{#4} &
    \comparisonimage{#5} &
    \comparisonimage{#6} &
    \comparisonimage{#7} &
    \comparisonimage{#8} &
    \comparisonimage{#9}
    \\[0.3pt]
}

\begin{subfigure}[t]{\textwidth}
\centering

\begin{tabular}{@{}*{9}{c}@{}}

{\fontsize{6.5pt}{7.5pt}\selectfont Ground Truth}
&
{\fontsize{6.5pt}{7.5pt}\selectfont Splatfacto}
&
{\fontsize{6.5pt}{7.5pt}\selectfont TransparentGS}
&
{\fontsize{6.5pt}{7.5pt}\selectfont Zip-NeRF}
&
{\fontsize{6.5pt}{7.5pt}\selectfont MS-NeRF}
&
{\fontsize{6.5pt}{7.5pt}\selectfont Ray Deformation}
&
{\fontsize{6.5pt}{7.5pt}\selectfont NU-NeRF}
&
{\fontsize{6.5pt}{7.5pt}\selectfont TNSR}
&
{\fontsize{6.5pt}{7.5pt}\selectfont \method (ours)}
\\[2pt]

\patternrow
    {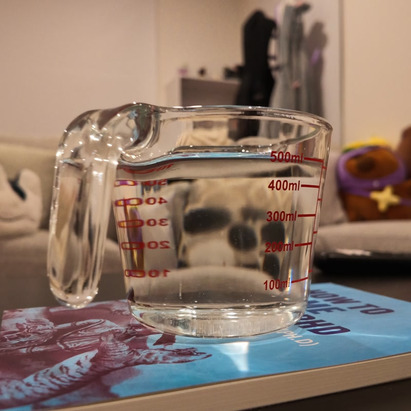}
    {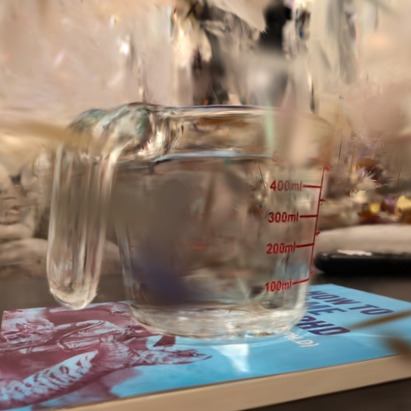}
    {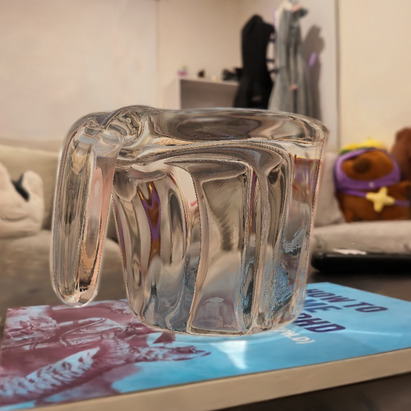}
    {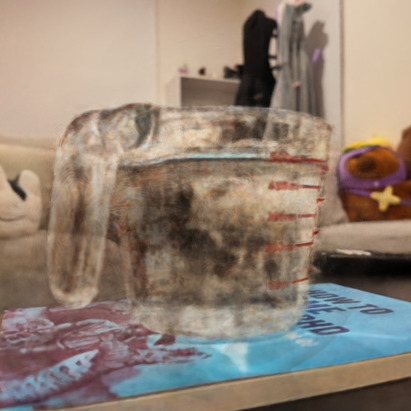}
    {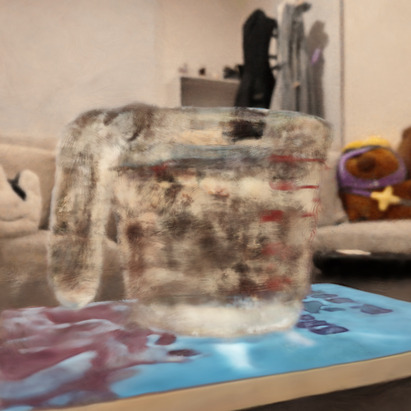}
    {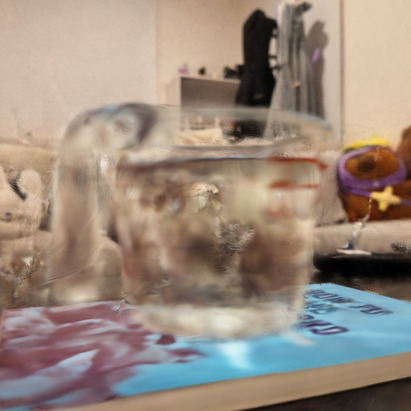}
    {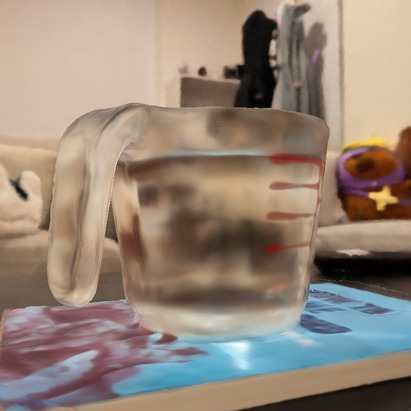}
    {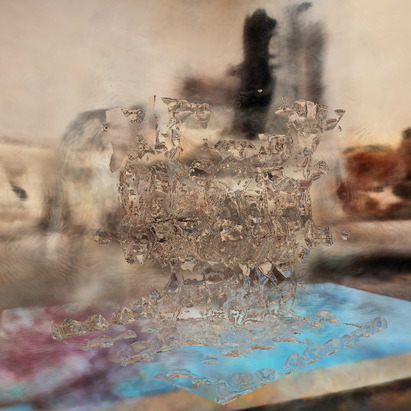}
    {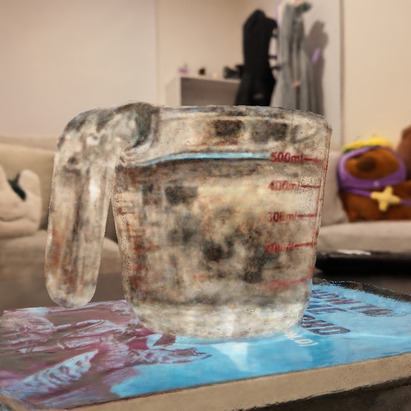}

\patternrow
    {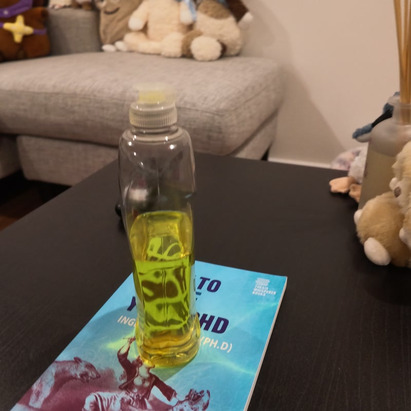}
    {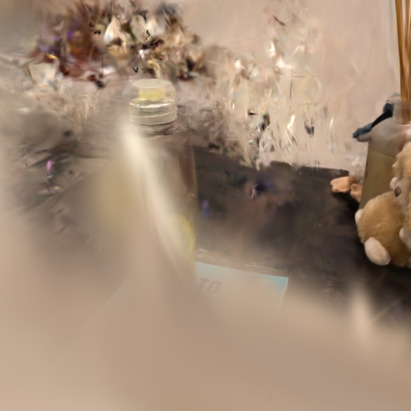}
    {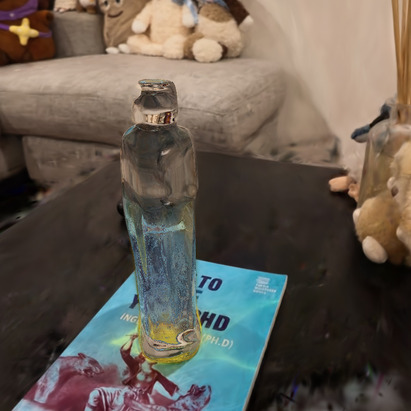}
    {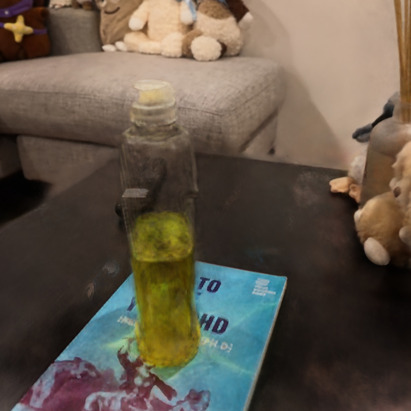}
    {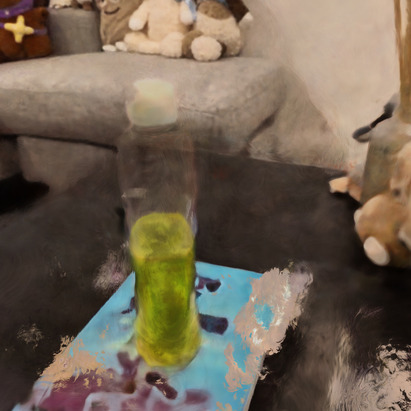}
    {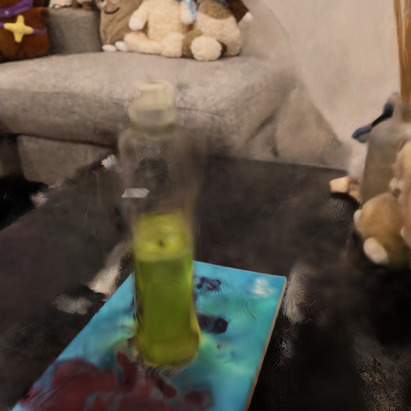}
    {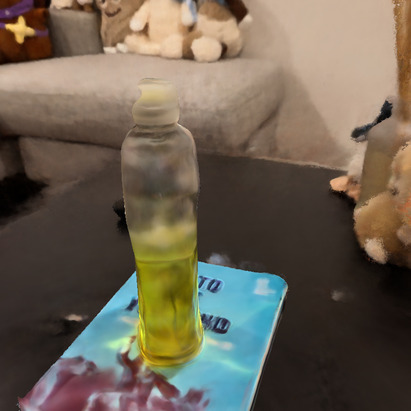}
    {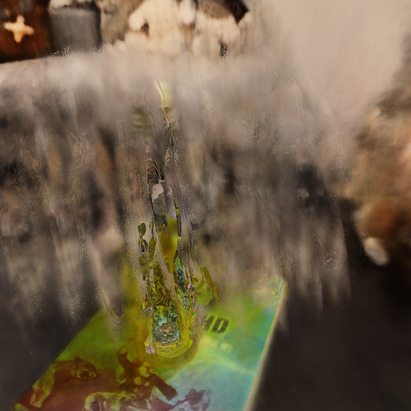}
    {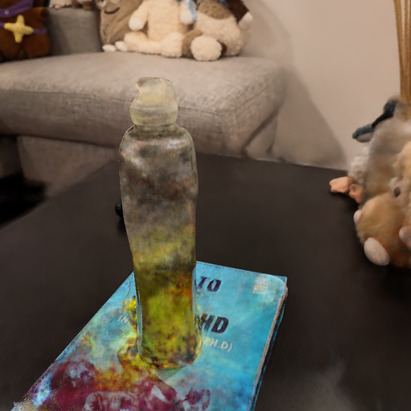}

\patternrow
    {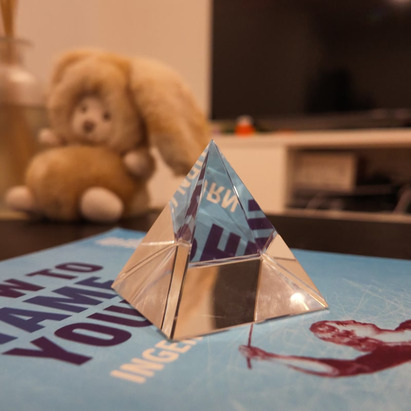}
    {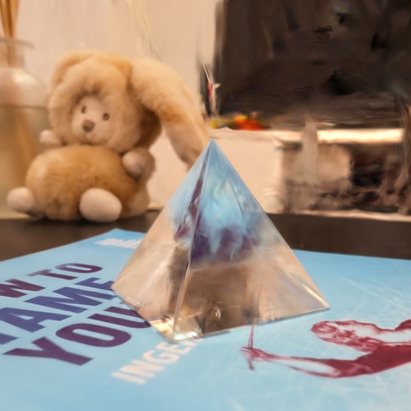}
    {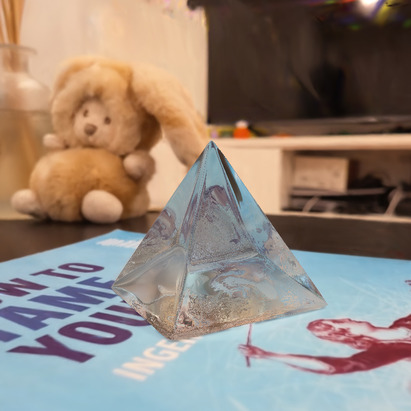}
    {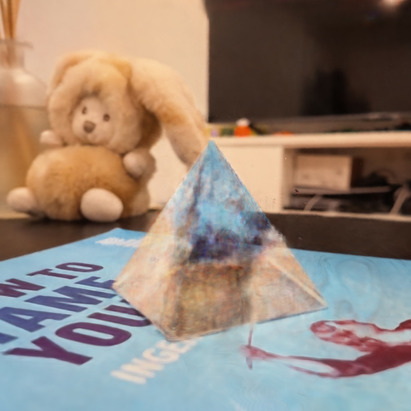}
    {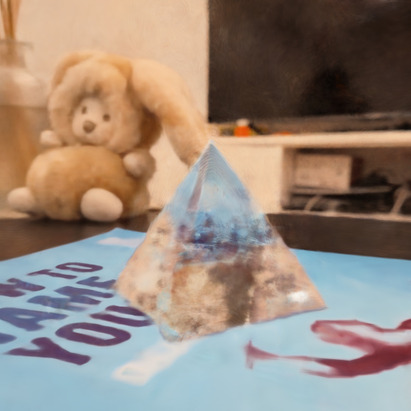}
    {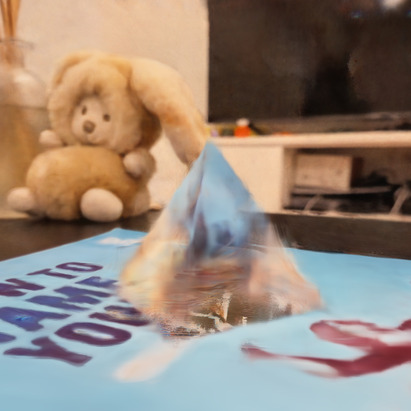}
    {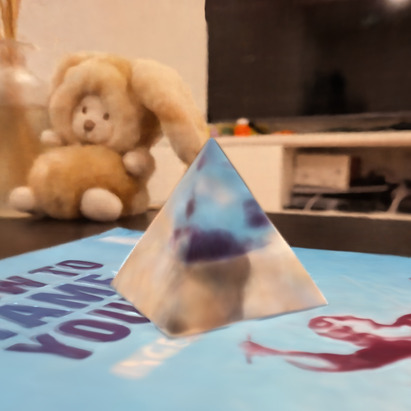}
    {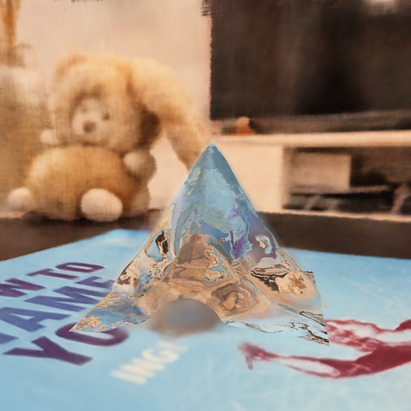}
    {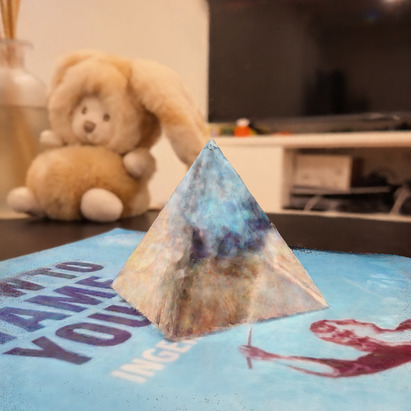}

\patternrow
    {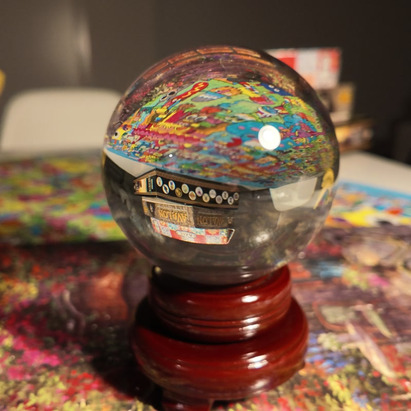}
    {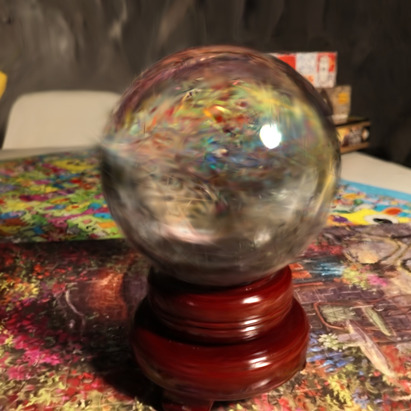}
    {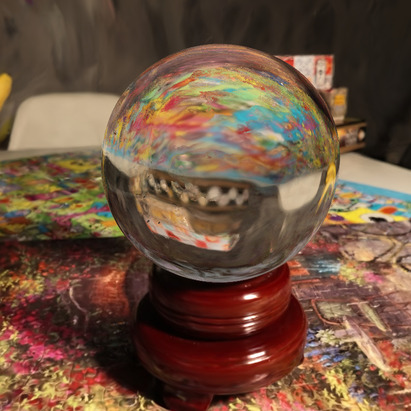}
    {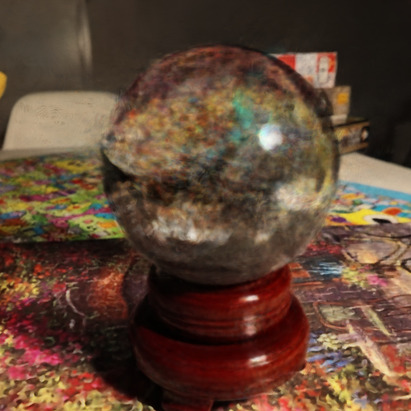}
    {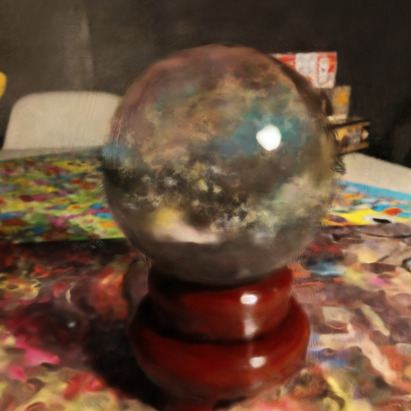}
    {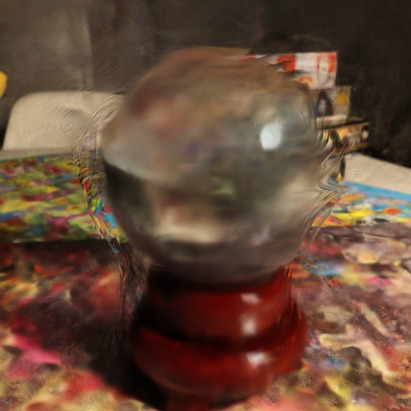}
    {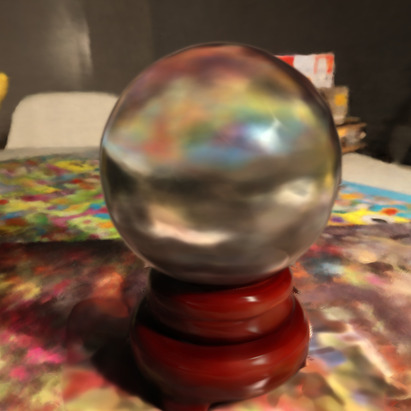}
    {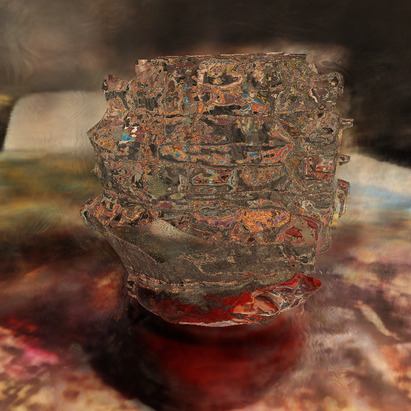}
    {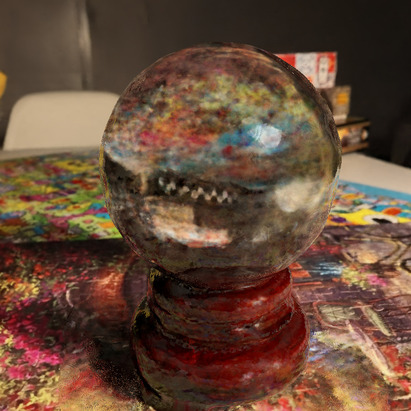}

\patternrow
    {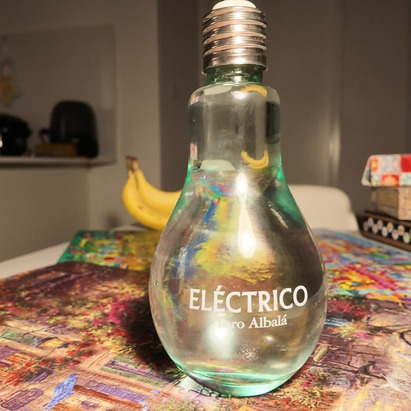}
    {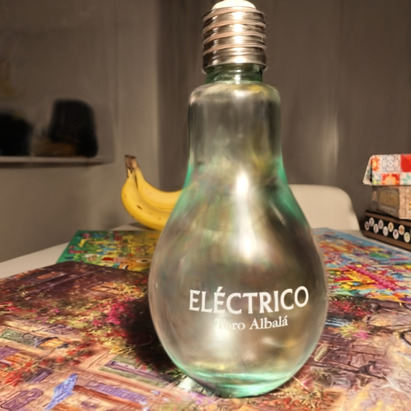}
    {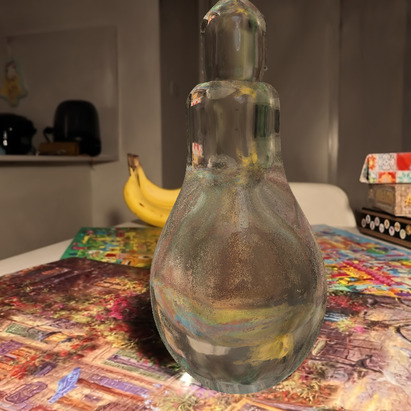}
    {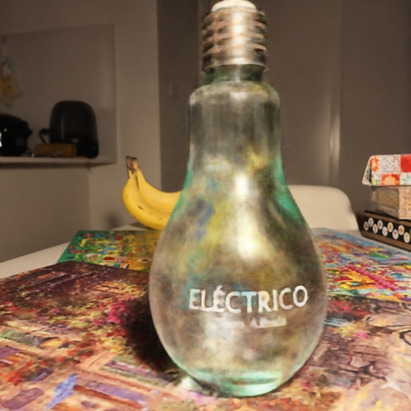}
    {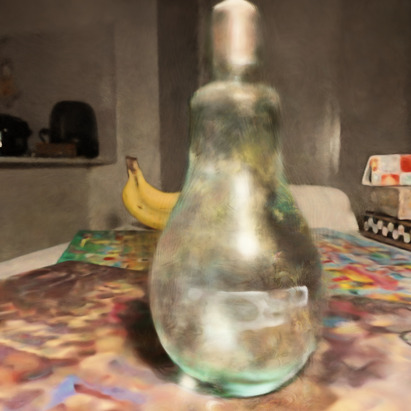}
    {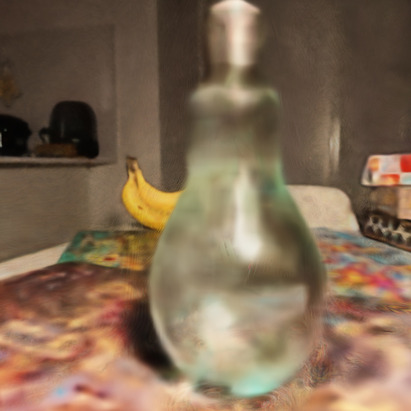}
    {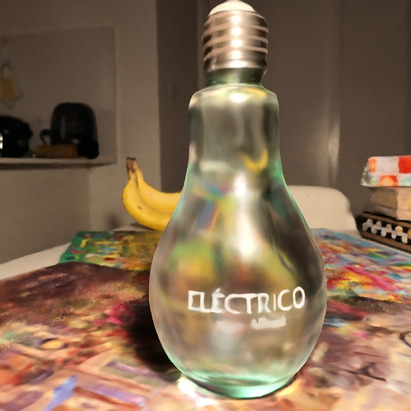}
    {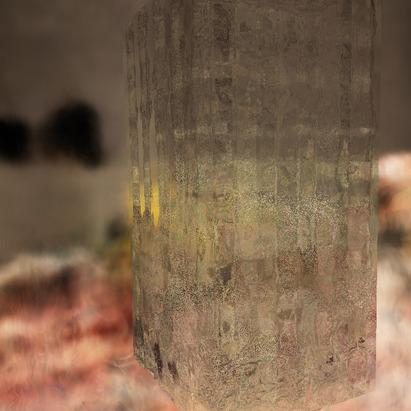}
    {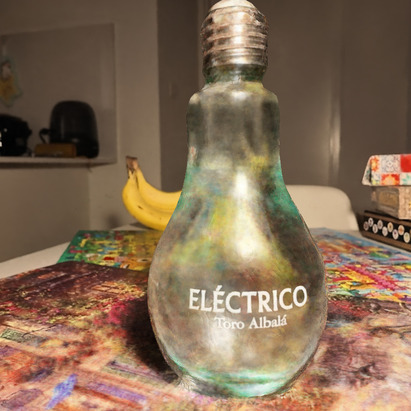}

\patternrow
    {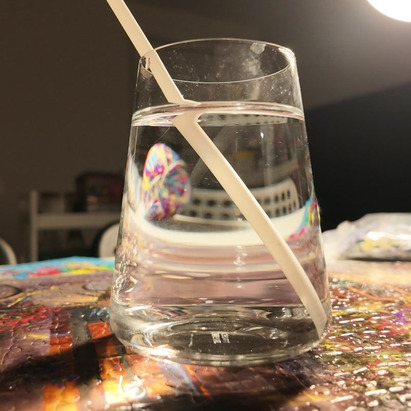}
    {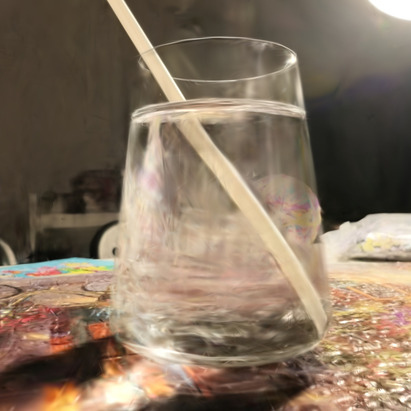}
    {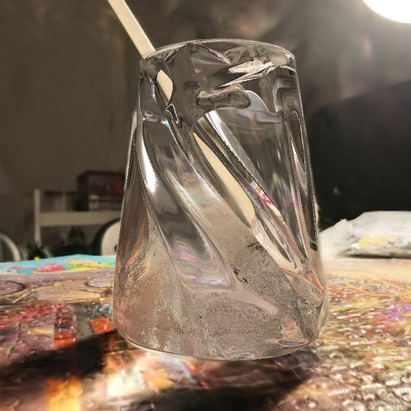}
    {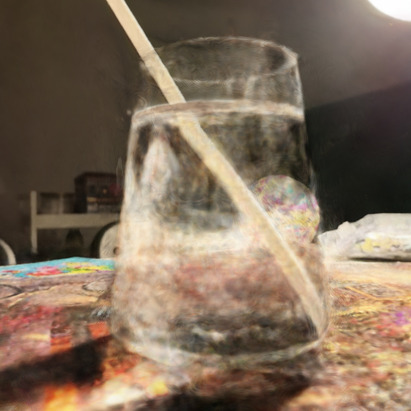}
    {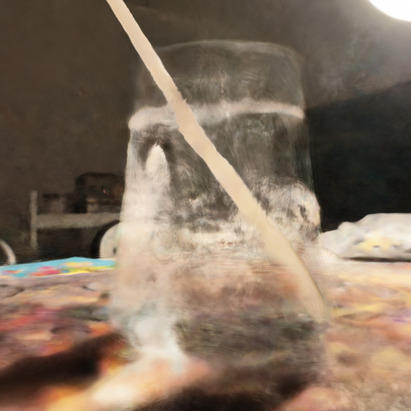}
    {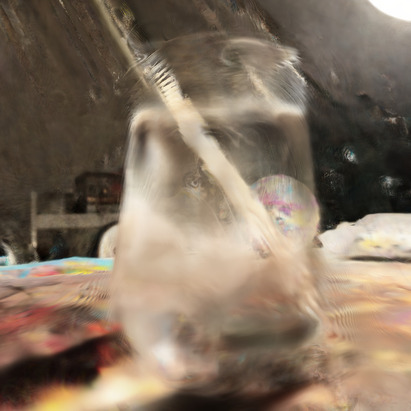}
    {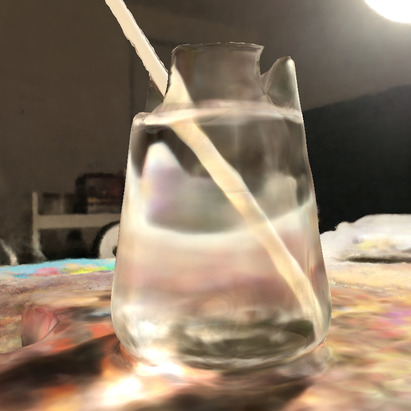}
    {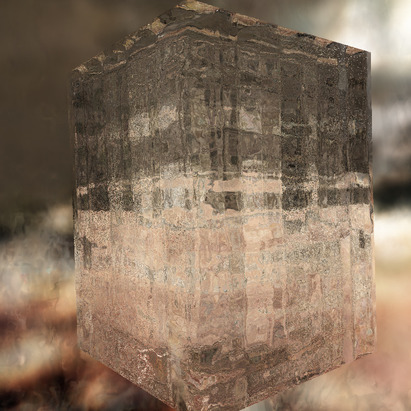}
    {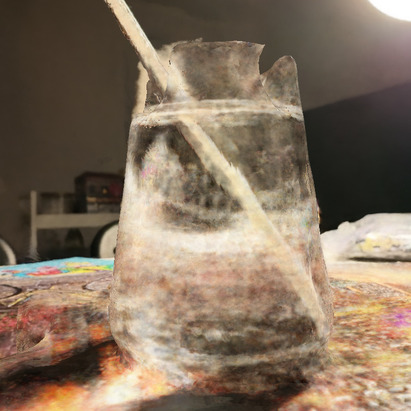}

\end{tabular}
\vspace{-4pt}
\caption{Indoor scene backgrounds.}
\label{fig:real_indoor_results}
\end{subfigure}

\vspace{3pt}

\begin{subfigure}[t]{\textwidth}
\centering

\begin{tabular}{@{}*{9}{c}@{}}

{\fontsize{6.5pt}{7.5pt}\selectfont Ground Truth}
&
{\fontsize{6.5pt}{7.5pt}\selectfont Splatfacto}
&
{\fontsize{6.5pt}{7.5pt}\selectfont TransparentGS}
&
{\fontsize{6.5pt}{7.5pt}\selectfont Zip-NeRF}
&
{\fontsize{6.5pt}{7.5pt}\selectfont MS-NeRF}
&
{\fontsize{6.5pt}{7.5pt}\selectfont Ray Deformation}
&
{\fontsize{6.5pt}{7.5pt}\selectfont NU-NeRF}
&
{\fontsize{6.5pt}{7.5pt}\selectfont TNSR}
&
{\fontsize{6.5pt}{7.5pt}\selectfont \method (ours)}
\\[2pt]

\hdrrow
    {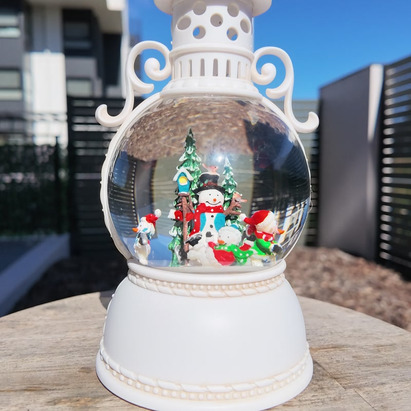}
    {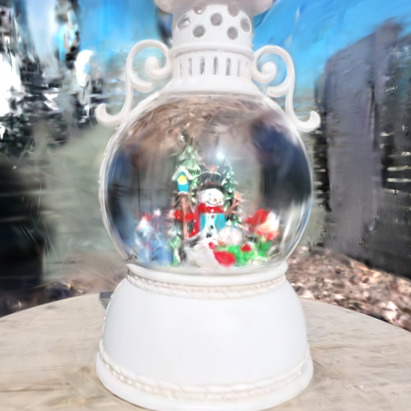}
    {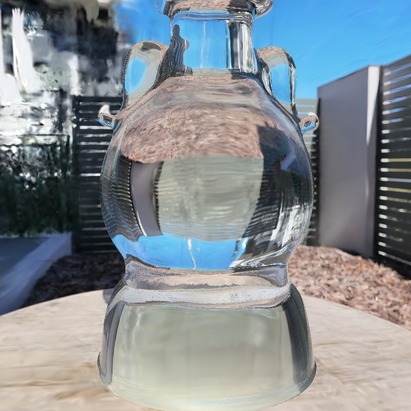}
    {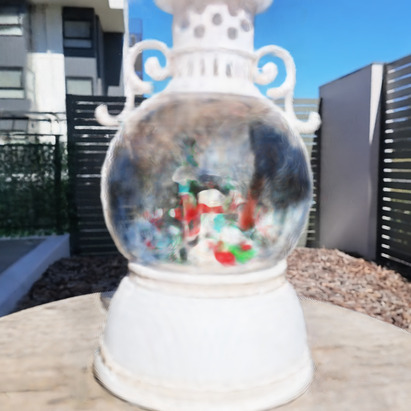}
    {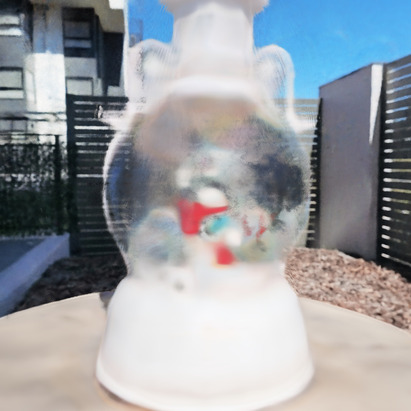}
    {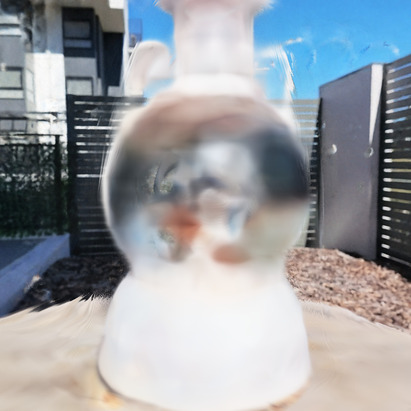}
    {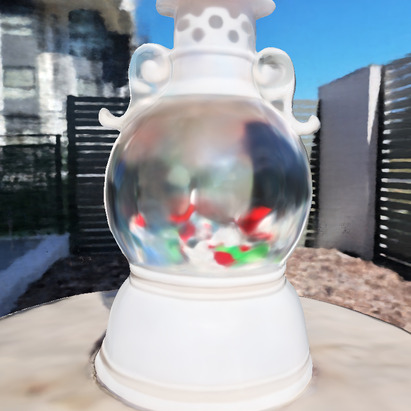}
    {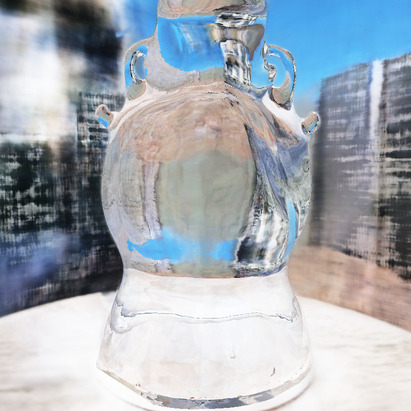}
    {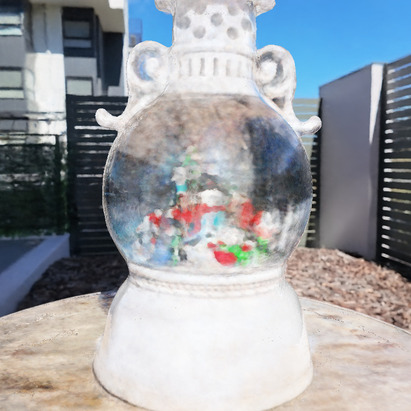}

\hdrrow
    {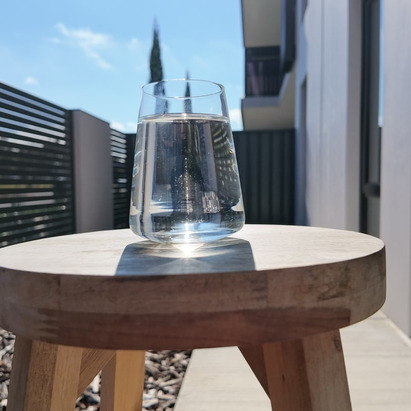}
    {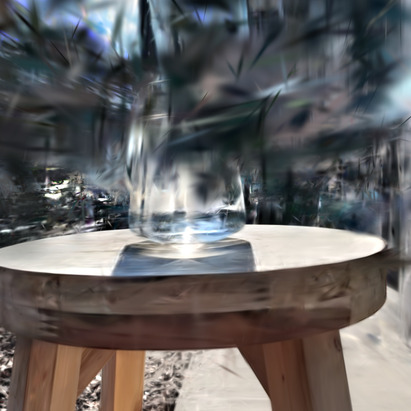}
    {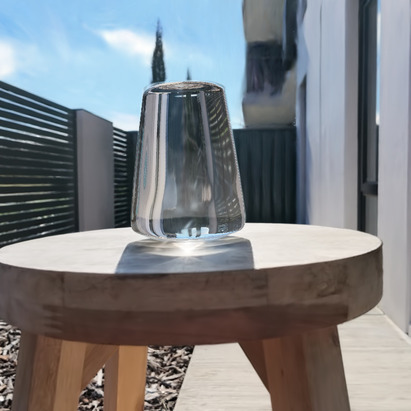}
    {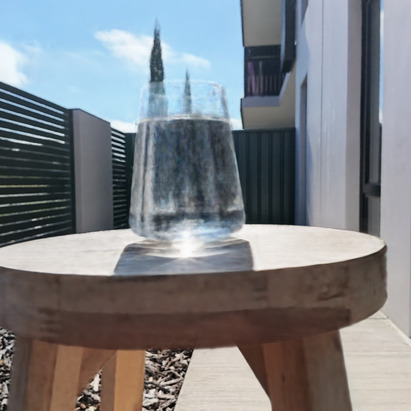}
    {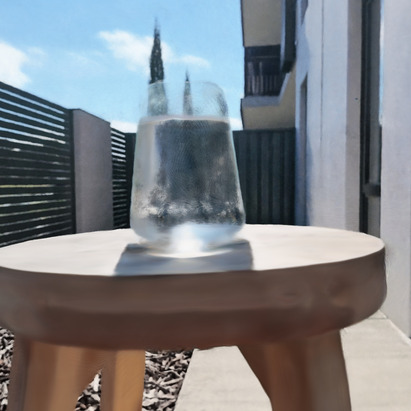}
    {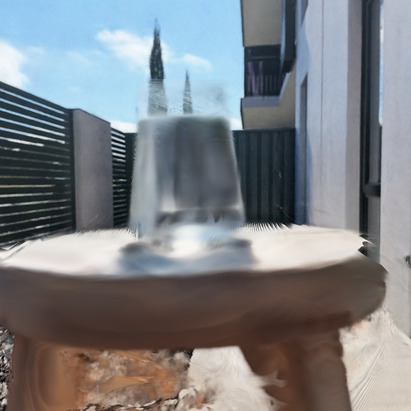}
    {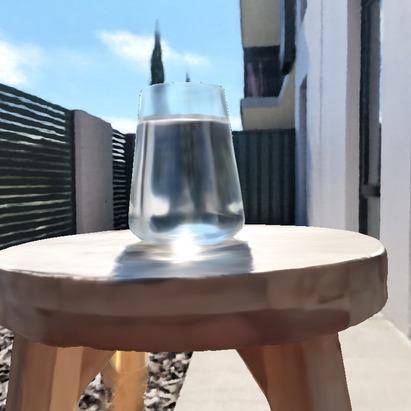}
    {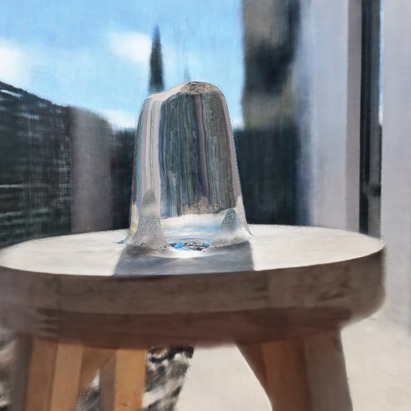}
    {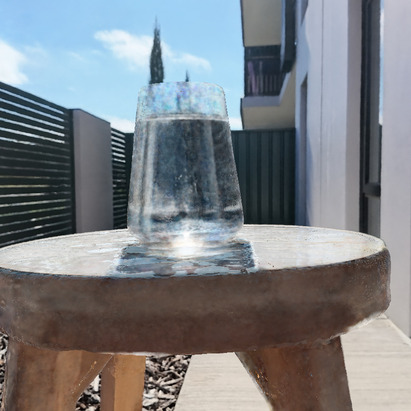}

\hdrrow
    {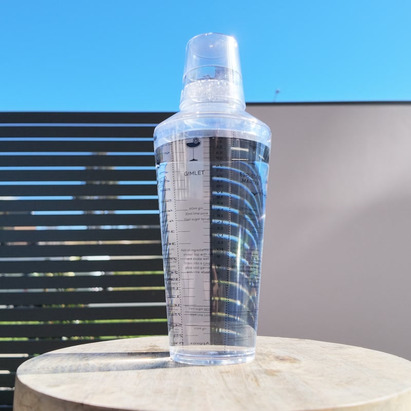}
    {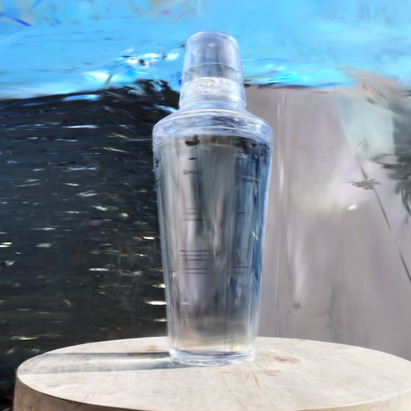}
    {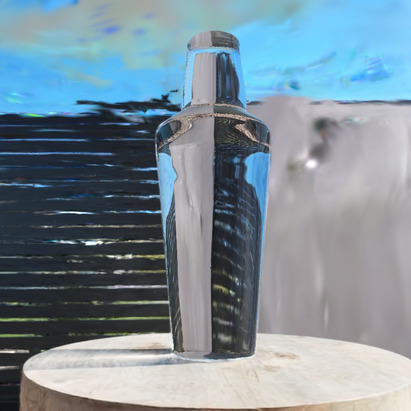}
    {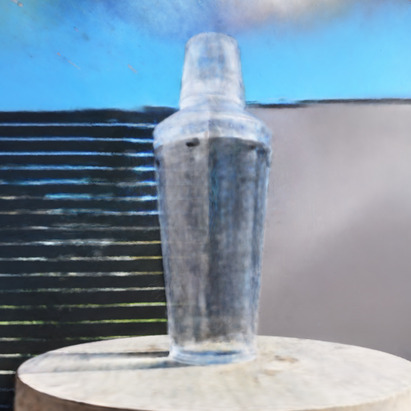}
    {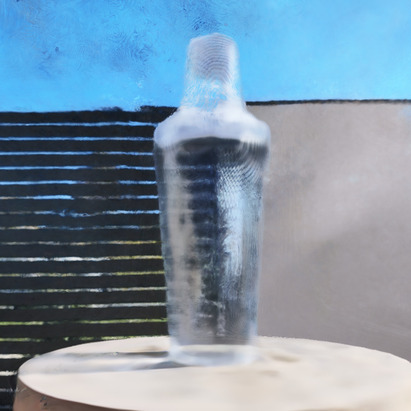}
    {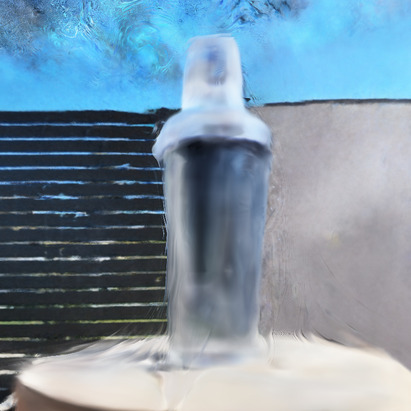}
    {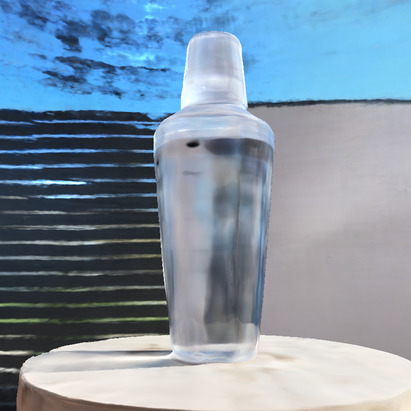}
    {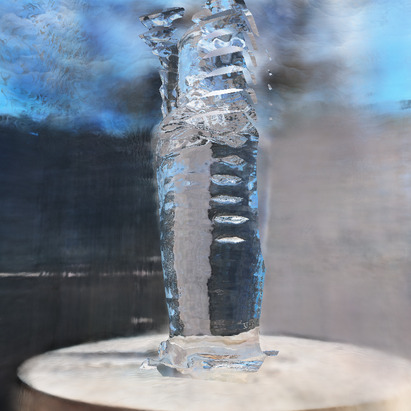}
    {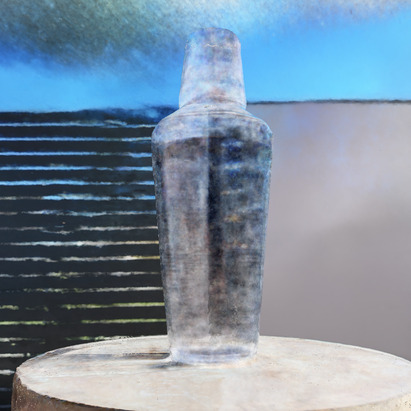}

\hdrrow
    {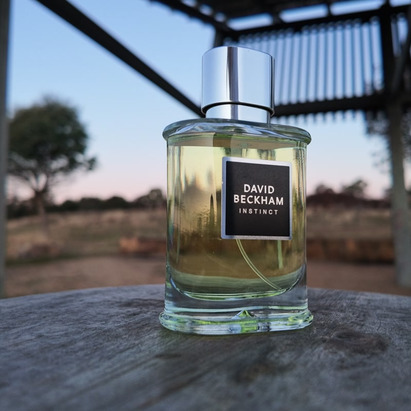}
    {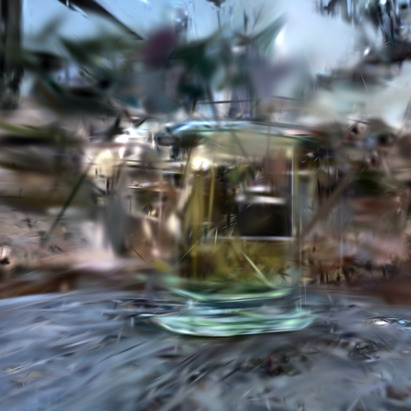}
    {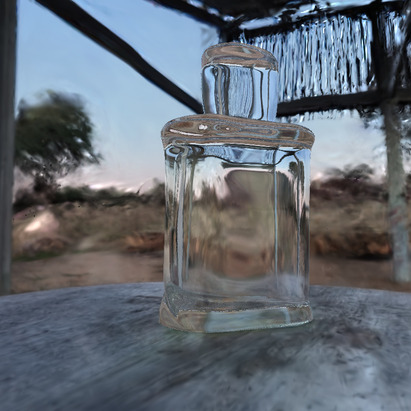}
    {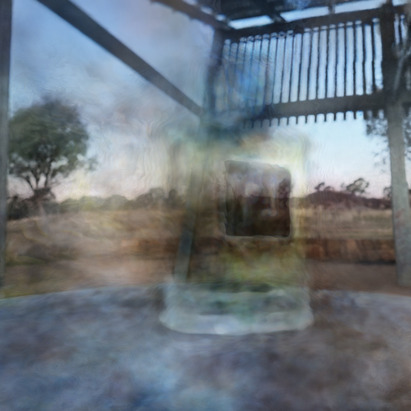}
    {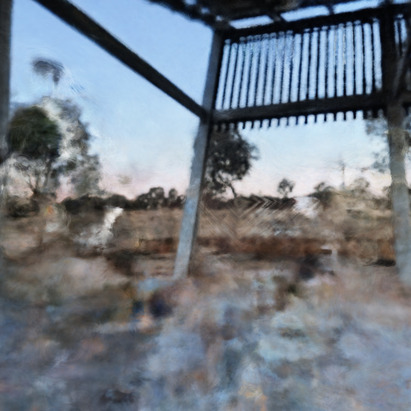}
    {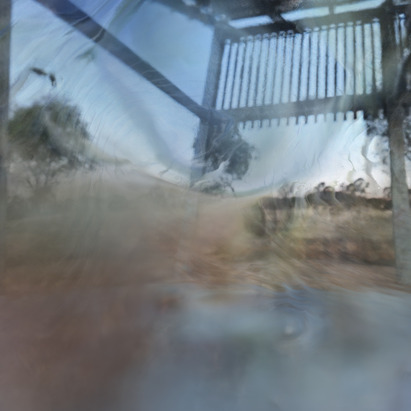}
    {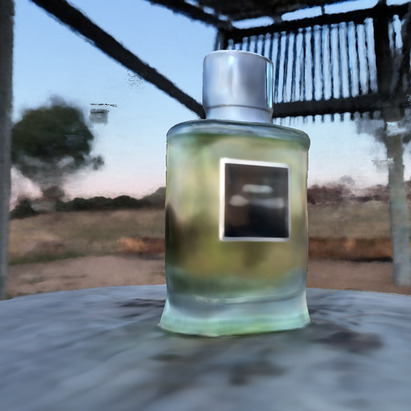}
    {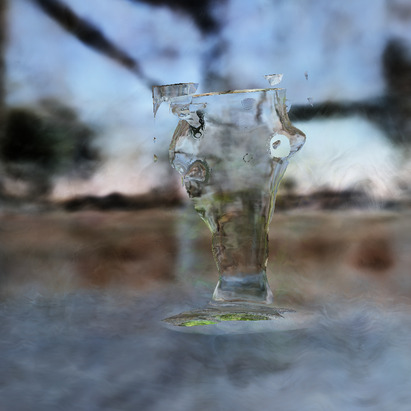}
    {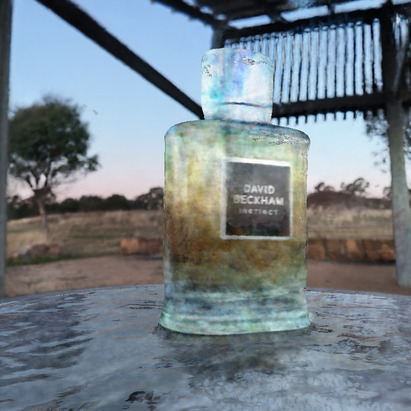}

\hdrrow
    {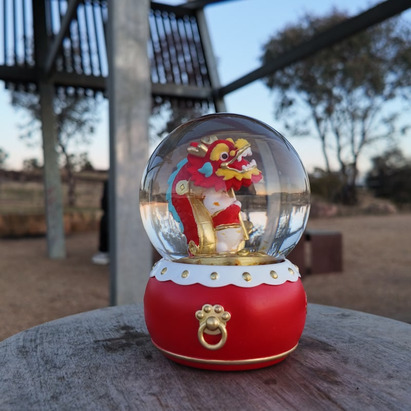}
    {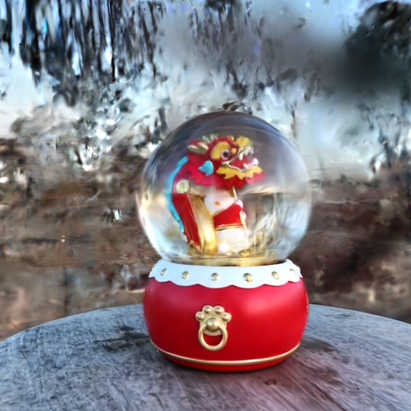}
    {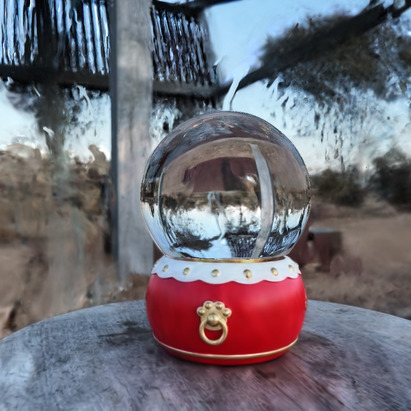}
    {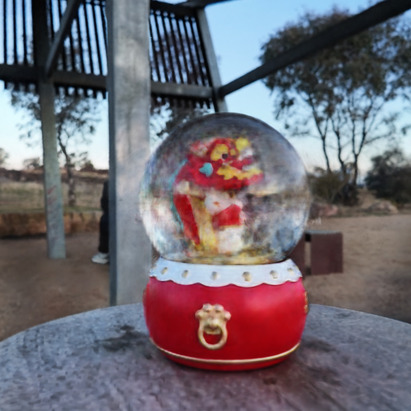}
    {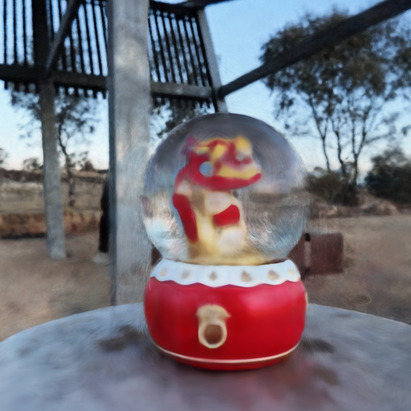}
    {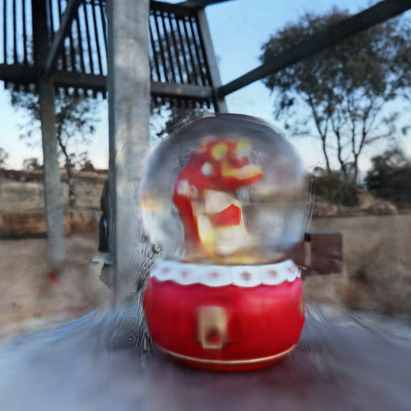}
    {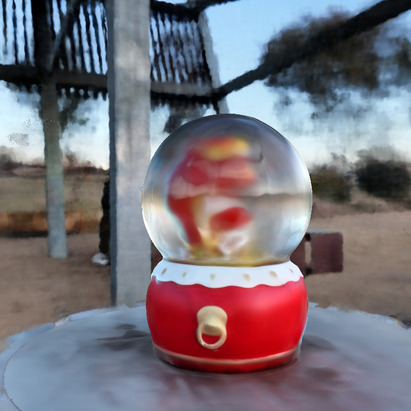}
    {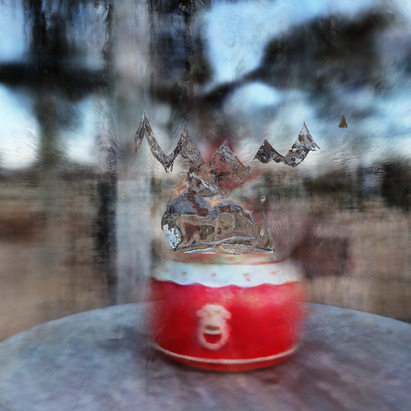}
    {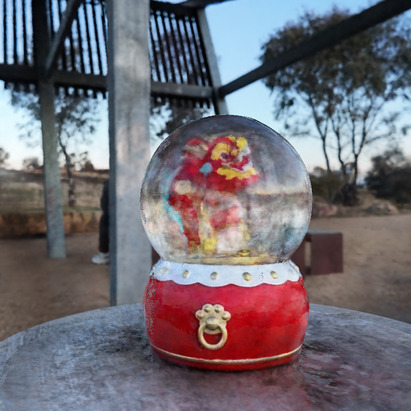}

\hdrrow
    {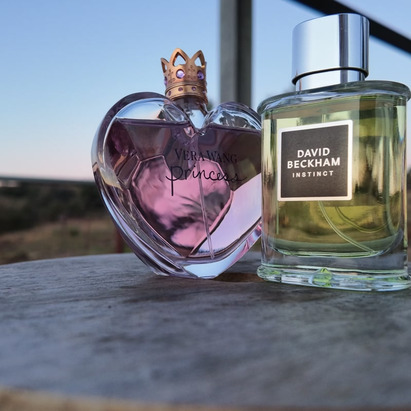}
    {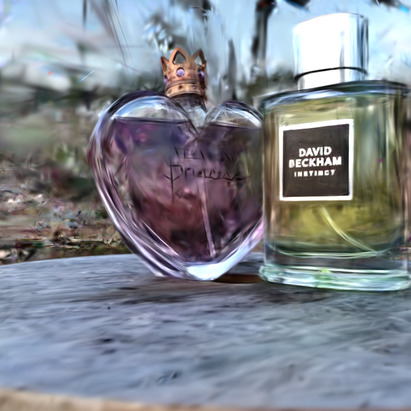}
    {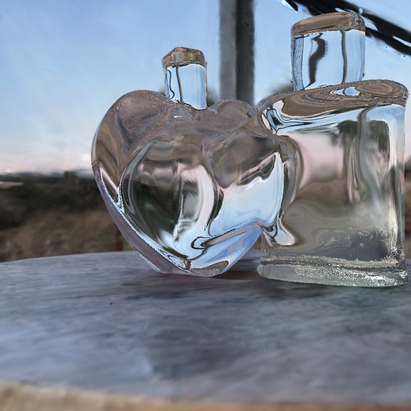}
    {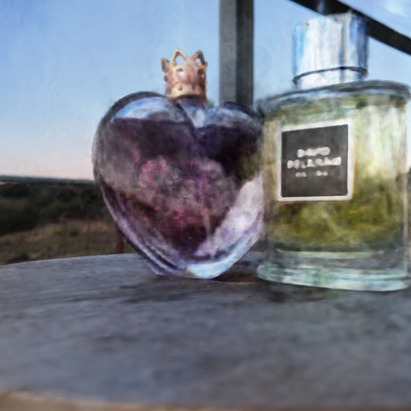}
    {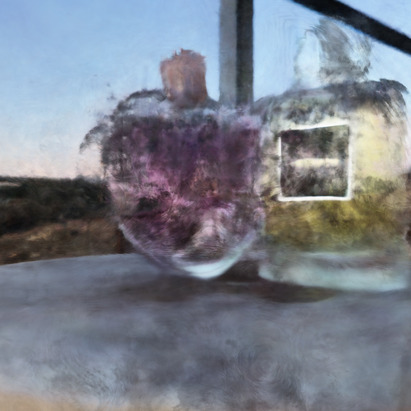}
    {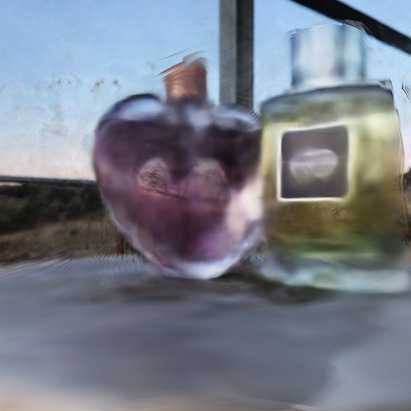}
    {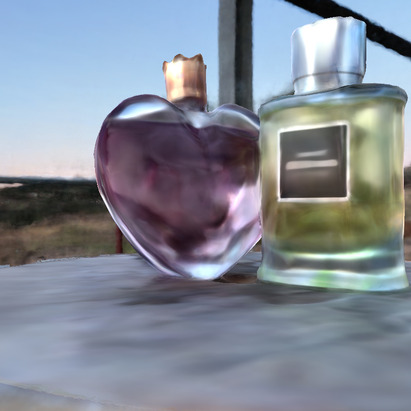}
    {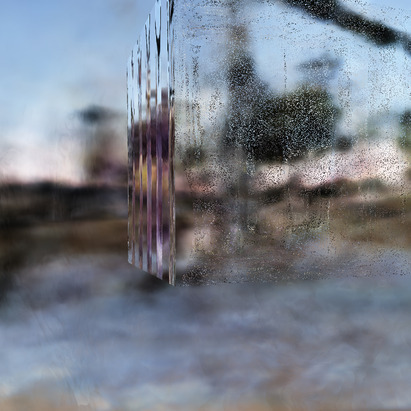}
    {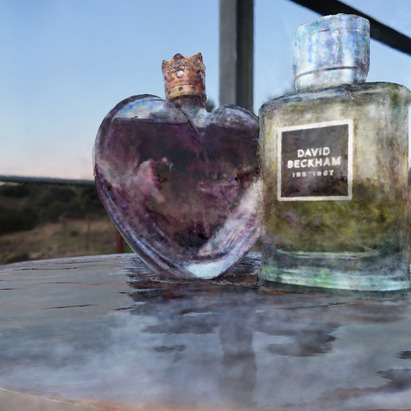}

\end{tabular}
\vspace{-3pt}
\caption{Outdoor scene backgrounds.}
\label{fig:real_outdoor_results}
\end{subfigure}
\vspace{-12pt}
\caption{
Additional qualitative comparison of novel view synthesis results on real scenes.
(a) shows results on indoor scenes, including living room and puzzle table backgrounds, while (b) shows results on outdoor scenes, including courtyard and lookout deck backgrounds.
}
\label{fig:real_qualitative_results}

\end{figure*}

\begin{figure*}[!t]
\centering

\setlength{\tabcolsep}{0.35pt}
\renewcommand{\arraystretch}{0}

\newcommand{\meshimage}[1]{%
    \includegraphics[width=0.098\linewidth]{#1}%
}

\newcommand{\meshrow}[9]{%
    \meshimage{#1} &
    \meshimage{#2} &
    \meshimage{#3} &
    \meshimage{#4} &
    \meshimage{#5} &
    \meshimage{#6} &
    \meshimage{#7} &
    \meshimage{#8} &
    \meshimage{#9}
    \\[0.3pt]
}

\begin{tabular}{@{}*{9}{c}@{}}

{\fontsize{6.5pt}{7.5pt}\selectfont Ground Truth} &
{\fontsize{6.5pt}{7.5pt}\selectfont Splatfacto} &
{\fontsize{6.5pt}{7.5pt}\selectfont TransparentGS} &
{\fontsize{6.5pt}{7.5pt}\selectfont Zip-NeRF} &
{\fontsize{6.5pt}{7.5pt}\selectfont MS-NeRF} &
{\fontsize{6.5pt}{7.5pt}\selectfont NeuS} &
{\fontsize{6.5pt}{7.5pt}\selectfont NU-NeRF} &
{\fontsize{6.5pt}{7.5pt}\selectfont TNSR} &
{\fontsize{6.5pt}{7.5pt}\selectfont \method (ours)}
\\[2pt]

\meshrow
    {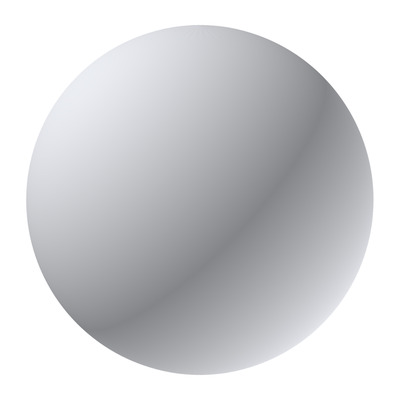}
    {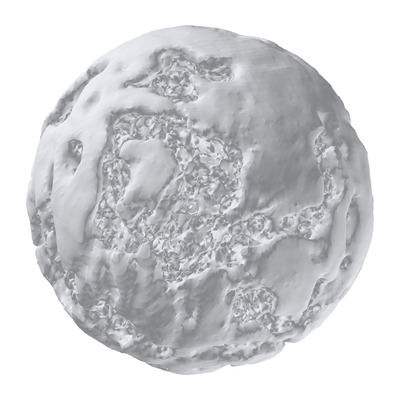}
    {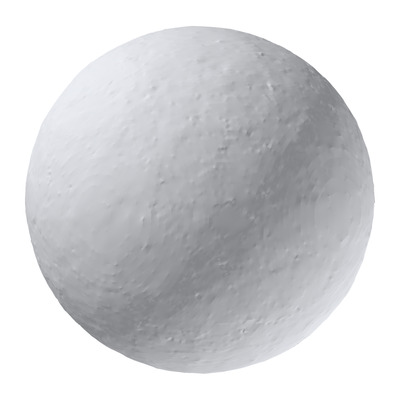}
    {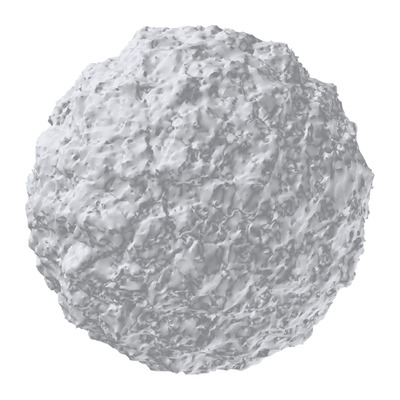}
    {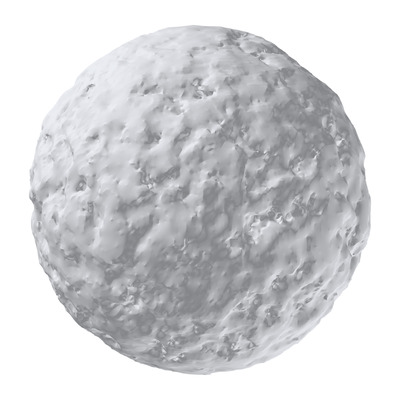}
    {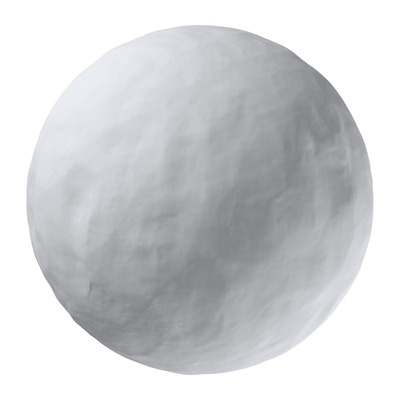}
    {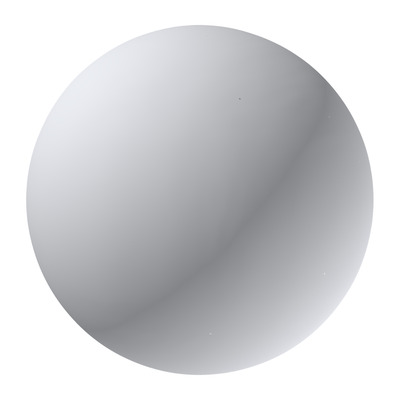}
    {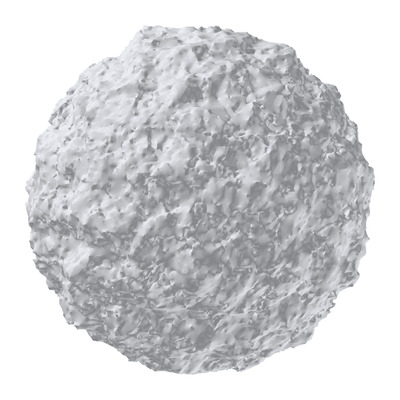}
    {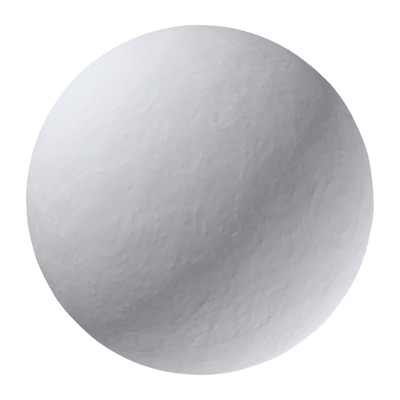}

\meshrow
    {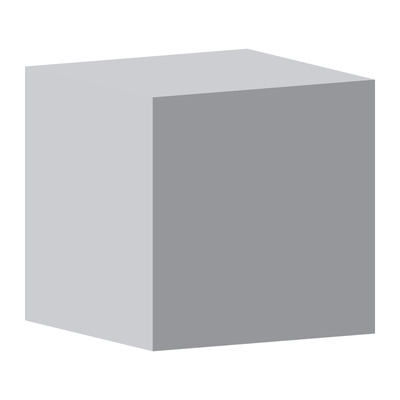}
    {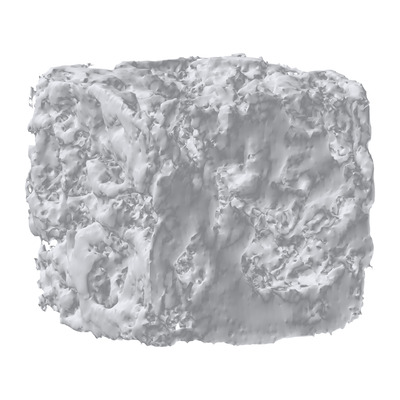}
    {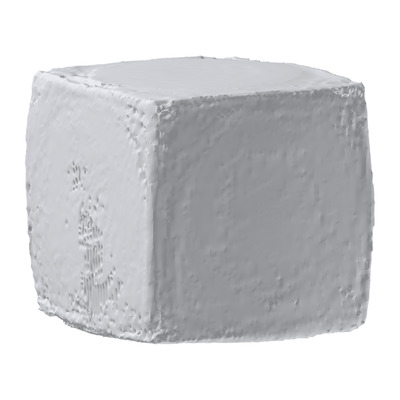}
    {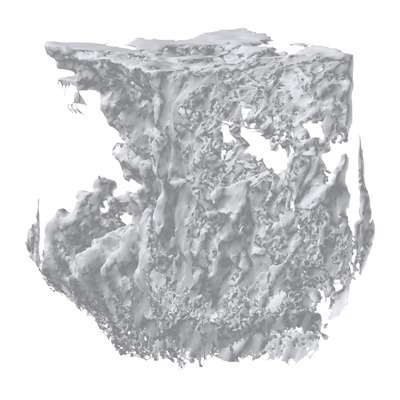}
    {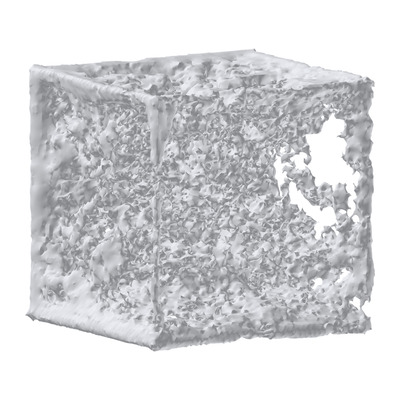}
    {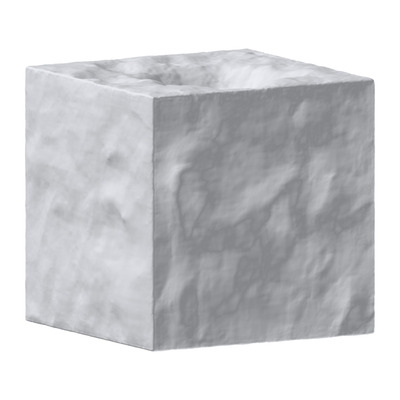}
    {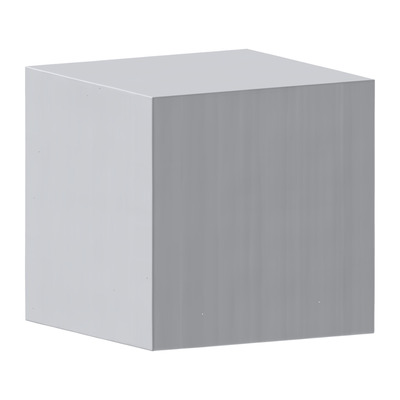}
    {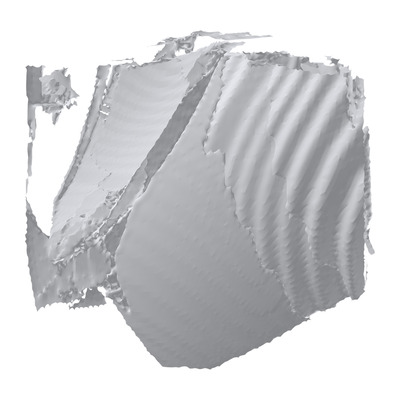}
    {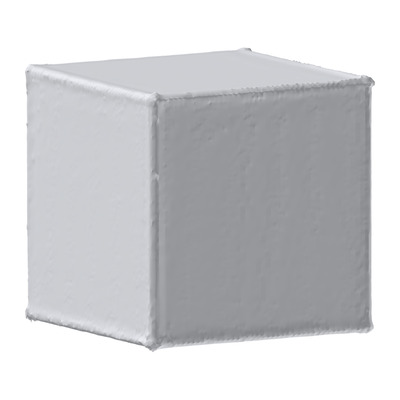}

\meshrow
    {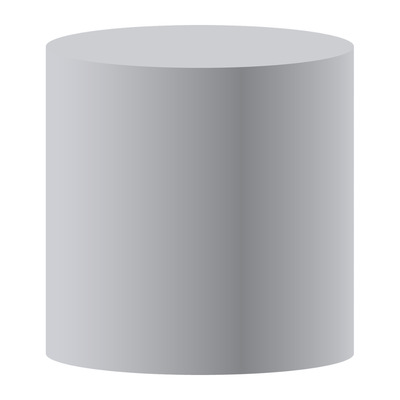}
    {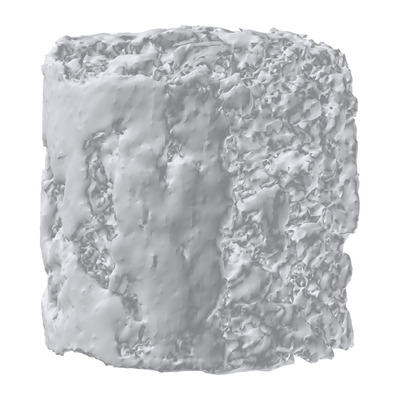}
    {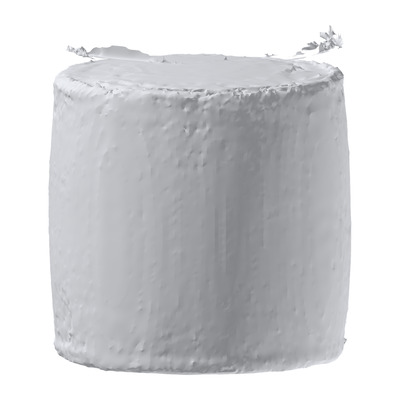}
    {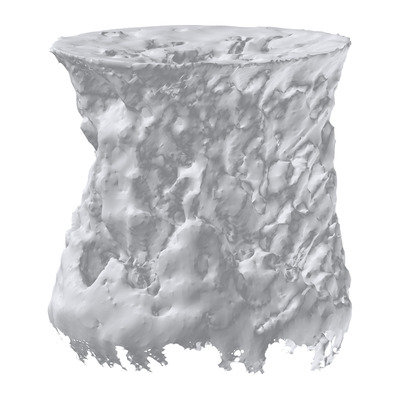}
    {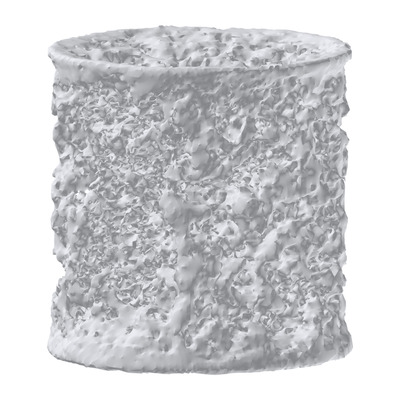}
    {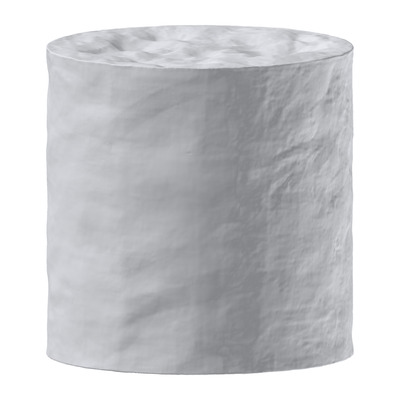}
    {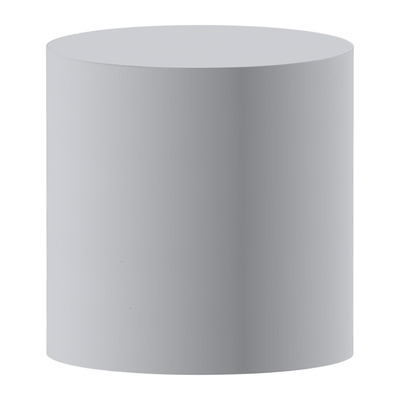}
    {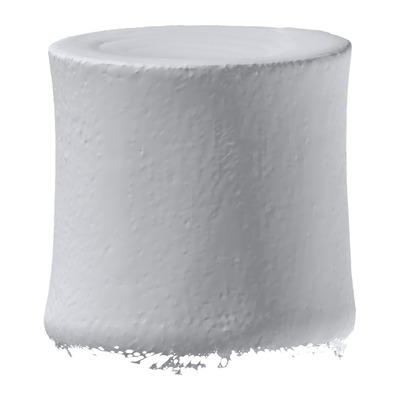}
    {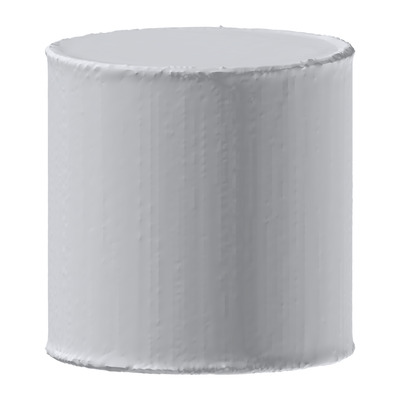}

\meshrow
    {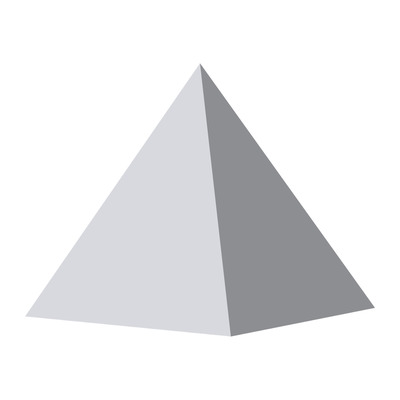}
    {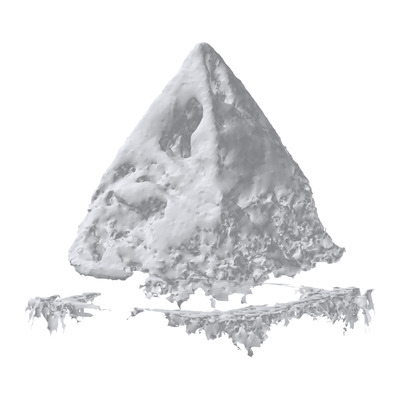}
    {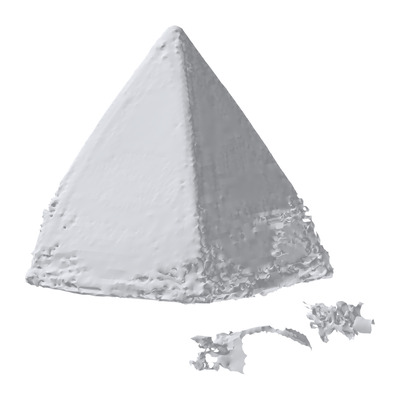}
    {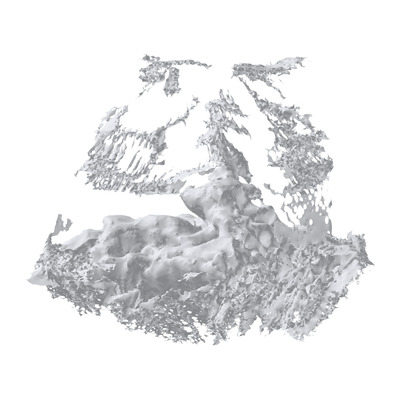}
    {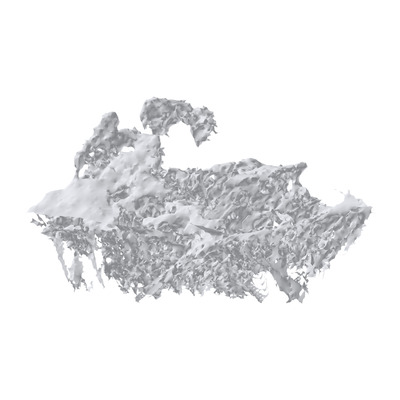}
    {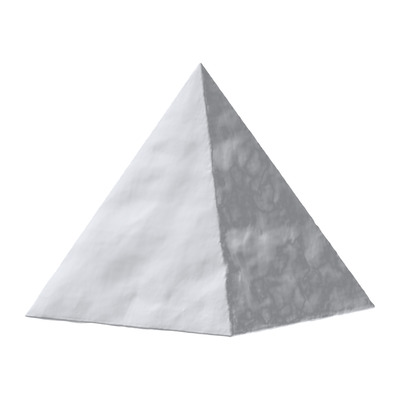}
    {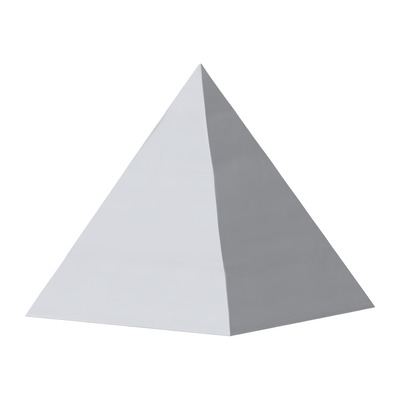}
    {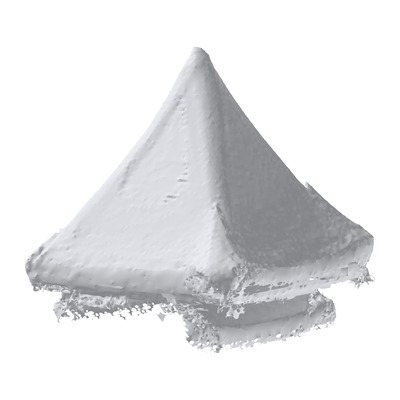}
    {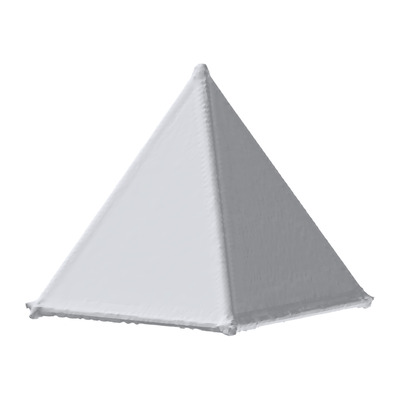}

\meshrow
    {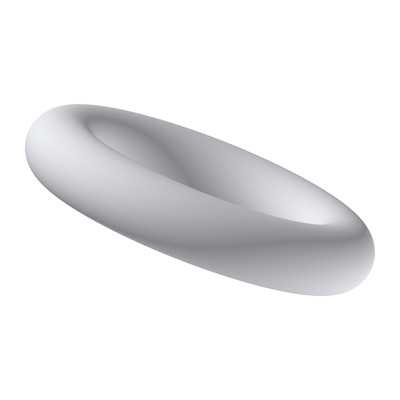}
    {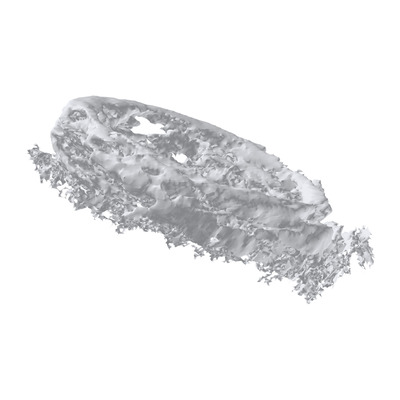}
    {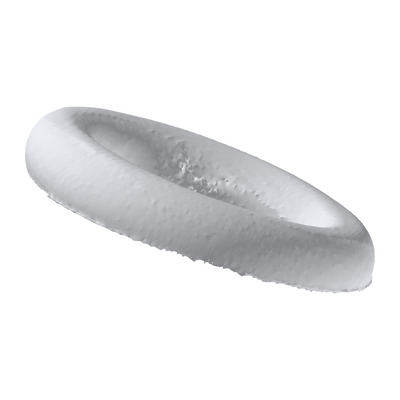}
    {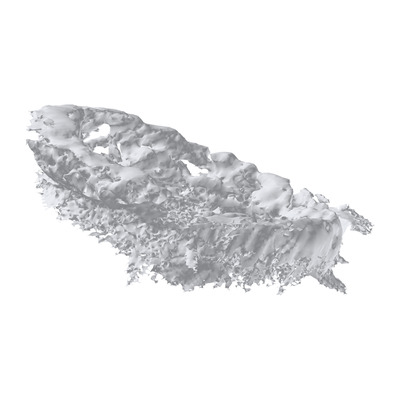}
    {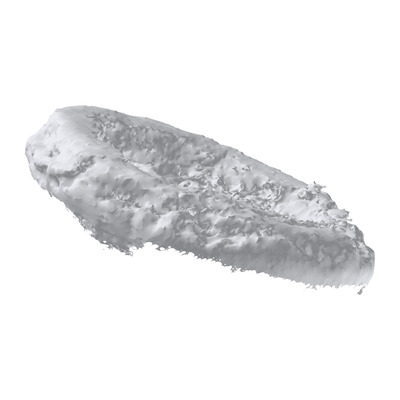}
    {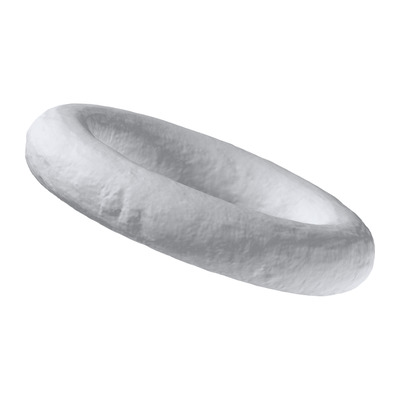}
    {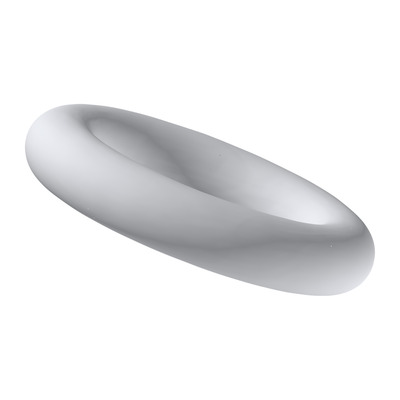}
    {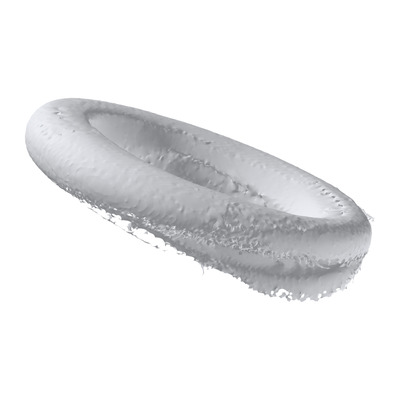}
    {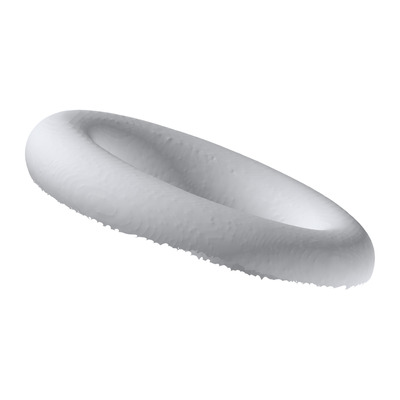}

\meshrow
    {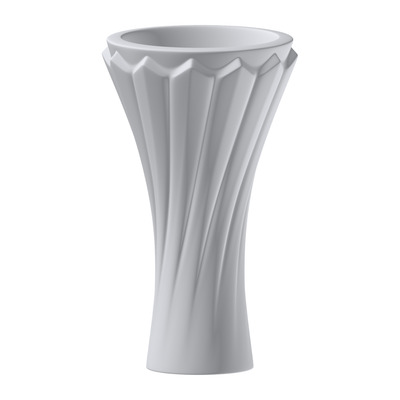}
    {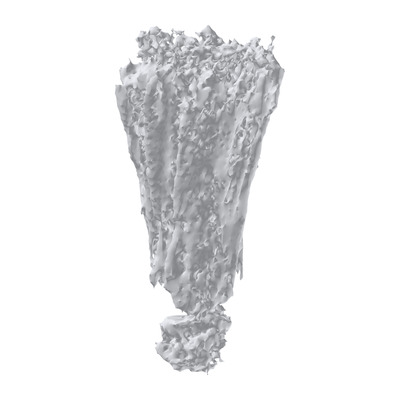}
    {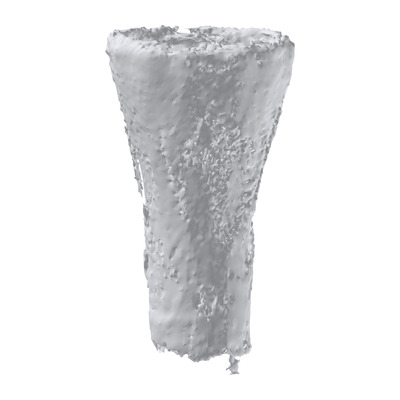}
    {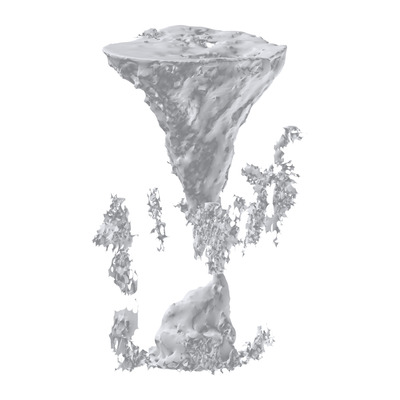}
    {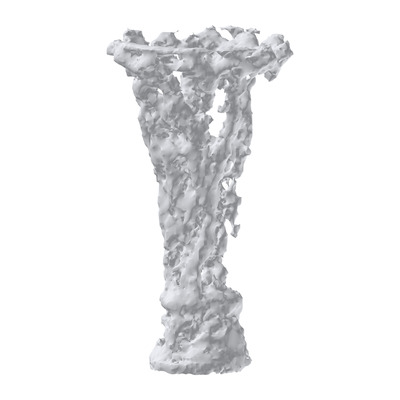}
    {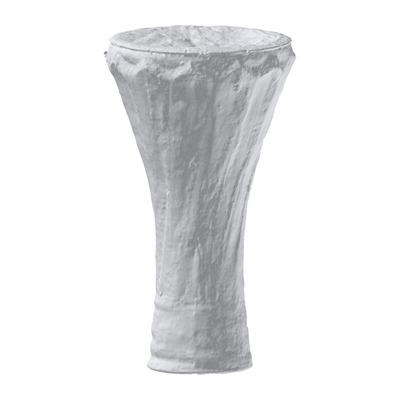}
    {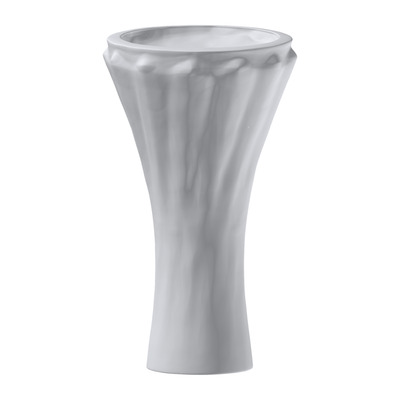}
    {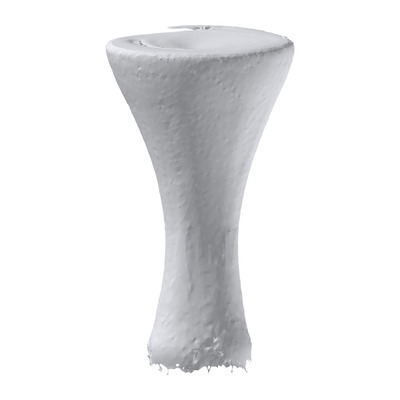}
    {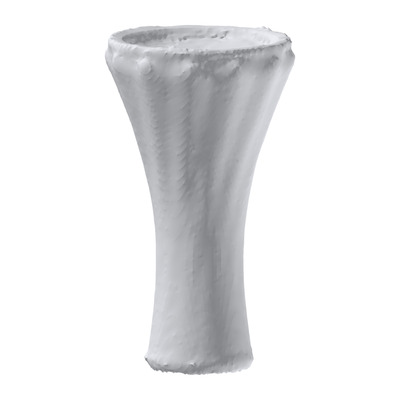}

\meshrow
    {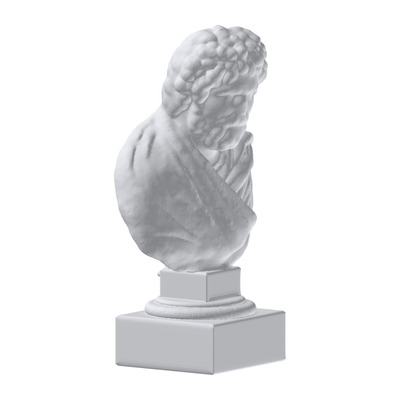}
    {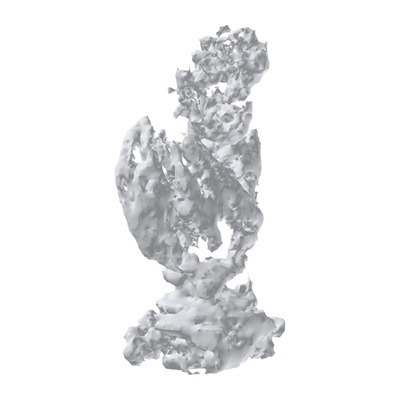}
    {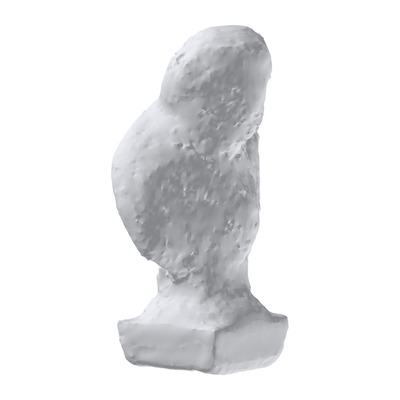}
    {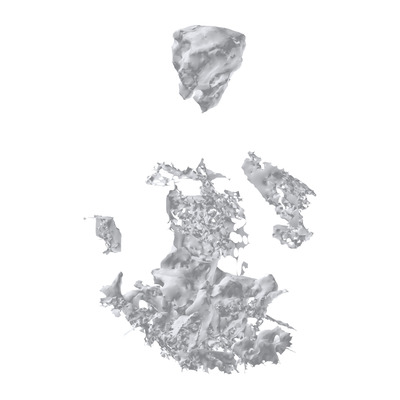}
    {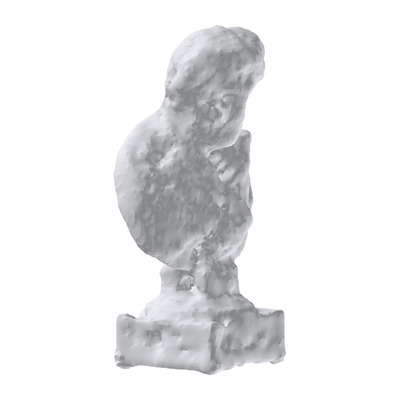}
    {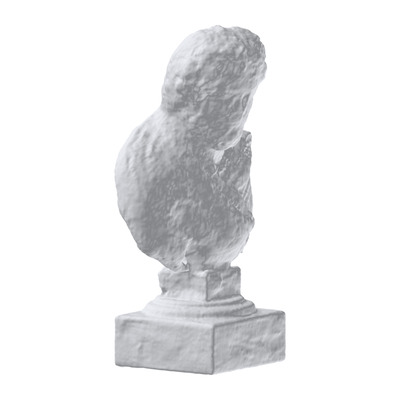}
    {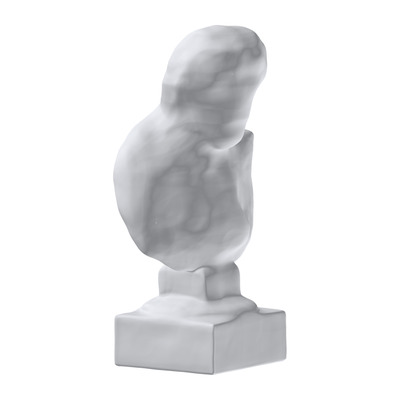}
    {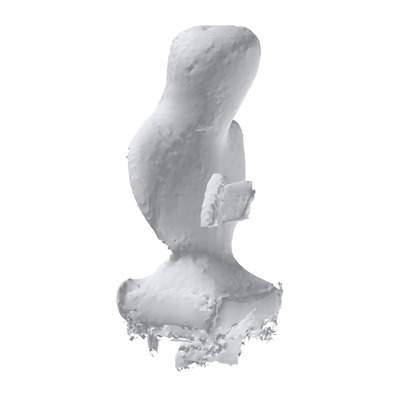}
    {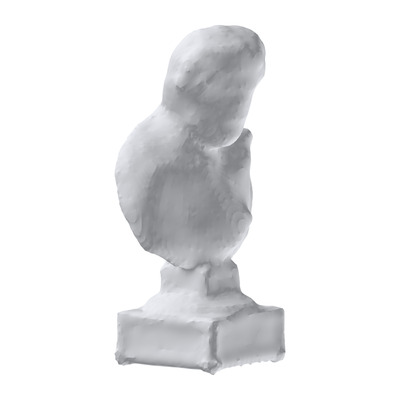}

\meshrow
    {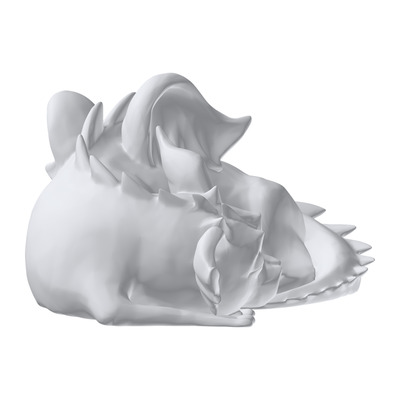}
    {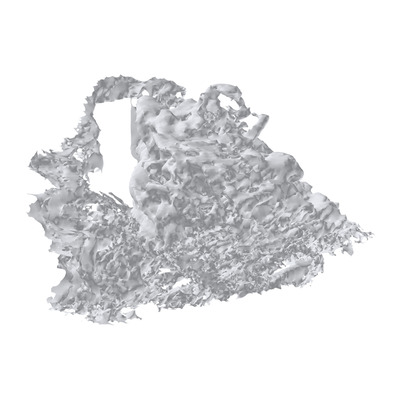}
    {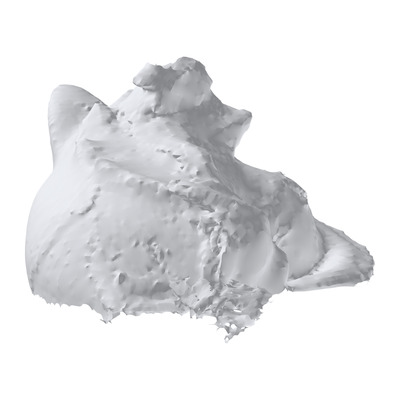}
    {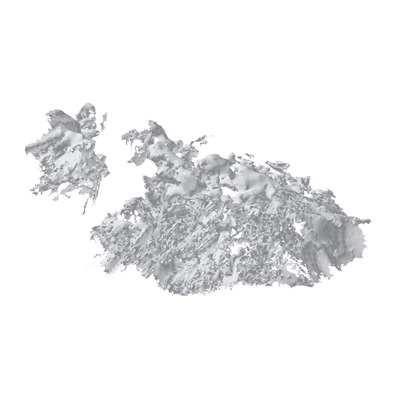}
    {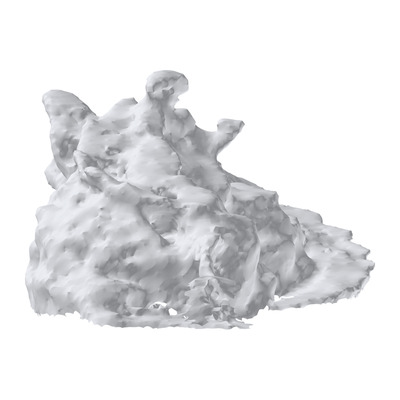}
    {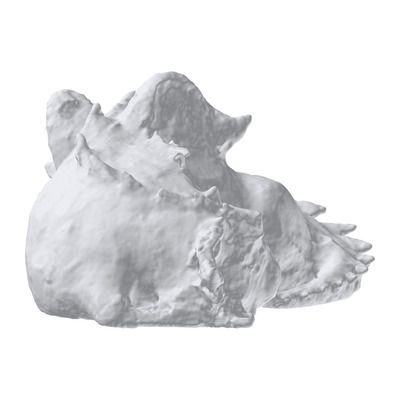}
    {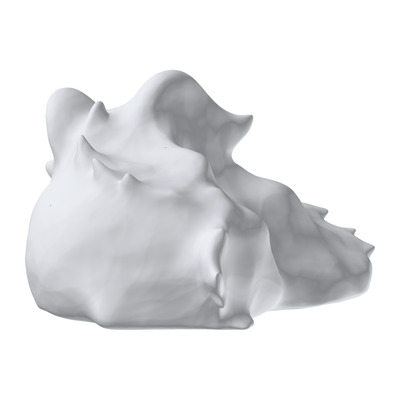}
    {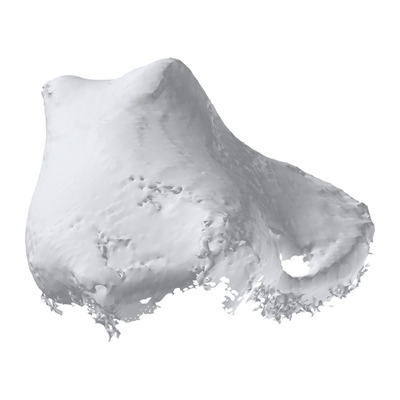}
    {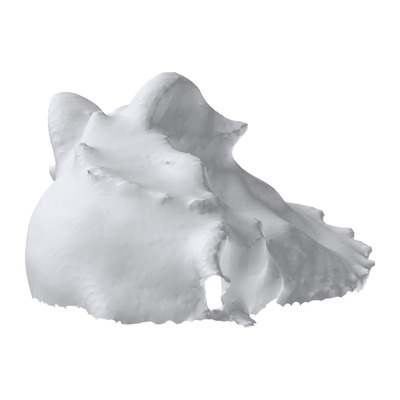}

\meshrow
    {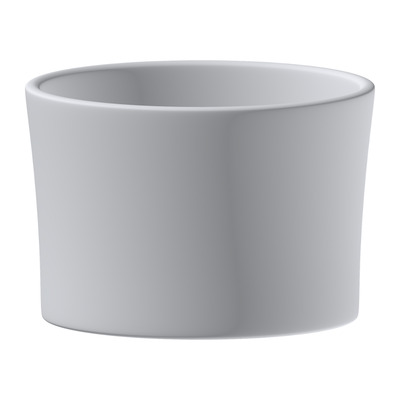}
    {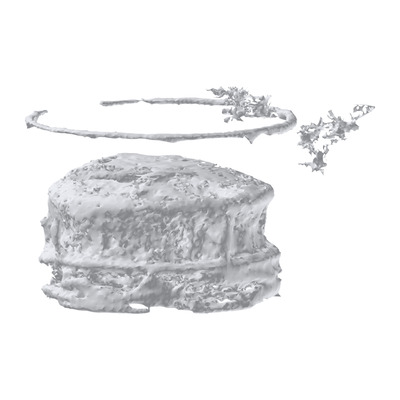}
    {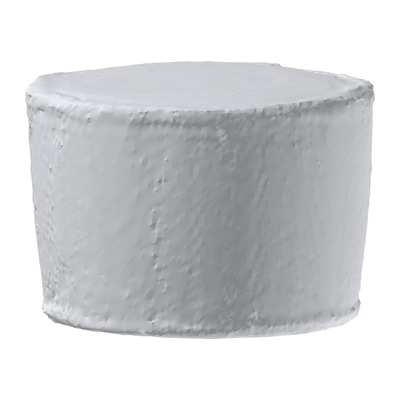}
    {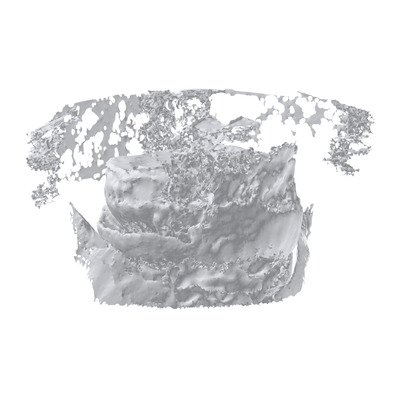}
    {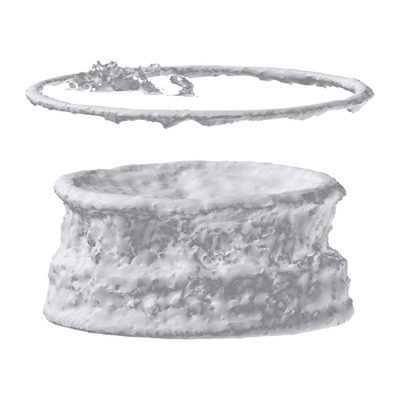}
    {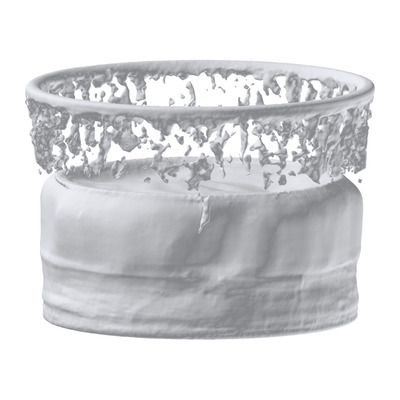}
    {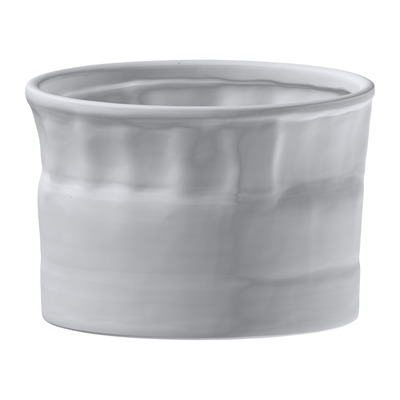}
    {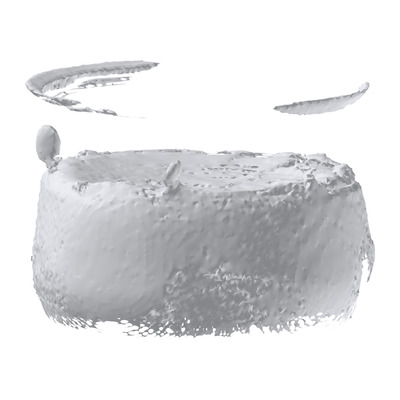}
    {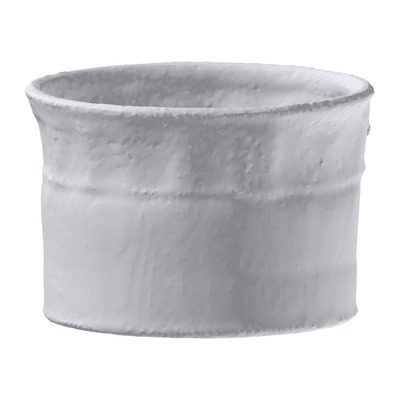}

\meshrow
    {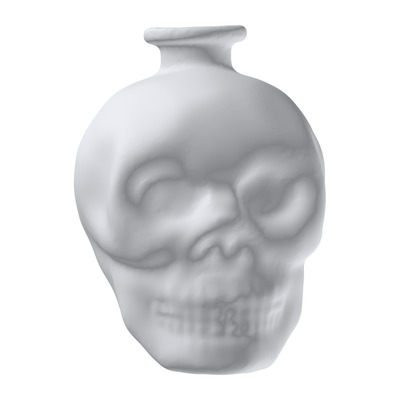}
    {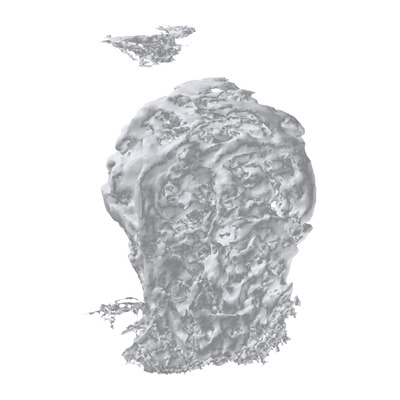}
    {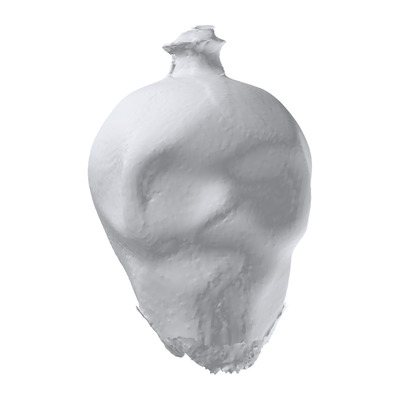}
    {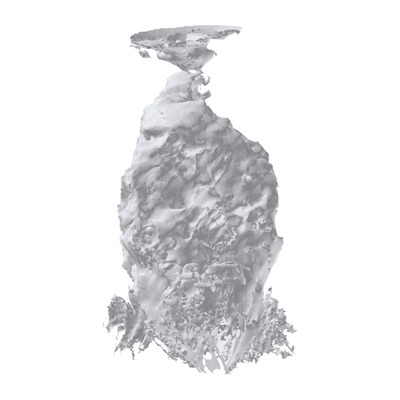}
    {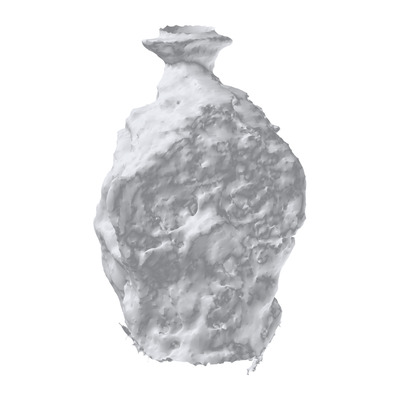}
    {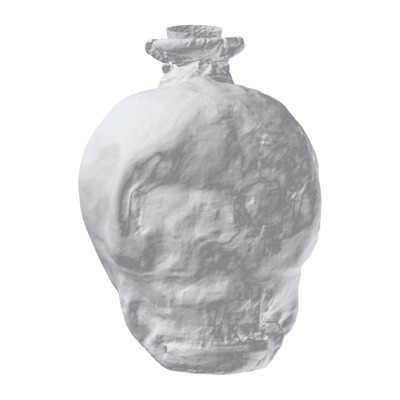}
    {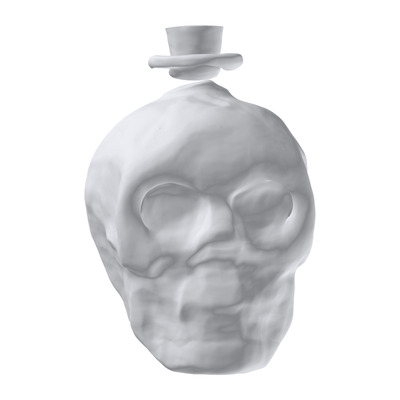}
    {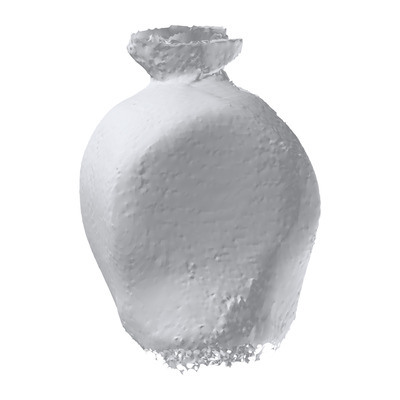}
    {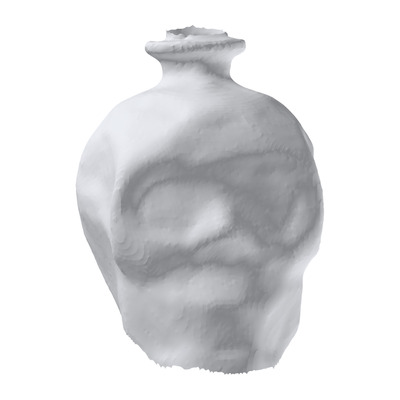}

\meshrow
    {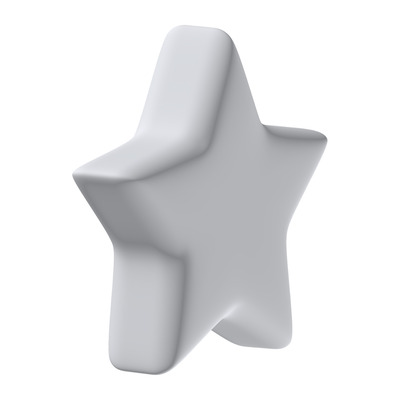}
    {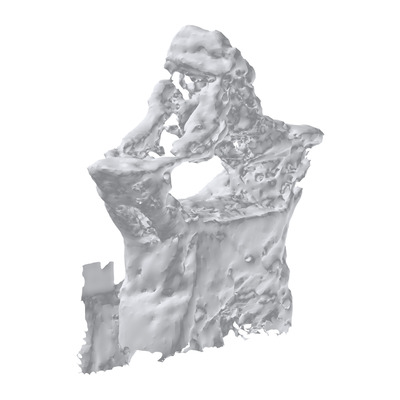}
    {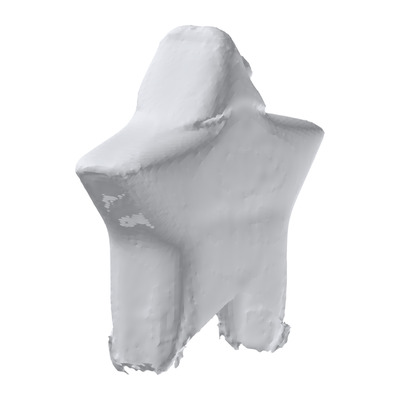}
    {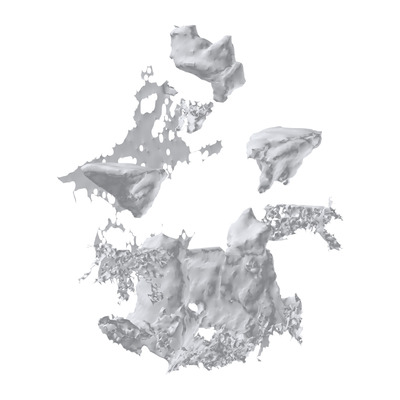}
    {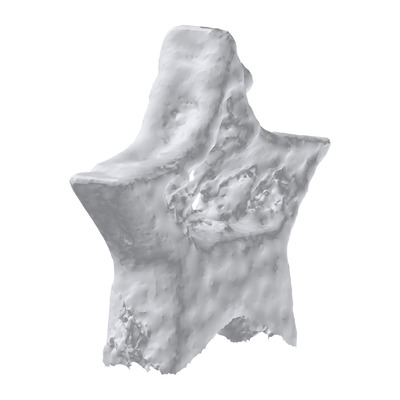}
    {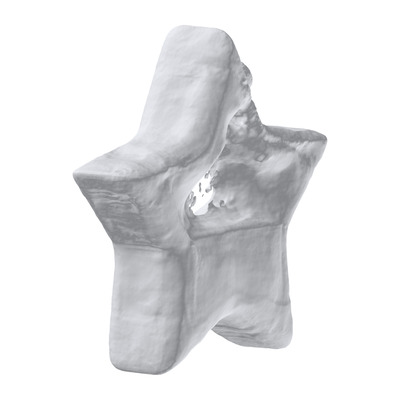}
    {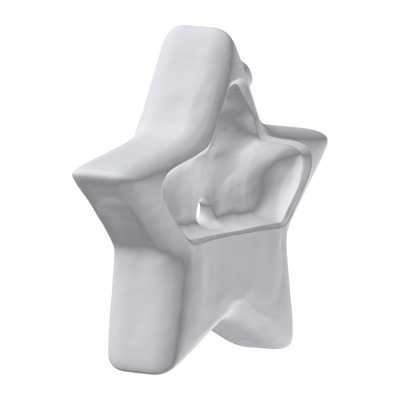}
    {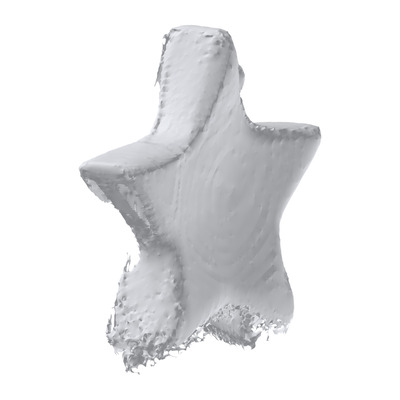}
    {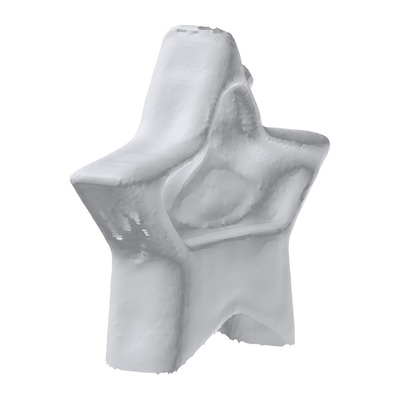}

\meshrow
    {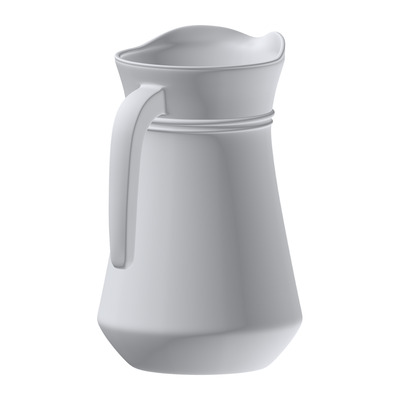}
    {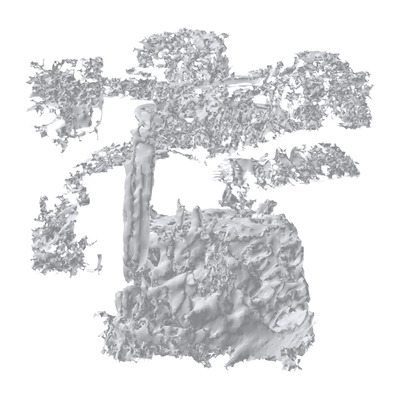}
    {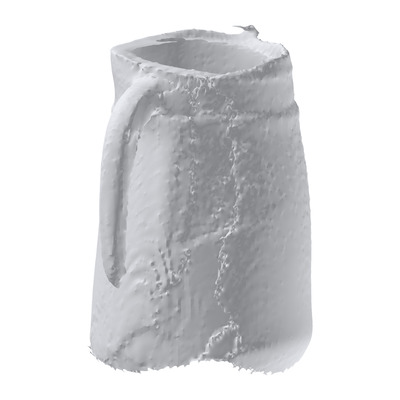}
    {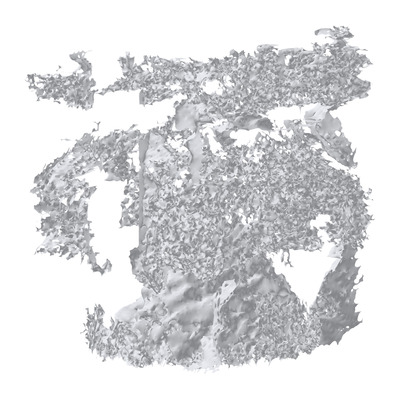}
    {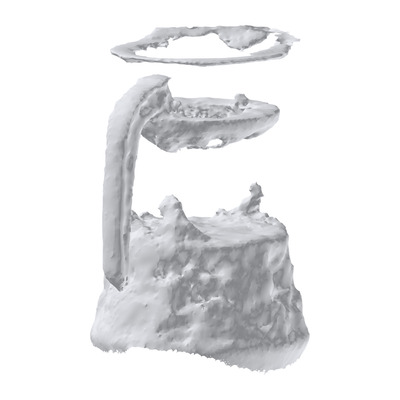}
    {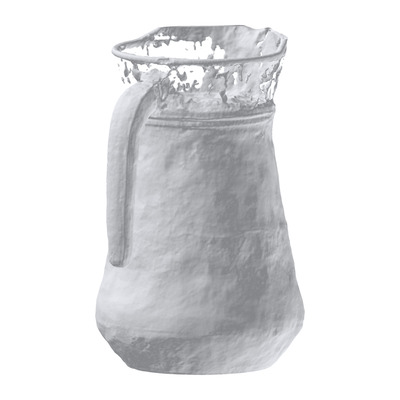}
    {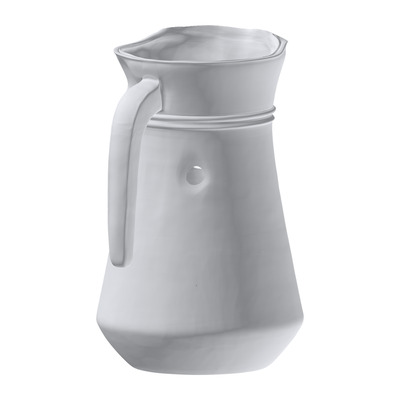}
    {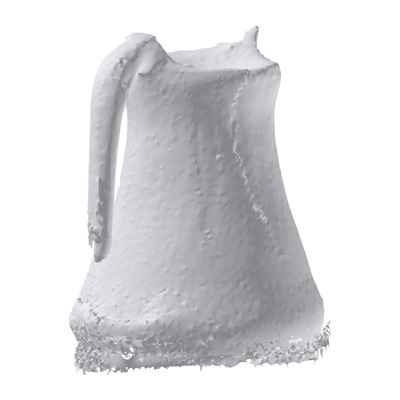}
    {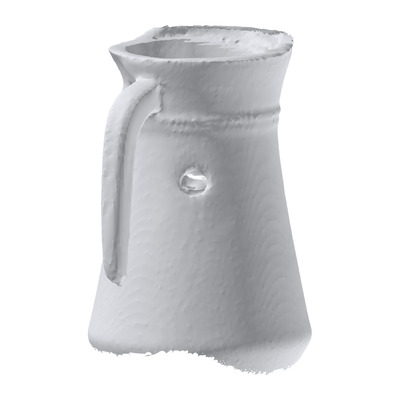}

\end{tabular}
\vspace{-4pt}
\caption{
Geometry visualization of the reconstruction results across different methods on
the test set from the \dataset{} dataset.
Inspecting the reconstructed geometry
directly exposes errors that are difficult to observe from the rendered RGB
images or the distance maps alone, such as fragmented surfaces and missing
structure.
}
\label{fig:mesh}
\end{figure*}